\documentclass{article}

    \PassOptionsToPackage{numbers, compress}{natbib}

     \usepackage[preprint]{neurips_2025}

\usepackage[utf8]{inputenc} 
\usepackage[T1]{fontenc}    
\usepackage{hyperref}       
\usepackage{url}            
\usepackage{booktabs}       
\usepackage{amsfonts}       
\usepackage{dsfont}
\usepackage{nicefrac}       
\usepackage{microtype}      
\usepackage{xcolor}         
\usepackage{bm}
\usepackage{comment}

\usepackage{amsmath}
\usepackage{amssymb}
\usepackage{mathtools}
\usepackage{amsthm}
\usepackage{algorithm}
\usepackage{algorithmic}

\newcommand{\deff}{\triangleq}

\newcommand{\norm}[1]{\|#1\|}

\newcommand{\eg}{\textit{e}.\textit{g}., \ }

\newtheorem{example}{Example}
\newtheorem{technicallemma}{Technical lemma}

\makeatletter
\newcommand*{\centernot}{%
  \mathpalette\@centernot
}
\def\@centernot#1#2{%
  \mathrel{%
    \rlap{%
      \settowidth\dimen@{$\m@th#1{#2}$}%
      \kern.5\dimen@
      \settowidth\dimen@{$\m@th#1=$}%
      \kern-.5\dimen@
      $\m@th#1\not$%
    }%
    {#2}%
  }%
}
\makeatother

\newcommand{\independent}{\perp\mkern-9.5mu\perp}
\newcommand{\notindependent}{\centernot{\independent}}

\usepackage[capitalize,noabbrev]{cleveref}

\theoremstyle{plain}
\newtheorem{theorem}{Theorem}[section]
\newtheorem{proposition}[theorem]{Proposition}
\newtheorem{lemma}[theorem]{Lemma}
\newtheorem{corollary}[theorem]{Corollary}
\theoremstyle{definition}
\newtheorem{definition}[theorem]{Definition}
\newtheorem{assumption}[theorem]{Assumption}
\theoremstyle{remark}
\newtheorem{remark}[theorem]{Remark}

\usepackage{todonotes}

\title{Handling Missing Data in Downstream Tasks With Distribution-Preserving Guarantees}

\author{%
  Rahul Bordoloi\thanks{Equal contribution.} \\
  Institute of Computer Science \\
  University of Rostock \\
  18051 Rostock, Germany \\
  \texttt{rahul.bordoloi@uni-rostock.de} \\
  \And
  Cl\'{e}mence R\'{e}da$^*$ \\
  Institute of Computer Science / Soda\\
  University of Rostock / Inria Saclay\\
  18051 Rostock, Germany / 91120 Palaiseau, France\\
  \texttt{clemence.reda@uni-rostock.de} \\
  \And
  Saptarshi Bej \\
  Department of Data Science \\
  IISER Thiruvananthapuram \\
  695551 Kerala, India \\
  \texttt{sbej7042@iisertvm.ac.in} \\
  \And
  Olaf Wolkenhauer \\
  Institute of Computer Science \\
  University of Rostock \\
  18051 Rostock, Germany \\
  \texttt{olaf.wolkenhauer@uni-rostock.de} \\
}

\begin{document}

\maketitle

\begin{abstract}
  Missing feature values are a significant hurdle for downstream machine-learning tasks such as classification. However, imputation methods for classification might be time-consuming for high-dimensional data, and offer few theoretical guarantees on the preservation of the data distribution and imputation quality, especially for not-missing-at-random mechanisms. First, we propose an imputation approach named F3I based on the iterative improvement of a K-nearest neighbor imputation, where neighbor-specific weights are learned through the optimization of a novel concave, differentiable objective function related to the preservation of the data distribution on non-missing values. F3I can then be chained to and jointly trained with any classifier architecture. Second, we provide a theoretical analysis of imputation quality and data distribution preservation by F3I for several types of missing mechanisms. Finally, we demonstrate the superior performance of F3I on several imputation and classification tasks, with applications to drug repurposing and handwritten-digit recognition data.
\end{abstract}

\section{Introduction}\label{sec:introduction}

Most machine-learning approaches assume full access to the features of the input data points. However, missing values might arise due to the incompleteness of public databases or measurement errors. Research on the imputation of missing values and inference on possibly missing data is motivated by the fact that naive approaches would not fare well. Indeed, ignoring samples with missing values might lead to severe data loss and meaningless downstream models, for classification or regression. Yet, replacing missing values with zeroes (or any ``simple'' univariate approach such as taking the mean or the median) can considerably distort the distribution of data values, as there is a more significant weight on the default value for missing entries, and then perhaps bias the training of a downstream model for classification or regression tasks. However, multivariate approaches are often time-consuming and prohibitive for high-dimensional data sets, as in biology.

The literature often distinguishes three main categories of missingness mechanisms~\cite{rubin1976inference} depending on the relationship between the probability $p^\text{miss}$ of a missing value and the data. 
The simplest one is Missing-Completely-At-Random (MCAR), where that probability is independent of the data. This can be applied when the measurement tools fail at some probability, regardless of the analyzed data. The second, more complex, setting is Missing-At-Random (MAR), where $p^\text{miss}$ depends solely on the observed (not missing) data. An example of MAR is when male patients drop out more often from a clinical study than female patients. Finally, the Missing-Not-At-Random setting (MNAR), where $p^\text{miss}$ depends on both the observed and missing data, is widely regarded as the most challenging setting for analysis because the actual values might not be identifiable. 

\section{Related work}\label{sec:related_work}

As previously mentioned, the fastest approaches to imputation are often univariate because they are simple operations applied feature-wise to a dataset. For instance, the missing value for a feature corresponding to a column of the data matrix might be replaced by the mean or the median of all non-missing values or even by zeroes in that column. Yet, such naive approaches might severely distort the distribution of values~\cite{morvan2024imputation}. 

That fact opened the path to multiple multivariate methods, such as MICE~\cite{van2011mice}, MissForest~\cite{stekhoven2012missforest}, RF-GAP~\cite{rhodes2023geometry} resorting to random forests; MIDAS~\cite{seu2022intelligent} using denoising auto-encoders; Optimal Transport-based algorithms~\cite{muzellec2020missing}; but also matrix factorizations~\cite{mazumder2010spectral}, penalized logistic regression methods~\cite{van2024imputation}, Bayesian network-based approaches (for instance, MIWAE~\cite{mattei2019miwae} for MAR mechanisms and its MNAR counterpart not-MIWAE~\cite{ipsen2021not}). 
Some recent works also provide a pipeline for the automated finetuning and refinement of imputers, such as MIRACLE~\cite{kyono2021miracle} or HyperImpute~\cite{jarrett2022hyperimpute}. 
 We dwell further on newer diffusion model and deep learning-based approaches in 
 Appendix (Section~\ref{app:relatedworks}).
 However, as the number of features increases, so does the computation time, making most of those approaches untractable on practical data sets. For instance, in genomic data, the feature set (genes) can amount to as many as $20,000$ genes in humans. 

Moreover, many published imputation methods come without any guarantee on the quality of the imputation or often on the more straightforward settings such as MCAR~\cite{mazumder2010spectral} and MAR~\cite{smieja2018processing}; with a few exceptions such as~\cite{tang2003analysis,mohan2018estimation,sportisse2020estimation} for pure imputation tasks, and NeuMiss networks~\cite{le2020neumiss}, which tackle a classification task in the presence of missing values. However, the MCAR and MAR settings are usually not applicable to real-life data, and~\cite{tang2003analysis,mohan2018estimation,sportisse2020estimation} rely on an assumption of data generation through low-rank or linear random models instead of simpler data distributions. 

Finally, nearest-neighbor imputers~\cite{troyanskaya2001missing} are known to be performant in practice and relatively fast~\cite{emmanuel2021survey,seu2022intelligent,joel2024performance}, at the price of some distortion in high-dimensional data sets~\cite{beretta2016nearest}. This observation led us to consider an improvement of a nearest-neighbor imputation that preserves the data distribution even in larger dimensions while remaining computationally fast.

\subsection{Contributions}

In Section~\ref{sec:contribution}, we describe an algorithm named Fast Iterative Improvement for Imputation (F3I) based on the distribution-preserving improvement of a $K$ nearest-neighbor imputer. F3I combines two ingredients: (1) a novel concave and differentiable objective function to quantify the preservation of data distribution during imputation; and (2) a fast routine to optimize that function through the weights in the nearest-neighbor imputation, by drawing a parallel with the problem of expert advice in online learning~\cite{cesa2006prediction,de2014follow,lattimore2020bandit}. This algorithm gives theoretical upper bounds on the imputation quality and the preservation of initial data distribution in some MCAR, MAR, and MNAR settings (Section~\ref{sec:analysis}). Furthermore, this imputation can also be chained to and jointly trained with a downstream task, \textit{e}.\textit{g}., classification, to increase its accuracy (Section~\ref{sec:joint_training}). Finally, we illustrate the performance of F3I compared to several baselines on standard data sets for imputation and classification (Section~\ref{sec:experiments_main}), and showcase its applications to drug repurposing and handwritten digit recognition (Appendix~\ref{sec:experiments}).

\subsection{Notation}

We denote $N$, $F$ and $K$ the number of samples, features, and nearest neighbors. $i$, $j$, $f$ are integer indices. For all other alphabet letters, $v$ is a scalar, $\bm{v}$ a vector, and $V$ a matrix. $\bm{v}^i$ is the $i^{th}$ column and $\bm{v}_j$ is the $j^{th}$ row of matrix $V$, and $v^j_i$ is the coefficient at position $(i,j)$ in $V$ for $i,j >0$. For any $K \geq 2$, $\triangle_K \triangleq \{\bm{p} \in [0,1]^K \mid \sum_{k \leq K} p_k = 1 \}$ is the simplex of dimension $K$. 
The initial data matrix with missing values is $X \in (\mathbb{R} \cup \{\texttt{NaN}\})^{N \times F}$, where $X^\star \in \mathbb{R}^{N \times F}$ is the full (unavailable) data matrix. Finally, $m^f_i \in \{0,1\}$ is the random variable that indicates whether the value at position $(i,f)$ is missing in the input data matrix, where $m^f_i=1$ means it is missing. 

\section{The Fast Iterative Improvement for Imputation (F3I) algorithm}\label{sec:contribution}

The key idea is that we would like to replace missing positions in $X$ with the most probable values by iteratively applying a ``good'' \textit{weighted combination of elements in the data set}, starting from a guess, \eg through K-nearest neighbor imputation. For each value $x^f_i$, we are looking to tune the weights $\bm{\alpha} = (\alpha_1, \alpha_2, \dots, \alpha_K)$ of a convex combination of the $K$ closest neighbors of $\bm{x}_i$ in a reference set $Z \in \mathbb{R}^{N \times F}$ without missing values, $\bm{z}_{i(1)}, \bm{z}_{i(2)}, \dots, \bm{z}_{i(K)}$, ordered by their increasing distance to $\bm{x}_i$. We denote that imputation improvement model $\texttt{Impute}(\bm{x}_i ; \bm{\alpha}, Z)$, which is fully described in Appendix (Algorithm~\ref{alg:imputation}). When the reference set is obvious, we define $\bm{x}(\bm{\alpha}) \triangleq \texttt{Impute}(\bm{x} ; \bm{\alpha}, Z)$.

\subsection{Theoretical assumptions}\label{subsec:assumption}

Assumptions are important--and all unrealistic to some extent--to control the behavior of the imputed values compared to the ground truth values, which are in essence unavailable, during the theoretical analysis to assess the imputation quality and the preservation of data distribution. Theoretical results on those controlled settings allow data-independent comparisons and are key to robust scientific advancement. In this section, we informally state our assumptions about the data generation procedure to derive theoretical guarantees. Formal statements can be found in Appendix (Section~\ref{sec:assumptions} and Algorithm~\ref{alg:data_generation}). In practice, those assumptions can be ignored, and F3I can be applied to real-life data, as shown in Section~\ref{sec:experiments_main}. First, we assume that each value in the full data matrix is drawn from independent fixed-variance Gaussian distributions 
(Assumption~\ref{as:x_distribution}). 

Second, the random indicator variables $m^f_i$ are then independently drawn according to the missingness mechanism with probability $p^\text{miss}$. If $m^f_i=1$, then the coefficient at position $(i,f)$, $x^f_i$ in $X$ is unavailable, otherwise, $x^f_i=(x^\star)^f_i$. We provide an analysis of our algorithm for three types of missingness mechanisms: a MCAR mechanism where the random indicator variables are drawn from Bernoulli law with fixed mean (Assumption~\ref{as:mcar}); a MAR mechanism where 
the probability of missingness depends only on observed values of the data set (Assumption~\ref{as:mar}); and, finally, a MNAR mechanism called Gaussian self-masking (Assumption $4$ from~\cite{le2020neumiss}) where the probability of $m^f_i=1$ is proportional to $\exp(-\zeta^{-2}((x^\star)^f_i-\mu_f)^2)$ where $\mu_f$ is specific to feature $f$ and $\zeta$ is fixed (Assumption~\ref{as:mnar}). 

Third, we also ensure that there are exactly $K$ neighbors for the initial guesses (Assumption~\ref{as:number_neighbors}), and that we know a constant upper bound on the norm of any feature vectors (Assumption~\ref{as:ub_norm}). Then, the goal of our algorithm F3I is to determine the proper weights in a nearest-neighbor imputation in a data-driven way that preserves the data distribution. But how do we define that property?

\subsection{A novel objective function for optimizing the preservation of the initial data distribution}

Ideally, if we had access to the true distribution $\mathcal{D}$ on feature vectors, we would like to set the weights $\bm{\alpha} \in \triangle_K$, the simplex of dimension $K$, such that the following quantity is maximized 
$\mathbb{E}_{\bm{x} \sim \mathcal{D}}\left[\mathds{1}(\Phi_{\Theta}(\texttt{Impute}\big(\bm{x}^0; \bm{\alpha},Z)) > \Phi_{\Theta}(\bm{x}^0)\big)\right]$, where $\mathds{1}$ is the Kronecker symbol, $\bm{x}^0 \in \mathbb{R}^F$ is an initial guess on the missing values in $\bm{x} \in (\mathbb{R} \cup \{\texttt{NA}\})^F$, and $\Phi_{\Theta}$ is the parametrized true data distribution according to Assumption~\ref{as:x_distribution}. That is, we want to choose $\bm{\alpha}$ so that the imputed values are more probable than the current guesses. Considering a \textit{full} data set of $N$ $F$-dimensional points $Z = \{(\bm{x}^0)_1, (\bm{x}^0)_2, \dots, (\bm{x}^0)_N\} \subset \mathbb{R}^F$ of initial guesses on the missing values in $X$ and approximating the true distribution $\mathcal{D}$ by a density kernel $D_0$ on $Z$, we would like to maximize $\mathds{1}\left(D_0((\bm{x}^0)_i(\bm{\alpha})) > D_0((\bm{x}^0)_i)\right)$ for each sample $\bm{x}_i$, $i \leq N$. That quantity can be approximated by 
\begin{eqnarray*}
  \max\left(0, \frac{D_0((\bm{x}^0)_i(\bm{\alpha}))}{D_0((\bm{x}^0)_i)}-1\right) \approx \log  \left(\frac{D_0((\bm{x}^0)_i(\bm{\alpha}))}{D_0((\bm{x}^0)_i)}\right)\;, 
\end{eqnarray*}
To restrict overfitting, we can add a $\ell_2$-regularization on the parameter $\bm{\alpha}$ with a regularization factor $\eta$. If we estimate a Gaussian kernel over the reference set $Z$, then the density kernel $D_0$ is defined as 
$D_0 : \bm{x} \in \mathbb{R}^F \mapsto 1/N \sum_{j \leq N} (\sqrt{2 \pi}h)^{-F} \exp(-\|\bm{x}-(\bm{x}^0)_j\|^2_2/(4h))$, where $h$ is the kernel width. Finally, we define for any $\bm{\alpha} \in \triangle_K$, $X \in \mathbb{R}^{N \times F}$ and $\eta \geq 0$, the function $G : \bm{\alpha}, X \mapsto 1/N \sum_{i \leq N} \log D_0(\bm{x}_i(\bm{\alpha}))/D_0(\bm{x}_{i}) - \eta \|\bm{\alpha}\|^2_2$.
An intuitive interpretation of $G$ is that if $G(\bm{\alpha}, X) \leq 0$, then the imputed points with $\bm{\alpha}$ are, on average, less probable than the previous imputations. 

We now show that $G$ can be maximized through standard convex optimization techniques. The proofs of the following propositions are located in Appendix (Section~\ref{subapp:G_lemmas}). First, G is continuous and infinitely differentiable (Proposition~\ref{lem:G_continuous}). A second less obvious result is that, if $\eta < 4K$, there always exists a bandwidth value $h$ in the definition of the Gaussian kernel in $D_0$ such that $G$ is also strictly concave in $\bm{\alpha}$ (Proposition~\ref{lem:G_strictly_concave}). The condition $\eta < 4K$ is not restrictive, as $\eta$ is the $\ell_2$-regularization factor and $K \geq 2$ is the number of neighbors. A practical value of $h$ can be computed by explicitly finding the smallest positive root of a specific cubic equation, for instance, by using Cardano's method~\cite{Cardano1968}. 
Moreover, another interesting property of G is that its gradient is Lipschitz-continuous with respect to $\bm{\alpha}$ (Proposition~\ref{lem:gradG_lipschitz}), which allows us to derive theoretical guarantees when G is combined with a loss function of a downstream task (Section~\ref{sec:joint_training}). 

Finally, imputation through the $\bm{\alpha}$ maximizing $G(\cdot, X)$ does not require performing a regression on a subset of the dataset, for example, by hiding some available values. This is an important property, as, in some cases, the number of available values is smaller than the total number of elements in the data matrix by several orders of magnitude, as in the collaborative filtering setting~\cite{koren2021advances}.

\subsection{A fast procedure to maximize the objective function}

Based on the function $G$ defined in the previous section, an approach to imputation consists of first imputing the missing values with K-nearest neighbors (K-NN) with uniform weights~\cite{troyanskaya2001missing}, and then recursively improving the imputed values by finetuning the weights in convex combinations of neighbors. Note that those neighbors might change for the same initial sample 
$\bm{x}_i$ across iterations, since the imputed values in that point are modified. At iteration $s$, the optimal weight vector $\bm{\alpha}^s$ is the solution to the maximization problem of $G(\cdot, X^{s-1})$, where $X^{s-1}$ is the data matrix with the imputed values obtained at the previous iteration. The neighbors among the reference set (which are the initial K-NN-imputed points) are obtained with a \emph{single} k-d tree~\cite{bentley1975multidimensional}. 

However, solving a full convex optimization problem at each iteration might be time-consuming. Similarly to prior works in other research fields~\cite{degenne2020gamification}, we advocate for learning the optimal weight vector on the fly by resorting to an online learner. We draw a parallel between the problem of finding the optimal weight vector in a K-nearest neighbor imputation and the problem of expert advice with K experts in online learning. The underlying idea is that we would like to put more credence on the $k^\text{th}$ closest neighbor if it allows us to improve the probability of the imputed values. This analogy permits the leverage of powerful online learners from the literature, for instance, AdaHedge~\cite{de2014follow} or 
EXP$3$~\cite{auer2002exp3}, to obtain theoretical guarantees while having a computationally fast imputation.

Those two ingredients are the keys to our main contribution F3I, described in Algorithm~\ref{alg:online_F3I}. A normalization step--for instance, with the $\ell_2$ norm--can be applied before the initial imputation step and reversed before returning the final data matrix to minimize bias induced by varying feature value ranges. For the sake of readability, we did not add that normalization in the pseudocode of F3I.

\begin{algorithm}[tb]
   \caption{The Fast Iterative Improvement for Imputation (F3I) algorithm.}
   \label{alg:online_F3I}
\begin{algorithmic}
   \STATE {\bfseries Input:} Data $X \in (\mathbb{R}\cup\{\texttt{N/A}\})^{N \times F}$
   \STATE {\bfseries Parameters:}  Maximum budget $T > 0$, number of neighbors $K \geq 2$, regularization factor $\eta > 0$
   \STATE {\bfseries Output:} Imputed data $\widehat{X} \in \mathbb{R}^{N \times F}$
   \STATE  $X^0 \gets \texttt{KNN\_imputer}$($X$, weights=$1/K\textbf{1}_K$)
   \STATE Build a k-d tree $\mathcal{T}$ on $Z = \{(\bm{x}^0)_1,(\bm{x}^0)_2, \dots, (\bm{x}^0)_N\}$
   \STATE  $\mathcal{L} \leftarrow (0, 0, \dots, 0) \in \mathbb{R}^K$ \textcolor{gray}{~ ~ ~ ~ ~ ~ ~ ~ ~ ~ ~ ~ ~ ~ ~ ~ ~ ~ ~ ~ ~ ~ ~ ~ ~ ~ ~ ~ ~ ~ ~ \# Initialize the AdaHedge learner}
   \FOR{$t=1, \dots, T$}
        \STATE $\bm{\alpha}^{t} \gets \mathcal{L}$  \textcolor{gray}{ ~ ~ ~ ~ ~ ~ ~ ~ ~ ~ ~ ~ ~ ~ ~ ~ ~ ~ ~ ~ ~ ~ ~ ~ ~ ~ ~ ~ ~ ~ ~ ~ ~ ~ ~ ~ ~ ~ ~ ~ ~ ~ ~ ~ ~ \# Get the predicted weight vector}
        \STATE $\bm{x}_i^t \gets \texttt{Impute}((\bm{x}^{t-1})_i ; \bm{\alpha}^{t}, Z)$ for all $i \leq N$ \textcolor{gray}{~ ~ ~ ~ ~ ~ ~ ~ ~ ~ ~ ~ ~ \# Apply Algorithm~\ref{alg:imputation} using $\mathcal{T}$}
        \STATE Update $\mathcal{L}$ with the loss $-\langle \bm{\alpha}^{t}, \ \nabla_{\bm{\alpha}} G(\bm{\alpha}^{t}, X^{t-1}) \rangle$ \textcolor{gray}{~ ~ ~ ~ ~ ~ ~ ~ ~ ~ \# Update the online learner $\mathcal{L}$}
        \STATE {\bfseries if} $G(\bm{\alpha}^{t}, X^{t-1}) \leq 0$ {\bfseries then break; end if} \textcolor{gray}{ ~ ~ ~ ~ ~ ~ ~ ~ ~ ~ ~ ~ ~ ~ ~ ~ ~ \# Early stopping criterion}
   \ENDFOR
   \STATE $\widehat{X} \gets X^{t}$ {\bfseries if} $t=T$, $X^{t-1}$ {\bfseries otherwise}
\end{algorithmic}
\end{algorithm}

What's the intuition behind the loss used to update the online learner? As we want to maximize the function $G(\cdot, X^{s-1})$ at iteration $s$, we set as the (possibly non-positive) ``loss'' for the $k^\text{th}$ weight, associated with the $k^\text{th}$ closest neighbor, $-\alpha^t_k \frac{\partial G}{\partial \alpha_k}(\bm{\alpha}^t, X^{s-1})$: the more $G(\bm{\alpha}, X^{s-1})$ increases as $\alpha_k$ increases, the more weight we would like to put on the $k^\text{th}$ closest neighbor. 

\section{Theoretical guarantees of F3I}\label{sec:analysis}


In a nutshell, F3I iteratively improves the imputed values by changing the weight vector that combines the $K$ neighbors among the naively imputed points for each sample. 
One of the most common metrics to evaluate the imputation quality is the Mean Squared Error (MSE) on the imputed values. 

\begin{definition}{\textnormal{Mean squared error.}}\label{def:mse} We define the mean squared error as
$\mathcal{L}^\text{MSE}(X^t, X^\star) \triangleq 
1/(NF) \sum_{i \leq N} \sum_{f \leq F} ((x^t)^f_i-(x^\star)^f_i)^2$. The root-mean-squared error (RMSE) is then defined as $\mathcal{L}^\text{RMSE}(X^t, X^\star) \triangleq \sqrt{\mathcal{L}^\text{MSE}(X^t, X^\star)}$.
\end{definition}

Note that F3I--and all of the baselines that we consider in our experimental study in Section~\ref{sec:experiments_main}--does not get access to the ground truth values and does not need to compute the mean squared error during training. However, we can still derive useful properties of F3I on the MSE. Imputation by convex combinations is theoretically supported by the following upper bound when the data distribution of true values follows Assumption~\ref{as:x_distribution} and one of the missingness mechanisms in Assumptions~\ref{as:mcar}-\ref{as:mnar}. 

\begin{theorem}{\textnormal{Bounds in high probability and in expectation on the MSE for F3I.}}\label{thm:mse_bounds} Under Assumptions~\ref{as:x_distribution}-\ref{as:ub_norm}, if $X^t$ is any imputed matrix at iteration $t \geq 1$
, $X^\star$ is the corresponding full (unavailable in practice) matrix, w.h.p. $1-1/N$, $\mathcal{L}^\text{MSE}(X^t, X^\star) \leq \mathcal{O}((\sigma^\text{miss})^2+\ln N/F)
$, where $\sigma^\text{miss}$ is linked to the variance of the data distribution and depends on the missingness mechanism.
\end{theorem}

In particular, this theorem means that the imputation quality decreases with the variance in the data, which is what we expect, as convex imputations would hardly be able to generate outlier data points. 
The proof and full expression of the bound in Theorem~\ref{thm:mse_bounds} are located in Appendix~\ref{subapp:mse}.

The other performance measure that we are interested in is the preservation of the data distribution, which we quantify with function $G$. However, function $G$ features the Gaussian kernel density $D_0$ estimated on the naively imputed points $\{(\bm{x}^0)_1, \dots, (\bm{x}^0)_N\}$. What we would want to optimize for is the ``true'' probability density $D_\star$ computed on the ground truth values $\{(\bm{x}^\star)_1, \dots, (\bm{x}^\star)_N\}$ which are of course unavailable at all times. 

Then we introduce function $G_\star : \bm{\alpha}, X \mapsto 1/N\sum_{i \leq N} \log D_\star(\bm{x}_i(\bm{\alpha}))/D_\star(\bm{x}_i)-\eta \|\bm{\alpha}\|^2_2$. We measure the imputation quality by the improvement in the probability of imputed points across iterations, that is, $\sum_{s=1}^t G_\star(\bm{\alpha}^s, X^{s-1})$, where $X^{s} \triangleq (\bm{x}^{s-1}_i(\bm{\alpha}^{s}))_{i \leq N}$ for $s \geq 1$ and $X^0$ is the data matrix imputed by the initial KNN imputer. Note that this quantity features a telescoping series and is then equivalent to comparing the final imputed values at time $s=t$ and the initial values at $s=0$ (Proposition~\ref{lem:tel_series} in Appendix). We compare this improvement with the imputation with the one incurred by the weight vector which \emph{a posteriori} maximizes the likelihood of imputed points for all previous iterations up to $t$, that is, $\mathcal{R}(t) \triangleq \max_{ \bm{\alpha} \in \triangle_K} \sum_{s=1}^t G_\star(\bm{\alpha}, X^{s-1})-G_\star(\bm{\alpha}^s, X^{s-1})$. In the online learning community, this measure is akin to the cumulative regret for the loss function $-G_\star$.~\footnote{However, that loss function is not necessarily non-negative.} 

In F3I, we use a no-regret learner named AdaHedge~\cite{de2014follow} to predict the weight vector at each iteration.

\begin{definition}{\textnormal{No-regret learners.}} A learner $\mathcal{L}$ over $\triangle_K$ is no-regret if for $t \geq 1$ and any sequence of bounded gains $\{g_s(\bm{\alpha})\}_{s \leq t}$ for any $\bm{\alpha} \in \triangle_K$, there exists $C \in \mathbb{R}^{+*}$ such that, if $\bm{\alpha}$ is the prediction of $\mathcal{L}$ at iteration $s \leq t$, then $\max_{\bm{\alpha} \in \triangle_K} \sum_{s=1}^{t} g_s(\bm{\alpha})-g_s(\bm{\alpha}^s) \leq C\sqrt{t}$.
\end{definition}

We denote $C^\text{AH}_G = \mathcal{O}(\sqrt{\log(K)})$ the constant associated with the regret bound incurred by AdaHedge on the objective function $G$. Combined with an upper bound on the difference between $G_\star$ and $G$ in high probability, we obtain the following upper bound on the imputation quality for F3I.

\begin{theorem}{\textnormal{High-probability upper bound on the imputation quality for F3I.}}\label{thm:regret_f3i} Under Assumptions~\ref{as:x_distribution}-\ref{as:ub_norm}, for $X \in (\mathbb{R}\cup\{\texttt{N/A}\})^{N \times F}$, $\mathcal{R}(t) \leq C^\text{AH}_G\sqrt{t} + H^\text{miss} h^{-1}t$ w.h.p. $1-1/N$,
where $H^\text{miss} = \mathcal{O}(F+\ln N)$ depends on the missingness mechanism and $h$ is chosen to guarantee that $G$ is concave in its first argument (Proposition~\ref{lem:G_strictly_concave}).
\end{theorem}

\begin{proof}The full proof is in Appendix~\ref{subapp:guarantees_F3I}. Applying the regret bound associated with AdaHedge (Lemma~\ref{lem:regret_adahedge} in Appendix) leads to an upper bound on quantity $\max_{\bm{\alpha} \in \triangle_K} \sum_{s=1}^t g_s(\bm{\alpha}) - g_s(\bm{\alpha}^s)$ which correspond to the difference in gain between the \textit{a posteriori} optimal weights $\bm{\alpha}$ and the weights predicted in F3I $\bm{\alpha}^s$, $s \leq t$, using the gain $g_s : \bm{\alpha} \mapsto \bm{\alpha}^\intercal \nabla_{\bm{\alpha}} G(\bm{\alpha}^s, X^{s-1})$ at iteration $s$. 

We denote $\widehat{\mathcal{R}}(\bm{\alpha}, t) \triangleq \sum_{s=1}^t G(\bm{\alpha}, X^{s-1})-G(\bm{\alpha}^s, X^{s-1})$ for any $\bm{\alpha} \in \triangle_K$. For any $\bm{\alpha} \in \triangle_K$, $\widehat{\mathcal{R}}(\bm{\alpha}, t) \leq \max_{\bm{\alpha} \in \triangle_K} \sum_{s=1}^t (\bm{\alpha}-\bm{\alpha}^s)^\intercal \nabla_{\bm{\alpha}} G(\bm{\alpha}^s, X^{s-1})$ by using the gradient trick on the concave function $G$. 
Finally, we derive a high probability upper bound on $|G_\star(\bm{\alpha}, X')-G(\bm{\alpha}, X')|$ for any $\bm{\alpha} \in \triangle_K$ and $X' \in \mathbb{R}^{N  \times F}$. We show that it suffices to find an upper bound $H^\text{miss}$ with high probability $1-1/N$ on $\max_{i \leq N} \|(\bm{x}^0)_i - (\bm{x}^\star)_i\|^2_2$, which is the norm of a random vector with independent, zero-mean subgaussian coordinates, which allows us to use Bernstein's inequality (Corollary~\ref{cor:concentration_x0_xstar_full2} with $\delta=1/N$).

Subsequently, we show that for all $\bm{x} \in \mathbb{R}^d$, $|\log(D_0(\bm{x})/D_\star(\bm{x}))| \leq H^\text{miss}/(4h)$ with probability $1-1/N$. Finally, the definitions of $G_0$ and $G_\star$ allow us to derive the second term of the sum in the upper bound. 
\end{proof}

The term in $\mathcal{O}(t)$ comes from the approximation in $\mathcal{O}(1)$ made between $G$ and $G_\star$ (Corollary~\ref{cor:concentration_x0_xstar_full}) at each round of F3I. Removing that term would perhaps require supplementary steps, \textit{e}.\textit{g}., considering $D_t$, the density computed on points $\{(\bm{x}^t)_1, (\bm{x}^t)_2, \dots, (\bm{x}^t)_N\}$ at iteration $t$, instead of $D_0$.

\section{Jointly training the imputation model and a model for a downstream task}\label{sec:joint_training}

As noticed by several prior works~\cite{le2021sa,morvan2024imputation,vo2024optimal}, a good imputation quality does not necessarily go hand in hand with an improved performance in a downstream task run on the imputed data set, \textit{e}.\textit{g}., for classification~\cite{le2021sa}, regression~\cite{ayme2023naive}, or structure learning~\cite{vo2024optimal}. That might explain why, in some cases, data sets imputed with naive constant imputations that are known to distort the initial data distribution might yield better performance metrics than those with more sophisticated approaches~\cite{morvan2024imputation}. In this section, we propose a generic approach that optimizes both for an imputation task and a specific downstream task, by learning the optimal (convex) imputation pattern for some model parameters.

Assuming that there is a convex, differentiable pointwise loss function $\ell$ for the downstream task, we now consider the maximization problem $\max_{\bm{\alpha} \in \triangle_K} \mathcal{G}(\bm{\alpha}, X ; \beta)$ on $X \in \mathbb{R}^{N \times F}$ on $\bm{\alpha}$, where
\begin{eqnarray}\label{eq:joint_training_obj}
    \mathcal{G}(\bm{\alpha}, X ; \beta) \triangleq (1-\beta)G(\bm{\alpha}, X) - \beta/N \sum_{i \leq N} \ell(\bm{x}_i(\bm{\alpha}))\;,
\end{eqnarray}

where $\beta \in [0,1]$ is a positive regularization parameter related to the importance of the downstream task. As reported in many papers on multi-task learning~\cite{zhao2018gradnorm,yu2020gradient,liu2021conflict}, simply replacing the gradient of G in the loss of the AdaHedge learner in F3I by the weighted sum of the gradient of G and $\ell$ might lead to optimization issues, for instance, stalling update due to orthogonal gradients. 

A recent method named PCGrad~\cite{yu2020gradient} performs gradient surgery for multi-task learning. In particular, PCGrad allows us to obtain theoretical guarantees on the performance of the training if the weighted sum $-\mathcal{G}$ is convex and $L$-Lipschitz continuous with $L>0$ and if both $\ell$ and $-G$ are convex and differentiable~\cite[Theorems 1-2]{yu2020gradient}. $-G$ is convex by Proposition~\ref{lem:G_strictly_concave} and differentiable by Proposition~\ref{lem:G_continuous}. Naturally, if $\nabla \ell$ is itself Lipschitz continuous with a positive Lipschitz constant, Proposition~\ref{lem:gradG_lipschitz} implies that this condition is verified for the objective function in Equation~\eqref{eq:joint_training_obj}. A simple example of such a loss function is the pointwise log loss $\ell(\bm{x})=-y\log C_{\bm{\omega}}(\bm{x})$ for the binary classification task, where $y$ is the true class in $\{0,1\}$ for sample $\bm{x}$ and $C_{\bm{\omega}} : \bm{x} \mapsto 1/(1+\exp(-\bm{\omega}^\intercal \bm{x}))$ is the sigmoid function of parameter $\bm{\omega}$. Proofs are in Appendix (Section~\ref{subapp:joint_training}).

Then, we modify F3I by changing the loss fed to the AdaHedge learner $\mathcal{L}$ in Line 10 in Algorithm~\ref{alg:online_F3I}. At iteration $s$, instead of using the loss $g_s(\bm{\alpha}) \triangleq -\left\langle \bm{\alpha}, \nabla_{\bm{\alpha}} G(\bm{\alpha}^s, X^{s-1})\right\rangle$, we consider $\overline{g}_s(\bm{\alpha}) \triangleq -\langle \bm{\alpha}, \mathcal{L}(\bm{\alpha}, X^{s-1})\rangle$, where $ \mathcal{L}(\bm{\alpha}, X^{s-1})$ is equal to
\begin{equation}\label{eq:pcgrad_update} (1-\beta) \nabla_{\bm{\alpha}} G^\text{PC}(\bm{\alpha}^s, X^{s-1})-\beta/N\sum_{i \leq N} \nabla_{\bm{\alpha}} \ell^\text{PC}((\bm{x}^{s-1})_i(\bm{\alpha}^s))\rangle\;,\end{equation}
and $\nabla_{\bm{\alpha}} G^\text{PC}$ and $\nabla_{\bm{\alpha}} \ell^\text{PC}$ are the gradient function of $G$ and $\ell$ with respect to their first argument corrected by the PCGrad procedure~\cite[Algorithm 1]{yu2020gradient}. We call PCGrad-F3I this joint training version of F3I. Under the conditions laid in the statement of Theorem $2$ in~\cite{yu2020gradient}, at any iteration $s \leq t$, if $(\bm{\alpha}^s)^\text{PC}$ and $\bm{\alpha}^s$ are respectively the parameters obtained after applying one PCGrad or a regular AdaHedge update to $\bm{\alpha}^{s-1}$, then $\mathcal{G}((\bm{\alpha}^{s})^\text{PC}, X^{s-1} ; \beta) \geq \mathcal{G}(\bm{\alpha}^{s}, X^{s-1} ; \beta)$. That is,

\begin{theorem}{\textnormal{High-probability upper bound on the joint imputation-downstream task performance.}}\label{thm:regret_f3i_joint} Under Assumptions~\ref{as:x_distribution}-\ref{as:ub_norm}, for any $X \in (\mathbb{R}\cup\{\texttt{N/A}\})^{N \times F}$, convex pointwise loss $\ell$ such that $\nabla \ell$ is Lipschitz-continuous, and $\beta \in [0,1]$, under the conditions in Theorem $2$ from~\cite{yu2020gradient}, $\max_{ \bm{\alpha} \in \triangle_K} \sum_{s=1}^t \mathcal{G}(\bm{\alpha}, X^{s-1} ; \beta)-\mathcal{G}(\bm{\alpha}^s, X^{s-1} ; \beta)  \leq C^\text{AH}_{(G,\ell)}\sqrt{t} + (1-\beta)H^\text{miss}h^{-1}t$ w.h.p. $1-1/N$, 
 where $H^\text{miss} = \mathcal{O}(F+\ln N)$ depends on the missingness mechanism, $h$ is chosen to guarantee that G is concave, and $ C^\text{AH}_{(G,\ell)}$ is the constant related to AdaHedge being applied with gains $\overline{g}_s(\cdot)$.
\end{theorem}

For $\beta=0$, this bound matches Theorem~\ref{thm:regret_f3i}, and for $\beta=1$, this is the classical AdaHedge regret bound (Theorem $8$ in~\cite{de2014follow}) with loss $\ell$.

\section{Experimental study}\label{sec:experiments_main}

This section is restricted to the comparison of our algorithmic contributions F3I and PCGrad-F3I to baselines for imputation-only and joint imputation-binary classification tasks on real-life data sets, due to space constraints. In Appendix (Section~\ref{sec:experiments}), we also empirically validate our theoretical results (Theorems~\ref{thm:mse_bounds},~\ref{thm:regret_f3i} and~\ref{thm:regret_f3i_joint}) and test the imputation and classification performance on additional real-life data sets for drug repurposing (with up to $9,000$ features), compared to other (older) baselines, and on synthetic data sets (with up to $20,000$ features) that comply with Assumptions~\ref{as:x_distribution}-\ref{as:ub_norm}, for all missingness mechanisms. Further information about hyperparameter tuning and computing infrastructure is also available in Section~\ref{sec:experiments}. 

\subsection{Imputation-only task}

First, we study the imputation quality--without any downstream task. We resorted to the framework HyperImpute~\cite{jarrett2022hyperimpute} to implement and run the benchmark for an imputation task across different performance metrics (including RMSE) on four standard data sets BreastCancer~\citep{breast_cancer_wisconsin_(diagnostic)_17}, Diabetes (from scikit-learn~\cite{pedregosa2011scikit}), HeartDisease~\citep{heart_disease_45}, Ionosphere~\citep{ionosphere} and a data set for drug repurposing, Gottlieb~\cite{luo2016drug}.  In this benchmark, we included recent methods from the literature which benefited from open-source, modular \texttt{scikit-learn}~\cite{pedregosa2011scikit}-like implementations: GAIN~\citep{yoon2018gain}, GRAPE~\citep{you_handling_2020}, HyperImpute~\citep{jarrett2022hyperimpute}, MIRACLE~\citep{kyono2021miracle}, NewImp~\citep{chen2024rethinking}, Remasker~\citep{du2023remasker} and TDM~\citep{zhao2023transformed}. We considered the scenario MNAR in the framework HyperImpute to add missing values. We report in Table~\ref{tab:imputation-benchmark} the corresponding numerical results across 10 runs with different random seeds. 

Those results show that, on the imputation task alone, F3I offers a good tradeoff between imputation quality (regardless of the performance metric) and computational efficacy (runtime). Indeed, F3I performs on par or better than the state-of-the-art, while remaining computationally efficient by several orders of magnitude. We also report numerical results across 100 runs restricted to the best baselines in Table~\ref{tab:imputation-benchmark-2} in Appendix (Section~\ref{sec:experiments}), which confirm these observations. 

\begin{table}
    \centering
    \caption{Average and standard deviation values of imputation quality metrics (rounded to the closest second decimal place) and runtime across 10 different random seeds. HeartDisease has native missing values, which is why the Wasserstein distance cannot be computed. RMSE: root mean square error. MAE: mean average error. WD: Wasserstein distance. Runtime is in seconds. TDM failed on the Gottlieb data set. Bold type is the top performer, underline denotes the second best (and average percentage of deterioration of performance across metrics compared to the top performer).}
    \label{tab:imputation-benchmark}
    \begin{tabular}{lrrrr}
        \toprule
        Data set & RMSE $\downarrow$ &  MAE $\downarrow$ & WD $\downarrow$ & Runtime $\downarrow$\\
       \midrule
       BreastCancer   & & & &  \\
      \midrule
       F3I (ours) &  \textbf{0.08$\pm$ 0.03} & \textbf{0.03$\pm$ 0.01} & \textbf{0.06$\pm$ 0.02}   & \textbf{ 0.14$\pm$ 0.04}\\
       GAIN  & 0.28$\pm$ 0.03 & 0.11$\pm$ 0.02 & 0.25$\pm$ 0.05  &   34$\pm$ 6\\
       GRAPE & 0.37$\pm$ 0.03 & 0.19$\pm$ 0.02 & 0.28$\pm$ 0.03 & 5,091$\pm$ 1,707\\
       HyperImpute & (+231\%) \underline{0.26$\pm$ 0.03}  & \underline{0.09$\pm$ 0.02}  & \underline{0.22$\pm$ 0.04}  &    \underline{7$\pm$ 2}\\
       MIRACLE & 4.32$\pm$ 0.35 & 4.22$\pm$ 0.33 &  10.00$\pm$ 1.04   & 77$\pm$ 4\\
       NewImp & 415$\pm$ 189 &  294$\pm$ 179 &  695$\pm$ 416 &  3,282$\pm$ 819\\
       Remasker & 0.27$\pm$ 0.03 & 0.11$\pm$ 0.02  & 0.25$\pm$ 0.04  & 393$\pm$ 32\\
       TDM & 0.35$\pm$ 0.04 & 0.21$\pm$ 0.03  &  0.40$\pm$ 0.05 &  532$\pm$ 22\\
      \midrule
       Diabetes   & & & &  \\
      \midrule
       F3I (ours) &  (+9\%) \underline{0.34$\pm$ 0.05} &  \underline{0.27$\pm$ 0.04} &  0.61$\pm$ 0.20    &  \textbf{0.09$\pm$ 0.01}\\
       GAIN  & 0.52$\pm$ 0.07 & 0.44$\pm$ 0.06&   0.82$\pm$ 0.20    & 41$\pm$ 8\\
       GRAPE & 0.43$\pm$ 0.04 & 0.35$\pm$ 0.04 & \textbf{0.54$\pm$ 0.09}  & 454$\pm$ 52\\
       HyperImpute & \textbf{0.32$\pm$ 0.05} & \textbf{0.25$\pm$ 0.04}  & \textbf{0.54$\pm$ 0.19}   &  \underline{19$\pm$ 6}\\
       MIRACLE & 5.96$\pm$ 0.66 &  5.82$\pm$ 0.63 &  13.36$\pm$ 3.25&   123$\pm$ 44\\
       NewImp & 2.40$\pm$ 0.94 &  1.76$\pm$ 0.68 &    3.85$\pm$ 1.22&   2,304$\pm$ 583\\
       Remasker & 0.42$\pm$ 0.04 & 0.35$\pm$ 0.04 & 0.78$\pm$ 0.15 & 39$\pm$ 2 \\
       TDM & 0.38$\pm$ 0.05  &0.31$\pm$ 0.04 & \underline{0.56$\pm$ 0.19} &  171$\pm$ 102\\
      \midrule
       Gottlieb & & & &  \\
      \midrule
       F3I (ours) & (+111\%) \underline{0.04$\pm$ 0.01} &  \underline{0.02$\pm$ 0.00} &  0.03$\pm$ 0.00   &   \textbf{2$\pm$ 1}\\
       GAIN  & \underline{0.04$\pm$ 0.01}  &  \underline{0.02$\pm$ 0.00}   &   \underline{0.02$\pm$ 0.00}    &  103$\pm$ 9\\
       GRAPE & 0.12$\pm$ 0.01 & 0.09$\pm$ 0.01 &  0.12$\pm$ 0.02 & 6,670$\pm$ 217\\
       HyperImpute & \textbf{0.03$\pm$ 0.00}   & \textbf{0.01$\pm$ 0.00}   &   \textbf{0.01$\pm$ 0.00}       & \underline{44$\pm$ 18}\\
       MIRACLE & 4.44$\pm$ 0.32  &   4.38$\pm$ 0.31  &   10.50$\pm$ 0.75  &   212$\pm$ 20\\
       NewImp & 131$\pm$ 72.4 & 71.3$\pm$ 47.2 &  170$\pm$ 112  & 12,933$\pm$ 621\\
       Remasker& 0.18$\pm$ 0.04 & 0.14$\pm$ 0.03 & 0.30$\pm$ 0.08  &3,016$\pm$ 42\\
       TDM & - & - & - & -\\
       \midrule
        HeartDisease &  &   &  & \\
      \midrule
       F3I (ours) & \textbf{0.14$\pm$ 0.05}  &   \textbf{0.07$\pm$ 0.03} & -    &   \textbf{0.10$\pm$ 0.02}\\
       GAIN  & 0.30$\pm$ 0.07   &   0.18$\pm$ 0.05 & -  &   34$\pm$ 7\\
       GRAPE & 0.54$\pm$ 0.03 & 0.38$\pm$ 0.04 & - & 536$\pm$ 439\\
       HyperImpute & (+210\%) \underline{0.24$\pm$ 0.07}  &   \underline{0.13$\pm$ 0.05} & -  &   \underline{17$\pm$ 5}\\
       MIRACLE & 4.91$\pm$ 0.44   &   4.72$\pm$ 0.41 & -  &   82$\pm$ 2\\
       NewImp & 308$\pm$ 176 & 197$\pm$ 127 & - & 2,404$\pm$ 270\\
       Remasker & 0.25$\pm$ 0.05 & 0.14$\pm$ 0.03 & -  & 32$\pm$ 2\\
       TDM & 0.48$\pm$ 0.04  &   0.36$\pm$ 0.03 &  - &  340$\pm$ 219\\
      \midrule
       Ionosphere & & & &  \\
      \midrule
       F3I (ours) & (+11\%) \underline{0.21$\pm$ 0.06} & \underline{0.15$\pm$ 0.05} & \underline{0.29$\pm$ 0.12}   &     \textbf{0.19$\pm$ 0.03}\\
       GAIN  & 0.47$\pm$ 0.05 &  0.34$\pm$ 0.05  &0.57$\pm$ 0.07   &  \underline{50$\pm$ 11}\\
       GRAPE & 0.48$\pm$ 0.03 & 0.39$\pm$ 0.04 &  0.49$\pm$ 0.12   & 765$\pm$ 110\\
        HyperImpute & \textbf{0.20$\pm$ 0.07} &  \textbf{0.13$\pm$ 0.05}  & \textbf{0.26$\pm$ 0.10} &     99$\pm$ 73\\
       MIRACLE & 5.22$\pm$ 0.54 & 5.14$\pm$ 0.53 &  12.4$\pm$ 1.35    &  96$\pm$ 2\\
       NewImp & 0.60$\pm$ 0.18 &  0.47$\pm$ 0.13 & 1.04$\pm$ 0.35 & 4,174$\pm$ 1,169\\
       Remasker  & 0.33$\pm$ 0.03 & 0.26$\pm$ 0.04 & 0.51$\pm$ 0.09 & 77$\pm$ 3 \\
       TDM & 0.39$\pm$ 0.04  &0.33$\pm$ 0.04&  0.57$\pm$ 0.10  &  697$\pm$ 13\\
       \bottomrule
    \end{tabular}
\end{table}

\subsection{Joint imputation-binary classification task}

Second, we implement the joint imputation-classification training with the log-loss function and sigmoid classifier $\ell(\bm{x}) \triangleq -y\log C_{\bm{\omega}}(\bm{x})$ mentioned in Section~\ref{sec:joint_training}, where $y \in \{0,1\}$ is the binary class associated with sample $\bm{x} \in \mathbb{R}^F$. To implement PCGrad-F3I, we chain the imputation phase by F3I with an MLP classifier, which returns logits. At time $t$, the imputation part applies at a fixed set of parameters $\bm{\omega}^t$  with the learner losses defined in Equation~\eqref{eq:pcgrad_update}. 

We compare the classification performance of PCGrad-F3I with \textit{imputation methods with separate classifier training}: imputing by the mean (Mean), a K nearest-neighbor algorithm with weights inversely proportional to the distance to neighbors (K-NN~\citep{troyanskaya2001missing}), or a random forest classifier (RF-GAP~\citep{rhodes2023geometry}) prior to applying the MLP classifier; or with \textit{imputation methods with \textnormal{joint} classifier training} like PCGradF3I: adding a NeuMiss block~\cite{le2020neumiss} to the MLP classifier, or training simultaneously GRAPE~\citep{you_handling_2020} and the MLP classifier. We use the same MLP architecture across all imputation techniques. 
The criterion for training the models is the log loss, and we split the samples into training ($70\%$), validation ($20\%$), and testing ($10\%$) sets, where the former two sets are used for training the MLP and hyperparameter finetuning (see Appendix in Section~\ref{subapp:drug_repurposing}), and the performance metric--Area Under the Curve (AUC)--is computed on the latter set. Further experimental details can be found in Appendix (Section~\ref{sec:experiments}). We consider the MNIST dataset~\cite{lecun1998mnist}, which comprises grayscale images of handwritten digits, and another drug repurposing data set, PREDICT~\cite{reda2023predict}. In MNIST, we restrict our study to images annotated with class 0 or 1 to get a binary classification problem. Moreover, we also include again the BreastCancer~\citep{breast_cancer_wisconsin_(diagnostic)_17} and Ionosphere~\citep{ionosphere} data sets with their native classification labels. In all data sets, we remove pixels at random with probability $50\%$ using a MCAR mechanism. Table~\ref{tab:joint_MNIST_PREDICT_main} displays the numerical results across $100$ iterations with different random seeds.~\footnote{We considered $60$ runs for the Ionosphere data set, due to an unsolvable error in our server.}  

\begin{table}
    \centering
    \caption{Average and standard deviation of Area Under the Curve (AUC) values (rounded to the closest second decimal place) across several runs on the joint imputation-classification task (MCAR scenario, $p^\text{miss}=0.5$). Bold type is the top performer, underline denotes the second best (and average percentage of deterioration of performance across data sets compared to the top performer).}
    \label{tab:joint_MNIST_PREDICT_main}
    \begin{tabular}{lrrrr}
        \toprule
    Imputation / Data  & BreastCancer & Ionosphere   & MNIST & PREDICT \\
       \midrule
       Joint classifier training & & & & \\
       \midrule
       GRAPE & 0.55$\pm$ 0.13 & 0.70$\pm$ 0.15  & \textbf{1.00 $\pm$0.00} &  0.49 $\pm$0.07 \\
       NeuMiss & \underline{0.62$\pm$ 0.18}  & 0.72$\pm$ 0.19  & \underline{0.99 $\pm$0.07} & 0.50 $\pm$0.01 \\
       PCGradF3I (ours) &  ($-2.1\%$) \textbf{0.70$\pm$ 0.14}  & \underline{0.79$\pm$ 0.17} & \underline{0.99 $\pm$0.09} & \underline{0.51 $\pm$0.01} \\
       \midrule
       Separate classifier training & & & & \\
       \midrule
       K-NN & 0.54$\pm$ 0.13 & 0.74$\pm$ 0.19  & 0.93 $\pm$0.17  & 0.47 $\pm$0.07\\
       Mean  & 0.52$\pm$ 0.07 & 0.73$\pm$ 0.16  & 0.64 $\pm$0.18 & 0.48 $\pm$0.00\\
       RF-GAP & 0.50 $\pm$0.00  &  \textbf{0.82 $\pm$0.21} & \textbf{1.00 $\pm$0.00} & \textbf{0.53 $\pm$0.13}\\
       \bottomrule
    \end{tabular}
\end{table}

The empirical performance of PCGrad-F3I on classification tasks is on par with the state-of-the-art, with a very small average deterioration of performance compared to the top baseline RF-GAP~\citep{rhodes2023geometry} ($-2.1\%$ in AUC). However, we argue that the fairest baselines to benchmark PCGradF3I for classification might be the imputation approaches with joint classifier training, in which case PCGradF3I comes on top. We also show in Appendix (Figures~\ref{fig:joint_MNIST}-\ref{fig:joint_MNIST3} in Section~\ref{sec:experiments}) that PCGradF3I preserves the correct shapes in MNIST across missingness proportions and mechanisms.


\section{Limitations}\label{sec:limitations}

We list here three limitations of our contribution. First, F3I is based on iterative improvements of a K-NN imputer. Yet, K-NN imputation is costly when the number of samples is very large. A solution is to use approximate neighbor-finding algorithms such as FAISS~\citep{johnson2019billion}, LSH~\citep{zhao2014locality,tsai2014locality} or Annoy~\citep{Github:annoy} and leverage the use of GPUs to accelerate F3I. Under Assumptions~\ref{as:x_distribution}-\ref{as:ub_norm}, our theoretical results still hold in that case. Second, F3I can only be used for continuous variables. Third, the theoretical guarantees derived in Theorems~\ref{thm:mse_bounds}-\ref{thm:regret_f3i_joint} require strong assumptions on the data distribution, which may not accurately reflect the statistical properties of data sets in practical applications. However, a
s shown in Section~\ref{sec:experiments_main} and~\ref{sec:experiments} in Appendix, F3I can still be applied to real-life data and be competitive. 

\section{Discussion}\label{sec:discussion}

We introduce an algorithm named F3I which iteratively improves a K-nearest neighbor imputation, by tuning the neighbor-associated weights. F3I is versatile as it  
can be jointly trained with any classification task to meaningfully impute values depending on the end goal. Moreover, F3I features theoretical guarantees on the imputation quality and the preservation of the data distribution across missingness mechanisms, including not-missing-at-random. 
Empirically, the performance of F3I is similar or better than the state-of-the-art across data sets, while being more computationally tractable. 
The experimental code and implementation of F3I are provided as supplementary material.

Combining online learning and density ratio estimation is a simple and flexible idea that could be improved further, notably to perhaps remove the linear term in the number of iterations in Theorem~\ref{thm:regret_f3i}. 
For instance, the density ratio estimation step might benefit from the classifier-based approach developed in BORE~\cite{tiao2021bore}, in particular in a version of F3I where a k-d tree would be rebuilt at every iteration to consider density $D_t$ on points $\{(\bm{x}^t)_1, \dots, (\bm{x}^t)_N\}$ instead of the density estimated on the naively imputed initial points $\{(\bm{x}^0)_1, \dots, (\bm{x}^0)_N\}$. 

\begin{ack}
Part of this work was completed during a secondment of C.R. at Soda Team, Inria Saclay, F‑91120 Palaiseau, France. The research leading to these results has received funding from the European Union’s HORIZON 2020 Programme under grant agreement no. 101102016 (RECeSS, HORIZON TMA MSCA Postdoctoral Fellowships - European Fellowships, C.R.). The funding sources have played no role in the design, the execution nor the analyses performed in this study.
\end{ack}

\bibliography{example_paper}

\begin{thebibliography}{75}
\providecommand{\natexlab}[1]{#1}
\providecommand{\url}[1]{\texttt{#1}}
\expandafter\ifx\csname urlstyle\endcsname\relax
  \providecommand{\doi}[1]{doi: #1}\else
  \providecommand{\doi}{doi: \begingroup \urlstyle{rm}\Url}\fi

\bibitem[Rubin(1976)]{rubin1976inference}
Donald~B Rubin.
\newblock Inference and missing data.
\newblock \emph{Biometrika}, 63\penalty0 (3):\penalty0 581--592, 1976.

\bibitem[Le~Morvan and Varoquaux(2024)]{morvan2024imputation}
Marine Le~Morvan and Ga{\"e}l Varoquaux.
\newblock Imputation for prediction: beware of diminishing returns.
\newblock \emph{arXiv preprint arXiv:2407.19804}, 2024.

\bibitem[{van Buuren} and Groothuis-Oudshoorn(2011)]{van2011mice}
Stef {van Buuren} and Karin Groothuis-Oudshoorn.
\newblock {mice}: Multivariate imputation by chained equations in r.
\newblock \emph{Journal of Statistical Software}, 45\penalty0 (3):\penalty0
  1--67, 2011.
\newblock \doi{10.18637/jss.v045.i03}.

\bibitem[Stekhoven and B{\"u}hlmann(2012)]{stekhoven2012missforest}
Daniel~J Stekhoven and Peter B{\"u}hlmann.
\newblock Missforest—non-parametric missing value imputation for mixed-type
  data.
\newblock \emph{Bioinformatics}, 28\penalty0 (1):\penalty0 112--118, 2012.

\bibitem[Rhodes et~al.(2023)Rhodes, Cutler, and Moon]{rhodes2023geometry}
Jake~S Rhodes, Adele Cutler, and Kevin~R Moon.
\newblock Geometry-and accuracy-preserving random forest proximities.
\newblock \emph{IEEE Transactions on Pattern Analysis and Machine
  Intelligence}, 45\penalty0 (9):\penalty0 10947--10959, 2023.

\bibitem[Seu et~al.(2022)Seu, Kang, and Lee]{seu2022intelligent}
Kimseth Seu, Mi-Sun Kang, and HwaMin Lee.
\newblock An intelligent missing data imputation techniques: A review.
\newblock \emph{JOIV: International Journal on Informatics Visualization},
  6\penalty0 (1-2):\penalty0 278--283, 2022.

\bibitem[Muzellec et~al.(2020{\natexlab{a}})Muzellec, Josse, Boyer, and
  Cuturi]{muzellec2020missing}
Boris Muzellec, Julie Josse, Claire Boyer, and Marco Cuturi.
\newblock Missing data imputation using optimal transport.
\newblock In \emph{International Conference on Machine Learning}, pages
  7130--7140. PMLR, 2020{\natexlab{a}}.

\bibitem[Mazumder et~al.(2010)Mazumder, Hastie, and
  Tibshirani]{mazumder2010spectral}
Rahul Mazumder, Trevor Hastie, and Robert Tibshirani.
\newblock Spectral regularization algorithms for learning large incomplete
  matrices.
\newblock \emph{The Journal of Machine Learning Research}, 11:\penalty0
  2287--2322, 2010.

\bibitem[van Loon et~al.(2024)van Loon, Fokkema, de~Vos, Koini, Schmidt, and
  de~Rooij]{van2024imputation}
Wouter van Loon, Marjolein Fokkema, Frank de~Vos, Marisa Koini, Reinhold
  Schmidt, and Mark de~Rooij.
\newblock Imputation of missing values in multi-view data.
\newblock \emph{Information Fusion}, page 102524, 2024.

\bibitem[Mattei and Frellsen(2019)]{mattei2019miwae}
Pierre-Alexandre Mattei and Jes Frellsen.
\newblock {MIWAE}: Deep generative modelling and imputation of incomplete data
  sets.
\newblock In Kamalika Chaudhuri and Ruslan Salakhutdinov, editors,
  \emph{Proceedings of the 36th International Conference on Machine Learning},
  volume~97 of \emph{Proceedings of Machine Learning Research}, pages
  4413--4423. PMLR, 09--15 Jun 2019.
\newblock URL \url{https://proceedings.mlr.press/v97/mattei19a.html}.

\bibitem[Ipsen et~al.(2021)Ipsen, Mattei, and Frellsen]{ipsen2021not}
Niels~Bruun Ipsen, Pierre-Alexandre Mattei, and Jes Frellsen.
\newblock not-{\{}miwae{\}}: Deep generative modelling with missing not at
  random data.
\newblock In \emph{International Conference on Learning Representations}, 2021.
\newblock URL \url{https://openreview.net/forum?id=tu29GQT0JFy}.

\bibitem[Kyono et~al.(2021)Kyono, Zhang, Bellot, and van~der
  Schaar]{kyono2021miracle}
Trent Kyono, Yao Zhang, Alexis Bellot, and Mihaela van~der Schaar.
\newblock Miracle: Causally-aware imputation via learning missing data
  mechanisms.
\newblock \emph{Advances in Neural Information Processing Systems},
  34:\penalty0 23806--23817, 2021.

\bibitem[Jarrett et~al.(2022)Jarrett, Cebere, Liu, Curth, and van~der
  Schaar]{jarrett2022hyperimpute}
Daniel Jarrett, Bogdan~C Cebere, Tennison Liu, Alicia Curth, and Mihaela
  van~der Schaar.
\newblock Hyperimpute: Generalized iterative imputation with automatic model
  selection.
\newblock In \emph{International Conference on Machine Learning}, pages
  9916--9937. PMLR, 2022.

\bibitem[{\'S}mieja et~al.(2018){\'S}mieja, Struski, Tabor, Zieli{\'n}ski, and
  Spurek]{smieja2018processing}
Marek {\'S}mieja, {\L}ukasz Struski, Jacek Tabor, Bartosz Zieli{\'n}ski, and
  Przemys{\l}aw Spurek.
\newblock Processing of missing data by neural networks.
\newblock \emph{Advances in neural information processing systems}, 31, 2018.

\bibitem[Tang et~al.(2003)Tang, Little, and Raghunathan]{tang2003analysis}
Gong Tang, Roderick~JA Little, and Trivellore~E Raghunathan.
\newblock Analysis of multivariate missing data with nonignorable nonresponse.
\newblock \emph{Biometrika}, 90\penalty0 (4):\penalty0 747--764, 2003.

\bibitem[Mohan et~al.(2018)Mohan, Thoemmes, and Pearl]{mohan2018estimation}
Karthika Mohan, Felix Thoemmes, and Judea Pearl.
\newblock Estimation with incomplete data: The linear case.
\newblock In \emph{Proceedings of the International Joint Conferences on
  Artificial Intelligence Organization}, 2018.

\bibitem[Sportisse et~al.(2020)Sportisse, Boyer, and
  Josse]{sportisse2020estimation}
Aude Sportisse, Claire Boyer, and Julie Josse.
\newblock Estimation and imputation in probabilistic principal component
  analysis with missing not at random data.
\newblock \emph{Advances in Neural Information Processing Systems},
  33:\penalty0 7067--7077, 2020.

\bibitem[Le~Morvan et~al.(2020)Le~Morvan, Josse, Moreau, Scornet, and
  Varoquaux]{le2020neumiss}
Marine Le~Morvan, Julie Josse, Thomas Moreau, Erwan Scornet, and Ga{\"e}l
  Varoquaux.
\newblock Neumiss networks: differentiable programming for supervised learning
  with missing values.
\newblock \emph{Advances in Neural Information Processing Systems},
  33:\penalty0 5980--5990, 2020.

\bibitem[Troyanskaya et~al.(2001)Troyanskaya, Cantor, Sherlock, Brown, Hastie,
  Tibshirani, Botstein, and Altman]{troyanskaya2001missing}
Olga Troyanskaya, Michael Cantor, Gavin Sherlock, Pat Brown, Trevor Hastie,
  Robert Tibshirani, David Botstein, and Russ~B Altman.
\newblock Missing value estimation methods for dna microarrays.
\newblock \emph{Bioinformatics}, 17\penalty0 (6):\penalty0 520--525, 2001.

\bibitem[Emmanuel et~al.(2021)Emmanuel, Maupong, Mpoeleng, Semong, Mphago, and
  Tabona]{emmanuel2021survey}
Tlamelo Emmanuel, Thabiso Maupong, Dimane Mpoeleng, Thabo Semong, Banyatsang
  Mphago, and Oteng Tabona.
\newblock A survey on missing data in machine learning.
\newblock \emph{Journal of Big data}, 8:\penalty0 1--37, 2021.

\bibitem[Joel et~al.(2024)Joel, Doorsamy, and Paul]{joel2024performance}
Luke~Oluwaseye Joel, Wesley Doorsamy, and Babu~Sena Paul.
\newblock On the performance of imputation techniques for missing values on
  healthcare datasets.
\newblock \emph{arXiv preprint arXiv:2403.14687}, 2024.

\bibitem[Beretta and Santaniello(2016)]{beretta2016nearest}
Lorenzo Beretta and Alessandro Santaniello.
\newblock Nearest neighbor imputation algorithms: a critical evaluation.
\newblock \emph{BMC medical informatics and decision making}, 16:\penalty0
  197--208, 2016.

\bibitem[Cesa-Bianchi and Lugosi(2006)]{cesa2006prediction}
Nicolo Cesa-Bianchi and G{\'a}bor Lugosi.
\newblock \emph{Prediction, learning, and games}.
\newblock Cambridge university press, 2006.

\bibitem[De~Rooij et~al.(2014)De~Rooij, Van~Erven, Gr{\"u}nwald, and
  Koolen]{de2014follow}
Steven De~Rooij, Tim Van~Erven, Peter~D Gr{\"u}nwald, and Wouter~M Koolen.
\newblock Follow the leader if you can, hedge if you must.
\newblock \emph{The Journal of Machine Learning Research}, 15\penalty0
  (1):\penalty0 1281--1316, 2014.

\bibitem[Lattimore and Szepesv{\'a}ri(2020)]{lattimore2020bandit}
Tor Lattimore and Csaba Szepesv{\'a}ri.
\newblock \emph{Bandit algorithms}.
\newblock Cambridge University Press, 2020.

\bibitem[Cardano et~al.(1968)Cardano, Witmer, and Ore]{Cardano1968}
Girolamo Cardano, T.~Richard Witmer, and {\O}ystein Ore.
\newblock \emph{Ars magna, or, The rules of algebra}.
\newblock Dover, New York, 1968.
\newblock ISBN 9780486678115; 0486678113.

\bibitem[Koren et~al.(2021)Koren, Rendle, and Bell]{koren2021advances}
Yehuda Koren, Steffen Rendle, and Robert Bell.
\newblock Advances in collaborative filtering.
\newblock \emph{Recommender systems handbook}, pages 91--142, 2021.

\bibitem[Bentley(1975)]{bentley1975multidimensional}
Jon~Louis Bentley.
\newblock Multidimensional binary search trees used for associative searching.
\newblock \emph{Commun. ACM}, 18\penalty0 (9):\penalty0 509–517, sep 1975.
\newblock ISSN 0001-0782.
\newblock \doi{10.1145/361002.361007}.
\newblock URL \url{https://doi.org/10.1145/361002.361007}.

\bibitem[Degenne et~al.(2020)Degenne, M{\'e}nard, Shang, and
  Valko]{degenne2020gamification}
R{\'e}my Degenne, Pierre M{\'e}nard, Xuedong Shang, and Michal Valko.
\newblock Gamification of pure exploration for linear bandits.
\newblock In \emph{International Conference on Machine Learning}, pages
  2432--2442. PMLR, 2020.

\bibitem[Auer et~al.(2002)Auer, Cesa-Bianchi, Freund, and
  Schapire]{auer2002exp3}
Peter Auer, Nicol\`{o} Cesa-Bianchi, Yoav Freund, and Robert~E. Schapire.
\newblock The nonstochastic multiarmed bandit problem.
\newblock \emph{SIAM Journal on Computing}, 32\penalty0 (1):\penalty0 48--77,
  2002.
\newblock \doi{10.1137/S0097539701398375}.
\newblock URL \url{https://doi.org/10.1137/S0097539701398375}.

\bibitem[Le~Morvan et~al.(2021)Le~Morvan, Josse, Scornet, and
  Varoquaux]{le2021sa}
Marine Le~Morvan, Julie Josse, Erwan Scornet, and Ga{\"e}l Varoquaux.
\newblock What’sa good imputation to predict with missing values?
\newblock \emph{Advances in Neural Information Processing Systems},
  34:\penalty0 11530--11540, 2021.

\bibitem[Vo et~al.(2024)Vo, Zhao, Le, Bonilla, and Phung]{vo2024optimal}
Vy~Vo, He~Zhao, Trung Le, Edwin~V Bonilla, and Dinh Phung.
\newblock Optimal transport for structure learning under missing data.
\newblock \emph{arXiv preprint arXiv:2402.15255}, 2024.

\bibitem[Ayme et~al.(2023)Ayme, Boyer, Dieuleveut, and Scornet]{ayme2023naive}
Alexis Ayme, Claire Boyer, Aymeric Dieuleveut, and Erwan Scornet.
\newblock Naive imputation implicitly regularizes high-dimensional linear
  models.
\newblock In \emph{International Conference on Machine Learning}, pages
  1320--1340. PMLR, 2023.

\bibitem[Chen et~al.(2018)Chen, Badrinarayanan, Lee, and
  Rabinovich]{zhao2018gradnorm}
Zhao Chen, Vijay Badrinarayanan, Chen-Yu Lee, and Andrew Rabinovich.
\newblock {G}rad{N}orm: Gradient normalization for adaptive loss balancing in
  deep multitask networks.
\newblock In Jennifer Dy and Andreas Krause, editors, \emph{Proceedings of the
  35th International Conference on Machine Learning}, volume~80 of
  \emph{Proceedings of Machine Learning Research}, pages 794--803. PMLR, 10--15
  Jul 2018.
\newblock URL \url{https://proceedings.mlr.press/v80/chen18a.html}.

\bibitem[Yu et~al.(2020)Yu, Kumar, Gupta, Levine, Hausman, and
  Finn]{yu2020gradient}
Tianhe Yu, Saurabh Kumar, Abhishek Gupta, Sergey Levine, Karol Hausman, and
  Chelsea Finn.
\newblock Gradient surgery for multi-task learning.
\newblock In H.~Larochelle, M.~Ranzato, R.~Hadsell, M.F. Balcan, and H.~Lin,
  editors, \emph{Advances in Neural Information Processing Systems}, volume~33,
  pages 5824--5836. Curran Associates, Inc., 2020.
\newblock URL
  \url{https://proceedings.neurips.cc/paper_files/paper/2020/file/3fe78a8acf5fda99de95303940a2420c-Paper.pdf}.

\bibitem[Liu et~al.(2021)Liu, Liu, Jin, Stone, and Liu]{liu2021conflict}
Bo~Liu, Xingchao Liu, Xiaojie Jin, Peter Stone, and Qiang Liu.
\newblock Conflict-averse gradient descent for multi-task learning.
\newblock \emph{Advances in Neural Information Processing Systems},
  34:\penalty0 18878--18890, 2021.

\bibitem[Wolberg et~al.(1993)Wolberg, Mangasarian, Street, and
  Street]{breast_cancer_wisconsin_(diagnostic)_17}
William Wolberg, Olvi Mangasarian, Nick Street, and W.~Street.
\newblock {Breast Cancer Wisconsin (Diagnostic)}.
\newblock UCI Machine Learning Repository, 1993.
\newblock {DOI}: https://doi.org/10.24432/C5DW2B.

\bibitem[Pedregosa et~al.(2011)Pedregosa, Varoquaux, Gramfort, Michel, Thirion,
  Grisel, Blondel, Prettenhofer, Weiss, Dubourg, Vanderplas, Passos,
  Cournapeau, Brucher, Perrot, and Duchesnay]{pedregosa2011scikit}
F.~Pedregosa, G.~Varoquaux, A.~Gramfort, V.~Michel, B.~Thirion, O.~Grisel,
  M.~Blondel, P.~Prettenhofer, R.~Weiss, V.~Dubourg, J.~Vanderplas, A.~Passos,
  D.~Cournapeau, M.~Brucher, M.~Perrot, and E.~Duchesnay.
\newblock Scikit-learn: Machine learning in {P}ython.
\newblock \emph{Journal of Machine Learning Research}, 12:\penalty0 2825--2830,
  2011.

\bibitem[Janosi et~al.(1989)Janosi, Steinbrunn, Pfisterer, and
  Detrano]{heart_disease_45}
Andras Janosi, William Steinbrunn, Matthias Pfisterer, and Robert Detrano.
\newblock {Heart Disease}.
\newblock UCI Machine Learning Repository, 1989.
\newblock {DOI}: https://doi.org/10.24432/C52P4X.

\bibitem[selva86(2024)]{ionosphere}
selva86.
\newblock Datasets.
\newblock \url{https://github.com/selva86/datasets/blob/master/Ionosphere.csv},
  2024.

\bibitem[Luo et~al.(2016)Luo, Wang, Li, Luo, Peng, Wu, and Pan]{luo2016drug}
Huimin Luo, Jianxin Wang, Min Li, Junwei Luo, Xiaoqing Peng, Fang-Xiang Wu, and
  Yi~Pan.
\newblock Drug repositioning based on comprehensive similarity measures and
  bi-random walk algorithm.
\newblock \emph{Bioinformatics}, 32\penalty0 (17):\penalty0 2664--2671, 2016.

\bibitem[Yoon et~al.(2018{\natexlab{a}})Yoon, Jordon, and Schaar]{yoon2018gain}
Jinsung Yoon, James Jordon, and Mihaela Schaar.
\newblock Gain: Missing data imputation using generative adversarial nets.
\newblock In \emph{International conference on machine learning}, pages
  5689--5698. PMLR, 2018{\natexlab{a}}.

\bibitem[You et~al.(2020)You, Ma, Ding, Kochenderfer, and
  Leskovec]{you_handling_2020}
Jiaxuan You, Xiaobai Ma, Yi~Ding, Mykel~J Kochenderfer, and Jure Leskovec.
\newblock Handling {Missing} {Data} with {Graph} {Representation} {Learning}.
\newblock In H.~Larochelle, M.~Ranzato, R.~Hadsell, M.~F. Balcan, and H.~Lin,
  editors, \emph{Advances in {Neural} {Information} {Processing} {Systems}},
  volume~33, pages 19075--19087. Curran Associates, Inc., 2020.
\newblock URL
  \url{https://proceedings.neurips.cc/paper_files/paper/2020/file/dc36f18a9a0a776671d4879cae69b551-Paper.pdf}.

\bibitem[Chen et~al.(2024)Chen, Li, Wang, Zhang, Xu, Jiang, Song, and
  Wang]{chen2024rethinking}
Zhichao Chen, Haoxuan Li, Fangyikang Wang, Odin Zhang, Hu~Xu, Xiaoyu Jiang,
  Zhihuan Song, and Hao Wang.
\newblock Rethinking the diffusion models for missing data imputation: A
  gradient flow perspective.
\newblock In \emph{The Thirty-eighth Annual Conference on Neural Information
  Processing Systems}, 2024.
\newblock URL \url{https://openreview.net/forum?id=fIz8K4DJ7w}.

\bibitem[Du et~al.(2023)Du, Melis, and Wang]{du2023remasker}
Tianyu Du, Luca Melis, and Ting Wang.
\newblock Remasker: Imputing tabular data with masked autoencoding.
\newblock \emph{arXiv preprint arXiv:2309.13793}, 2023.

\bibitem[Zhao et~al.(2023)Zhao, Sun, Dezfouli, and
  Bonilla]{zhao2023transformed}
He~Zhao, Ke~Sun, Amir Dezfouli, and Edwin~V Bonilla.
\newblock Transformed distribution matching for missing value imputation.
\newblock In \emph{International Conference on Machine Learning}, pages
  42159--42186. PMLR, 2023.

\bibitem[LeCun et~al.(1998)LeCun, Cortes, and Burges]{lecun1998mnist}
Y.~LeCun, C.~Cortes, and C.J.C. Burges.
\newblock The mnist database of handwritten digits.
\newblock
  \url{https://drive.google.com/file/d/1eEKzfmEu6WKdRlohBQiqi3PhW_uIVJVP/view},
  1998.

\bibitem[Réda(2023{\natexlab{a}})]{reda2023predict}
Clémence Réda.
\newblock Predict drug repurposing dataset. doi: 10.5281/zenodo.7983090,
  2023{\natexlab{a}}.
\newblock URL \url{https://doi.org/10.5281/zenodo.7983090}.

\bibitem[Johnson et~al.(2019)Johnson, Douze, and J{\'e}gou]{johnson2019billion}
Jeff Johnson, Matthijs Douze, and Herv{\'e} J{\'e}gou.
\newblock Billion-scale similarity search with gpus.
\newblock \emph{IEEE Transactions on Big Data}, 7\penalty0 (3):\penalty0
  535--547, 2019.

\bibitem[Zhao et~al.(2014)Zhao, Lu, and Mei]{zhao2014locality}
Kang Zhao, Hongtao Lu, and Jincheng Mei.
\newblock Locality preserving hashing.
\newblock In \emph{Proceedings of the AAAI conference on artificial
  intelligence}, volume~28, 2014.

\bibitem[Tsai and Yang(2014)]{tsai2014locality}
Yi-Hsuan Tsai and Ming-Hsuan Yang.
\newblock Locality preserving hashing.
\newblock In \emph{2014 IEEE International Conference on Image Processing
  (ICIP)}, pages 2988--2992. IEEE, 2014.

\bibitem[Bernhardsson(2018)]{Github:annoy}
Erik Bernhardsson.
\newblock \emph{Annoy: Approximate Nearest Neighbors in C++/Python}, 2018.
\newblock URL \url{https://pypi.org/project/annoy/}.
\newblock Python package version 1.13.0.

\bibitem[Tiao et~al.(2021)Tiao, Klein, Seeger, Bonilla, Archambeau, and
  Ramos]{tiao2021bore}
Louis~C Tiao, Aaron Klein, Matthias~W Seeger, Edwin~V Bonilla, Cedric
  Archambeau, and Fabio Ramos.
\newblock Bore: Bayesian optimization by density-ratio estimation.
\newblock In \emph{International Conference on Machine Learning}, pages
  10289--10300. PMLR, 2021.

\bibitem[Tashiro et~al.(2021{\natexlab{a}})Tashiro, Song, Song, and
  Ermon]{tashiro2021csdi}
Yusuke Tashiro, Jiaming Song, Yang Song, and Stefano Ermon.
\newblock Csdi: Conditional score-based diffusion models for probabilistic time
  series imputation.
\newblock \emph{Advances in Neural Information Processing Systems},
  34:\penalty0 24804--24816, 2021{\natexlab{a}}.

\bibitem[Luo et~al.(2018)Luo, Cai, Zhang, Xu, et~al.]{luo2018multivariate}
Yonghong Luo, Xiangrui Cai, Ying Zhang, Jun Xu, et~al.
\newblock Multivariate time series imputation with generative adversarial
  networks.
\newblock \emph{Advances in neural information processing systems}, 31, 2018.

\bibitem[Tashiro et~al.(2021{\natexlab{b}})Tashiro, Song, Song, and
  Ermon]{tashiro_csdi_2021}
Yusuke Tashiro, Jiaming Song, Yang Song, and Stefano Ermon.
\newblock {CSDI}: {Conditional} {Score}-based {Diffusion} {Models} for
  {Probabilistic} {Time} {Series} {Imputation}.
\newblock In M.~Ranzato, A.~Beygelzimer, Y.~Dauphin, P.~S. Liang, and
  J.~Wortman Vaughan, editors, \emph{Advances in {Neural} {Information}
  {Processing} {Systems}}, volume~34, pages 24804--24816. Curran Associates,
  Inc., 2021{\natexlab{b}}.
\newblock URL
  \url{https://proceedings.neurips.cc/paper_files/paper/2021/file/cfe8504bda37b575c70ee1a8276f3486-Paper.pdf}.

\bibitem[Chen et~al.(2023)Chen, Deng, Fang, Li, Yang, Zhang, Rasul, Zhe,
  Schneider, and Nevmyvaka]{chen2023provably}
Yu~Chen, Wei Deng, Shikai Fang, Fengpei Li, Nicole~Tianjiao Yang, Yikai Zhang,
  Kashif Rasul, Shandian Zhe, Anderson Schneider, and Yuriy Nevmyvaka.
\newblock Provably convergent schr{\"o}dinger bridge with applications to
  probabilistic time series imputation.
\newblock In \emph{International Conference on Machine Learning}, pages
  4485--4513. PMLR, 2023.

\bibitem[Zheng and Charoenphakdee(2022)]{zheng2022diffusion}
Shuhan Zheng and Nontawat Charoenphakdee.
\newblock Diffusion models for missing value imputation in tabular data.
\newblock In \emph{NeurIPS 2022 First Table Representation Workshop}, 2022.
\newblock URL \url{https://openreview.net/forum?id=4q9kFrXC2Ae}.

\bibitem[Jolicoeur-Martineau et~al.(2024)Jolicoeur-Martineau, Fatras, and
  Kachman]{pmlr-v238-jolicoeur-martineau24a}
Alexia Jolicoeur-Martineau, Kilian Fatras, and Tal Kachman.
\newblock Generating and imputing tabular data via diffusion and flow-based
  gradient-boosted trees.
\newblock In Sanjoy Dasgupta, Stephan Mandt, and Yingzhen Li, editors,
  \emph{Proceedings of The 27th International Conference on Artificial
  Intelligence and Statistics}, volume 238 of \emph{Proceedings of Machine
  Learning Research}, pages 1288--1296. PMLR, 02--04 May 2024.
\newblock URL
  \url{https://proceedings.mlr.press/v238/jolicoeur-martineau24a.html}.

\bibitem[Ouyang et~al.(2025)Ouyang, Xie, Li, and Cheng]{ouyang2025missdiff}
Yidong Ouyang, Liyan Xie, Chongxuan Li, and Guang Cheng.
\newblock Missdiff: Training diffusion models on tabular data with missing
  values, 2025.
\newblock URL \url{https://openreview.net/forum?id=PyyoSwPaSa}.

\bibitem[Peis et~al.(2022)Peis, Ma, and Hern{\'a}ndez-Lobato]{peis2022missing}
Ignacio Peis, Chao Ma, and Jos{\'e}~Miguel Hern{\'a}ndez-Lobato.
\newblock Missing data imputation and acquisition with deep hierarchical models
  and hamiltonian monte carlo.
\newblock In Alice~H. Oh, Alekh Agarwal, Danielle Belgrave, and Kyunghyun Cho,
  editors, \emph{Advances in Neural Information Processing Systems}, 2022.
\newblock URL \url{https://openreview.net/forum?id=xpR25Tsem9C}.

\bibitem[Du et~al.(2024)Du, Melis, and Wang]{du2024remasker}
Tianyu Du, Luca Melis, and Ting Wang.
\newblock Remasker: Imputing tabular data with masked autoencoding.
\newblock In \emph{The Twelfth International Conference on Learning
  Representations}, 2024.
\newblock URL \url{https://openreview.net/forum?id=KI9NqjLVDT}.

\bibitem[He et~al.(2021)He, Chen, Xie, Li, Doll{\'{a}}r, and Girshick]{mae}
Kaiming He, Xinlei Chen, Saining Xie, Yanghao Li, Piotr Doll{\'{a}}r, and
  Ross~B. Girshick.
\newblock Masked autoencoders are scalable vision learners.
\newblock \emph{CoRR}, abs/2111.06377, 2021.
\newblock URL \url{https://arxiv.org/abs/2111.06377}.

\bibitem[Muzellec et~al.(2020{\natexlab{b}})Muzellec, Josse, Boyer, and
  Cuturi]{pmlr-v119-muzellec20a}
Boris Muzellec, Julie Josse, Claire Boyer, and Marco Cuturi.
\newblock Missing data imputation using optimal transport.
\newblock In Hal~Daumé III and Aarti Singh, editors, \emph{Proceedings of the
  37th International Conference on Machine Learning}, volume 119 of
  \emph{Proceedings of Machine Learning Research}, pages 7130--7140. PMLR,
  13--18 Jul 2020{\natexlab{b}}.
\newblock URL \url{https://proceedings.mlr.press/v119/muzellec20a.html}.

\bibitem[Li et~al.(2019)Li, Jiang, and Marlin]{li2018learning}
Steven Cheng-Xian Li, Bo~Jiang, and Benjamin Marlin.
\newblock Learning from incomplete data with generative adversarial networks.
\newblock In \emph{International Conference on Learning Representations}, 2019.
\newblock URL \url{https://openreview.net/forum?id=S1lDV3RcKm}.

\bibitem[Yoon et~al.(2018{\natexlab{b}})Yoon, Jordon, and van~der
  Schaar]{pmlr-v80-yoon18a}
Jinsung Yoon, James Jordon, and Mihaela van~der Schaar.
\newblock {GAIN}: Missing data imputation using generative adversarial nets.
\newblock In Jennifer Dy and Andreas Krause, editors, \emph{Proceedings of the
  35th International Conference on Machine Learning}, volume~80 of
  \emph{Proceedings of Machine Learning Research}, pages 5689--5698. PMLR,
  10--15 Jul 2018{\natexlab{b}}.
\newblock URL \url{https://proceedings.mlr.press/v80/yoon18a.html}.

\bibitem[Hamilton et~al.(2017)Hamilton, Ying, and Leskovec]{graphsage}
William~L. Hamilton, Rex Ying, and Jure Leskovec.
\newblock Inductive representation learning on large graphs.
\newblock \emph{CoRR}, abs/1706.02216, 2017.
\newblock URL \url{http://arxiv.org/abs/1706.02216}.

\bibitem[Vershynin(2018)]{vershynin2018high}
Roman Vershynin.
\newblock \emph{High-dimensional probability: An introduction with applications
  in data science}, volume~47.
\newblock Cambridge university press, 2018.

\bibitem[Réda(2023{\natexlab{b}})]{reda2023transcript}
Clémence Réda.
\newblock Transcript drug repurposing dataset. doi: 10.5281/zenodo.7982976,
  2023{\natexlab{b}}.
\newblock URL \url{https://doi.org/10.5281/zenodo.7982976}.

\bibitem[Virtanen et~al.(2020)Virtanen, Gommers, Oliphant, Haberland, Reddy,
  Cournapeau, Burovski, Peterson, Weckesser, Bright, {van der Walt}, Brett,
  Wilson, Millman, Mayorov, Nelson, Jones, Kern, Larson, Carey, Polat, Feng,
  Moore, {VanderPlas}, Laxalde, Perktold, Cimrman, Henriksen, Quintero, Harris,
  Archibald, Ribeiro, Pedregosa, {van Mulbregt}, and {SciPy 1.0
  Contributors}]{2020SciPy-NMeth}
Pauli Virtanen, Ralf Gommers, Travis~E. Oliphant, Matt Haberland, Tyler Reddy,
  David Cournapeau, Evgeni Burovski, Pearu Peterson, Warren Weckesser, Jonathan
  Bright, St{\'e}fan~J. {van der Walt}, Matthew Brett, Joshua Wilson, K.~Jarrod
  Millman, Nikolay Mayorov, Andrew R.~J. Nelson, Eric Jones, Robert Kern, Eric
  Larson, C~J Carey, {\.I}lhan Polat, Yu~Feng, Eric~W. Moore, Jake
  {VanderPlas}, Denis Laxalde, Josef Perktold, Robert Cimrman, Ian Henriksen,
  E.~A. Quintero, Charles~R. Harris, Anne~M. Archibald, Ant{\^o}nio~H. Ribeiro,
  Fabian Pedregosa, Paul {van Mulbregt}, and {SciPy 1.0 Contributors}.
\newblock {{SciPy} 1.0: Fundamental Algorithms for Scientific Computing in
  Python}.
\newblock \emph{Nature Methods}, 17:\penalty0 261--272, 2020.
\newblock \doi{10.1038/s41592-019-0686-2}.

\bibitem[Arthur and Vassilvitskii(2006)]{arthur2006k}
David Arthur and Sergei Vassilvitskii.
\newblock k-means++: The advantages of careful seeding.
\newblock Technical report, Stanford, 2006.

\bibitem[Gao et~al.(2022)Gao, Zhou, Xin, Min, and Du]{gao2022dda}
Chu-Qiao Gao, Yuan-Ke Zhou, Xiao-Hong Xin, Hui Min, and Pu-Feng Du.
\newblock Dda-skf: predicting drug--disease associations using similarity
  kernel fusion.
\newblock \emph{Frontiers in Pharmacology}, 12:\penalty0 784171, 2022.

\bibitem[Akiba et~al.(2019)Akiba, Sano, Yanase, Ohta, and
  Koyama]{akiba2019optuna}
Takuya Akiba, Shotaro Sano, Toshihiko Yanase, Takeru Ohta, and Masanori Koyama.
\newblock {O}ptuna: A next-generation hyperparameter optimization framework.
\newblock In \emph{The 25th ACM SIGKDD International Conference on Knowledge
  Discovery \& Data Mining}, pages 2623--2631, 2019.

\bibitem[Bergstra et~al.(2011)Bergstra, Bardenet, Bengio, and
  K\'{e}gl]{hyperparamTPE}
James Bergstra, R\'{e}mi Bardenet, Yoshua Bengio, and Bal\'{a}zs K\'{e}gl.
\newblock Algorithms for hyper-parameter optimization.
\newblock In J.~Shawe-Taylor, R.~Zemel, P.~Bartlett, F.~Pereira, and K.Q.
  Weinberger, editors, \emph{Advances in Neural Information Processing
  Systems}, volume~24. Curran Associates, Inc., 2011.
\newblock URL
  \url{https://proceedings.neurips.cc/paper_files/paper/2011/file/86e8f7ab32cfd12577bc2619bc635690-Paper.pdf}.

\bibitem[Hutter et~al.(2014)Hutter, Hoos, and Leyton-Brown]{pmlr-v32-hutter14}
Frank Hutter, Holger Hoos, and Kevin Leyton-Brown.
\newblock An efficient approach for assessing hyperparameter importance.
\newblock In Eric~P. Xing and Tony Jebara, editors, \emph{Proceedings of the
  31st International Conference on Machine Learning}, volume~32 of
  \emph{Proceedings of Machine Learning Research}, pages 754--762, Bejing,
  China, 22--24 Jun 2014. PMLR.
\newblock URL \url{https://proceedings.mlr.press/v32/hutter14.html}.

\end{thebibliography}
\bibliographystyle{unsrtnat}

\appendix

\section{Other related works and comments on F3I}\label{app:relatedworks}

\paragraph{Related works~ } Relevant methods developed for multivariate time-series data might also be adapted to single-timepoint data sets: Conditional Score-based Diffusion Models (CDSI)~\cite{tashiro2021csdi}, Generative Adversarial Networks~\cite{luo2018multivariate,yoon2018gain} or Last Observation Carried Forward (LOCF), where the last non-missing value is duplicated until the next non-missing time point. However, those methods are best-suited in the case where temporal connections can be made, and we restrict our study to single-timepoint data. 

Due to diffusion models' impressive generative modeling capacity, many imputation approaches based on them have been developed in recent years. One of the first approaches in this direction was Conditional Score-Based Diffusion models~\cite{tashiro_csdi_2021}, which attempts to impute values at missing time points in multivariate time-series data by conditioning the diffusion model on the observed time points and then denoising from white noise using the conditional diffusion model. Another similar approach~\cite{chen2023provably} constructs and solves a Schrödinger Bridge problem with conditional constraints to impute missing values in time-series data, where the conditional constraints are derived from the observed values. However, both these approaches are only available for imputation in multivariate time-series data, which is out of the scope of our paper. 

An interesting approach for tabular data~\cite{zheng2022diffusion} consists of removing the temporal transformer layers from CSDI~\cite{tashiro_csdi_2021} and constructing embeddings for categorical variables, after which the same diffusion procedure as CSDI is applied. Another diffusion-based approach~\cite{pmlr-v238-jolicoeur-martineau24a} constructs multiple copies of the dataset with different noise samples added, on which gradient-boosted trees are trained to perform the reverse diffusion process to reconstruct the missing data values. The NewImp~\cite{chen2024rethinking} model, on the other hand, attempts to learn the missing value distribution by learning the score function of the joint distribution using a Denoising Score Matching approach and then by stepwise reconstructing the missing values using an ODE simulation of the joint distribution based on these learned scores. Another similar approach, MissDiff~\cite{ouyang2025missdiff}, attempts to learn a score-based model by modifying the loss in a conventional Variance Preserving SDE for diffusion models to only consider the observed values.

A variational autoencoder-based approach~\cite{peis2022missing} generates the missing data by training multiple separate variational autoencoders to encode the latent space of each feature and the dependencies among them. This training is helpful in the presence of diverse feature types. New approaches also explore the use of transformer models in missing data imputation. For instance, ReMasker~\cite{du2024remasker} is an extension of the Masked AutoEncoder~\cite{mae} where the encoder and decoder models consist of a sequence of transformer layers, and the model learns by masking additionally available values and learning to predict them, similar to CSDI~\cite{tashiro_csdi_2021}. Another approach based on a modification of OTImputer~\cite{pmlr-v119-muzellec20a} attempts to minimize the Wasserstein distance between latent representations of the original and imputed data set generated via a neural network, where the latent representation generating neural network is fit to maximize the Mutual Information to prevent model collapse. 

Other generative approaches involve using Generative Adversarial Networks. The simplest GAN-based approach~\cite{li2018learning} jointly trains two pairs of generator-discriminator networks, one to predict the data and the other to predict the missing masks. Another GAN-based approach masks certain values in the data set besides the missing ones. It uses the generator to generate all masked values, after which the discriminator is trained to predict the mask matrix from the imputed data set~\cite{pmlr-v80-yoon18a}. 

Finally, GRAPE is a somewhat more novel approach~\cite{you_handling_2020}, which consists of representing the data set as a labeled bipartite graph, with the two node types representing the samples and the features and each edge label representing the value of the particular feature for that sample. One then uses a GraphSAGE~\cite{graphsage} inspired Graph Neural Network to predict the missing edge labels, which are the imputed values.

\paragraph{Comparison of F3I to optimal transport-based imputation~ } Authors of OTImputer in~\cite{muzellec2020missing} leverage optimal transport (OT) to define a loss function based on Sinkhorn divergences for imputation. This loss function, like $G$, aims at quantifying the gap in data distribution between any two random batches of samples from the data matrix and can also be iteratively minimized through a gradient descent approach. The imputation is performed feature-wise. However, this approach requires the input of several parameters, among which $t_\text{max}$, the budget for the number of improvements, which is always fully exhausted (contrary to F3I where an early stopping criterion exists); $m$ the size of the randomly sampled batches and $K$ the number of batches which are evaluated. Moreover, contrary to F3I, the OT imputer does not provide theoretical guarantees on the imputation quality.

\paragraph{Out-of-sample imputation for F3I~ } For a new sample $\bm{x} \in (\mathbb{R} \cup \{\texttt{N/A}\})^{F}$, is there a way not to re-run the full F3I procedure? If we assume that the new sample comes from the data set, the simplest idea is to apply on $\bm{x}$ the initial imputer and successively the imputation improvement model in Algorithm~\ref{alg:imputation} with the weight vector 
$\bm{\alpha}^t$, where $t$ is the final step of F3I. However, this out-of-sample imputation loses the theoretical guarantees in Section~\ref{sec:analysis}. 


\section{Theoretical assumptions}\label{sec:assumptions}

In this section, we state the formal expression of our assumptions about the data generation procedure to derive theoretical guarantees from our algorithm. We explicitly write the pseudo-code for the imputation improvement model in Algorithm~\ref{alg:imputation} and the full data generation procedure in Algorithm~\ref{alg:data_generation}.

\begin{algorithm}[tb]
   \caption{Imputation improvement model \texttt{Impute}$(\cdot; \bm{\alpha}, Z)$}
   \label{alg:imputation}
\begin{algorithmic}
   \STATE {\bfseries Input:} Guess for a $F$-dimensional $\bm{x} \in \mathbb{R}^{F}$, missing indicator for that sample $\bm{m} \in \{0,1\}^{F}$
   \STATE {\bfseries Parameters:} Number of neighbors $K$, weights $\bm{\alpha} \in \triangle_K$, reference set $Z = \{\bm{z}^i\}_{0 \leq i \leq N} \subset \mathbb{R}^F$
   \STATE {\bfseries Output:} Improved guess $\widetilde{\bm{x}} \in \mathbb{R}^F$
   \STATE \textcolor{gray}{\# Neighbor indices by increasing Chebychev distance:  $\overset{1,2,\dots,K}{\arg\,\min}_{j \leq N}$ selects the $K$ elements in $1,2,\dots,N$ with smallest values (with a k-d tree for instance)}
   \STATE $(n_1, n_2, \dots, n_K) \gets \overset{1,2,\dots,K}{\arg\,\min}_{j \leq N} \max_{f \leq F} |x_{f}-z^j_{f}|$
   \STATE $\widetilde{x}_f \gets \sum_{k \leq K} \alpha_k z^{n_k}_{f}$ for any $f \leq F$, $m^f = 1$
\end{algorithmic}
\end{algorithm}

First, we assume that each value in the full data matrix is drawn from independent fixed-variance Gaussian distributions.

\begin{assumption}{\textnormal{Independent Gaussian distributions.}}\label{as:x_distribution} There exist $\bm{\mu} \in \mathbb{R}^F$ and $\sigma > 0$ such that, for any sample $i \leq N$ and any feature $f \leq F$, $(x^\star)^f_i \sim_{\text{iid}} \mathcal{N}(\mu_f, \sigma^2)$.
\end{assumption}

The random indicator variables $m^f_i$ are then independently drawn according to the missingness mechanism with probability $p^\text{miss}$. If $m^f_i=1$, then the coefficient at position $(i,f)$ $x^f_i$ in $X$ is unavailable, otherwise, $x^f_i=(x^\star)^f_i$. We will provide an analysis of our algorithm for three types of missingness mechanisms: 

\begin{assumption}{\textnormal{MCAR mechanism: Bernouilli distribution.}}\label{as:mcar}
The random indicator variables for missing values $m^f_i$ are drawn \textit{iid} from $\mathcal{B}(p^\text{miss}(\bm{x}))$, where $p^\text{miss}(\bm{x})  \in (0,1)$ is a constant value for any $\bm{x}$.
\end{assumption}

\begin{assumption}{\textnormal{MAR mechanism.}}\label{as:mar}
We assume a subset $\mathcal{F}_o$ of size $F_o < F$ features is always observed. We denote $(x^\star)^\text{$\mid$ obs}_i \triangleq (x^\star_i[f])_{f \in \mathcal{F}_o}$. Then there exist a function $p^\text{miss}$, $\forall \bm{x} \in \mathbb{R}^{F_o}$, $\mathbb{P}(m^f_i=1 \mid (x^\star)^\text{$\mid$ obs}_i=\bm{x}) = p^\text{miss}(\bm{x}) \;.$ 
\end{assumption}

\begin{assumption}{\textnormal{MNAR mechanism: Gaussian self-masking (Assumption $4$ from~\cite{le2020neumiss}).}}\label{as:mnar}
The probability of event $\{m^f_i=1\}$ depends on $(x^\star)^f_i$: $\exists K_f \in (0,1) , \ \forall x \in \mathbb{R}$, $\mathbb{P}(m^f_i=1 \mid (x^\star)^f_i=x) = K_f e^{-\frac{1}{\sigma^2}(x-\mu_f)^2} = p^\text{miss}(\bm{x})\;.$
\end{assumption}

We also ensure that there are exactly $K$ neighbors for the initial simple guesses and that we know a (constant) upper bound on the norm of any feature vectors.

\begin{algorithm}[tb]
   \caption{Data generation procedure according to Assumptions~\ref{as:x_distribution}-\ref{as:ub_norm}}
   \label{alg:data_generation}
\begin{algorithmic}
   \STATE {\bfseries Input:} $N$ number of samples, $F$ number of features
   \STATE {\bfseries Output:} Initial data $X^\text{miss} \in (\mathbb{R}\cup\{\texttt{NaN}\})^{N \times F}$ and naively imputed $X^0 \in \mathbb{R}^{N \times F}$
    \STATE \textcolor{gray}{\# Generation of the complete data set}
    \FOR{$i=1,2,\dots,N$}
        \FOR{$f=1,2,\dots,F$}
             \STATE $(\bm{x}^\star)_i  \sim \mathcal{N}_F(\mu_f, \sigma^2) $ (where $\Theta \deff (\bm{\mu}=(\mu_1, \dots, \mu_F), \sigma^2\bm{I}_{F \times F}) \in \mathbb{R}^F \times \mathbb{R}^{F \times F}$)
        \ENDFOR
    \ENDFOR
    \STATE \textcolor{gray}{\# Missingness mechanism}
    \FOR{$i=1,2,\dots,N$}
        \FOR{$f=1,2,\dots,F$}
             \STATE $m^f_i  \sim_\text{iid} p^\text{miss}((\bm{x}^\star)_i,f) $
             \IF{$m^f_i=1$}
                \STATE {\bfseries then} $(x^\text{miss})^f_i \gets \texttt{NaN}$
                 \STATE {\bfseries else} $(x^\text{miss})^f_i \gets (x^\star)^f_i$
            \ENDIF
        \ENDFOR
    \ENDFOR
    \STATE \textcolor{gray}{\# Create the naively imputed data set}
    \FOR{$i=1,2,\dots,N$}
        \FOR{$f=1,2,\dots,F$}
             \IF{$m^f_i=1$}
               \STATE \textcolor{gray}{\# K-nearest neighbor imputation with uniform weights}
               \STATE \textcolor{gray}{\# $k^\text{th}$ closest neighbor for $(\bm{x}^\text{miss})_i^f$ is denoted $\mathcal{K}((\bm{x}^\text{miss})_i,f,k)$}
               \STATE {\bfseries then} $(x^0)^f_i \gets \frac{1}{K}\sum_{k \leq K} x^f_{\mathcal{K}((\bm{x}^\text{miss})_i,f,k)} = \frac{1}{K}\sum_{k \leq K} (x^\star)^f_{\mathcal{K}((\bm{x}^\text{miss})_i,f,k)}$ 
               \STATE {\bfseries else} $(x^0)^f_i \gets (x^\star)^f_i$
            \ENDIF
        \ENDFOR
    \ENDFOR
\end{algorithmic}
\end{algorithm}

\begin{assumption}{\textnormal{Number of neighbors $K$.}}\label{as:number_neighbors}
In the remainder of the paper, if $\{ i \leq N \mid m^f_i=0 \}$ is the set of data point indices for which the feature $f$ is not missing, then $K \leq \min_{f \leq F} |\{ i \leq N \mid m^f_i=0 \}|$. Without a loss of generality, $\min_{f \leq F} |\{ i \leq N \mid m^f_i=0 \}| \geq 2$ (otherwise, we can ignore the corresponding feature).
\end{assumption}

\begin{assumption}{\textnormal{Upper bound on any of the $(\|\bm{x}_i\|^2_2)_{i \leq n}$.}}\label{as:ub_norm}
We assume a constant $S > 0$ exists, such that for any $i \leq N$, $\|\bm{x}_i\|^2_2 \leq S$ (ignoring potential missing values). Up to renormalization, we assume that $S=1$. Moreover, the initial imputation step (the ``simple guess'') preserves that condition, meaning that for any $i \leq N$ and $t \geq 0$, $\|\bm{x}^t_i\|^2_2 \leq S$, where $X^0$ is the imputed data matrix with the initial imputation step, and $X^t$ for $t\geq1$ is obtained through Algorithm~\ref{alg:imputation}.
\end{assumption}

\begin{remark}\textnormal{Assumption~\ref{as:ub_norm} can hold.}
    Indeed, Assumption~\ref{as:ub_norm} is satisfied by the K-nearest neighbor imputation with uniform weights, where the imputed value equals the mean of all feature-wise values from the $K$ neighbors.
\end{remark}

\begin{assumption}{Assumptions on $G$ and $l$ for Theorem~\ref{thm:regret_f3i_joint}}\label{as:pcgrad}
    Let $\ell$ be a convex pointwise loss such that $\nabla\ell$ is Lipschitz-continuous and $\beta\in[0,1]$. Define $\mathcal{G}$ as in~\eqref{eq:joint_training_obj}. Also define $\bm{H}(\mathcal{G};\bm{\alpha},\bm{\alpha}')=\int_0^1\nabla\mathcal{G}(\bm{\alpha},X;\beta)^\intercal\nabla^2\mathcal{G}(\bm{\alpha}+a(\bm{\alpha}'-\bm{\alpha}),X;\beta)da$. Let $\bm{g_1}=\nabla_\alpha(1-\beta)G(\bm{\alpha})$, $\bm{g}_2=-\nabla_\alpha\frac{\beta}{N}\sum_{i\leq N}\ell(\bm{x_i(\alpha)})$, $\phi_{12}$ be the angle between $\bm{g}_1$ and $\bm{g}_2$,
    and $\bm{g}=\bm{g}_1+\bm{g}_2$. Let $\bm{\alpha}^t$ be the updated value of $\bm{\alpha}$ and let $\lambda$ be the step-size for this update. Lemma~\ref{lem:pcagrad_loss} shows that $-\mathcal{G}$ is also Lipschitz-continuous w.r.t. $\bm{\alpha}$, so let this Lipschitz constant be $L$. We assume that there exists $w\leq L$ such that $\bm{H}(-\mathcal{G},\bm{\alpha},\bm{\alpha}^t)\geq w\norm{g}_2^2$. We also assume that $\cos{\phi_{12}}\leq \frac{2\norm{g_1}_2\norm{g_2}_2}{\norm{g_1}_2^2\norm{g_2}_2^2}$, $w\geq(1-\cos^2(\phi_{12})\frac{\norm{g_1-g_2}_2^2}{\norm{g_1+g_2}_2^2}W$ and $\lambda\geq\frac{2}{w-(1-\cos^2(\phi_{12}))\frac{\norm{g_1-g_2}_2^2}{\norm{g_1+g_2}_2^2}W}$.
\end{assumption}

\section{Properties of the objective function G}\label{subapp:G_lemmas}

\begin{proposition}{\textnormal{Continuity and derivability of $G$.}}\label{lem:G_continuous} $G$ is continuous and infinitely derivable with respect to $\bm{\alpha} \in \triangle_K$.\end{proposition}

\begin{proof}
    $G$ is a composition and sum of indefinitely derivable functions on their respective domains, which are compatible: $\log$ on $\mathbb{R}^{+*}$, $\exp$ on $\mathbb{R}$ of image domain $\mathbb{R}^{+*}$, $\|\cdot\|_2$ and the linear imputation model (Algorithm~\ref{alg:imputation}) on $\mathbb{R}$ with image domain $\mathbb{R}$.
\end{proof}

\begin{proposition}{\textnormal{Strict concavity of $G$ in $\bm{\alpha}$.}}~\label{lem:G_strictly_concave} Assume that $\eta < 4KN$. Then there exists $h_0>0$ such that for all $h \geq h_0$
, $G$ is strictly concave in $\bm{\alpha}$. 
\end{proposition}

    We aim to show that a value of $h_0$ always exists such that, for $h \geq 0$, the Hessian matrix of $G$ with respect to $\bm{\alpha}$ is negative (semi-)definite. First, we compute the Hessian matrix of $G$.

    \begin{lemma}{\textnormal{Gradient of $G$ with respect to $\bm{\alpha}$.}}\label{lem:gradient_G} The gradient $\nabla_{\bm{\alpha}} G(\bm{\alpha}, X) \in \mathbb{R}^{K}$ at $\bm{\alpha} \in \mathbb{R}^K$ and fixed $X \in \mathbb{R}^{N \times F}$ is 
\[ \nabla_{\bm{\alpha}} G(\bm{\alpha}, X) = - \sum_{i \leq N} \frac{D_0(\bm{x}_i(\bm{\alpha}))^{-1}}{2hN^2(\sqrt{2 \pi}h)^F} \left( \sum_{j \leq N} e^{-\frac{1}{4h}\|\bm{x}_i(\bm{\alpha})-(\bm{x}^0)_j\|^2_2}\left(\bm{x}_i(\bm{\alpha})-(\bm{x}^0)_j\right)\right)^\intercal \widetilde{Z}^{n_i}  - 2\eta \bm{\alpha} \;,\]
where $n^i \triangleq (n^i_1, n^i_2, \dots, n^i_K)$ is the set of indices of the K-nearest neighbors of $\bm{x}_{i}$ among the reference set $Z=\{(\bm{x}^0)_1, (\bm{x}^0)_2, \dots, (\bm{x}^0)_N\}$, $\widetilde{Z}^{n_i} \in \mathbb{R}^{F \times K}$ where the $k^\text{th}$ column of $\widetilde{Z}^{n_i}$ is defined as $(\widetilde{\bm{z}}^{n_i})^f_k = 0$ if $m^f_i=0$, $(\widetilde{z}^{n_i})^f_k = (x^0)^f_{n^i_k}$ otherwise. That is, $(\widetilde{\bm{z}}^{n_i})_k$ is equal to the $k^\text{th}$ closest neighbor of $\bm{x}_i$ (by increasing order of distance) on missing coordinates of $\bm{x}_i$, and equal to zero otherwise.
    \end{lemma}

\begin{proof}
    The gradient of $G$ at $\bm{\alpha} \in \triangle_K$ for a fixed $X \in \mathbb{R}^{N \times F}$ is
    \begin{eqnarray*}
        \nabla_{\bm{\alpha}} G(\bm{\alpha}, X) & = & \sum_{i \leq N} \frac{D_0(\bm{x}_i(\bm{\alpha}))^{-1}}{N} \nabla_{\bm{\alpha}} D_0(\bm{x}_i(\bm{\alpha}))  - 0 - 2\eta \bm{\alpha} \\ 
        \nabla_{\bm{\alpha}} D_0(\bm{x}_i(\bm{\alpha})) & = & -\frac{1}{4hN(\sqrt{2 \pi}h)^F} \sum_{j \leq N} e^{-\frac{1}{4h}\|\bm{x}_i(\bm{\alpha})-(\bm{x}^0)_j\|^2_2} \nabla_{\bm{\alpha}} \|\bm{x}_i(\bm{\alpha})-(\bm{x}^0)_j\|^2_2\\
        \nabla_{\bm{\alpha}} \|\bm{x}_i(\bm{\alpha})-(\bm{x}^0)_j\|^2_2 & = & 2\left(\bm{x}_i(\bm{\alpha})-(\bm{x}^0)_j\right) \nabla_{\bm{\alpha}} \bm{x}_i(\bm{\alpha}) \text{ and } \nabla_{\bm{\alpha}} \bm{x}_i(\bm{\alpha}) = \widetilde{Z}^{n_i}\;.
    \end{eqnarray*} 
\end{proof}

\begin{lemma}{\textnormal{Hessian matrix of $G$ with respect to $\bm{\alpha}$.}}\label{lem:hessian_G} Let us denote for any $i,j \leq N$ and $\bm{\alpha} \in \triangle_K$
\begin{itemize}
    \item $u^{ij}_{\bm{\alpha}} \triangleq e^{-\frac{1}{4h}\|\bm{x}_i(\bm{\alpha})-(\bm{x}^0)_j\|^2_2}$ and $U^i_{\bm{\alpha}} \triangleq \sum_{j \leq N} u^{ij}_{\bm{\alpha}} = N(\sqrt{2\pi}h)^F D_0(\bm{x}_i(\bm{\alpha}))$,
    \item $S^i_{\bm{\alpha}} \triangleq \sum_{j \leq N} u^{ij}_{\bm{\alpha}} (\bm{x}_i(\bm{\alpha})-(\bm{x}^0)_j)$ and $T^{i}_{\bm{\alpha}} \triangleq \sum_{j \leq N} u^{ij}_{\bm{\alpha}} (\|\bm{x}_i(\bm{\alpha})-(\bm{x}^0)_j\|^2_2 - 2h)$.
\end{itemize}
Then the coefficient at position $(k, q)$ of Hessian matrix $\nabla^2_{\bm{\alpha}} G(\bm{\alpha}, X) \in \mathbb{R}^{K \times K}$ at $\bm{\alpha}$ and fixed X is
\begin{eqnarray*}
    \frac{\partial^2 G(\bm{\alpha}, X)}{\partial \alpha_k \partial \alpha_q}  & = & \sum_{i \leq N} \left( \frac{T^{i}_{\bm{\alpha}}}{N U^i_{\bm{\alpha}}} - \frac{(S^i_{\bm{\alpha}})^\intercal S^i_{\bm{\alpha}}}{4h^2 (U^i_{\bm{\alpha}})^2} \right)  (\widetilde{\bm{z}}^{n_i})_q^\intercal  (\widetilde{\bm{z}}^{n_i})_k - \eta \mathds{1}(q=k) \;.\\
\end{eqnarray*}
\end{lemma}
\begin{proof}
    According to Lemma~\ref{lem:gradient_G}, for any $k \leq K$
    \[ \frac{\partial G(\bm{\alpha}, X)}{\partial \alpha_k} = - \sum_{i \leq N} \frac{D_0(\bm{x}_i(\bm{\alpha}))^{-1}}{2hN^2(\sqrt{2 \pi}h)^F} (S^i_{\bm{\alpha}})^\intercal (\widetilde{\bm{z}}^{n_i})_k - 2\eta\alpha_k\;. \]
    This implies that
    \begin{eqnarray*}
        \frac{\partial^2 G(\bm{\alpha}, X)}{\partial \alpha_k \partial \alpha_q} + 2\eta \mathds{1}(q=k) & = & \sum_{i \leq N}  \frac{-D_0(\bm{x}_i(\bm{\alpha}))^{-1}}{2hN^2(\sqrt{2 \pi}h)^F} \Big(\left(\frac{\partial S^i_{\bm{\alpha}}}{\partial \alpha_q}\right)^\intercal (\widetilde{\bm{z}}^{n_i})_k\\
        & & -\frac{(S^i_{\bm{\alpha}})^\intercal (\widetilde{\bm{z}}^{n_i})_k}{D_0(\bm{x}_i(\bm{\alpha}))}\frac{\partial D_0(\bm{x}_i(\bm{\alpha}))}{\partial \alpha_q}\Big)
    \end{eqnarray*}
    And then
    \begin{eqnarray*}
        \frac{\partial S^i_{\bm{\alpha}}}{\partial \alpha_q} & =& \frac{-1}{4h} \sum_{j \leq N} e^{-\frac{1}{4h}\|\bm{x}_i(\bm{\alpha})-(\bm{x}^0)_j\|^2_2} \Big( \frac{\partial \|\bm{x}_i(\bm{\alpha})-(\bm{x}^0)_j\|^2_2}{\partial \alpha_q} \Big)^\intercal (\bm{x}_i(\bm{\alpha})-(\bm{x}^0)_j) \\
        &+& \sum_{j \leq N} e^{-\frac{1}{4h}\|\bm{x}_i(\bm{\alpha})-(\bm{x}^0)_j\|^2_2} \frac{\partial \bm{x}_i(\bm{\alpha})}{\partial \alpha_q} \\
        & =& \frac{-1}{2h} \sum_{j \leq N} e^{-\frac{1}{4h}\|\bm{x}_i(\bm{\alpha})-(\bm{x}^0)_j\|^2_2} (\widetilde{\bm{z}}^{n_i})^\intercal_q (\bm{x}_i(\bm{\alpha})-(\bm{x}^0)_j)^\intercal \left(\bm{x}_i(\bm{\alpha})-(\bm{x}^0)_j\right) \\
        &+& \sum_{j \leq N} e^{-\frac{1}{4h}\|\bm{x}_i(\bm{\alpha})-(\bm{x}^0)_j\|^2_2} (\widetilde{\bm{z}}^{n_i})^\intercal_q
    \end{eqnarray*}
    That is,
    \begin{eqnarray*}
       \frac{\partial S^i_{\bm{\alpha}}}{\partial \alpha_q} & =&  (\widetilde{\bm{z}}^{n_i})^\intercal_q \underbrace{\left(\sum_{j \leq N} e^{-\frac{1}{4h}\|\bm{x}_i(\bm{\alpha})-(\bm{x}^0)_j\|^2_2} \Big(1-(2h)^{-1}\|\bm{x}_i(\bm{\alpha})-(\bm{x}^0)_j\|^2_2\Big) \right)}_{=-2hT^i_{\bm{\alpha}}} \\
        \frac{\partial D_0(\bm{x}_i(\bm{\alpha}))}{\partial \alpha_q} & =& -\frac{(S^i_{\bm{\alpha}})^\intercal (\widetilde{\bm{z}}^{n_i})_q}{2hN(\sqrt{2 \pi}h)^F} \text{ according to Lemma~\ref{lem:gradient_G}}\;.
    \end{eqnarray*} 
    Moreover, since $S^i_{\bm{\alpha}}, (\widetilde{\bm{z}}^{n_i})_k \in \mathbb{R}^F$ for any $k \leq K$
    \[ (S^i_{\bm{\alpha}})^\intercal (\widetilde{\bm{z}}^{n_i})_k (S^i_{\bm{\alpha}})^\intercal (\widetilde{\bm{z}}^{n_i})_q = \underbrace{(S^i_{\bm{\alpha}})^\intercal (\widetilde{\bm{z}}^{n_i})_q}_{= (\widetilde{\bm{z}}^{n_i})_q^\intercal S^i_{\bm{\alpha}}} (S^i_{\bm{\alpha}})^\intercal (\widetilde{z}^{n^i})_k =   (\widetilde{\bm{z}}^{n_i})_q^\intercal (S^i_{\bm{\alpha}})^\intercal S^i_{\bm{\alpha}} (\widetilde{\bm{z}}^{n_i})_k \;.\]
\end{proof}

Then, to show that $G$ is (strictly) concave, it is enough to show that the Hessian matrix $\nabla^2_{\bm{\alpha}} G(\bm{\alpha}, X)$ is negative semi-definite (or definite). We assume that $\eta < 4S^2K = 4K$ (using Assumption~\ref{as:ub_norm}), which is the case for most realistic settings.

\begin{proof}
Let us denote $\bm{x}^{ij}_{\bm{\alpha}} \triangleq x^i_{\bm{\alpha}} - (\bm{x}^0)_j$ for any $i,j \leq N$. Then
    \begin{eqnarray}\label{eq:hessian_rewrite}
        \frac{T^{i}_{\bm{\alpha}}}{NU^i_{\bm{\alpha}}} - \frac{(S^i_{\bm{\alpha}})^\intercal S^i_{\bm{\alpha}}}{4h^2 (U^i_{\bm{\alpha}})^2}  & = &  \frac{T^{i}_{\bm{\alpha}}}{NU^i_{\bm{\alpha}}} - \frac{1}{4h^2}\left( \sum_{j,j' \leq N} \frac{u^{ij}_{\bm{\alpha}}}{U^i_{\bm{\alpha}}} \frac{u^{ij'}_{\bm{\alpha}}}{U^i_{\bm{\alpha}}} (\bm{x}^{ij}_{\bm{\alpha}})^\intercal \bm{x}^{ij'}_{\bm{\alpha}} \right) \;.
    \end{eqnarray}
    Now consider $(\bm{x}^{ij}_{\bm{\alpha}})^\intercal \bm{x}^{ij'}_{\bm{\alpha}}  = \langle \bm{x}_i(\bm{\alpha}) - (\bm{x}^0)_j, \ \bm{x}_i(\bm{\alpha}) - (\bm{x}^0)_{j'}\rangle$. From the triangle equality, we have, for all $j,j'\leq N$
    \[  \langle \bm{x}_i(\bm{\alpha}) - (\bm{x}^0)_j, \ \bm{x}_i(\bm{\alpha}) - (\bm{x}^0)_{j'}\rangle = \frac{1}{2}(\|\bm{x}_i(\bm{\alpha}) - (\bm{x}^0)_j\|^2_2 + \|\bm{x}_i(\bm{\alpha}) - (\bm{x}^0)_{j'}\|^2_2 - \| (\bm{x}^0)_j-(\bm{x}^0)_{j'}\|^2_2) \;.\]
    We plug this inequality into Equation~\eqref{eq:hessian_rewrite}
    \begin{align*}
        \frac{T^{i}_{\bm{\alpha}}}{NU^i_{\bm{\alpha}}} - \frac{(S^i_{\bm{\alpha}})^\intercal S^i_{\bm{\alpha}}}{4h^2 (U^i_{\bm{\alpha}})^2}  & =  \frac{T^{i}_{\bm{\alpha}}}{NU^i_{\bm{\alpha}}} + \frac{1}{8h^2} \sum_{j,j' \leq N} \frac{u^{ij}_{\bm{\alpha}}}{U^i_{\bm{\alpha}}} \frac{u^{ij'}_{\bm{\alpha}}}{U^i_{\bm{\alpha}}} \| (\bm{x}^0)_j-(\bm{x}^0)_{j'}\|^2_2 \\
        & - \underbrace{\frac{1}{8h^2}\sum_{j,j'\leq N}\frac{u^{ij}_\alpha}{U^i_\alpha} \frac{u^{ij'}_\alpha}{U^i_\alpha}(\|\bm{x}_i(\bm{\alpha}) - (\bm{x}^0)_j\|^2_2 + \|\bm{x}_i(\bm{\alpha}) - (\bm{x}^0)_{j'}\|^2_2)}_{\geq 0}\\
        & \leq \frac{T^{i}_{\bm{\alpha}}}{NU^i_{\bm{\alpha}}} + \frac{1}{8h^2} \sum_{j,j' \leq N} \frac{u^{ij}_{\bm{\alpha}}}{U^i_{\bm{\alpha}}} \frac{u^{ij'}_{\bm{\alpha}}}{U^i_{\bm{\alpha}}} \| (\bm{x}^0)_j-(\bm{x}^0)_{j'}\|^2_2\;.
    \end{align*}


Obviously $\frac{u^{ij}_{\bm{\alpha}}}{U^i_{\bm{\alpha}}}\leq 1$. Moreover, since Assumption~\ref{as:ub_norm} gives $\|(\bm{x}^0)_j\|^2_2 \leq S$ and $\|\bm{x}_j(\bm{\alpha})\|^2_2 \leq S$  (using Jensen's inequality) for any $j \leq N$,
\begin{align*}
      \frac{T^{i}_{\bm{\alpha}}}{N U^i_{\bm{\alpha}}} - \frac{(S^i_{\bm{\alpha}})^\intercal S^i_{\bm{\alpha}}}{4h^2 (U^i_{\bm{\alpha}})^2}  &\leq \frac{T^{i}_{\bm{\alpha}}}{NU^i_{\bm{\alpha}}} + \frac{2N^2S}{8h^2}=\frac{T^{i}_{\bm{\alpha}}}{NU^i_{\bm{\alpha}}} + \frac{N^2S}{4h^2}\\
     & =\sum_{j\leq N}\frac{u^{ij}_{\bm{\alpha}}}{N U^i_{\bm{\alpha}}}(\|\bm{x}^i(\bm{\alpha})-(\bm{x}^0)_j\|^2_2 - 2h) + \frac{NS}{4h^2}\\
    &\leq \frac{1}{N} \sum_{j\leq N}  \|\bm{x}^i(\bm{\alpha})-(\bm{x}^0)_j\|^2_2 - 2h + \frac{N^2S}{4h^2} \\
    &\leq \frac{1}{N} \sum_{j\leq N}  (\|\bm{x}^i(\bm{\alpha})||^2_2 + \|(\bm{x}^0)_j\|^2_2) - 2h + \frac{N^2S}{4h^2} \\
    & \leq 2S-2h+N^2S(4h^2)^{-1}\;.
\end{align*}

We set $C(h) \triangleq  h^{-2}( -2h^3 + 2Sh^2 + N^2S/4 )$, and fix $\bm{v} \in\mathbb{R}^K$. Then
    \begin{align*}
        \bm{v}^\intercal\nabla^2_{\bm{\alpha}} G(\bm{\alpha}, X)\bm{v} &= -\eta\|\bm{v}\|^2_2 + \sum_{i\leq N}\left( \frac{T^{i}_{\bm{\alpha}}}{NU^i_{\bm{\alpha}}} - \frac{(S^i_{\bm{\alpha}})^\intercal S^i_{\bm{\alpha}}}{4h^2 (U^i_{\bm{\alpha}})^2} \right) (\bm{v}^\intercal(\widetilde{Z}^{n_i})^\intercal \widetilde{Z}^{n_i} \bm{v})\\
        & \leq -\eta\|\bm{v}\|^2_2 + C(h)\sum_{i\leq N}\|\widetilde{Z}^{n_i} \bm{v}\|^2_2
    \end{align*}
     so, using Technical lemma~\ref{lem:norm_Z} (proven below), 
    \begin{equation*}
        \bm{v}^\intercal\nabla^2_{\bm{\alpha}} G(\bm{\alpha}, X)\bm{v} \leq \|\bm{v}\|^2_2(2KSC(h)-\eta) = \|\bm{v}\|^2_2 \times h^{-2}\left(- 4KSh^3 + (4S^2K-\eta) h^2  + \frac{KN^2S^2}{2}\right)\;.
    \end{equation*}
   Then we choose $h> 0$ such that $\bm{v}^\intercal\nabla^2_{\bm{\alpha}} G(\bm{\alpha}, X)\bm{v} < 0$. That is
   \begin{equation}\label{eq:on_h}
        - 4KSh^3 + (4S^2K-\eta) h^2  + \frac{KN^2S^2}{2} < 0 \Leftrightarrow - 2h^3 + \frac{4S^2K-\eta}{2KS} h^2  + \frac{N^2S}{4} < 0 \;.
   \end{equation}
Under the assumption of $\eta < 4S^2K$, this is equivalent to analyzing the following cubic equation
\[ -2h^3 + bh^2 + c = 0 \text{ where } b,c > 0\;.\]

The cubic equation above admits three roots and at least one real root. We show that at least one real root is positive (thus, corresponds to a valid bandwidth). To show that there exists $h > 0$ such that $-2h^3 +b h^2 + c<0$, it is enough to show that there exists $h' \in\mathbb{R}$ such that on $[h', +\inf)$, continuous and infinitely derivable function $x \mapsto -2x^3 + bx^2 + c$ is strictly decreasing. We have $\frac{\text{d} }{\text{d} h}(-2h^3 +b h^2 + c)=-6h^2+2bh$ with roots $0$ and $b/3$, and $\frac{\text{d}^2}{\text{d}^2 h}(-2h^3 +b h^2 + c)=-12h+2b$, and then  $-12 \times 0+2b=2b > 0$ and  $-12 \times \frac{b}{3}+2b=-2b < 0$. The analysis of the behavior of $x \mapsto -2x^3 + bx^2 + c$ then shows that the condition is fulfilled for $h'=b/3 > 0$.

Finally, the value of $h$ can be found through the known closed-form expressions of roots of the rightmost cubic polynomial in $h$ in Equation~\eqref{eq:on_h}.
\end{proof}

\begin{proposition}{\textnormal{The gradient of $G(\cdot, X)$ for any $X \in \mathbb{R}^{N \times F}$ is Lipschitz-continuous.}}\label{lem:gradG_lipschitz} There exists a positive constant 
$H$ such that
\[ \|\nabla_{\bm{\alpha}} G(\bm{\alpha}, X)-\nabla_{\bm{\alpha}} G(\bm{\alpha}', X)\|_2 \leq H \|\bm{\alpha}-\bm{\alpha}'\|_2\;.\] 
\end{proposition}

\begin{proof}
According to Lemma~\ref{lem:gradient_G}, and using notation from Lemma~\ref{lem:hessian_G}, for any $X \in \mathbb{R}^{N \times F}$ and $\bm{\alpha} \in \triangle_K$
\[ \nabla_{\bm{\alpha}} G(\bm{\alpha}, X) =  - \frac{1}{2hN}\sum_{i,j \leq N} \frac{u^{ij}_{\bm{\alpha}} }{U^i_{\bm{\alpha}}} \left(\bm{x}_i(\bm{\alpha})-(\bm{x}^0)_j\right)^\intercal \widetilde{Z}^{n_i}  - 2\eta \bm{\alpha}\;. \]

Then for any $\bm{\alpha}, \bm{\alpha}' \in \triangle_K$
\begin{eqnarray*}
    & & \|\nabla_{\bm{\alpha}} G(\bm{\alpha}, X)-\nabla_{\bm{\alpha}} G(\bm{\alpha}', X)\|_2\\
    & & =  \|- \frac{1}{2hN}\sum_{i,j \leq N} \Big( \frac{u^{ij}_{\bm{\alpha}} }{U^i_{\bm{\alpha}}} \left(\bm{x}_i(\bm{\alpha})-(\bm{x}^0)_j\right) - \frac{u^{ij}_{\bm{\alpha}'} }{U^i_{\bm{\alpha}'}} \left(\bm{x}_i(\bm{\alpha}')-(\bm{x}^0)_j\right) \Big)^\intercal \widetilde{Z}^{n_i}  - 2\eta (\bm{\alpha}-\bm{\alpha}')\|_2\\
    & & \leq \frac{1}{2hN} \sum_{i\leq N} \| \sum_{j \leq N} \Big( \frac{u^{ij}_{\bm{\alpha}} }{U^i_{\bm{\alpha}}} \left(\bm{x}_i(\bm{\alpha})-(\bm{x}^0)_j\right) - \frac{u^{ij}_{\bm{\alpha}'} }{U^i_{\bm{\alpha}'}} \left(\bm{x}_i(\bm{\alpha}')-(\bm{x}^0)_j\right) \Big)^\intercal \widetilde{Z}^{n_i} \|_2 + 2\eta \|\bm{\alpha}-\bm{\alpha}'\|_2\\
    & &\leq    \frac{1}{2hN} \sum_{i \leq N} \| \sum_{j \leq N} \frac{u^{ij}_{\bm{\alpha}} }{U^i_{\bm{\alpha}}} \left(\bm{x}_i(\bm{\alpha})-(\bm{x}^0)_j\right) - \frac{u^{ij}_{\bm{\alpha}'} }{U^i_{\bm{\alpha}'}} \left(\bm{x}_i(\bm{\alpha}')-(\bm{x}^0)_j\right) \|_2   \sqrt{2KS} + 2\eta \|\bm{\alpha}-\bm{\alpha}'\|_2\\
    & & \text{ (using the Cauchy-Schwartz inequality and Technical lemma~\ref{lem:norm_Z}) }\\
    & &=  \frac{\sqrt{KS}}{\sqrt{2}hN} \sum_{i \leq N} \| \sum_{j \leq N} \frac{u^{ij}_{\bm{\alpha}} }{U^i_{\bm{\alpha}}} \bm{x}_i(\bm{\alpha})- \frac{u^{ij}_{\bm{\alpha}'} }{U^i_{\bm{\alpha}'}} \bm{x}_i(\bm{\alpha}') + \Big(\frac{u^{ij}_{\bm{\alpha}'} }{U^i_{\bm{\alpha}'}} -\frac{u^{ij}_{\bm{\alpha}} }{U^i_{\bm{\alpha}}}\Big) (\bm{x}^0)_j \|_2  + 2\eta \|\bm{\alpha}-\bm{\alpha}'\|_2\\
     & & =_{U^i_{\bm{\alpha}}= \sum_{j} u^{ij}_{\bm{\alpha}} } \frac{\sqrt{KS}}{\sqrt{2}hN} \sum_{i \leq N} \| \bm{x}_i(\bm{\alpha})-  \bm{x}_i(\bm{\alpha}') + \sum_{j \leq N}\Big(\frac{u^{ij}_{\bm{\alpha}'} }{U^i_{\bm{\alpha}'}} -\frac{u^{ij}_{\bm{\alpha}} }{U^i_{\bm{\alpha}}}\Big) (\bm{x}^0)_j  \|_2  + 2\eta \|\bm{\alpha}-\bm{\alpha}'\|_2\\
     & & \leq \frac{\sqrt{KS}}{\sqrt{2}hN} \sum_{i \leq N} \big( \sqrt{S} \| \bm{\alpha} - \bm{\alpha}'\|_2 + \|\sum_{j \leq N} \Big(\frac{u^{ij}_{\bm{\alpha}'} }{U^i_{\bm{\alpha}'}} -\frac{u^{ij}_{\bm{\alpha}} }{U^i_{\bm{\alpha}}}\Big) (\bm{x}^0)_j \|_2 \big)  + 2\eta \|\bm{\alpha}-\bm{\alpha}'\|_2\\
     & & \leq  \frac{S\sqrt{K}}{\sqrt{2}hN} \sum_{i \leq N} \big( \| \bm{\alpha} - \bm{\alpha}'\|_2 + \sqrt{2F}\sqrt{\sum_{j \leq N} \Big(\frac{u^{ij}_{\bm{\alpha}'} }{U^i_{\bm{\alpha}'}} -\frac{u^{ij}_{\bm{\alpha}} }{U^i_{\bm{\alpha}}}\Big)^2}  \big)  + 2\eta \|\bm{\alpha}-\bm{\alpha}'\|_2\;,\\
     & & \text{ (using Cauchy-Schwartz, Assumption~\ref{as:ub_norm}, and $\|X^0\|_F \leq \sqrt{2FS}$) }
\end{eqnarray*}

All that remains is to show that $f_i : \bm{\alpha} \in \triangle_K \mapsto (\frac{u^{ij}_{\bm{\alpha}}}{U^i_{\bm{\alpha}}})_{j \leq N} \in \triangle_N$ (which is a bounded space) is Lipschitz-continuous in $\bm{\alpha}$. For starters, if all coordinates of $f_i$ $f_{i,j} : \bm{\alpha} \in \triangle_K \mapsto u^{ij}_{\bm{\alpha}}/U^i_{\bm{\alpha}} \in [0,1]$ are Lipschitz-continuous, each with constant $L_j$, then it is easy to show that $f_i$ is Lipschitz-continuous (always with respect to the $\ell_2$-norm) with constant $L=\sqrt{\sum_{j \leq N} L_j^2}$. Let's consider now any $j \leq N$. For any pair of points $\bm{\alpha}_1,\bm{\alpha}_2 \in \triangle_K$, let's introduce the linear path $\gamma : t \in [0,1] \mapsto t\bm{\alpha}_1 + (1-t)\bm{\alpha}_2$.  $f_{i,j} \circ \gamma$ is well-defined, continuous on the closed space $[0,1]$, differentiable on $(0,1)$, then by the mean-value theorem
\[ f_{i,j}(\bm{\alpha}_1) - f_{i,j}(\bm{\alpha}_2) = f_{i,j} \circ \gamma(1)-f_{i,j} \circ \gamma(0) \leq \sup_{t' \in [0,1]} \underbrace{\nabla_{t}  f_{i,j} \circ \gamma(t')}_{=(\nabla_{\bm{\alpha}} f_{i,j})(\gamma(t'))^\intercal (\bm{\alpha}_1-\bm{\alpha}_2)} (1-0)\;. \]

Meaning that, using the Cauchy-Schwartz inequality
`\[ | f_{i,j}(\bm{\alpha}_1) - f_{i,j}(\bm{\alpha}_2) | \leq \| \sup_{t \in [0,1]} (\nabla_{\bm{\alpha}} f_{i,j})(\gamma(t)) \|_2 \|\bm{\alpha}_1-\bm{\alpha}_2\|_2 \leq \underbrace{\| \sup_{\bm{\alpha} \in \triangle_K} \nabla_{\bm{\alpha}} f_{i,j}(\bm{\alpha}) \|_2}_{=L_j} \|\bm{\alpha}_1-\bm{\alpha}_2\|_2 \;.\]

Let's show that the value of $L_j$ is bounded (we are not interested in finding the tightest value of $L_j$, simply that $L_j < \infty$). Then for any $\bm{\alpha} \in \triangle_K$

\begin{eqnarray*}
    \nabla_{\bm{\alpha}} f_{i,j}(\bm{\alpha}) & = & \left(-\frac{u^{ij}_{\bm{\alpha}}}{2hU^i_{\bm{\alpha}}}(\bm{x}_i(\bm{\alpha})-(\bm{x}^0)_j)-\frac{u^{ij}_{\bm{\alpha}}}{(U^i_{\bm{\alpha}})^2}\sum_{\ell \leq N} \frac{-1}{2h}u^{i\ell}_{\bm{\alpha}}(\bm{x}_i(\bm{\alpha})-(\bm{x}^0)_{\ell})\right)^\intercal \widetilde{Z}^{n_i}\\
    & = & -\frac{u^{ij}_{\bm{\alpha}}}{2hU^i_{\bm{\alpha}}} \left((\bm{x}_i(\bm{\alpha})-(\bm{x}^0)_j)-\sum_{\ell \leq N} \frac{u^{i\ell}_{\bm{\alpha}}}{U^i_{\bm{\alpha}}}(\bm{x}_i(\bm{\alpha})-(\bm{x}^0)_{\ell})\right)^\intercal \widetilde{Z}^{n_i}\;,\\
    & & \text{Using the Cauchy-Schwartz inequality and Technical lemma~\ref{lem:norm_Z}}\\
    \| \nabla_{\bm{\alpha}} f_{i,j}(\bm{\alpha}) \|_2 & \leq & \frac{1}{2h}\underbrace{\left|\frac{u^{ij}_{\bm{\alpha}}}{U^i_{\bm{\alpha}}}\right|}_{\leq 1} \left\|(\bm{x}_i(\bm{\alpha})-(\bm{x}^0)_j)-\sum_{\ell \leq N} \frac{u^{i\ell}_{\bm{\alpha}}}{U^i_{\bm{\alpha}}}(\bm{x}_i(\bm{\alpha})-(\bm{x}^0)_{\ell})\right\|_2 \sqrt{2KS}\\
    && \text{ Using the triangular inequality on the $\ell_2$-norm }\\
    & \leq & \frac{1}{2h}\left(\underbrace{\|\bm{x}_i(\bm{\alpha})-(\bm{x}^0)_j\|_2}_{\leq \sqrt{2S}}+\sum_{\ell \leq N} \underbrace{\left|\frac{u^{i\ell}_{\bm{\alpha}}}{U^i_{\bm{\alpha}}}\right|}_{\leq 1}\underbrace{\|\bm{x}_i(\bm{\alpha})-(\bm{x}^0)_{\ell}\|_2}_{\leq \sqrt{2S}}\right) \sqrt{2KS}\\
    & \leq &  \frac{1}{2h}(N+1)\sqrt{2S}\sqrt{2KS} = \frac{(N+1)}{h}S\sqrt{K}\;.
\end{eqnarray*}

All in all, $f_{i,j}$ is $\frac{(N+1)}{h}S\sqrt{K}$-Lipschitz continuous in $\bm{\alpha}$ and then $f_i$ is $\frac{(N+1)}{h}S\sqrt{NK}$-Lipschitz continuous in $\bm{\alpha}$. Finally

\begin{eqnarray*}
    \|\nabla_{\bm{\alpha}} G(\bm{\alpha}, X)-\nabla_{\bm{\alpha}} G(\bm{\alpha}', X)\|_2 & \leq & \frac{S\sqrt{K}}{\sqrt{2}hN} \sum_{i \leq N} \big( \| \bm{\alpha} - \bm{\alpha}'\|_2 + \sqrt{2F} \| f_i(\bm{\alpha}')-f_i(\bm{\alpha}) \|_2  \big)\\
    & & + 2\eta \|\bm{\alpha}-\bm{\alpha}'\|_2\\
    &\leq& \frac{S\sqrt{K}}{\sqrt{2}hN} \sum_{i \leq N} \big( 1 + \sqrt{2F} \frac{(N+1)}{h}S\sqrt{NK}  \big)\| \bm{\alpha} - \bm{\alpha}'\|_2\\
    && + 2\eta \|\bm{\alpha}-\bm{\alpha}'\|_2\\
    & \leq& \underbrace{\left( \frac{S\sqrt{K}}{\sqrt{2}h}  \big( 1 + \sqrt{2F} \frac{(N+1)}{h}S\sqrt{K}  \big)  + 2\eta \right)}_{\triangleq H} \|\bm{\alpha}-\bm{\alpha}'\|_2
\end{eqnarray*}

Then $\nabla_{\bm{\alpha}} G(\bm{\alpha}, X)$ is $H$-Lipschitz continuous in $\bm{\alpha}$ with respect to the $\ell_2$-norm.
\end{proof}
\color{black}


\section{Bounds on the mean squared error}\label{subapp:mse}

We recall that the loss function associated with the mean squared error (MSE) is defined as the average of the MSE between each true sample and its corresponding imputed sample at iteration $t$ (Definition~\ref{def:mse})
\[ \mathcal{L}^\text{MSE}(X^t, X^\star) \triangleq \frac{1}{N} \sum_{i \leq N} \text{MSE}((\bm{x}^t)_i, (\bm{x}^\star)_i) = \frac{1}{N} \sum_{i \leq N} \frac{1}{F} \sum_{f \leq F} ((x^t)^f_i-(x^\star)^f_i)^2\;.\]

\begin{theorem}{\textnormal{Bounds in high probability and in expectation on the MSE for F3I (Theorem~\ref{thm:mse_bounds}).}} Under Assumptions~\ref{as:x_distribution}-\ref{as:ub_norm}, if $X^t$ is any imputed matrix at iteration $t \geq 1$ (after the initial imputation step), $X^\star$ is the corresponding full (unavailable in practice) matrix
\[ \mathcal{L}^\text{MSE}(X^t, X^\star) \leq C^\text{miss}/F \quad \text{ with high probability $1-1/N$}, \text{ where }C^\text{miss} = \mathcal{O}((\sigma^\text{miss})^2F+\ln N) \;, \]
and $\sigma^\text{miss}$ is linked to the variance of the data distribution and depends on the missingness mechanism.
\end{theorem}

\begin{proof}
First, we denote $\mathcal{K}(\bm{x}, X^0, k)$ the index of the $k^\text{th}$ nearest neighbor to vector $\bm{x}$ among the rows of $X^0$, that is, $\{(\bm{x}^0)_1, (\bm{x}^0)_2, \dots, (\bm{x}^0)_N\}$. The selection of neighbors does not depend on $f$ after the initial imputation step at $t=0$. We recall that for any step $t \geq 1$, $(x^t)^f_i \triangleq (x^\star)^f_i$ if $m^f_i = 0$, $\sum_{k \leq K} \alpha^t_k (x^0)^f_{\mathcal{K}((\bm{x}^{t-1})_i, X^0, k)}$ otherwise. Then, for any $i \leq N$
\begin{eqnarray*}
    \text{MSE}(\bm{x}^t_i, \bm{x}^\star_i) & = & \frac{1}{F} \sum_{f \leq F} \Big(\sum_{k \leq K} \alpha^t_k (x^0)^f_{\mathcal{K}(x^{t-1}_i, X^0, k)}-(x^\star)^f_i \Big)^2\\
    & \leq & \frac{1}{F} \sum_{k \leq K} \alpha^t_k \sum_{f \leq F} \Big((x^0)^f_{\mathcal{K}(x^{t-1}_i, X^0, k)}-(x^\star)^f_i \Big)^2\\
    & & \text{ (using Jensen's inequality on convex function $x \mapsto x^2$ and $\sum_{k \leq K} \alpha^t_k = 1$)}\\
    & \leq & \frac{1}{F} \sum_{k \leq K} \alpha^t_k \| (\bm{x}^0)_{\mathcal{K}(x^{t-1}_i, X^0, k)} - (\bm{x}^\star)_i \|^2_2\;.
\end{eqnarray*}

Applying Corollary~\ref{cor:concentration_x0_xstar_full2} (proven below) with $\delta=1/N$ and using $ \sum_{k \leq K} \alpha^t_k = 1$ yields
\[ \forall i \leq N, \quad \text{MSE}(\bm{x}^t_i, \bm{x}^\star_i) \leq  \frac{1}{F} \times 1 \times C^\text{miss}_{1/N^3} \quad 
\text{w.p.} \quad 1-1/N\;. \]

Then we conclude by noticing that $\mathcal{L}^\text{MSE}(X^t, X^\star) \leq \frac{N}{N}\frac{1}{F}C^\text{miss}_{1/N^3} \quad $ 
w.p. $\quad 1-1/N$.
\end{proof}

\section{Regret analysis of F3I}\label{subapp:guarantees_F3I}

\begin{theorem}{\textnormal{High-probability upper bound on the imputation quality for F3I (Theorem~\ref{thm:regret_f3i}).}} Under Assumptions~\ref{as:x_distribution}-\ref{as:ub_norm}, for any initial matrix $X \in (\mathbb{R}\cup\{\texttt{N/A}\})^{N \times F}$,
   \begin{eqnarray*}
     \max_{ \bm{\alpha} \in \triangle_K} \sum_{s=1}^t G_\star(\bm{\alpha}, X^{s-1})-G_\star(\bm{\alpha}^s, X^{s-1}) & \leq & C^\text{AH}_G\sqrt{t} + H^\text{miss} h^{-1}t \;,
\end{eqnarray*} 
 with probability $1-1/N$, where $H^\text{miss} = \mathcal{O}(F+\ln N)$ is another value which depends on the missingness mechanism 
 and $h$ is chosen to guarantee that $G$ is concave in its first argument (Proposition~\ref{lem:G_strictly_concave}). $C^\text{AH}_G = \mathcal{O}(\sqrt{\log(K)})$ is the constant associated with the regret bound on the gain in F3I incurred by AdaHedge.
\end{theorem}

\begin{proof}
We set the gain in F3I to $g^s(\bm{\alpha}) \triangleq \sum_{k \leq K} \alpha_k \frac{\partial G}{\partial \alpha_k}(\bm{\alpha}^s, X^{s-1})$ for $s \leq t$ and $\bm{\alpha} \in \triangle_K$. We use the ``gradient trick'' to transfer the regret bound from a linear loss function to the convex loss $-G(\cdot, X^{t-1})$ for any $t\geq 1$, for all $\bm{\alpha}, \bm{\alpha}^s \in \triangle_K$, $s \leq t$,
\[ \sum_{s=1}^t G(\bm{\alpha}, X^{s-1})-G(\bm{\alpha}^s, X^{s-1}) \leq \sum_{s=1}^t (\bm{\alpha}-\bm{\alpha}^s)^\intercal \nabla_{\bm{\alpha}} G(\bm{\alpha}^s, X^{s-1})\;.\]
and note that for any $\bm{\alpha} \in \triangle_K$,
\begin{align*}
    \sum_{s=1}^t g^s(\bm{\alpha})-g^s(\bm{\alpha}^s) & =  \sum_{s=1}^t \bm{\alpha}^\intercal \nabla_{\bm{\alpha}} G(\bm{\alpha}^s, X^{s-1}) - (\bm{\alpha}^s)^\intercal \nabla_{\bm{\alpha}} G(\bm{\alpha}^s, X^{s-1})\\
    & = \sum_{s=1}^t (\bm{\alpha}- \bm{\alpha}^s)^\intercal \nabla_{\bm{\alpha}} G(\bm{\alpha}^s, X^{s-1})\:.
\end{align*}
Then applying the regret bound of AdaHedge (Technical lemma~\ref{lem:regret_adahedge}, proven below) to that gain yields at time $t > 1$ (rightmost term) and the gradient trick on the function $G$ which is concave in its first argument (leftmost term) with Proposition~\ref{lem:G_strictly_concave}
\begin{eqnarray}\label{eq:diff_G}
    \forall \bm{\alpha} \in \triangle_K, \ \sum_{s=1}^t  G(\bm{\alpha}, X^{s-1})-G(\bm{\alpha}^s, X^{s-1}) \leq 2 \delta_t \sqrt{t \log(K) } + 16\delta_t \left(2+ \frac{\log K}{3}\right)\;,
\end{eqnarray}
where $\delta_t \triangleq \max_{s \leq t} \left(\max_{k \leq K} \frac{\partial G}{\partial \alpha_k}(\bm{\alpha}^s, X^{s-1}) - \min_{q \leq K} \frac{\partial G}{\partial \alpha_q}(\bm{\alpha}^s, X^{s-1})\right)$. Now we go from $G$ to $G_\star$ point-wise. Corollary~\ref{cor:concentration_x0_xstar_full2} with $\delta = 1/N$ states that under Assumptions~\ref{as:x_distribution}-\ref{as:ub_norm}, there exists $C^\text{miss}_{1/N^3}=\mathcal{O}(F+\ln N)$ such that for any $i \leq N$, $\|(\bm{x}^0)_i-(\bm{x}^\star)_i\|^2_2 \leq C^\text{miss}_{1/N^3}$ with probability $1-1/N$. By triangle inequality, 
\[ \forall \bm{x} \in \mathbb{R}^F \ \forall i \leq N, \ \|\bm{x}-(\bm{x}^\star)_i\|^2_2 - \|\bm{x}-(\bm{x}^0)_i\|^2_2 \leq \|(\bm{x}^0)_i - (\bm{x}^\star)_i\|^2_2 \leq C^\text{miss}_{1/N^3} \text{ w.p. } 1-1/N\;. \]
Then, with probability $1-1/N$,
\begin{eqnarray*}
    \forall \bm{x} \in \mathbb{R}^F, &\|\bm{x}-(\bm{x}^\star)_i\|^2_2 &\leq \|\bm{x}-(\bm{x}^0)_i\|^2_2+C^\text{miss}_{1/N^3}\\
    \implies &-\frac{1}{4h}\|\bm{x}-(\bm{x}^\star)_i\|^2_2 &\geq -\frac{1}{4h}\|\bm{x}-(\bm{x}^0)_i\|^2_2-\frac{C^\text{miss}_{1/N^3}}{4h}\\
    \implies &e^{-\frac{1}{4h}\|\bm{x}-(\bm{x}^\star)_i\|^2_2} &\geq e^{-\frac{C^\text{miss}_{1/N^3}}{4h}}e^{-\frac{1}{4h}\|\bm{x}-(\bm{x}^0)_i\|^2_2}\\
    \implies &\sum_{i\leq N}e^{-\frac{1}{4h}\|\bm{x}-(\bm{x}^\star)_i\|^2_2} &\geq e^{-\frac{C^\text{miss}_{1/N^3}}{4h}}\sum_{i\leq N}e^{-\frac{1}{4h}\|\bm{x}-(\bm{x}^0)_i\|^2_2}\\
    \implies&\log\left(\frac{\sum_{i\leq N}e^{-\frac{1}{4h}\|\bm{x}-(\bm{x}^0)_i\|^2_2}}{\sum_{i\leq N}e^{-\frac{1}{4h}\|\bm{x}-(\bm{x}^\star)_i\|^2_2}}\right) = \log\left(\frac{D_0(\bm{x})}{D_\star(\bm{x})}\right) &\leq \frac{C^\text{miss}_{1/N^3}}{4h}\;.
\end{eqnarray*}
Symmetrically (by switching the roles of $(\bm{x}^\star)_i$ and $(\bm{x}^0)_i$ in the previous inequalities), we obtain with probability $1-1/N$
\begin{eqnarray*}
    \forall \bm{x} \in \mathbb{R}^F, \ \log\left(\frac{D_\star(\bm{x})}{D_0(\bm{x})}\right) & = & -\log\left(\frac{D_0(\bm{x})}{D_\star(\bm{x})}\right) \leq C^\text{miss}_{1/N^3}/(4h)\\
    \implies \Big|\log\left(\frac{D_0(\bm{x})}{D_\star(\bm{x})}\right)\Big| & \leq & C^\text{miss}_{1/N^3}/(4h) \;.
\end{eqnarray*}

That is, for any $\bm{\alpha} \in \triangle_K$ and $X \in \mathbb{R}^{N \times F}$, with probability $1-1/N$
\begin{align}\label{eq:G_Gstar}
   | (G-G_\star)(\bm{\alpha}, X) | &=\frac{1}{N}\sum_{i\leq N}\log\left(\frac{D_0(\texttt{Impute}(\bm{x}_i; \bm{\alpha}))}{D_\star(\texttt{Impute}(\bm{x}_i; \bm{\alpha}))}\right)-\log\left(\frac{D_0(\bm{x}_i)}{D_\star(\bm{x}_i)}\right) \leq \frac{C^\text{miss}_{1/N^3}}{2h}\;.
\end{align}

Finally, we combine Equations~\eqref{eq:diff_G}-\eqref{eq:G_Gstar} to obtain for any $\bm{\alpha} \in \triangle_K$, with probability $1-1/N$
\[ \sum_{s=1}^t  G_\star(\bm{\alpha}, X^{s-1})-G_\star(\bm{\alpha}^s, X^{s-1}) \leq \frac{C^\text{miss}_{1/N^3}}{h}t + \underbrace{2 \delta_t \sqrt{t \log(K) } + 16\delta_t \left(2+ \frac{\log K}{3}\right)}_{=C^\text{AH}_G \sqrt{t}}\;. \]



\end{proof}

\section{Joint training on a downstream task}\label{subapp:joint_training}

\begin{lemma}{\textnormal{Any loss $\ell$ with a Lipschitz continuous gradient allows the use of PCGrad~\cite{yu2020gradient} combined with F3I.}}\label{lem:pcagrad_loss} If $\nabla \ell$ is $L$-Lipschitz continuous with a finite $L>0$ with respect to its single argument, then for any matrix $X \in \mathbb{R}^{N \times F}$, $\bm{\alpha} \mapsto \nabla_{\bm{\alpha}} \Big( (1-\beta) G(\bm{\alpha}, X) - \frac{\beta}{N} \sum_{i \leq N} \ell(\texttt{Impute}(\bm{x}^i, \bm{\alpha})) \Big)$ is also Lipschitz continuous with a positive finite constant.
\end{lemma}

\begin{proof}
    Note that Proposition~\ref{lem:gradG_lipschitz} establishes that the gradient of $G$ with respect to $\bm{\alpha}$ is $H$-Lipschitz continuous with $H >0$. Then for all $\bm{\alpha}, \bm{\alpha}' \in \triangle_K$
\begin{eqnarray*}
     & & \left\| \nabla_{\bm{\alpha}} \Big((1-\beta)G(\bm{\alpha}, X)+\frac{\beta}{N}\sum_{i \leq N} \ell(\bm{x}_i(\bm{\alpha})) - \bigg( (1-\beta)G(\bm{\alpha}', X)+\frac{\beta}{N}\sum_{i \leq N} \ell(\bm{x}_i(\bm{\alpha}')) \bigg) \Big) \right\|_2\\
    & \leq & (1-\beta)\left\|  \nabla_{\bm{\alpha}} G(\bm{\alpha}, X) - \nabla_{\bm{\alpha}} G(\bm{\alpha}', X)\right\|_2 + \frac{\beta}{N}\sum_{i \leq N} \left\| \nabla_{\bm{\alpha}} \ell(\bm{x}_i(\bm{\alpha})) - \nabla_{\bm{\alpha}} \ell(\bm{x}_i(\bm{\alpha}')) \right\|_2\\
    & \leq & H(1-\beta)\|\bm{\alpha} - \bm{\alpha}'\|_2 + \frac{L\beta}{N} \sum_{i \leq N} \|\bm{x}_i(\bm{\alpha})- \bm{x}_i(\bm{\alpha}')\|_2\\
    & \leq & (H(1-\beta)-L\beta\sqrt{KS})\|\bm{\alpha} - \bm{\alpha}'\|_2 \;.
\end{eqnarray*}

The last step holds because of the fact that, for any $i \leq N$, if $\mathcal{K}(\bm{x}_i, X^0, k)$ is the index of the $k^\text{th}$ nearest neighbor of $\bm{x}_i$ among $\{(\bm{x}^0)_1, \dots, (\bm{x}^0)_N\} \subseteq \mathbb{R}^F$ and $\bm{x}_i^{\mathcal{M}_i}$ is the vector restricted to columns $f$ such that $x^f_i$ is missing
\begin{eqnarray*}
    \|\bm{x}_i(\bm{\alpha})- \bm{x}_i(\bm{\alpha}')\|^2_2  & = & \Big\|\sum_{k \leq K} (\alpha_k-\alpha'_k)(\bm{x}^0)^{\mathcal{M}_i}_{\mathcal{K}(\bm{x}_i,X^0, k)}\Big\|^2_2\\
    & = & \sum_{f \in \mathcal{M}_i} \Big( \sum_{k \leq K} (\alpha_k-\alpha'_k)(x^0)^f_{\mathcal{K}(x_i,X^0, k)} \Big)^2\\
    & = &  \sum_{f \in \mathcal{M}_i} \langle \bm{\alpha}-\bm{\alpha}', [(x^0)^{f}_{\mathcal{K}(\bm{x}_i,X^0, 1)}, ..., (x^0)^{f}_{\mathcal{K}(\bm{x}_i,X^0, K)}]^\intercal \rangle^2\\
    & \leq & \| \bm{\alpha} - \bm{\alpha}' \|^2_2 \sum_{f \in \mathcal{M}_i}  \| [(x^0)^{f}_{\mathcal{K}(\bm{x}_i,X^0, 1)}, ..., (x^0)^{f}_{\mathcal{K}(\bm{x}_i,X^0, K)}]^\intercal \|^2_2\\ 
    & = & \| \bm{\alpha} - \bm{\alpha}' \|^2_2 \sum_{f \in \mathcal{M}_i}  \sum_{k \leq K} ((x^0)^{f}_{\mathcal{K}(\bm{x}_i,X^0, k)})^2\\ 
    &  \leq & \| \bm{\alpha} - \bm{\alpha}' \|^2_2  \sum_{k \leq K} \sum_{f \leq F}  ((x^0)^{f}_{\mathcal{K}(\bm{x}_i,X^0, k)})^2\\
    & = &  \| \bm{\alpha} - \bm{\alpha}' \|^2_2  \sum_{k \leq K} \| (\bm{x}^0)_{\mathcal{K}(\bm{x}_i,X^0, k)} \|^2_2\\
    & \leq & \| \bm{\alpha} - \bm{\alpha}' \|^2_2 KS \text{ using Assumption~\ref{as:ub_norm} } \;.
\end{eqnarray*}
The first inequality is obtained by applying the Cauchy-Schwartz inequality $|\mathcal{M}_i|$ times, since the selection of neighbors does not depend on $\bm{\alpha}$. Note that $(x_i(\bm{\alpha}))^f = (x_i(\bm{\alpha}'))^f$ for any $f \not\in \mathcal{M}_i$.
\end{proof}

\begin{example}{\textnormal{A simple example of a convex loss function $\ell$ with a Lipschitz-continuous gradient function.}}\label{lem:pcgrad_example}
    The pointwise log loss $\ell(\bm{x})=-y\log C_{\bm{\omega}}(\bm{x})$ for the binary classification task is convex and such that $\nabla_{\bm{x}} \ell$ is Lipschitz continuous, where $y$ is the true class in $\{0,1\}$ for sample $\bm{x}$ and $C_{\bm{\omega}} : \bm{x} \mapsto 1/(1+\exp(-\bm{\omega}^\intercal \bm{x}))$ is the sigmoid function of parameter $\bm{\omega}$.
\end{example}

\begin{proof}
    $\ell$ is continuous and twice differentiable on $\mathbb{R}^F$. Knowing that $\nabla_{\bm{x}} C_{\bm{\omega}}(\bm{x}) = C_{\bm{\omega}}(\bm{x})(1-C_{\bm{\omega}}(\bm{x}))\bm{\omega}^\intercal$, the Hessian matrix of $\ell$ in its single argument is
    \[ \forall \bm{x} \in \mathbb{R}^F \ \forall y \in \{0,1\}, \ \nabla^2_{\bm{x}} \ell(\bm{x}) = yC_{\bm{\omega}}(\bm{x})(1-C_{\bm{\omega}}(\bm{x}))\bm{\omega}\bm{\omega}^\intercal\;.\]

In particular, it is easy to see that $\ell$ is convex, because for any $\bm{v} \in \mathbb{R}^F$,
\[ \bm{v}^\intercal \nabla^2_{\bm{x}} \ell(\bm{x}) \bm{v} = \underbrace{yC_{\bm{\omega}}(\bm{x})(1-C_{\bm{\omega}}(\bm{x}))}_{\geq 0}(\bm{v}^\intercal \bm{\omega})^2 \geq 0\;. \]
Then for any $\bm{x} \in \mathbb{R}^F$ and $y \in \{0,1\}$,
    \begin{eqnarray*}
        \| \nabla^2_{\bm{x}} \ell(\bm{x})\|^2_{F} & =  & \underbrace{yC_{\bm{\omega}}(\bm{x})(1-C_{\bm{\omega}}(\bm{x}))}_{\leq 1 \times 1/4} \sum_{f,f' \leq F} (\omega^f)^2 \leq \frac{1}{4}\|\bm{\omega}\|^2_F\;.
    \end{eqnarray*}
    
Similarly to the proof of Proposition~\ref{lem:gradG_lipschitz}, proving that $\nabla_{\bm{x}} \ell$ is Lipschitz continuous in each of its $F$ coordinates will be enough to prove that $\nabla_{\bm{x}} \ell$ is Lipschitz continuous as well. For any pair of points $\bm{x}_1, \bm{x}_2 \in \mathbb{R}^F$, we introduce the linear path $\gamma' : t \in [0,1] \mapsto t\bm{x}_1 + (1-t)\bm{x}_2$. For any $f \leq F$, $t \in [0,1] \mapsto \big(\nabla_{\bm{x}} \ell(\gamma'(t))\big)^f \in \mathbb{R}$ is well-defined, continuous on the closed space $[0,1]$, differentiable on $(0,1)$. Then applying the mean value theorem to this function yields
\begin{align*}
   |\big(\nabla_{\bm{x}} \ell(\bm{x}_1)\big)^f-\big(\nabla_{\bm{x}} \ell(\bm{x}_2)\big)^f| & \leq \| \sup_{\bm{x} \in \mathbb{R}^F}  \big( \nabla^2_{\bm{x}} \ell(\bm{x}) \big)^f \|_2 \|\bm{x}_1-\bm{x}_2\|_2  \leq \frac{\|\bm{\omega}\|_2}{2} \|\bm{x}_1-\bm{x}_2\|_2\;. 
\end{align*}
Then $\nabla_{\bm{x}} \ell$ is Lipschitz-continuous with constant $\sqrt{\sum_{f \leq F} \frac{1}{4}\|\bm{\omega}\|^2_2} = \frac{\|\bm{\omega}\|^2_F}{2} > 0$.
\end{proof}

\begin{theorem}{\textnormal{High-probability upper bound on the joint imputation-downstream task performance (Theorem~\ref{thm:regret_f3i_joint}).}}Under Assumptions~\ref{as:x_distribution}-\ref{as:ub_norm}, for any initial matrix $X \in (\mathbb{R}\cup\{\texttt{N/A}\})^{N \times F}$, convex pointwise loss $\ell$ such that $\nabla \ell$ is Lipschitz-continuous, and $\beta \in [0,1]$, under the conditions mentioned in Theorem $2$ from~\cite{yu2020gradient}
   \begin{align*}
     & \max_{ \bm{\alpha} \in \triangle_K} \sum_{s=1}^t (1-\beta)\Big(G_\star(\bm{\alpha}, X^{s-1})-G_\star(\bm{\alpha}^s, X^{s-1})\Big) - \frac{\beta}{N}\sum_{i \leq N} \Big( \ell((\bm{x}^{s-1})_i(\bm{\alpha}))-\ell((\bm{x}^{s-1})_i(\bm{\alpha}^s)) \Big)\\
     & \leq C^\text{AH}_{(G,\ell)}\sqrt{t} + (1-\beta)H^\text{miss}h^{-1}t \;,
\end{align*} 
 with probability $1-1/N \in (0,1)$, where $H^\text{miss} = \mathcal{O}(F+\ln N)$ depends on the missingness mechanism 
 and $ C^\text{AH}_{(G,\ell)}$ is the constant related to AdaHedge being applied with gains $\overline{g}_s(\cdot)$.
\end{theorem}

\begin{proof}
    Similarly to the proof of Theorem~\ref{thm:regret_f3i}, the application of the AdaHedge regret bound (Technical lemma~\ref{lem:regret_adahedge}), and the gradient trick on the concave function $(1-\beta)G(\cdot, X)+\beta \ell(\cdot)$
    \begin{align*}
     & \sum_{s=1}^t (1-\beta)\Big(G(\bm{\alpha}^\text{PC}, X^{s-1})-G((\bm{\alpha}^s)^\text{PC}, X^{s-1})\Big) + \frac{\beta}{N}\sum_{i \leq N} \Big( \ell((\bm{x}^{s-1})_i(\bm{\alpha}^s)^\text{PC})-\ell((\bm{x}^{s-1})_i(\bm{\alpha}^\text{PC})) \Big)\\
     & \leq C^\text{AH}_{(G,\ell)}\sqrt{t} \;,
\end{align*} 
where $\bm{\alpha}^\text{PC}$ and $(\bm{\alpha}^s)^\text{PC}$ are the parameters updated with PCGrad~\cite{yu2020gradient}. Assuming the three conditions in Theorem $2$ from~\cite{yu2020gradient} are all satisfied, which only depend on functions $G$ and $\ell$, then for $\bm{\theta} \in \{\bm{\alpha}, \bm{\alpha}^s\}$
\[ (1-\beta)G(\bm{\theta}, X^{s-1})- \frac{\beta}{N}\sum_{i \leq N} \ell((\bm{x}^{s-1})_i(\bm{\theta})) \leq (1-\beta)G(\bm{\theta}^\text{PC}, X^{s-1})- \frac{\beta}{N}\sum_{i \leq N} \ell((\bm{x}^{s-1})_i(\bm{\theta}^\text{PC}))  \;.\]
Finally, we apply the pointwise approximation in high probability of $G_\star$ by $G$ (Corollary~\ref{cor:concentration_x0_xstar_full2} for $\delta=1/N$) that yields for any $\bm{\alpha} \in \triangle_K$
     \begin{align*}
     & \sum_{s=1}^t (1-\beta)\Big(G_\star(\bm{\alpha}, X^{s-1})-G_\star(\bm{\alpha}^s, X^{s-1})\Big) + \frac{\beta}{N}\sum_{i \leq N} \Big( \ell((\bm{x}^{s-1})_i(\bm{\alpha}^s))-\ell((\bm{x}^{s-1})_i(\bm{\alpha})) \Big)\\
     & \leq C^\text{AH}_{(G,\ell)}\sqrt{t} + (1-\beta)C^\text{miss}_{1/N^3}h^{-1}t \;.
\end{align*} 
\end{proof}

To implement PCGrad-F3I, we also need to compute $\nabla_{\bm{\alpha}} \ell((\bm{x}^{s-1})_i(\bm{\alpha}^s))$ at each iteration $s$ for each point $i$. By the chain rule, 
\[ \nabla_{\bm{\alpha}} \ell((\bm{x}^{s-1})_i(\bm{\alpha}^s)) = \nabla_{\bm{x}} \ell(\bm{x})_{\mid \bm{x} = (\bm{x}^{s-1})_i(\bm{\alpha}^s)} \nabla_{\bm{\alpha}} (\bm{x}^{s-1})_i(\bm{\alpha})_{\mid \bm{\alpha}=\bm{\alpha}^s} \;,\]

In particular, we give the gradient at any $\bm{\alpha}$ and $i \leq N$ for the log-loss with sigmoid classifier below
\begin{lemma}{\textnormal{Gradient $\nabla_{\bm{\alpha}} \ell((\bm{x}^{s-1})_i(\bm{\alpha}))$ for Example~\ref{lem:pcgrad_example}.}} The gradient at any $\bm{\alpha}$ for the log-loss $\ell$ with sigmoid classifier $C_{\bm{\omega}}$ where the true class for sample $\bm{x} \in \mathbb{R}^F$ is $y \in \{0,1\}$ is
    \[ \nabla_{\bm{\alpha}} \ell(\bm{x}(\bm{\alpha})) = -y (1-C_{\bm{\omega}}(\bm{x}(\bm{\alpha})))\bm{\omega}^\intercal \widetilde{Z}^{n_i}_s\;,\]
where $\widetilde{Z}^{n_i}_s \in \mathbb{R}^{F \times K}$ is the matrix which $k^\text{th}$ column is defined as $(\widetilde{z}^{n_i}_s)^k_f = 0$ if $m^f_i=0$, and otherwise, $(\widetilde{z}^{n_i}_s)^k_f$ is the value of the feature $f$ for the $k^\text{th}$ closest neighbor to $(\bm{x}^{s-1})_i$ among rows $(\bm{x}^0)_2,\dots,(\bm{x}^0)_N\}$  of $X^0$ $\{(\bm{x}^0)_1,(\bm{x}^0)_2,\dots,(\bm{x}^0)_N\}$ (see Lemma~\ref{lem:gradient_G}). 
\end{lemma}

\section{Technical lemmas}\label{subapp:technical_lemmas}

We consider below for any $i \leq N$ the matrix $\widetilde{Z}^{n_i} \in \mathbb{R}^{F \times K}$ where the $k^\text{th}$ column of $\widetilde{Z}^{n_i}$ is defined as $(\widetilde{\bm{z}}^{n_i})^k_f = 0$ if $m^f_i=0$, otherwise $(\widetilde{z}^{n_i})^k_f$ is the value of the feature $f$ for the $k^\text{th}$ closest neighbor to $(\bm{x}^{s-1})_i$ among rows of $X^0$ $\{(\bm{x}^0)_1,(\bm{x}^0)_2,\dots,(\bm{x}^0)_N\}$. That is, $(\widetilde{\bm{z}}^{n_i})^k$ is equal to the $k^\text{th}$ closest neighbor of $\bm{x}_i$ (by increasing order of distance) on missing coordinates of $\bm{x}_i$, and equal to zero otherwise. To upper-bound norms involving matrix $\widetilde{Z}^{n_i}$, we use the following lemma
\begin{technicallemma}{\textnormal{Upper bound on $\ell_2$ norms on $\widetilde{Z}^{n_i}$.}}\label{lem:norm_Z} For any $i \leq N$ and any vectors $\bm{v} \in \mathbb{R}^{K}$ and $\bm{u} \in \mathbb{R}^F$, \[\|\widetilde{Z}^{n_i}\bm{v}\|_2 \leq \sqrt{2KS}\|\bm{v}\|_2 \text{ and } \|\bm{u}^\intercal \widetilde{Z}^{n_i}\|_2 \leq \sqrt{2KS}\|\bm{u}\|_2\;.\]
\end{technicallemma}
\begin{proof}
    Using the Cauchy-Schwartz inequality applied respectively $F$ and $K$ times and if $\|M\|_F = \sqrt{\sum_{i} \sum_{j} |m^j_i|^2} = \sqrt{\sum_{i} \|\bm{m}_i\|^2_2} = \sqrt{\sum_{j} \|\bm{m}^j\|^2_2}$ is the Frobenius matrix norm of matrix $M$, then $\|\widetilde{Z}^{n_i}\|^2_F \leq 2KS$ and
    \begin{eqnarray*}
        \|\widetilde{Z}^{n_i}\bm{v}\|^2_2 & = &  \sum_{f \leq F} \langle (\widetilde{Z}^{n_i})_f^\intercal, \bm{v} \rangle ^2  \leq  \underbrace{\sum_{f \leq F} \|(\widetilde{Z}^{n_i})_f^\intercal\|^2_2}_{=_\text{def} \|\widetilde{Z}^{n_i}\|^2_F} \|\bm{v}\|^2_2  \leq  2KS\|\bm{v}\|^2_2 \;.\\
        \|\bm{u}^\intercal \widetilde{Z}^{n_i}\|^2_2 & = &  \sum_{k \leq K} \langle \bm{u}, (\widetilde{Z}^{n_i})^k \rangle ^2  \leq  \|\bm{u}\|^2_2 \underbrace{\sum_{k \leq K}  \|(\widetilde{Z}^{n_i})^k\|^2_2}_{=_\text{def} \|\widetilde{Z}^{n_i}\|^2_F}  \leq  2KS\|\bm{u}\|^2_2 \;.
    \end{eqnarray*}
\end{proof}

\color{black}

\begin{technicallemma}{\textnormal{Regret of AdaHedge.}}\label{lem:regret_adahedge} On the online learning problem with $K$ elements, using gains $\bm{\alpha} \mapsto g^s(\bm{\alpha}) \triangleq \sum_{k \leq K} \alpha_k U_k$ for $s \leq t$, and denoting $ \delta_t \triangleq \max_{s \leq t} \left( \max_{k \leq K} U_k - \min_{q \leq K} U_q \right) $, the regret at time $t > 1$ incurred by AdaHedge with predictions $(\bm{\alpha}^s)_{s \leq t}$ is
\[\max_{\bm{\alpha} \in \triangle_K}  \sum_{s=1}^t g^s(\bm{\alpha})-g^s(\bm{\alpha}^s) \leq 2 \delta_t \sqrt{t \log(K) } + 16\delta_t (2+ \log(K)/3)\;.\] 
\end{technicallemma}

\begin{proof}
    This statement stems directly from Theorem $8$ and Corollary $17$ in~\cite{de2014follow} applied to the loss $\ell^s = -g^s$, and using the fact that $\alpha_k \leq 1$ for any $k \leq K$.
\end{proof}

In several proofs, we need an upper bound on $ \|(\bm{x}^0)_j - (\bm{x}^\star)_i\|^2_2$ for any pair $i,j \leq N$, where $X^0$ is the initially K-nearest neighbor-imputed matrix and $X^\star$ is the corresponding full matrix. This is our most important lemma for analyzing F3I. This result still holds for any missingness mechanism such that the random variable $(x^0)^f_i-(x^\star)^f_i$ is a zero-mean subgaussian for any $i \leq N$ and $f \leq F$, independent across \emph{features}. We show below that this statement includes all three mechanisms mentioned in Assumptions~\ref{as:mcar}-\ref{as:mnar} as described in Algorithm~\ref{alg:data_generation}.

\begin{technicallemma}{\textnormal{Concentration bound on the norm of the difference between $(\bm{x}^0)_j$ and $(\bm{x}^\star)_i$.}}\label{lem:concentration_x0_xstar} Under any assumption in Assumptions~\ref{as:mcar}-\ref{as:mnar}, if we consider a subset of features $\mathcal{F} \subseteq \{1,2,\dots,F\}$ such that $\bm{x}^{\mid \mathcal{F}}$ is the restriction of $\bm{x} \in \mathbb{R}^F$ to features in $\mathcal{F}$, then 
\[ \forall c \geq \frac{4\ln N}{(\sigma^\text{miss})^2} \Big(1+\sqrt{1+\frac{4(\sigma^\text{miss})^2|\mathcal{F}|}{\ln N}}\Big) \ \forall i,j \leq N, \  \|(\bm{x}^0)^{\mid \mathcal{F}}_j - (\bm{x}^\star)^{\mid \mathcal{F}}_i\|^2_2 \leq  (\sigma^\text{miss})^{2}(|\mathcal{F}|+c)\;,\] 
with probability $1-\exp \left(-\frac{(\sigma^\text{miss}c)^{2}}{4(8 |\mathcal{F}|+ c)}+2\ln N\right) \in [0,1]$, where $\sigma^\text{miss} \triangleq \max(\sigma_2,\sigma^\text{GSM})$, where for an initial K-nearest imputation with uniform weights,
\[ \sigma_2 \triangleq \sigma\sqrt{1+1/K} \quad \text{ (Assumptions~\ref{as:mcar}-\ref{as:mar}) ~ and }\quad  \sigma^\text{GSM} \triangleq \sigma\sqrt{(K+3)/3K} \quad \text{ (Assumption~\ref{as:mnar})}\;.\]
\end{technicallemma}

\begin{proof}
We summarized the procedure according to which the data matrices $X^\star$ and $X^0$ are generated in Algorithm~\ref{alg:data_generation}. In particular, we assumed that $(x^\star)^f_i \sim_\text{iid} \mathcal{N}(\mu_f, \sigma^2)$ for any $i \leq N, f \leq F$ and fixed $\sigma >0$ (Assumption~\ref{as:x_distribution}), $\mu=(\mu_1,\dots,\mu_F) \in \mathbb{R}^F$, and that $K \leq \min_{f \leq F} |\{ i \leq N \mid m^f_i=0 \}|$ (Assumption~\ref{as:number_neighbors}), where the last term is the number of samples which do not miss the value of feature $f$ in the data set. Based on this, we can assume the following independence relationships for any $i, j \leq F$, where $i \neq j$, and $f, f' \leq F$, where $f \neq f'$
\begin{eqnarray}
    & & (x^\star)^f_i \independent (x^\star)^f_j \label{eq:indep_star_ij}\\
     & & (x^0)^f_i \independent (x^\star)^f_i \mid m^f_i=1 \label{eq:indep_x0_xstar_i_m1}\\
    & & (x^0)^f_i \notindependent (x^\star)^f_i \mid m^f_i=0 \quad  \text{ (since $\big( (x^0)^f_i \mid m^f_i = 0 \big) = (x^\star)^f_i$)} \label{eq:notindep_x0_xstar_i_m0}\\
    & & (x^0)^f_i \notindependent (x^0)^f_j \mid m^f_i=1, m^f_j =1 \quad \text{ (the two points can share a neighbor)} \label{eq:notindep_x0_ij_m1}\\
    & & (x^0)^f_i \notindependent (x^0)^f_j \mid m^f_i=1, m^f_j =0 \quad  \text{ ($(\bm{x}^\star)_j$ can be a neighbor of $\bm{x}_i$ for $f$)} \label{eq:notindep_x0_ij_m1m0}\\
   & & (x^0)^f_i \independent (x^0)^f_j \mid m^f_i=0, m^f_j =0 \label{eq:indep_x0_ij_m0m0}\\
    & & (x^\star)^f_i \independent (x^\star)^{f'}_j \label{eq:indep_star_f}\\
    & &  (x^\star)^f_i \independent (x^\star)^{f'}_j \label{eq:indep_star_ffp}\\
    & &  (x^\star)^f_i \independent (x^\star)^{f'}_i \text{ and }   (x^0)^f_i \independent (x^0)^{f'}_i \label{eq:indep_x0_xstar_ii_ffp}\;.
\end{eqnarray}

What is the distribution of random variable $\big( (x^0)^f_i \mid m^f_i = m\big)$ for $m \in \{0,1\}$? If $m^f_i=0$, that is, if the value for the feature $f$ and sample $i$ is not missing in the input matrix $X$, then $\big((x^0)^f_i \mid m^f_i = 0\big)$ follows the same law as $(x^\star)^f_i$. Otherwise, if $m^f_i=1$, then at the initial imputation step, 
$(x^0)^f_i \mid m^f_i = 1$ is the arithmetic mean of \textit{exactly} $K$ independent random variables of distribution $\mathcal{N}(\mu_f, \sigma^2)$ (by Independence~\eqref{eq:indep_star_ij}).~\footnote{Due to the upper bound on $K$ (Assumption~\ref{as:number_neighbors}).} All in all,
\begin{eqnarray}\label{eq:x0_distr}
    \big((x^0)^f_i \mid m^f_i = 0\big) = (x^\star)^f_i \sim \mathcal{N}(\mu_f,\sigma^2) \quad \text{ and } \quad \big((x^0)^f_i \mid m^f_i = 1\big) \sim \mathcal{N}(\mu_f, \sigma^2/K)\;.
\end{eqnarray}

Let us denote $\sigma_0 \triangleq \sigma$ and $\sigma_1 \triangleq \sigma/\sqrt{K}$ and $\sigma_2 \triangleq \sqrt{\sigma^2_0 + \sigma_1^2}$ and $\sigma_3 \triangleq \sqrt{\sigma^2_0 - \sigma_1^2}$ and $p^\text{miss}_{if} \triangleq \mathbb{P}(m^f_i=1)$. Let us now consider the distribution of the random variable $\big((x^0)^f_i-(x^\star)^f_i\big)$ for any $i,f$
\begin{eqnarray*}
    \forall i \leq N, \forall f \leq F, & & \big((x^0)^f_i-(x^\star)^f_i \mid m^f_i=0\big) = 0\\
   & & \big((x^0)^f_i-(x^\star)^f_i \mid m^f_i=1\big) \sim \mathcal{N}(0, \sigma_2^2) \quad \text{ (by Independence~\ref{eq:indep_x0_xstar_i_m1})\;.}
\end{eqnarray*}

Similarly, for any $i,j \leq N$
\begin{eqnarray*}
    & &  \forall j \neq i, \forall f \leq F, \ \big((x^0)^f_j-(x^\star)^f_i \mid m^f_j=0\big) \sim \mathcal{N}(0, 2\sigma^2) \quad \text{ (by Independence~\ref{eq:indep_star_ij})}\\
   & & \big((x^0)^f_j-(x^\star)^f_i \mid m^f_j=1\big) \sim \begin{cases} \mathcal{N}(0, \sigma_2^2) & \text{ if } \forall k \leq K, i \neq \mathcal{K}(x^f_j, X^0, k)\\ \mathcal{N}(0,\sigma_3^2) & \text{ otherwise} \end{cases}\;,
\end{eqnarray*}
because in the last case, $(x^0)^f_j-(x^\star)^f_i = \frac{1}{K} \sum_{q \neq k} (x^\star)^f_{\mathcal{K}(x^f_j,X^0,q)} + (\frac{1}{K}-1)(x^\star)^f_i$. Let us denote now $p_{ij} \triangleq \mathbb{P}\big(\forall k \leq K, i \neq \mathcal{K}(x^f_j, X^0, k) \mid m^f_j = 1\big)$. The law of total probability gives
\begin{eqnarray}
   \forall i \leq N, \ \forall f \leq F, \ \forall x \neq 0, & & \mathbb{P}\Big(\big((x^0)^f_i-(x^\star)^f_i\big) = x\Big) = p^\text{miss}_{if} \mathcal{N}(x ; 0,\sigma_2^2) \label{eq:distribution_diff_ii}\\
   & & \mathbb{P}\Big(\big((x^0)^f_i-(x^\star)^f_i\big) = 0\Big) = 1 + p^\text{miss}_{if}\underbrace{\Big(\mathcal{N}(0 ; 0,\sigma_2^2)-1\Big)}_{=1/\sqrt{2\pi \sigma_2^2}-1} \nonumber \\
   \forall i \neq j, \ \forall f \leq F, \ \forall x \in \mathbb{R}, & & \mathbb{P}\Big(\big((x^0)^f_j-(x^\star)^f_i\big) = x\Big) = (1-p^\text{miss}_{if})\mathcal{N}(0, 2\sigma_0^2)  \quad \label{eq:distribution_diff_ij} \\
   & + & p^\text{miss}_{if}\Big(p_{ij}\mathcal{N}(x ; 0,\sigma_2^2)+(1-p_{ij})\mathcal{N}(x ; 0,\sigma_3^2)\Big)\;. \nonumber
\end{eqnarray}

Then, we show that the random variable $(x^0)^f_j-(x^\star)^f_i$ is a zero-mean $\sigma^\text{miss}$-subgaussian variable under Assumptions~\ref{as:mcar}-\ref{as:mnar}, where $\sigma^\text{miss}$ depends on the missingness mechanism and the initial imputation algorithm. We recall that a zero-mean $\sigma$-subgaussian variable $X$ satisfies $\mathbb{E}[e^{\lambda X}] \leq e^{\sigma^2 \lambda^2 / 2}$ for all $\lambda \in \mathbb{R}$, with equality for any zero-mean Gaussian random variable of variance $\sigma^2$. 

\begin{lemma}{\textnormal{$(x^0)^f_j-(x^\star)^f_i$ is a zero-mean $\sigma_2$-subgaussian random variable under Assumption~\ref{as:mcar}.}}\label{lem:subgaussian_mcar} For all $i \neq j \leq N$, $f \leq F$, under the MCAR assumption, $p^\text{miss}_{jf} = p \in (0,1)$ is a constant and then $(x^0)^f_j-(x^\star)^f_i$ is a zero-mean $\sigma_2$-subgaussian random variable. 
\end{lemma}

\begin{proof}
    First, let us denote $X^f_{ij} \triangleq (x^0)^f_j-(x^\star)^f_i$ for $i,j \leq N$ and $f \leq F$. Then using Equation~\eqref{eq:distribution_diff_ii}, it is clear that $X^f_{ii}$ is centered for any $i \leq N$. Similarly, due to Equation~\eqref{eq:distribution_diff_ij} for any $i \neq j \leq N$
    \[ \mathbb{E}[X^f_{ij}] = (1-p)\underbrace{\mathbb{E}[X^f_{ij} \mid m^f_j = 0]}_{=0} + p\underbrace{\mathbb{E}[X^f_{ij} \mid m^f_j = 1]}_{=0} = 0\;.\]
    Moreover, 
    \[ \forall \lambda \in \mathbb{R}, \ \mathbb{E}[e^{\lambda X^f_{ii}}] = 1 \times \Big( 1+p\left(1/\sqrt{2\pi \sigma^2_2}-1\right)\Big) + p \mathbb{E}_{Y \sim \mathcal{N}(0,\sigma^2_2)}[e^{\lambda Y}] - p \times 1 \times \left(1/\sqrt{2 \pi \sigma^2_2}\right)\;,\]
    and then
    \[ \forall \lambda \in \mathbb{R}, \ \exp(\sigma^2_2 \lambda^2/2) - \mathbb{E}[e^{\lambda X^f_{ii}}] = p-1 + (1-p)\exp(\sigma^2_2 \lambda^2 / 2) \geq 0\;.\]
    Second, we notice that $\sigma_3 \leq \sigma_1 \leq \sigma_0 \leq \sigma_0\sqrt{2} \leq \sigma_2$ (since $K>1$). It is easy to see that any $\sigma'$-subgaussian variable is also a $\sigma''$-subgaussian variable, where $\sigma' \leq \sigma''$. Then $X^f_{ij} \mid m^f_j = 1$ and $X^f_{ij} \mid m^f_j = 0$ are both $\sigma_2$-subgaussian. Then $X^f_{ij}$ is $\sigma_2$-subgaussian for any $i,j \leq N$.
\end{proof}

\begin{lemma}{\textnormal{$(x^0)^f_j-(x^\star)^f_i$ is a zero-mean $\sigma_2$-subgaussian random variable under Assumption~\ref{as:mar}.}}\label{lem:subgaussian_mar} For all $i \neq j \leq N$, $f \leq F$, under the MAR assumption, the missingness depends on a fixed subset of \textnormal{always observed} values $F^O \subset \{1,2,\dots,F\}$: $\mathbb{P}(m^f_j = 1 \mid x^\star_j) = h\big((x^\star)^{F^O}_j,f\big)$ where $(x^\star)^{F^O}_i$ is the restriction of $(x^\star)_i$ to rows in $F^O$ and $h$ some deterministic function.~\footnote{With an abuse in notation as we denote $x^\star_i$ both the random variable and its realization.} Then $(x^0)^f_j-(x^\star)^f_i$ is a zero-mean $\sigma_2$-subgaussian random variable.
\end{lemma}

\begin{proof}
Under Assumption~\ref{as:mar}, for all $j \leq N$ and for all $f \in F^O$, $p^\text{miss}_{jf}=0$ and for all $f \not\in F^O$,
\[ p^\text{miss}_{jf} = \int_{x_{f'}, f' \in F^O} h\big([(x^\star)^{f'}_j = x^{f'}, \ f' \in F^O],f\big) \Pi_{f' \in F^O} \mathcal{N}(x^{f'}; \mu_{f'}, \sigma^2)dx\;.\]
By Independence~\eqref{eq:indep_star_ffp} and similarly to the proof of Lemma~\ref{lem:subgaussian_mcar}, $X^f_{ji}$ is then a zero-mean $\sigma_2$-subgaussian random variable for any $f \leq F$ and $i,j \leq N$. 
\color{black}
\end{proof}

\begin{lemma}{\textnormal{$(x^0)^f_j-(x^\star)^f_i$ is a zero-mean $\sigma^\text{GSM}$-subgaussian random variable under Assumption~\ref{as:mnar}.}}\label{lem:subgaussian_mnar}  For all $i \neq j \leq N$, $f \leq F$, under the Gaussian self-masking mechanism from Assumption $4$ in~\cite{le2020neumiss}, the probability of $x^f_i$ missing is given by 
\[ \forall x \in \mathbb{R}, \ \mathbb{P}(m^f_i=1 \mid (x^\star)^f_i = x) = p^\text{miss}_{i}(x,f) = K_f\exp\Big(-\frac{(x-\mu_f)^2}{\sigma^2}\Big) \text{ with } K_f \in (0,1)\;.\]
Then $(x^0)^f_j-(x^\star)^f_i$ is a zero-mean $\sigma^\text{GSM}$-subgaussian random variable, where $\sigma^\text{GSM} \triangleq \sigma\sqrt{\frac{K+3}{3K}}$.
\end{lemma}

\begin{proof}
For all $i \leq N$, $f \leq F$, and for any $x \neq 0$, by the law of total probability
\[\mathbb{P}(X^f_{ii}=x)  = \mathbb{P}(X^f_{ii}=x|m^f_i=1)\mathbb{P}(m^f_i=1) + \underbrace{\mathbb{P}(X^f_{ii}=x|m^f_i=0)}_{=0 \text{ because } x \neq 0}\mathbb{P}(m^f_i=0)\;.\]
Then since $\mathbb{P}(X^f_{ii}=x|m^f_i=1,(x^\star)^f_i=y) = \mathbb{P}((x^0)^f_i=x+y|m^f_i=1)$, using Equation~\ref{eq:x0_distr}
\begin{align*}
  \mathbb{P}(X^f_{ii}=x) & = \int_{y \in \mathbb{R}} \mathbb{P}(X^f_{ii}=x|m^f_i=1,(x^\star)^f_i=y)\mathbb{P}(m^f_i=1|(x^\star_i)^f=y)\mathbb{P}((x^\star)^f_i=y)dy\\
  & = \int_{y \in \mathbb{R}} \mathcal{N}(x+y; \mu_f, \sigma^2/K)p^\text{miss}_i(y,f)\mathcal{N}(y; \mu_f, \sigma^2)dy \triangleq I_{i,f}(x)
  \end{align*}
  \begin{align*}
\forall x \in \mathbb{R}, & \ I_{i,f}(x)  = \int_{y \in \mathbb{R}} \frac{K_f\sqrt{K}}{2\pi\sigma^2}\exp\left(-\frac{(x+y-\mu_f)^2}{\frac{2\sigma^2}{K}}\right)\exp\left(-\frac{(y-\mu_f)^2}{\sigma^2}\right)\exp\left(-\frac{(y-\mu_f)^2}{2\sigma^2}\right)dy\\
   & = \frac{K_f\sqrt{K}}{2\pi\sigma^2}\int_{y \in \mathbb{R}}\exp\left(-\frac{Kx^2+2Kx(y-\mu_f)+K(y-\mu_f)^2+2(y-\mu_f)^2+(y-\mu_f)^2}{2\sigma^2}\right) dy\\
   & = \frac{K_f\sqrt{K}}{2\pi\sigma^2}\int_{y \in \mathbb{R}}\exp\bigg(-\frac{1}{2\sigma^2}\Big((K+3)(y-\mu_f)^2+2\sqrt{K+3}(y-\mu_f)\frac{Kx}{\sqrt{K+3}}\\
   & +\frac{K^2}{K+3}x^2-\frac{K^2}{K+3}x^2+Kx^2\Big)\bigg) dy\\
   & = \frac{K_f\sqrt{K}}{2\pi\sigma^2}\int_{y \in \mathbb{R}}\exp\bigg(-\frac{1}{2\sigma^2}\Big((y\sqrt{K+3}-\mu_f+\frac{Kx}{\sqrt{K+3}})^2-\frac{3Kx^2}{K+3}\Big)\bigg) dy\\
   & = \frac{K_f\sqrt{K}}{2\pi\sigma^2}\exp\left(-\frac{3Kx^2}{2\sigma^2(K+3)}\right)\int_{y \in \mathbb{R}}\exp\left(-\frac{K+3}{2\sigma^2}\left(y-\frac{\mu_f}{\sqrt{K+3}}+\frac{Kx}{K+3}\right)^2\right) dy\\
   & = \frac{K_f\sqrt{K}}{2\pi\sigma^2}\exp\left(-\frac{3Kx^2}{2\sigma^2(K+3)}\right)\sqrt{\frac{2\pi\sigma^2}{K+3}}\\
   & = K_f\sqrt{\frac{K}{2\pi\sigma^2(K+3)}}\exp\left(-\frac{3K}{2\sigma^2(K+3)}x^2\right)\;.
\end{align*}
When $x=0$, $X^f_{ii}$ follows the second law described at Equation~\eqref{eq:distribution_diff_ii} and then
\begin{align*}
  \mathbb{P}(X^f_{ii}=0) & = \mathbb{P}(X^f_{ii}=0|m^f_i=1)\mathbb{P}(m^f_i=1) + \underbrace{\mathbb{P}(X^f_{ii}=0|m^f_i=0)}_{=1}\underbrace{\mathbb{P}(m^f_i=0)}_{=1-\mathbb{P}(m^f_i=1)}\\
  & = \int_{y \in \mathbb{R}} \mathbb{P}(X^f_{ii}=0|m^f_i=1,(x^\star)^f_i=y)\mathbb{P}(m^f_i=1 \mid (x^\star)^f_i = y)\mathbb{P}((x^\star)^f_i = y)dy\\
  & + 1 - \int_{y \in \mathbb{R}} \mathbb{P}(m^f_i=1 \mid (x^\star)^f_i = y)\mathbb{P}((x^\star)^f_i = y)dy\\
  & = \int_{y \in \mathbb{R}} \mathcal{N}(y;\mu_f, \sigma^2/K) p^\text{miss}_i(y,f)\mathcal{N}(y;\mu_f,\sigma^2)dy + 1 - \int_{y \in \mathbb{R}} p^\text{miss}_i(y,f)\mathcal{N}(y;\mu_f,\sigma^2)dy\\
  & = I_{i,f}(0) + 1 - \frac{K_f}{\sqrt{2\pi \sigma^2}}\int_{y \in \mathbb{R}} e^{-\frac{3}{2\sigma^2}(y-\mu_f)^2} dy\\
  & =  K_f\sqrt{\frac{K}{2\pi\sigma^2(K+3)}} + 1 - \frac{K_f}{\sqrt{2\pi \sigma^2}}\sqrt{\frac{2 \pi \sigma^2}{3}} = 1 + K_f\left(\sqrt{\frac{K}{2\pi\sigma^2(K+3)}}-\frac{1}{\sqrt{3}}\right)\;.
\end{align*}

That is
\begin{eqnarray}\label{eq:distribution_diff_GSM}
    \forall x \neq 0, \ \mathbb{P}((x^0)^f_i-(x^\star)^f_i=x) & = & \frac{K_f}{\sqrt{3}} \times \mathcal{N}(x; 0, (\sigma^\text{GSM})^2) \text{ where } \sigma^\text{GSM} \triangleq \sigma\sqrt{\frac{K+3}{3K}}\\
    \mathbb{P}((x^0)^f_i-(x^\star)^f_i=0) & = & 1 + K_f\left(\sqrt{\frac{K}{2\pi\sigma^2(K+3)}}-\frac{1}{\sqrt{3}}\right)\:. \nonumber
\end{eqnarray}


For the zero-mean variable X following the distribution described in Equation~\eqref{eq:distribution_diff_GSM}, the moment-generating function (MGF) of $X$ is given by
\begin{align*}
    \mathbb{E}[e^{tX}] =&\int\mathbb{P}(X=x)e^{tx}dx\\
    =&\int_{x\neq0}e^{tx}\frac{K_f}{\sqrt{6\pi(\sigma^\text{GSM})^2}}\exp\left(-\frac{x^2}{2(\sigma^\text{GSM})^2}\right)dx\\
    =& \frac{K_f}{\sqrt{3}}\mathbb{E}_{Y\sim\mathcal{N}(0,(\sigma^\text{GSM})^2)}[e^{tY}]=\frac{K_f}{\sqrt{3}}\exp\left(\frac{(\sigma^\text{GSM})^2t^2}{2}\right)\;.
\end{align*}
Choose $s=\sigma^\text{GSM}$. Then,
\begin{align*}
    \exp(\frac{s^2t^2}{2})-\mathbb{E}[e^{tX}]=&\exp\left(\frac{(\sigma^\text{GSM})^2t^2}{2}\right)-\frac{K_f}{\sqrt{3}}\exp\left(\frac{(\sigma^\text{GSM})^2t^2}{2}\right)\\
    =&\exp\left(\frac{(\sigma^\text{GSM})^2t^2}{2}\right)\Big(1-\frac{K_f}{\sqrt{3}}\Big)\;.
\end{align*}
Clearly the minimum value, achieved at $t=0$, is $1-\frac{K_f}{\sqrt{3}} \geq 0$ since $K_f \in (0,1)$ by definition. 
All in all, $X$ is a zero-mean $\sigma^\text{GSM}$-subgaussian variable.
\end{proof}

\begin{lemma}{\textnormal{If $X$ is $s$-subgaussian, then $\beta X$ is $\beta s$-gaussian when $\beta > 0$.}}\label{lem:subgaussian_mult} For any $\beta > 0$ and $X$ zero-mean $s$-subgaussian, $\beta X$ is zero-mean $\beta s $-subgaussian.
\end{lemma}

\begin{proof}
    If $X$ is a zero-mean $s$-subgaussian, then $\mathbb{E}[\beta X] = 0$ and
\[ \forall t > 0, \ \mathbb{P}(|\beta X| \geq t ) = \mathbb{P}(|X| \geq \beta^{-1} t ) \leq 2\exp\left(-\frac{t^2}{2(\beta s)^2}\right) \;,\]
and using Proposition 2.5.2 from~\cite{vershynin2018high}, $\beta X$ is a (zero-mean) $\beta s$-subgaussian variable.
\end{proof}

Finally, we determine a concentration bound on $\|(\bm{x}^0)^{\mid \mathcal{F}}_j - (\bm{x}^\star)^{\mid \mathcal{F}}_i\|^2_2$ for any $i,j \leq N$ and $\mathcal{F} \subseteq \{1,2,\dots,F\}$ under any of the Assumptions~\ref{as:mcar}-\ref{as:mnar}. Let us set $\sigma^\text{miss} \triangleq \max(\sigma_2, \sigma^\text{GSM})$ and introduce the $|\mathcal{F}|$-dimensional random vector $\widetilde{X}^{\mid \mathcal{F}}_{ji} \triangleq (\bm{x}^0)^{\mid \mathcal{F}}_j - (\bm{x}^\star)^{\mid \mathcal{F}}_i$ for any $i,j \leq N$. The $|\mathcal{F}|$ coefficients of $\widetilde{X}^{\mid \mathcal{F}}_{ji}$ follow the distribution described in Equations~\eqref{eq:distribution_diff_ii}-\eqref{eq:distribution_diff_ij}. Then, the random vector $(\sigma^\text{miss})^{-1}\widetilde{X}^{\mid \mathcal{F}}_{ji}$ has $|\mathcal{F}|$ independent $1$-subgaussian zero-mean coefficients. The independence holds by Independence~\eqref{eq:indep_x0_xstar_ii_ffp} and~\eqref{eq:indep_star_ij}. The coefficients are $1$-subgaussian due to Lemma~\ref{lem:subgaussian_mult}. Using Theorem 3.1.1 from~\cite{vershynin2018high}, which relies on Bernstein's inequality applied to the random variables $(\sigma^\text{miss})^{-1}\widetilde{X}^{f}_{ji}$, for any feature $f \in \mathcal{F}$ and samples $i,j \leq N$, for any constant $c> 0$
\begin{eqnarray*}
    & & \mathbb{P}[ (\sigma^\text{miss})^{-2}\|\widetilde{X}^{\mid \mathcal{F}}_{ji}\|^2_2 \geq  |\mathcal{F}|+c] \leq   \exp \Big(-\frac{c^2}{4(8|\mathcal{F}|+c)}\Big)  \\ 
   & \implies &  \mathbb{P}[\|(\bm{x}^0)^{\mid \mathcal{F}}_j - (\bm{x}^\star)^{\mid \mathcal{F}}_{i}\|^2_2 \geq (\sigma^\text{miss})^{2}(|\mathcal{F}|+c) ]  \leq   \exp \left(-\frac{(\sigma^\text{miss}c)^{2} }{4(8 |\mathcal{F}|+ c)}\right)\\
   & \implies &   \mathbb{P}[\cup_{i,j \leq N} \{ \|(\bm{x}^0)^{\mid \mathcal{F}}_j - (\bm{x}^\star)^{\mid \mathcal{F}}_i\|^2_2 \geq  (\sigma^\text{miss})^{2}(|\mathcal{F}|+c) \}]  \leq \exp \left(-\frac{(\sigma^\text{miss}c)^{2}}{4(8 |\mathcal{F}|+ c)}+2\ln N\right)\;,\\
\end{eqnarray*}

by applying an union bound on $\{1,2,\dots,N\}^2$. And then for any positive constant $c$ such that $2\ln N-(\sigma^\text{miss}c)^2/(4(8|\mathcal{F}|+c)) \leq 0$,
\begin{eqnarray*}
 & & \mathbb{P}[\cap_{i,j \leq N} \ \|(\bm{x}^0)^{\mid \mathcal{F}}_j - (\bm{x}^\star)^{\mid \mathcal{F}}_i\|^2_2 \leq (\sigma^\text{miss})^{2}(|\mathcal{F}|+c)]\\
 & =  & 1-\mathbb{P}[\cup_{i,j \leq N} \ \|(\bm{x}^0)^{\mid \mathcal{F}}_j - (\bm{x}^\star)^{\mid \mathcal{F}}_i\|^2_2 \geq (\sigma^\text{miss})^{2}(|\mathcal{F}|+c)]\\
  & \geq &  1- \exp \left(-\frac{(\sigma^\text{miss}c)^{2}}{4(8 |\mathcal{F}|+ c)}+2\ln N\right)\;.
\end{eqnarray*}

A positive such $c$ always exists, which can be shown by choosing $c$ such that
\[ c \geq \frac{4\ln N}{(\sigma^\text{miss})^2} \Big(1+\sqrt{1+4(\sigma^\text{miss})^2|\mathcal{F}|/\ln N}\Big) > 0 \implies 2\ln N-(\sigma^\text{miss}c)^2/(4(8|\mathcal{F}|+c)) \leq 0 \;.\]
Note that, similarly, by union bound on $\{1,2,\dots,N\}$, for such a $c$,
\begin{eqnarray*}
  \mathbb{P}[\cap_{i \leq N} \ \|(\bm{x}^0)^{\mid \mathcal{F}}_i - (\bm{x}^\star)^{\mid \mathcal{F}}_i\|^2_2 \leq (\sigma^\text{miss})^{2}(|\mathcal{F}|+c)] \geq  1- \exp \left(-\frac{(\sigma^\text{miss}c)^{2}}{4(8 |\mathcal{F}|+ c)}+\ln N\right) \in [0,1]\;.
\end{eqnarray*}
\end{proof}

\begin{corollary}{\textnormal{First concentration bound on $\|(\bm{x}^0)_j - (\bm{x}^\star)_i\|^2_2$.}}\label{cor:concentration_x0_xstar_full} Under any assumption in Assumptions~\ref{as:mcar}-\ref{as:mnar}, then 
\[ \forall c \geq \frac{4\ln N}{(\sigma^\text{miss})^2} \Big(1+\sqrt{1+\frac{4(\sigma^\text{miss})^2F}{\ln N}}\Big) \quad \forall i,j \leq N, \quad  \|(\bm{x}^0)_j - (\bm{x}^\star)_i\|^2_2 \leq  (\sigma^\text{miss})^{2}(F+c)\;,\] 
with probability $1-\exp \left(-\frac{(\sigma^\text{miss}c)^{2}}{4(8 F+ c)}+2\ln N\right) \in [0,1]$, where $\sigma^\text{miss} \triangleq \max(\sigma_2,\sigma^\text{GSM}) \propto \sigma$ is defined in Technical lemma~\ref{lem:concentration_x0_xstar}. 
\end{corollary}

\begin{proof}
    This statement holds by application of Lemma~\ref{lem:concentration_x0_xstar} with $\mathcal{F}=\{1,2,\dots,F\}$.
\end{proof}

\begin{corollary}{\textnormal{Second concentration bound on $\|(\bm{x}^0)_j - (\bm{x}^\star)_i\|^2_2$ and $\|(\bm{x}^0)_i - (\bm{x}^\star)_i\|^2_2$.}}\label{cor:concentration_x0_xstar_full2} Under any assumption in Assumptions~\ref{as:mcar}-\ref{as:mnar}, for $\sigma^\text{miss} \triangleq \max(\sigma_2,\sigma^\text{GSM}) \propto \sigma$ (Technical lemma~\ref{lem:concentration_x0_xstar}), let us denote
\[ C^\text{miss}_\delta \triangleq  (\sigma^\text{miss})^{2}F+2\ln(1/\delta)\left(1+\sqrt{1 + 8(\sigma^\text{miss})^2F/\ln(1/\delta)}\right) \text{ for }\delta \leq 1/N\;.\]
Then, with probability $1-\delta \in (0,1)$, for all $i,j \leq N$, $\|(\bm{x}^0)_j - (\bm{x}^\star)_i\|^2_2 \leq C^\text{miss}_{\delta/N^2}$. 
\end{corollary}

\begin{proof}
    We solve the following equation in $c>0$ from Corollary~\ref{cor:concentration_x0_xstar_full},
    \[ \delta = \exp \left(-\frac{(\sigma^\text{miss}c)^{2}}{4(8 F+ c)}+2\ln N\right) \Leftrightarrow -(\sigma^\text{miss})^2c^2 + (4\ln(N^2/\delta)) c + 32F\ln(N^2/\delta) = 0\;.  \]
    This equation has two real roots, one positive root being
    \[ c_\delta \triangleq \frac{4\ln(N^2/\delta)}{(\sigma^\text{miss})^2 }\left(1+\sqrt{1 + 8(\sigma^\text{miss})^2F /\ln(N^2/\delta)}\right)\;. \]
    Applying Corollary~\ref{cor:concentration_x0_xstar_full} with $c=c_\delta$ when $\delta \in (0,1)$ yields  for any $j,i \leq N$, with probability $1-\frac{\delta}{N^2} \in (0,1)$, 
\[ \|(\bm{x}^0)_j - (\bm{x}^\star)_i\|^2_2 \leq C^\text{miss}_{\delta/N^2}\;.\]
Applying an upper bound on $\{1,2,\dots, N\}^2$ yields the expected result.
\end{proof}

\begin{proposition}\textnormal{Iterative improvement from $X^0$ until $X^t$.}\label{lem:tel_series} For $G_\circ \in \{G, G_\star\} $, for any data matrix $X \in (\mathbb{R} \cup \{\texttt{NaN}\})^{N \times F}$ and $(\bm{\alpha}^s)_{s \leq t} \in (\triangle_K)^t$
    \[ \sum_{s=1}^t G_\circ(\bm{\alpha}^s, X^{s-1}) = \frac{1}{N} \sum_{i \leq N} \log D_\circ((\bm{x}^t)_i(\bm{\alpha}^t))/\log D_\circ((\bm{x}^0)_i) - \eta  \sum_{s=1}^t \|\bm{\alpha}^s\|^2_2\;. \]
\end{proposition}

\begin{proof}
    For any $G_\circ \in \{G, G_\star\} $ and $t \geq 1$, since $(\bm{x}^{s})_i = (\bm{x}^{s-1})_i(\bm{\alpha}^{s})$ for $s < t$ and $i \leq N$
    \begin{eqnarray*}
        \sum_{s=1}^t G_\circ(\bm{\alpha}^s, X^{s-1}) & = & \sum_{s=1}^t \frac{1}{N} \sum_{i \leq N} \log \frac{D_\circ((\bm{x}^{s-1})_i(\bm{\alpha}^s))}{D_\circ((\bm{x}^{s-1})_i)} - \eta \|\bm{\alpha}^s\|^2_2\\
        & = & \frac{1}{N} \sum_{i \leq N} \log \frac{D_\circ((\bm{x}^{t})_i(\bm{\alpha}^t))}{D_\circ((\bm{x}^{0})_i)} - \eta \sum_{s=1}^t \|\bm{\alpha}^s\|^2_2\;.
    \end{eqnarray*}
\end{proof}

\section{Experimental study}\label{sec:experiments}

We compare our algorithmic contributions F3I and PCGrad-F3I to baselines for imputation and joint imputation-classification tasks. We considered as baselines the imputation by the mean value, the MissForest algorithm~\cite{stekhoven2012missforest}, K-nearest neighbor (KNN) imputation with uniform weights and distance-proportional weights, where the weight is inversely proportional to the distance to the neighbor~\cite{troyanskaya2001missing}, an Optimal Transport-based imputer~\cite{muzellec2020missing} and finally not-MIWAE~\cite{ipsen2021not}. 

We consider synthetic data sets produced by Algorithm~\ref{alg:data_generation}, public drug repurposing data sets and the MNIST data set for handwritten-digit recognition~\cite{lecun1998mnist}, along with the three missingness mechanisms corresponding to Assumptions~\ref{as:mcar}-\ref{as:mnar} for different missingness frequencies in $\{0.1, 0.25, 0.5, 0.75, 0.9\}$ across the full matrix. The missingness frequencies aim at approximating the actual expected probability of a missing value. 

\begin{remark}\textnormal{Implementation of the missingness mechanisms.} The implementations of the MCAR and MAR mechanisms come from~\cite{muzellec2020missing} (with \texttt{opt=`logistic'}). For a MCAR mechanism or MAR mechanism implemented by~\cite{muzellec2020missing}, it corresponds to the random probability of missing data. For the MNAR Gaussian self-masking and feature $f$, using the notation in Assumption~\ref{as:mnar}, we sample $K_f$ from $\mathcal{N}(\frac{3.5}{3}p^\text{miss}(1-p^\text{miss}), 0.1)$ and we clip $K_f$ in $[0.01, 0.99]$ whenever necessary. Empirically, as long as $p^\text{miss}$, the expected missingness frequency is not too extreme (\textit{i}.\textit{e}., far from the bounds of $[0,1]$), the empirical probability of missingness is close to $p^\text{miss}$. However, controlling this probability more finely in the case of a MNAR mechanism might break the not-missing-at-random property.
\end{remark}

\begin{remark}\textnormal{Computational resources.}
    The experiments on synthetic data (Subsection~\ref{subapp:synthetic}) were run on a personal laptop (processor 13th Gen Intel(R) Core(TM) i7-13700H, 20 cores @5GHz, RAM 32GB). The experiments on drug repurposing (Subsection~\ref{subapp:drug_repurposing}) were run on remote cluster servers (processor QEMU Virtual v2.5+, 48 cores @2.20GHz, RAM 500GB, and processor Intel Core i7-8750H, 20 cores @2.50GHz, RAM 7.7GB for the TRANSCRIPT drug repurposing data set~\cite{reda2023transcript}). No GPU was used in our experiments. 
\end{remark}

\begin{remark}\textnormal{Numerical considerations.} To ensure the stability of the optimization procedure, we compute directly the logarithm of the kernel density $D_0$ using function $\texttt{kernel\_density(}\cdot\texttt{,h=h,kernel=`gaussian')}$ from the k-d tree class in the Python package \texttt{scikit-learn}~\cite{pedregosa2011scikit}. 
\end{remark}

\begin{remark}\textnormal{Time complexity for the imputation steps in F3I (Algorithm~\ref{alg:online_F3I}).} The time complexity of running the KNN imputer~\cite{troyanskaya2001missing} with uniform weights and building the k-d tree on $N$ $F$-dimensional points is $\mathcal{O}(FN\log N)$, both steps being performed once. For each input point $\bm{x}$, Algorithm~\ref{alg:imputation} first queries $K$ nearest neighbors (each query has a time complexity of $\mathcal{O}(\log N)$) and then performs the imputation in at most $FK$ operations, for a total time complexity across all points of $\mathcal{O}(NK(\log N + F))$. 
\end{remark}

\subsection{Synthetic Gaussian data sets}\label{subapp:synthetic}

The data matrices $X \in \mathbb{R}^{N \times F}$ with $N=50$ samples and $F=100$ features are generated according to Algorithm~\ref{alg:data_generation} (Lines 4-18), with the multivariate mean parameter $\bm{\mu} \sim \mathcal{N}(\bm{0}_F, \nu^2\bm{I}_{F \times F})$ where $\nu=0.1$ and covariance parameter $\Sigma = \sigma^2\bm{I}_{F \times F}$. Hyperparameter values are reported in Table~\ref{tab:hyperparams}. We use $K=5$ neighbors here for all algorithms for which it is relevant.

\begin{table}[tb]
    \centering
    \caption{Hyperparameters for F3I (Algorithm~\ref{alg:online_F3I}) and its baselines, unless otherwise specified (as some of those hyperparameters might be finetuned in our experiments). K is the number of neighbors in F3I. The names of the hyperparameters match the corresponding argument names in their  implementation in Python (official, in scikit-learn~\citep{pedregosa2011scikit} or in HyperImpute~\citep{jarrett2022hyperimpute}, if present). The $k$ in TDM is not a number of neighbors or anything equivalent.}
    \label{tab:hyperparams}
    \begin{tabular}{ll}
        \toprule
        \textbf{Imputer} & \textbf{Hyperparameters} \\
        \midrule
        F3I & \texttt{n\_neighbors}$=K$, \texttt{max\_iter}$=500$, \texttt{$\eta$}$=0.001$, \texttt{S}$=1$, \texttt{$\beta$}$=0$\\
        MissForest~\cite{stekhoven2012missforest} & \texttt{n\_estimators}$=K$, \texttt{max\_depth}$=10$, \texttt{max\_size}$=0.5$,\\
        & \texttt{max\_iters}$=500$, \texttt{$\beta$}$=0$\\
        KNN~\cite{troyanskaya2001missing} (uniform) & \texttt{n\_neighbors}$=K$, \texttt{distance}=`nan\_euclidean' \\
        KNN~\cite{troyanskaya2001missing} (distance) & \texttt{n\_neighbors}$=K$, \texttt{distance}=`nan\_euclidean'\\
        Optimal Transport~\cite{muzellec2020missing} & \texttt{eps}$=0.01$, \texttt{lr}$=0.01$, \texttt{max\_iters}$=500$, \texttt{batch\_size}$=128$,\\
        & \texttt{n\_pairs}$=1$, \texttt{noise}$=0.1$, \texttt{scaling}$=0.9$\\
        not-MIWAE~\cite{ipsen2021not} & \texttt{n\_latent}$=\lfloor F/2 \rfloor$, \texttt{n\_hidden}$=150$\\
        GAIN~\cite{yoon2018gain} & \texttt{batch\_size}$=128$, \texttt{n\_epochs}$=100$, \texttt{hint\_rate}$= 0.8$, \\
        & \texttt{loss\_alpha}$ = 10$ \\
        GRAPE~\cite{you_handling_2020} & \texttt{node\_dim}$=64$, \texttt{edge\_dim}$=16$, \texttt{nepochs}$=20,000$\\
        HyperImpute~\cite{jarrett2022hyperimpute} & \texttt{imputation\_order}$= 2$, \texttt{baseline\_imputer}$ = 0$, \\
        & \texttt{optimizer}$=$'simple',  \texttt{class\_threshold}$ = 5$,\\
        & \texttt{optimize\_thresh}$= 5,000$, \texttt{n\_inner\_iter}$ = 40$, \\
        & \texttt{select\_patience}$ = 5$\\
        MIRACLE~\cite{kyono2021miracle} & \texttt{lr}$ = 0.001$, \texttt{batch\_size}$ = 1,024$, \texttt{num\_outputs}$ = 1$, \\
        & \texttt{n\_hidden}$ = 32$, \texttt{reg\_lambda}$ = 1$, \texttt{reg\_beta}$ = 1$, \texttt{reg\_m}$ = 1.0$, \\
      & \texttt{window}$ = 10$, \texttt{max\_steps}$= 400$, \texttt{seed\_imputation}$ =$'mean'\\
        NewImp~\cite{chen2024rethinking} & \texttt{entropy\_reg}$=10$, \texttt{eps}$=0.01$, \texttt{lr}$=0.01$,
                 \texttt{opt}$=$'Adam', \texttt{niter}$=50$,\\
                 & \texttt{kernel\_func}$=$'xRBF',
                 \texttt{mlp\_hidden}$=[256, 256]$,\\
                 & \texttt{score\_net\_epoch}$=2,000$, \texttt{score\_net\_lr}$=0.001$, \\
                 & \texttt{score\_loss\_type}$=$'dsm',
                 \texttt{bandwidth}$=10$, \\
                 & \texttt{sampling\_step}$=500$, 
                 \texttt{batchsize}$=128$, \texttt{n\_pairs}$=1$, \\
                 & \texttt{noise}$=0.1$, \texttt{scaling}$=0.9$\\
        Remasker~\cite{du2023remasker} & \texttt{batch\_size}$=64$, \texttt{max\_epochs}$=600$, \texttt{accum\_iter}$=1$, \\
        & \texttt{mask\_ratio}$=0.5$, \texttt{embed\_dim}$=32$, \texttt{depth}$=6$, \texttt{decoder\_depth}$=4$,\\
        & \texttt{num\_heads}$=4$, \texttt{mlp\_ratio}$=4$, \texttt{encode\_func}$=$'linear', \\
        & \texttt{weight\_decay}$=0.05$, \texttt{lr}$=$None, \texttt{blr}$=0.001$, \texttt{min\_lr}$=0.00001$, \\
        & \texttt{warmup\_epochs}$=40$\\
        TDM~\cite{zhao2023transformed} & \texttt{k}$= 2$, \texttt{depth}$=3$, \texttt{im\_lr}$=0.01$, \texttt{proj\_lr}$=0.01$, \texttt{opt}$=$'RMSprop', \\
        & \texttt{niter}$=2,000$, \texttt{batchsize}$=128$, \texttt{n\_pairs}$=1$, \texttt{noise}$=0.1$ \\
        RF-GAP~\cite{rhodes2023geometry} &  \texttt{prox\_method}$=$'rfgap'\\
        \bottomrule
    \end{tabular}
\end{table}

\subsubsection{Validation of theoretical results (single imputation task)} 

We first show on synthetic data that both Theorems~\ref{thm:mse_bounds} and~\ref{thm:regret_f3i} are experimentally validated, and look at the behavior of $\bm{\alpha}$ across imputation steps in F3I for all missingness mechanisms.

\begin{figure}
    \centering
    \includegraphics[width=0.32\linewidth]{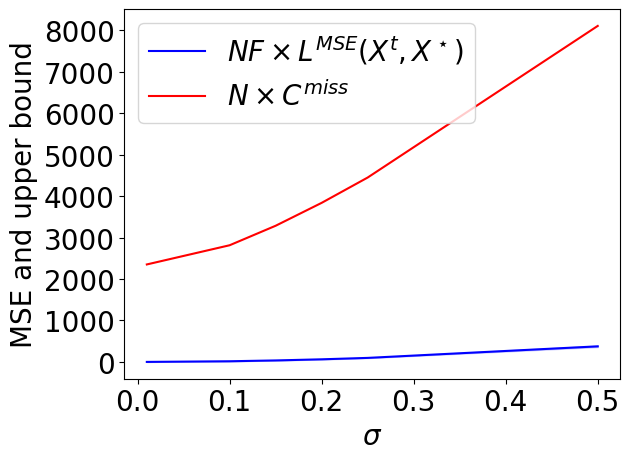}
     \includegraphics[width=0.32\linewidth]{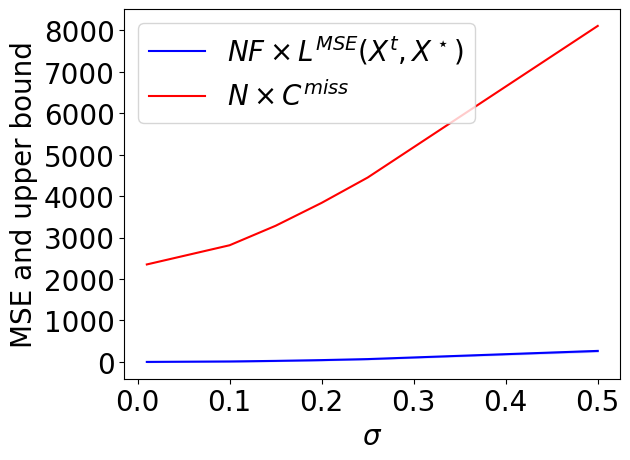}
     \includegraphics[width=0.32\linewidth]{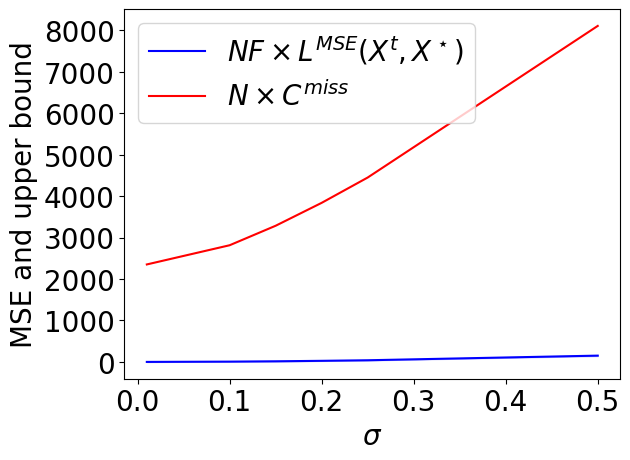}
    \caption{Emprical validation of Theorem~\ref{thm:mse_bounds} by comparing the value of the upper bound $N \times C^\text{miss}$ and $NF \times \mathcal{L}^\text{MSE}(X^t, X^\star)$ where $t$ is the final round for F3I. Left: MCAR setting. Center: MAR setting. Right: MNAR setting. }
    \label{fig:thm1}
\end{figure}

\begin{figure}
    \centering
    \includegraphics[width=0.32\linewidth]{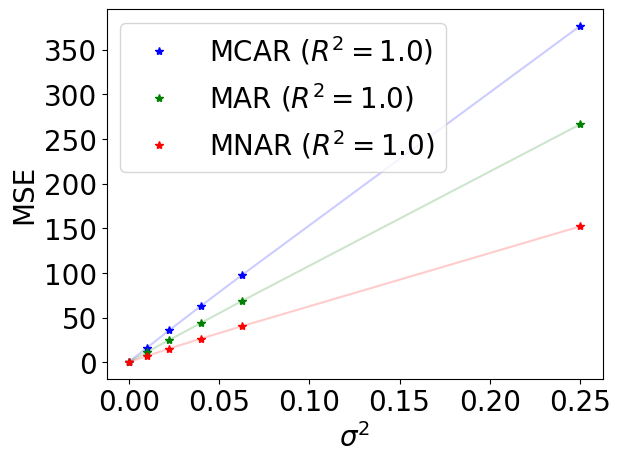}
    \caption{$NF \times \mathcal{L}^\text{MSE}(X^t, X^\star)$ is linear in $\sigma^2$ regardless of the missingness mechanism (numerical values are reported in Table~\ref{tab:thm1}). }
    \label{fig:thm1-dep-sigma}
\end{figure}

\paragraph{Empirical validation of Theorem~\ref{thm:mse_bounds}~ } We fix the missingness frequency to $25\%$ for all three missingness mechanisms. First, in Figure~\ref{fig:thm1}, for each missingness type in Assumptions~\ref{as:mcar}-\ref{as:mnar}, we ran F3I on $100$ randomly generated synthetic data matrices with each $\sigma \in \{0.01, 0.1, 0.15, 0.2, 0.25, 0.5\}$ instead of $\sigma=0.1$ and reported the mean-squared error (MSE) loss $\mathcal{L}^\text{MSE}(X^t,X^\star)$ (where $X^t$ is the last imputed data set in F3I) along with the corresponding $\sigma$-dependent upper bound $C^\text{miss} = \mathcal{O}((\sigma^\text{miss})^2F + \ln N)$. The exact definition of $C^\text{miss}$ is in the proof of Theorem~\ref{thm:mse_bounds}, in Appendix~\ref{subapp:mse}. The upper bound is largely above the MSE value for each iteration. This might be because concentration bounds derived from Bernstein's inequality are not very tight. From Table~\ref{tab:thm1} which reports the numerical values shown on Figure~\ref{fig:thm1}, we notice that there is a correlation between $\sigma$ and $\mathcal{L}^\text{MSE}(X^t,X^\star)$. Moreover, $C^\text{miss}$ recovers interesting dependencies as an upper bound of $\mathcal{L}^\text{MSE}(X^t,X^\star)$. Indeed, we also observe empirically that $\mathcal{L}^\text{MSE}(X^t,X^\star)$ is roughly linear in $\sigma^2 \approx (\sigma^\text{miss})^2$ regardless of the missingness mechanism (see Figure~\ref{fig:thm1-dep-sigma}), which matches the upper bound given by Theorem~\ref{thm:mse_bounds}.

\begin{table}[tb]
    \centering
    \caption{Empirical validation of Theorem~\ref{thm:mse_bounds} by comparing the value of the upper bound $N \times C^\text{miss}$ and the average and standard deviation value of $NF \times \mathcal{L}^\text{MSE}(X^t, X^\star)$ across iterations where $t$ is the final round for F3I. All values are rounded to the closest second decimal place. Theorem~\ref{thm:mse_bounds} states that $\mathcal{L}^\text{MSE}(X^t, X^\star) \leq C^\text{miss}/F$ with probability $1-1/50$.}
    \label{tab:thm1}
    \begin{tabular}{llrr}
    \toprule
   Missingness type &  $\sigma$ & $NF \times \mathcal{L}^\text{MSE}(X^t, X^\star)$ & $N \times C^\text{miss}$\\
    \midrule
       MCAR (Assumption~\ref{as:mcar})  &  0.01 & 0.63$\pm$ 0.38 & 2,352.60\\
        &  0.10 & 16.36$\pm$ 0.93 & 2,816.02\\
        &  0.15 & 36.12$\pm$ 1.89 & 3,286.57\\
        &  0.20 & 63.26$\pm$ 3.14 & 3,839.30\\
        &  0.25 & 97.44$\pm$ 4.56 & 4,450.16\\
        &  0.50 & 376.29$\pm$ 16.76  & 8,109.04\\
        \midrule
        MAR (Assumption~\ref{as:mar})   & 0.01 & 0.42$\pm$ 0.21  & 2,352.60\\
        &  0.10 & 11.53$\pm$ 0.78 & 2,816.02\\
        &  0.15 & 25.19$\pm$ 1.58 & 3,286.57\\
        &  0.20 & 44.07$\pm$ 2.65 & 3,839.30\\
        &  0.25 & 68.50$\pm$ 4.08 & 4,450.16\\
        &  0.50 & 266.19$\pm$ 15.22 & 8,109.04\\
        \midrule
         MNAR (Assumption~\ref{as:mnar})  & 0.01 & 0.37$\pm$ 0.23 & 2,352.60\\
        &  0.10 & 7.13$\pm$ 0.64 & 2,816.02\\
        &  0.15 & 15.50$\pm$ 1.22 & 3,286.57\\
        &  0.20 & 26.53$\pm$ 1.86 & 3,839.30\\
        &  0.25 & 40.36$\pm$ 2.66 & 4,450.16\\
        &  0.50 & 152.03$\pm$ 9.56 & 8,109.04\\
         \bottomrule
    \end{tabular}
\end{table}

\begin{figure}
    \centering
   \includegraphics[width=0.32\linewidth]{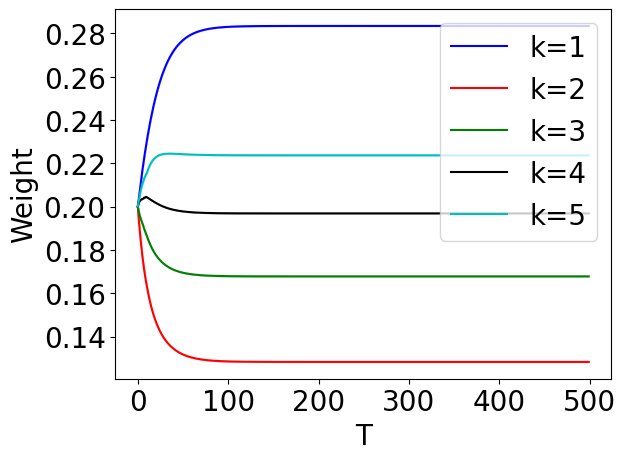}
    \includegraphics[width=0.32\linewidth]{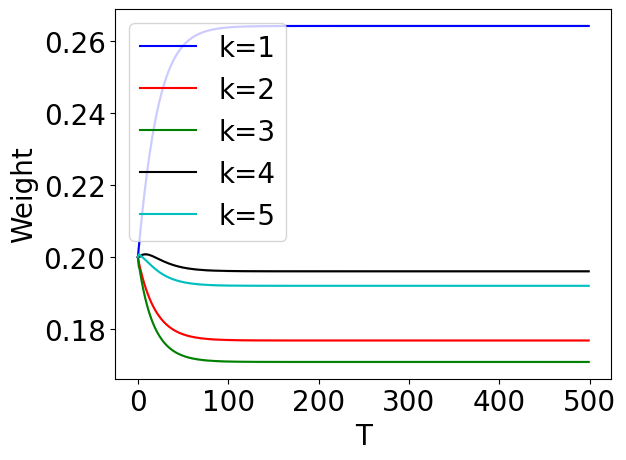}
     \includegraphics[width=0.32\linewidth]{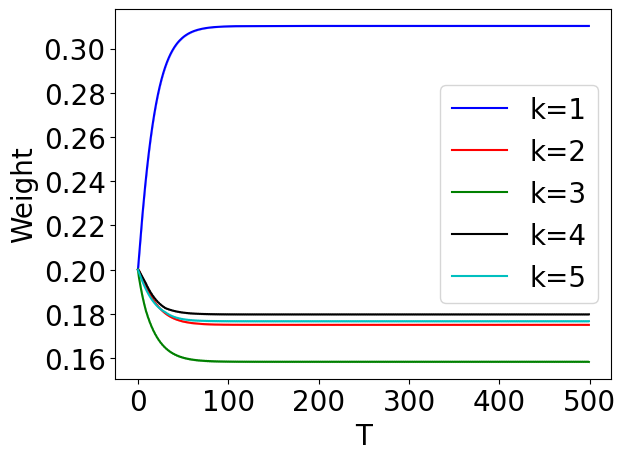}
    \caption{Evolution of the weight of each of the $K$-nearest neighbors for each sample as computed by F3I, where the $k$ neighbor is the $k^\text{th}$-nearest point, depending on the round $T$. Left: MCAR setting. Center: MAR setting. Right: MNAR setting.}
    \label{fig:evolution-alpha}
\end{figure}

\paragraph{Behavior of $\bm{\alpha}^t$ depending on imputation round $t$~ } Second, we look at the evolution of $\bm{\alpha}^t$ depending on the round $t$, knowing that at $t=0$, $\bm{\alpha}^0=\frac{1}{K}\bm{1}_K$ is a uniform weight vector. Figure~\ref{fig:evolution-alpha} displays the evolution of weight $(\alpha^t)_k$ for each $k$-nearest neighbor in iteration $t$ in F3I. Surprisingly enough, the optimal weight vector is not proportional to the rank of the neighbor; that is, the closer the neighbor, the higher the weight, which often motivates some heuristics about k-nearest neighbor algorithms. Optimality (preserving the data distribution) puts higher weights on the first \emph{and last} closest neighbors. 

\begin{figure}
    \centering
   \includegraphics[width=0.32\linewidth]{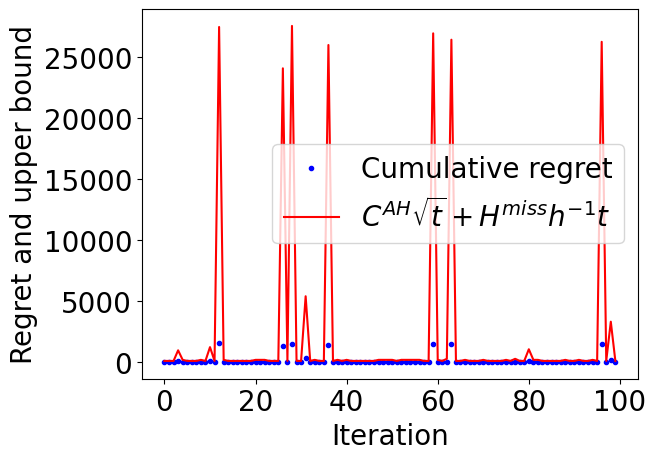}
     \includegraphics[width=0.32\linewidth]{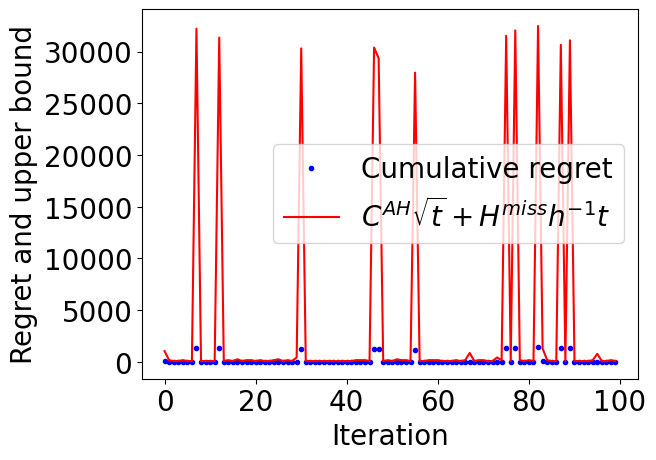}
     \includegraphics[width=0.32\linewidth]{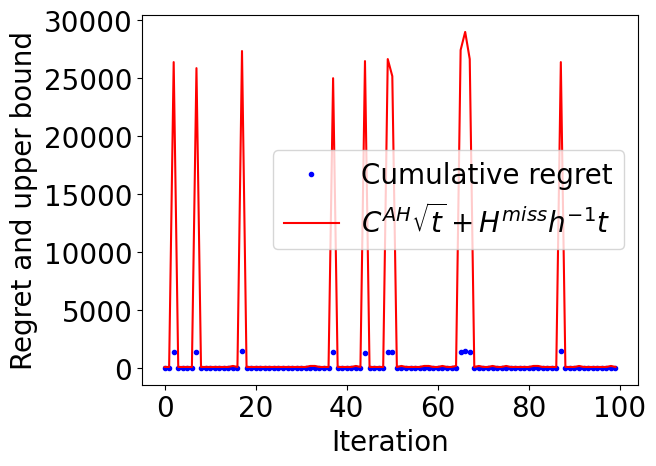}
    \caption{Cumulative regret for F3I and upper bound from Theorem~\ref{thm:regret_f3i} across $100$ iterations. The blue points are always below the red lines. Left: MCAR setting. Center: MAR setting. Right: MNAR setting.}
    \label{fig:theorem2}
\end{figure}

\paragraph{Empirical validation of Theorem~\ref{thm:regret_f3i}~ } Third, we look at the upper bound for the expected cumulative regret stated in Theorem~\ref{thm:regret_f3i}. Using a missingness frequency of $25\%$ again, we run $100$ times F3I on synthetic data sets for all three missingness mechanisms and track the values of $\max_{\bm{\alpha} \in \triangle_K} \sum_{s=1}^t G_*(\bm{\alpha}, X^{s-1})-G_*(\bm{\alpha}^s, X^{s-1})$ and its upper bound $C^\text{AH}\sqrt{t} + H^\text{miss}h^{-1}t$ across iterations, where $t$ is the final step of F3I (that can change across iterations). We compute the value of $\max_{\bm{\alpha} \in \triangle_K} \sum_{s=1}^t G_*(\bm{\alpha}, X^{s-1})$ by solving the related convex problem with function \texttt{minimize} in Python package \texttt{scipy.optimize}~\cite{2020SciPy-NMeth} after running F3I
\[ \min_{\bm{\alpha} \in \mathbb{R}^K} -\sum_{s=1}^t G_\star(\bm{\alpha}, X^{s-1}) \quad \text{ such that } \quad \forall k \leq K, \ \alpha_k \geq 0 \quad \text{ and } \quad \sum_{k \leq K} \alpha_k = 1\;,\]
where $G_\star$ is computed with respect to the true complete points $\{(\bm{x}^\star)_1, \dots, (\bm{x}^\star)_N\}$ and $(\bm{x}^{s})_i = \texttt{Impute}((\bm{x}^{s-1})_i; \bm{\alpha}^{s})$ if $s \geq 1$ and $X^0$ is the naively imputed matrix. Figure~\ref{fig:theorem2} and Table~\ref{tab:theorem2} show that the upper bound is always valid across those experiments. Some random data sets among the 100 might be harder than the others, incurring larger regret. However, Figure~\ref{fig:theorem2} shows that the upper bound adapts to these instances. The large gap between the empirical and theoretical peaks in hardness might be due, as for Theorem~\ref{thm:mse_bounds}, to the conservative estimates given by Bernstein's inequality.

\begin{table}[tb]
    \centering
    \caption{Empirical validation of Theorem~\ref{thm:regret_f3i} by comparing the value of the upper bound and the average and standard deviation value of the cumulative regret $\max_{\bm{\alpha} \in \triangle_K} \sum_{s=1}^t G_*(\bm{\alpha}, X^{s-1})-G_*(\bm{\alpha}^s, X^{s-1})$ across iterations where $t$ is the final round for F3I (the maximum number of rounds is set to $500$). All values are rounded to the closest second decimal place, except for the time round $t$, which is rounded to the closest integer. Theorem~\ref{thm:regret_f3i} states that $\max_{\bm{\alpha} \in \triangle_K} \sum_{s=1}^t G_*(\bm{\alpha}, X^{s-1})-G_*(\bm{\alpha}^s, X^{s-1}) \leq C^\text{AH}\sqrt{t} + H^\text{miss}h^{-1}t$ with probability $1-1/2,500 \approx 0.9996$.}
    \label{tab:theorem2}
    \begin{tabular}{llrr}
    \toprule
   Missingness type &  $t$ & Cumulative regret on $G$ & $C_G^\text{AH}\sqrt{t} + H^\text{miss}h^{-1}t$\\
    \midrule
       MCAR  $\quad$(Assumption~\ref{as:mcar})  &  24$\pm$ 78 & 113.76$\pm$ 371.39 & 2,067.60$\pm$ 6,734.50\\
        MAR $\quad$~ (Assumption~\ref{as:mar})   & 40$\pm$ 111 & 153.06$\pm$ 417.40 & 3,542.96$\pm$ 9,659.89\\
         MNAR $\quad$(Assumption~\ref{as:mnar})  & 35$\pm$ 96 & 158.61$\pm$ 438.45 & 3,015.86$\pm$ 8,336.243\\
         \bottomrule
    \end{tabular}
\end{table}

\begin{figure}
    \centering
    \includegraphics[width=\textwidth]{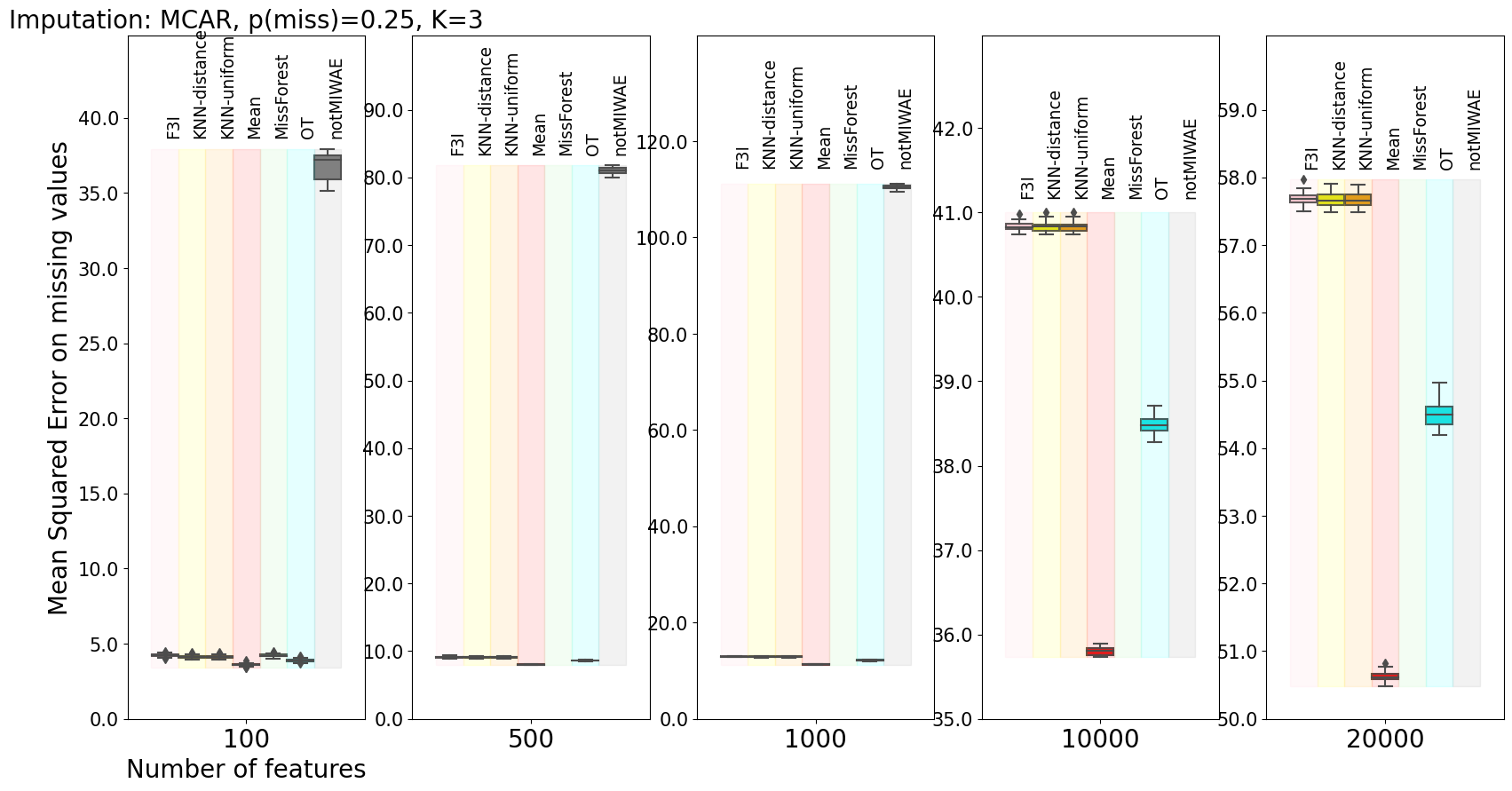}
    \caption{Imputation on $2$ synthetic data sets $\times$ $10$ different random seeds for generating missing values for F3I, K-nearest neighbor imputers~\cite{troyanskaya2001missing} (uniform or distance-based weights), mean imputation, MissForest~\cite{stekhoven2012missforest}, Optimal-Transport imputer~\cite{muzellec2020missing} and not-MIWAE~\cite{ipsen2021not}.}
    \label{fig:notmiwae_worst}
\end{figure}

\subsubsection{Empirical performance (single imputation task)}

Now we compare the MSE of F3I to its baselines on synthetic data sets generated with Algorithm~\ref{alg:data_generation}. Note that the definition of the MSE is slightly different from Definition~\ref{def:mse} as we only compute the gaps in the positions of missing values
\[ \overline{\mathcal{L}}^\text{MSE}(X^t, X^\star) \triangleq \frac{1}{N} \sum_{i \leq N} \frac{1}{|\{f \mid m^f_i = 1\}|} \sum_{f, m^f_i = 1} ((x^t)^f_i-(x^\star)^f_i)^2 \leq F\mathcal{L}^\text{MSE}(X^t, X^\star) \;. \]
We consider the following baselines: imputation by the mean value, random-forest-based imputation by MissForest~\cite{stekhoven2012missforest}, traditional KNN imputation~\cite{troyanskaya2001missing} with uniform and distance-based weights, Optimal Transport-based imputation~\cite{muzellec2020missing} and the Bayesian network approach not-MIWAE~\cite{ipsen2021not}. We consider a number of neighbors (for F3I and KNN) or of estimators (for MissForest) $K=3$, a number of features $F \in \{100;500;1,000;10,000;20,000\}$ with $N=50$ samples, and a missingness frequency $p^\text{miss} \propto \{0.10,0.25,0.50,0.75\}$. We generate $2$ random data sets and perform $10$ iterations of each algorithm on each data set, for a total of $20$ values per combination of parameters ($F$, missingness mechanism), where the missingness mechanism is either MCAR, MAR, or MNAR. 

\begin{figure}
    \centering
    \includegraphics[width=\textwidth]{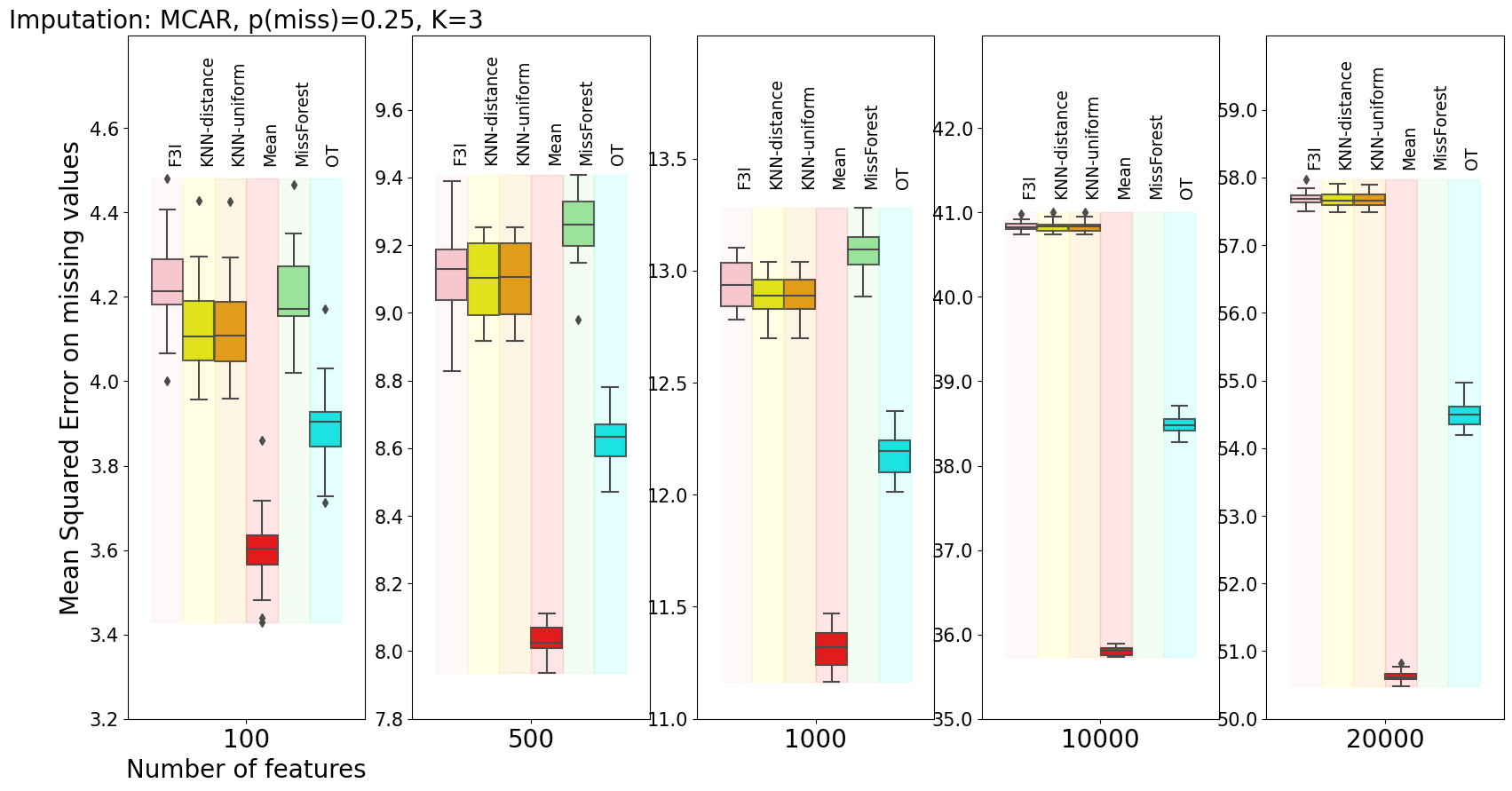}
    \includegraphics[width=\textwidth]{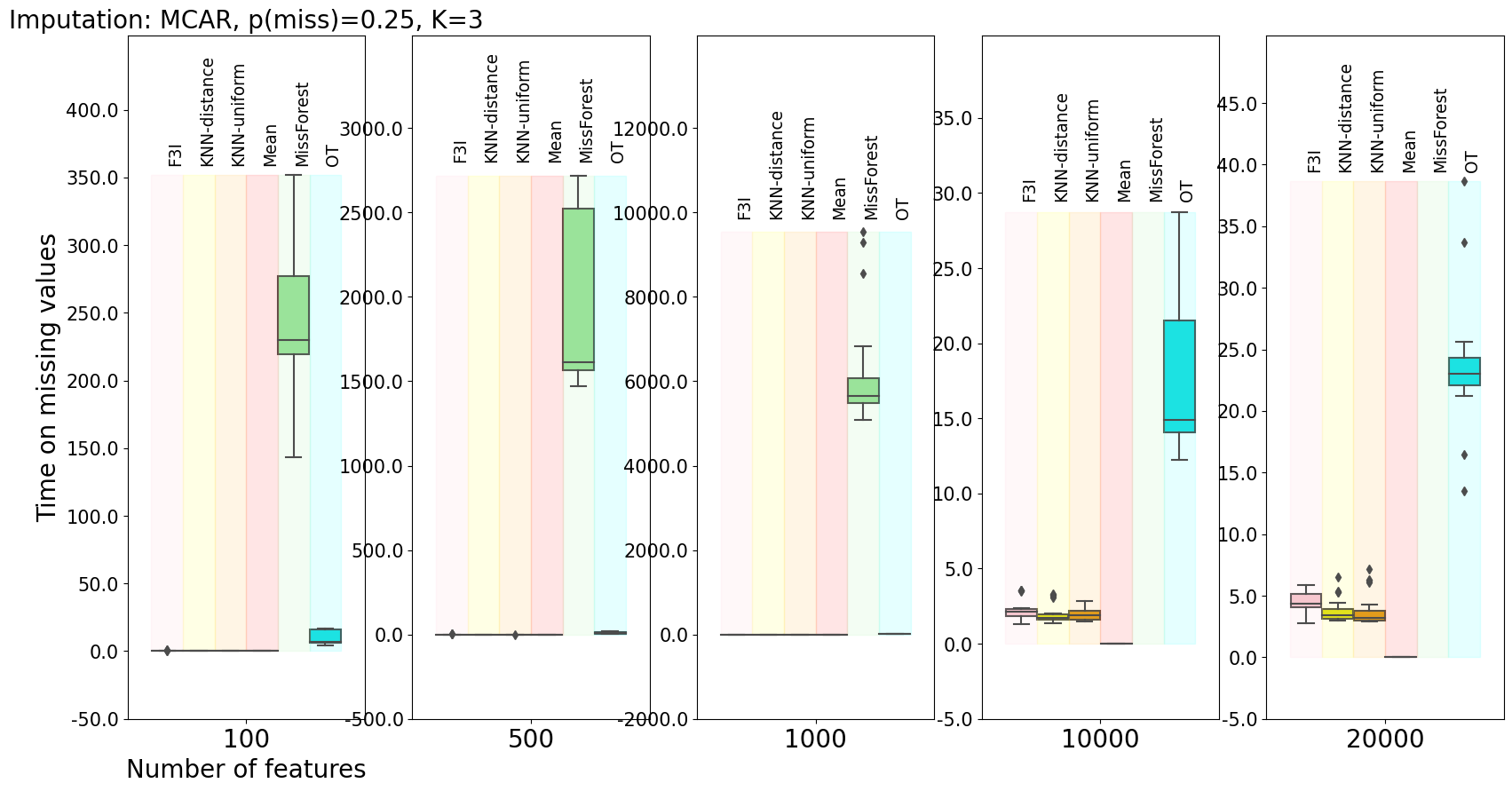}
    \caption{Imputation on $2$ synthetic data sets $\times$ $10$ different random seeds for generating missing values for F3I, K-nearest neighbor imputers~\cite{troyanskaya2001missing} (uniform or distance-based weights), mean imputation, MissForest~\cite{stekhoven2012missforest} and Optimal-Transport imputer~\cite{muzellec2020missing}.}
    \label{fig:synthetic_mcar_025}
\end{figure}

\begin{figure}
    \centering
    \includegraphics[width=\textwidth]{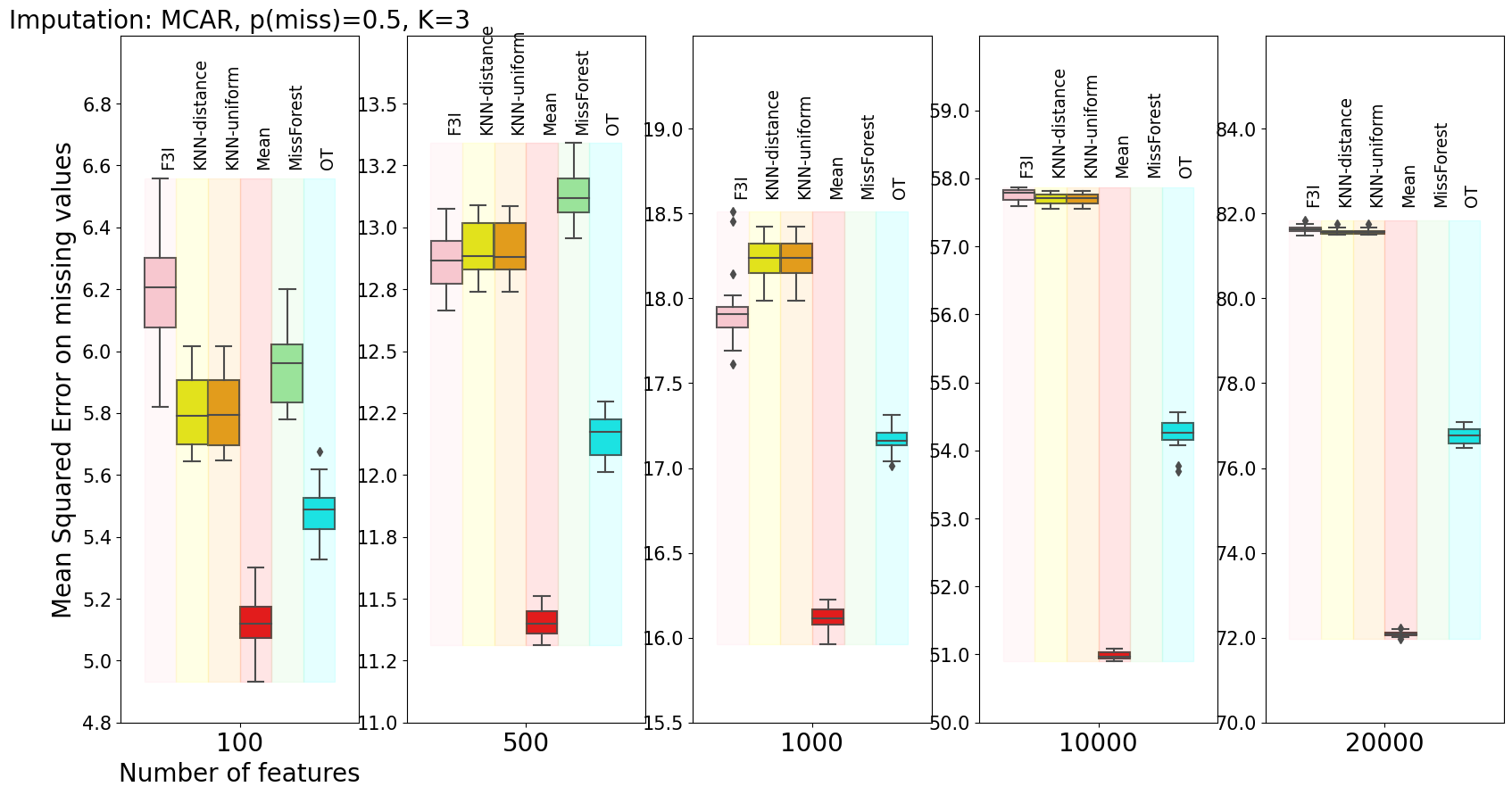}
    \includegraphics[width=\textwidth]{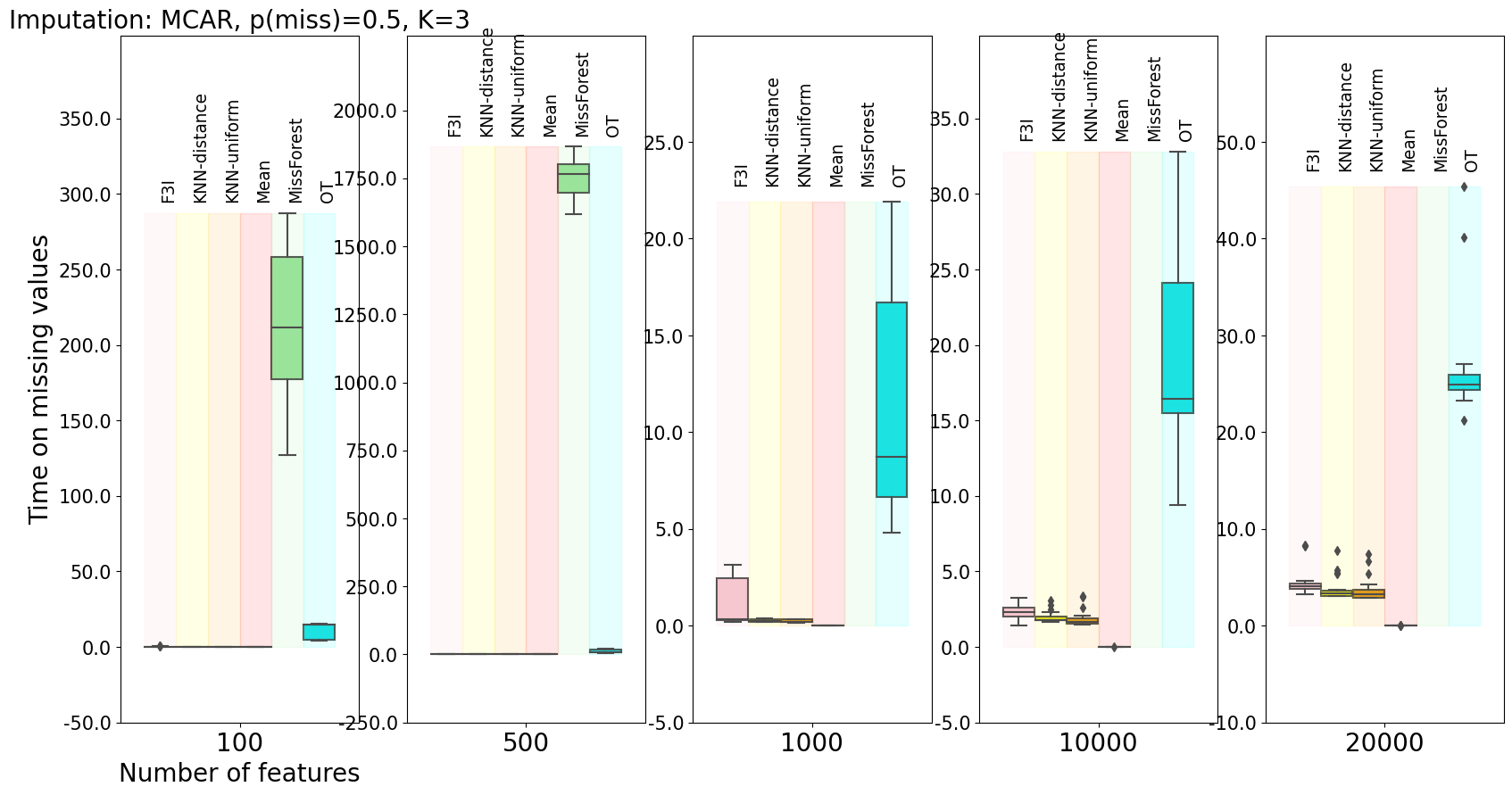}
    \caption{Imputation on $2$ synthetic data sets $\times$ $10$ different random seeds for generating missing values for F3I, K-nearest neighbor imputers~\cite{troyanskaya2001missing} (uniform or distance-based weights), mean imputation, MissForest~\cite{stekhoven2012missforest} and Optimal-Transport imputer~\cite{muzellec2020missing}.}
    \label{fig:synthetic_mcar_05}
\end{figure}

\begin{figure}
    \centering
    \includegraphics[width=\textwidth]{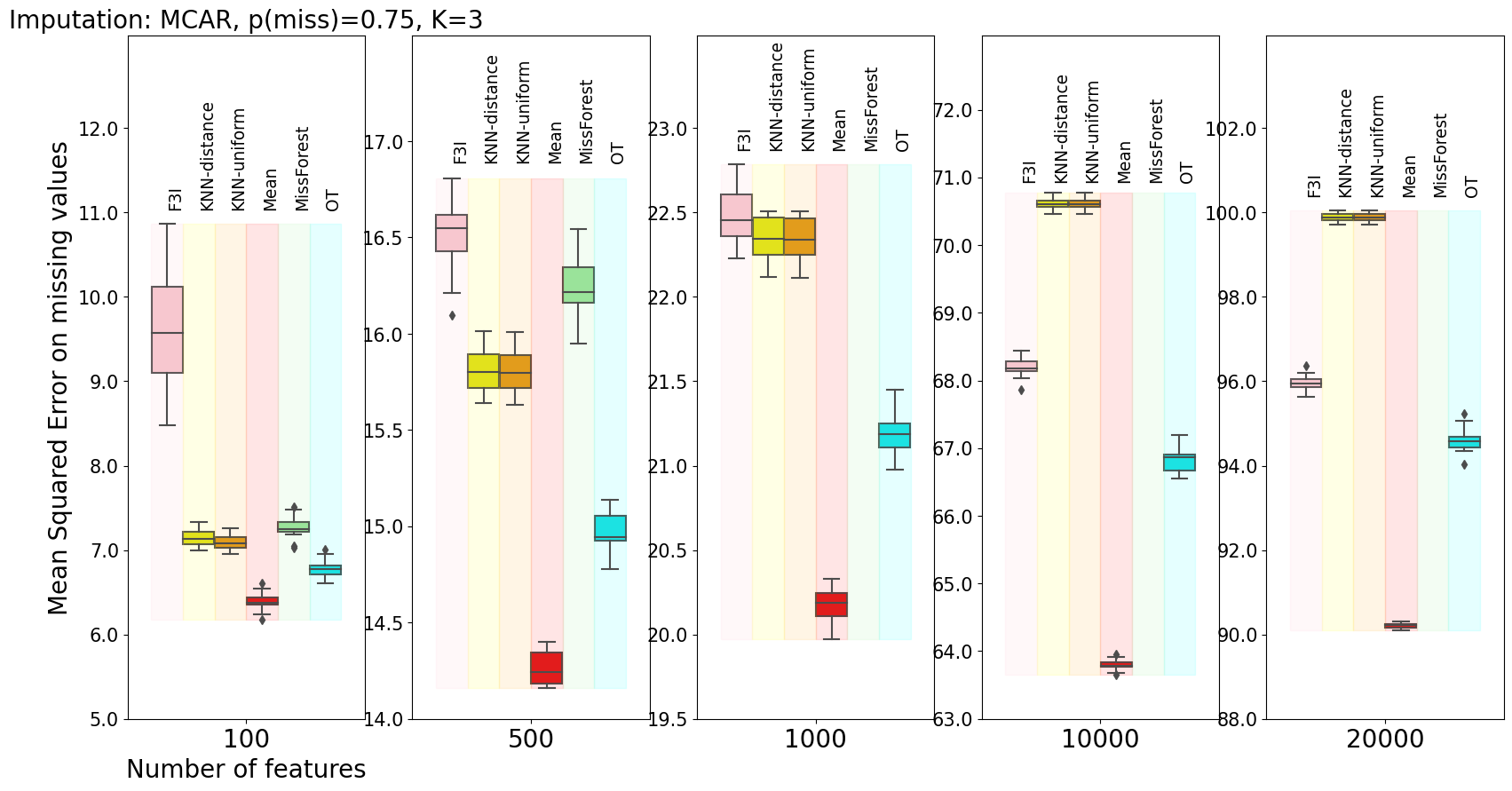}
    \includegraphics[width=\textwidth]{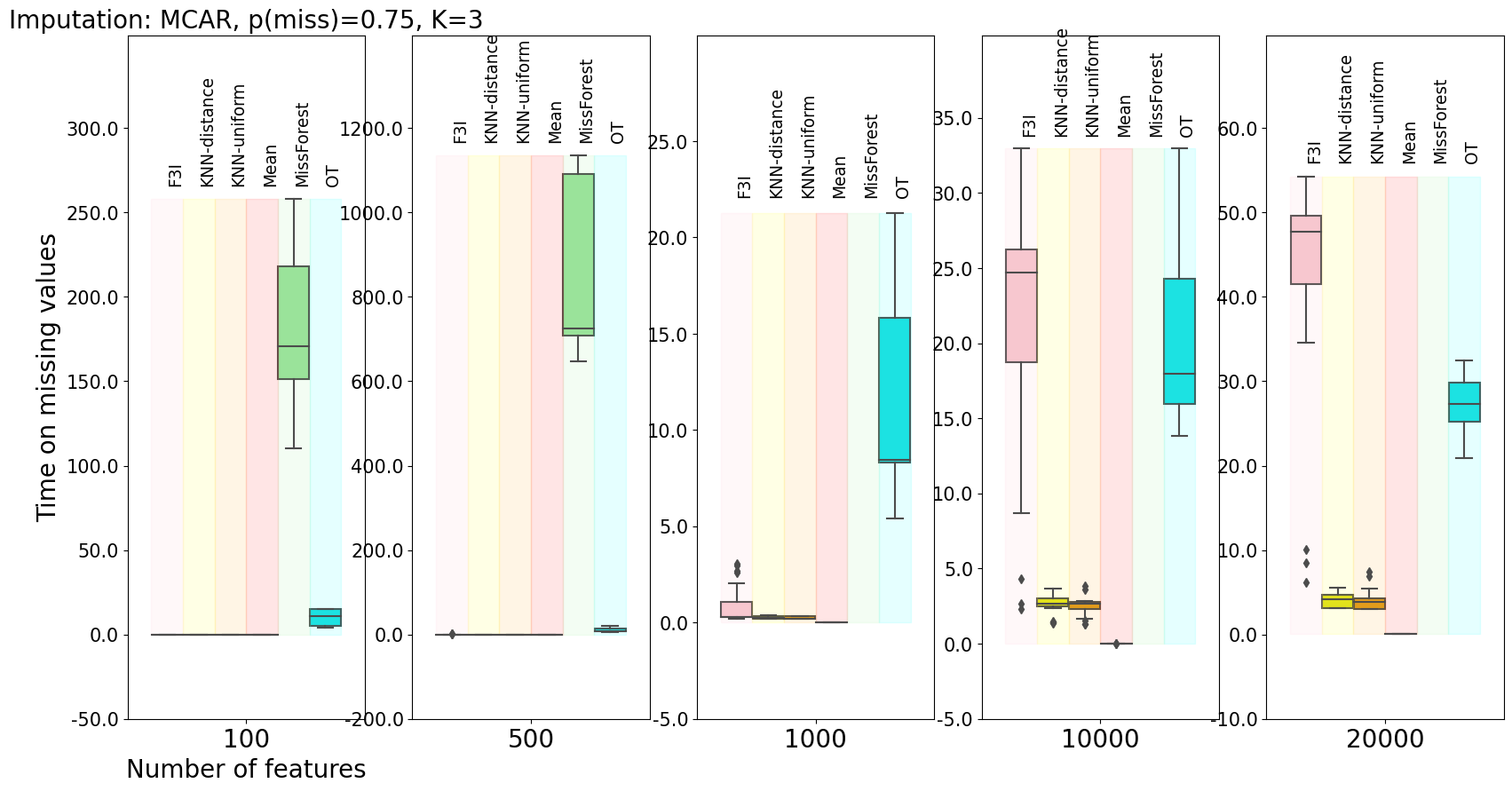}
    \caption{Imputation on $2$ synthetic data sets $\times$ $10$ different random seeds for generating missing values for F3I, K-nearest neighbor imputers~\cite{troyanskaya2001missing} (uniform or distance-based weights), mean imputation, MissForest~\cite{stekhoven2012missforest} and Optimal-Transport imputer~\cite{muzellec2020missing}.}
    \label{fig:synthetic_mcar_075}
\end{figure}

\begin{figure}
    \centering
    \includegraphics[width=\textwidth]{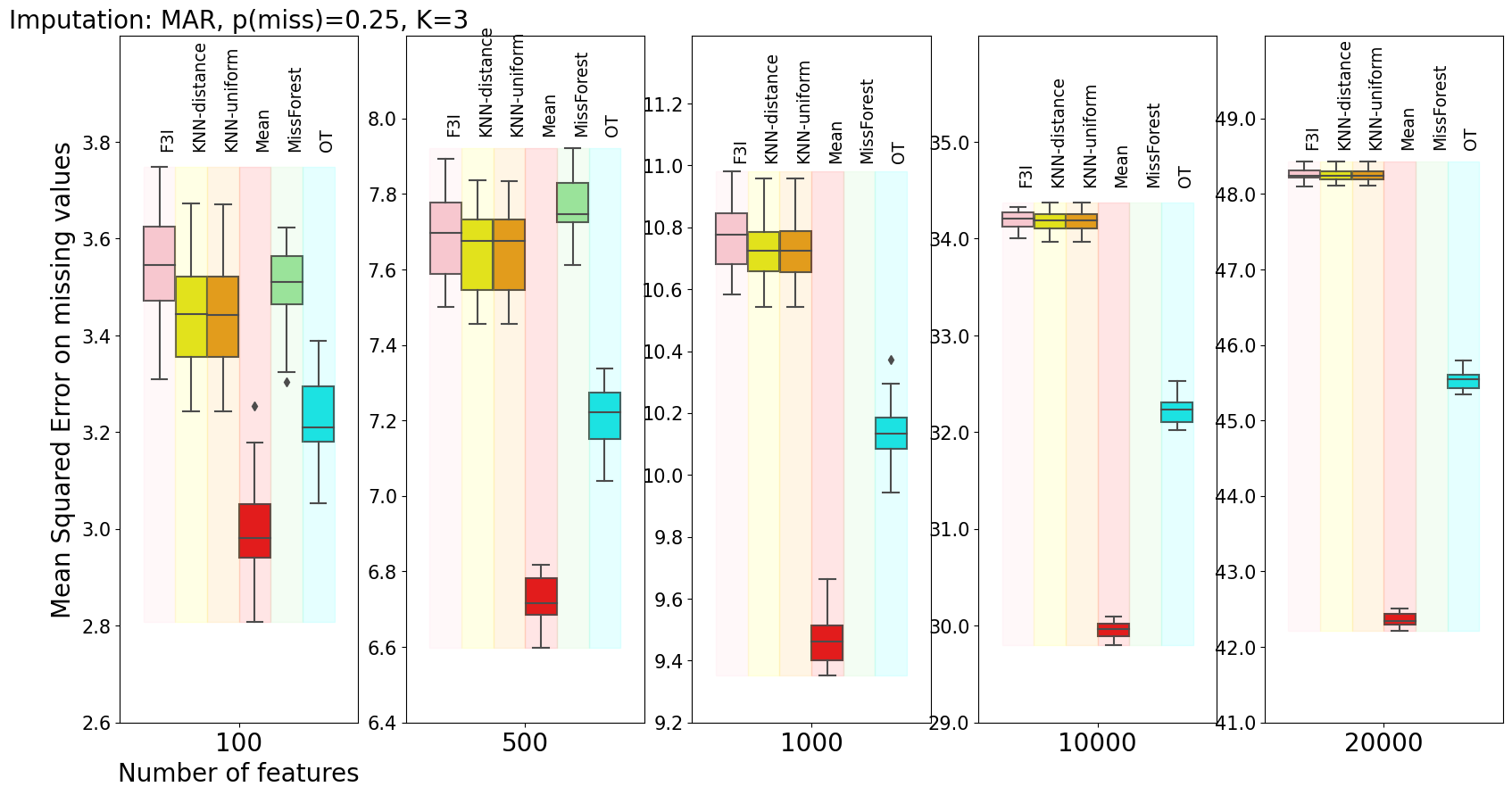}
    \includegraphics[width=\textwidth]{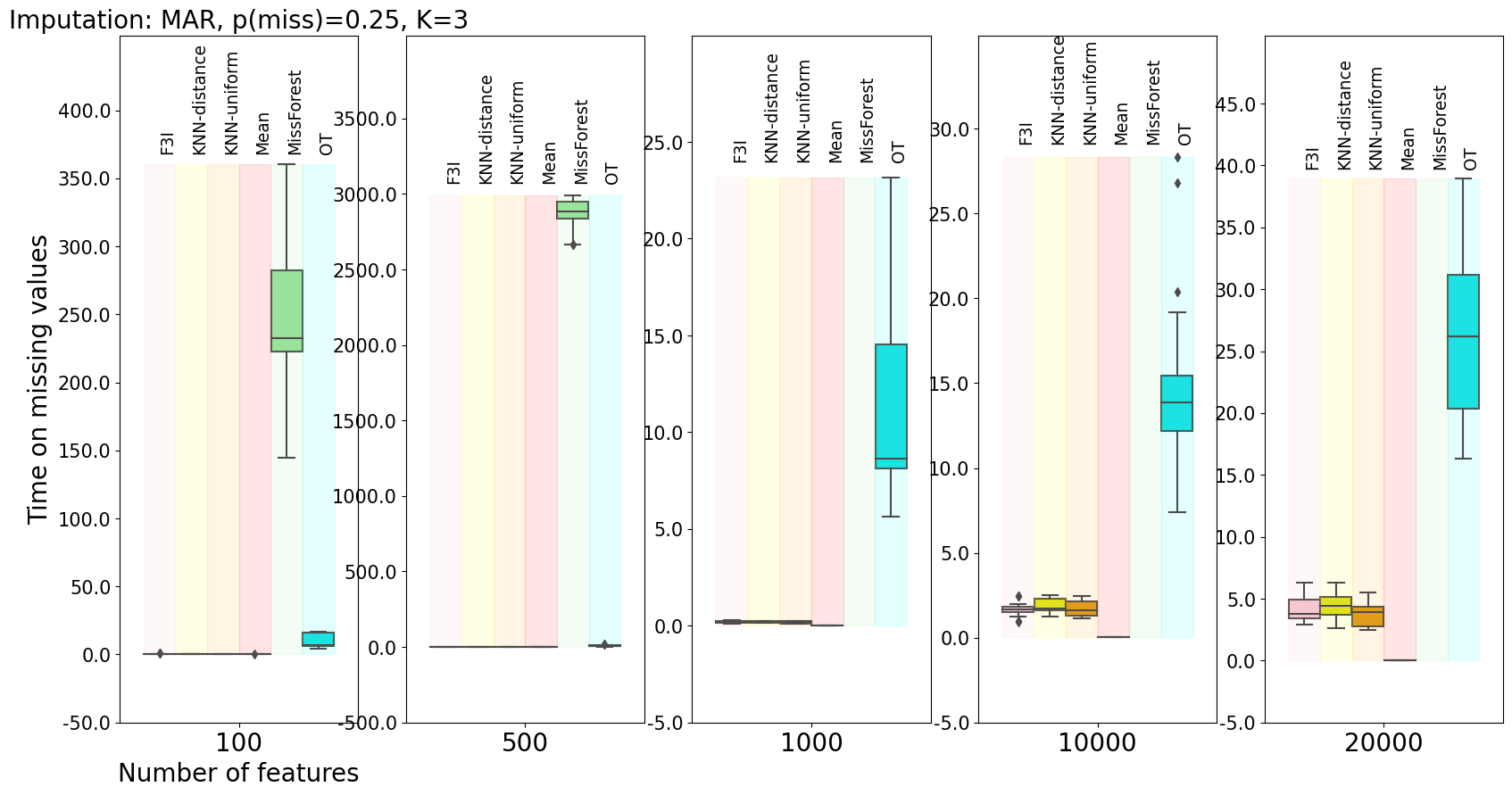}
    \caption{Imputation on $2$ synthetic data sets $\times$ $10$ different random seeds for generating missing values for F3I, K-nearest neighbor imputers~\cite{troyanskaya2001missing} (uniform or distance-based weights), mean imputation, MissForest~\cite{stekhoven2012missforest} and Optimal-Transport imputer~\cite{muzellec2020missing}.}
    \label{fig:synthetic_mar_025}
\end{figure}

\begin{figure}
    \centering
    \includegraphics[width=\textwidth]{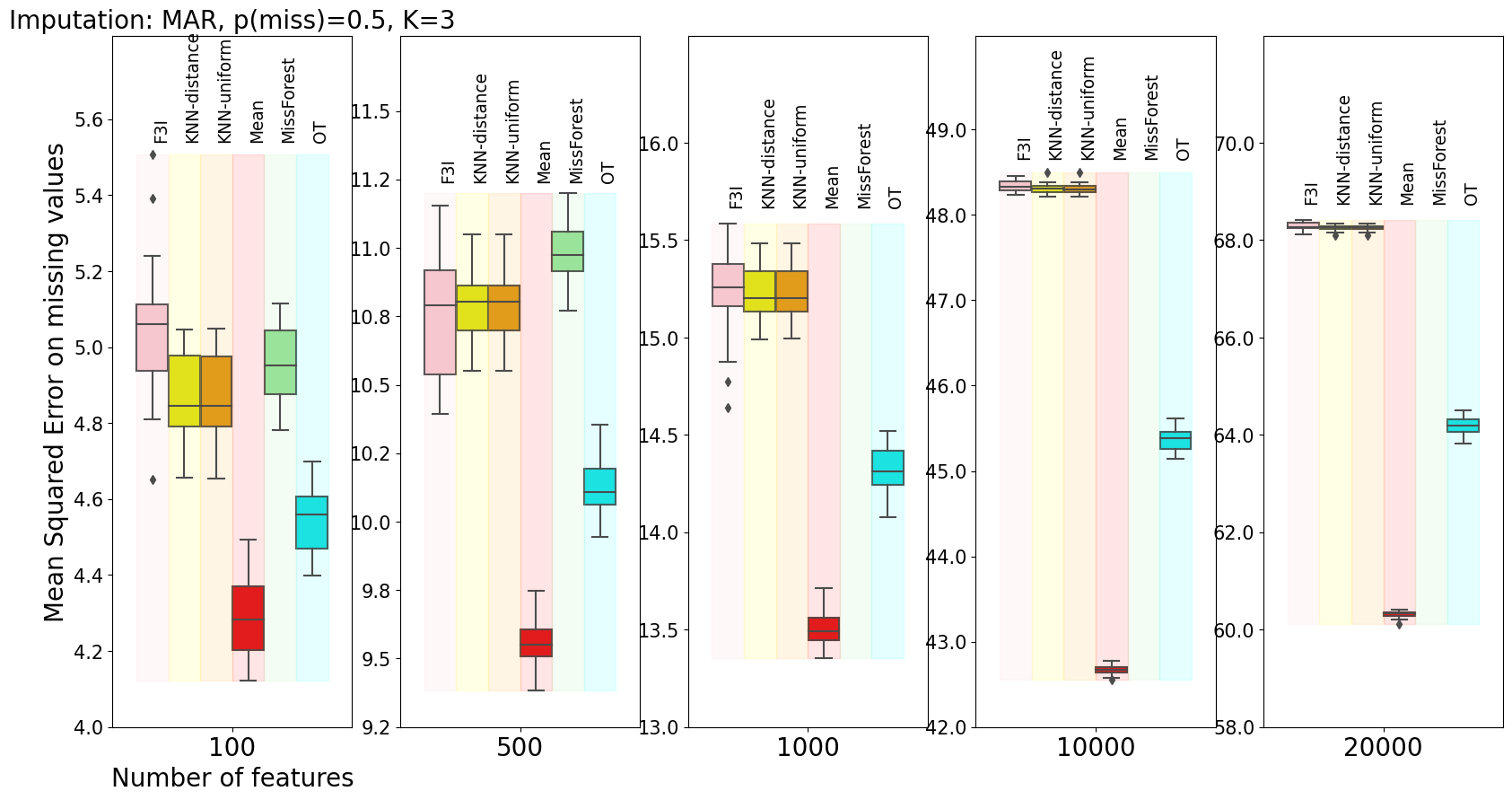}
    \includegraphics[width=\textwidth]{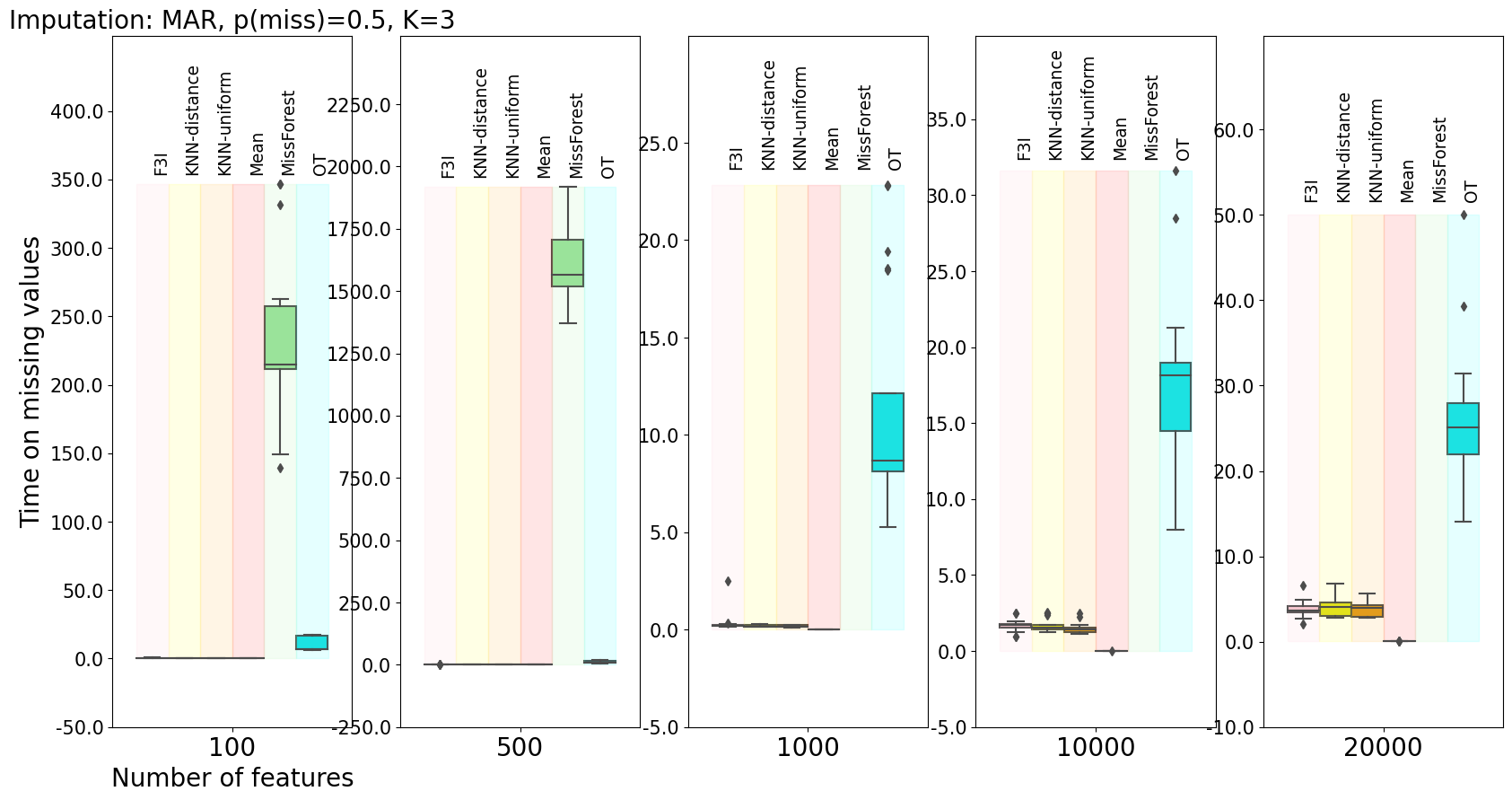}
    \caption{Imputation on $2$ synthetic data sets $\times$ $10$ different random seeds for generating missing values for F3I, K-nearest neighbor imputers~\cite{troyanskaya2001missing} (uniform or distance-based weights), mean imputation, MissForest~\cite{stekhoven2012missforest} and Optimal-Transport imputer~\cite{muzellec2020missing}.}
    \label{fig:synthetic_mar_05}
\end{figure}

\begin{figure}
    \centering
    \includegraphics[width=\textwidth]{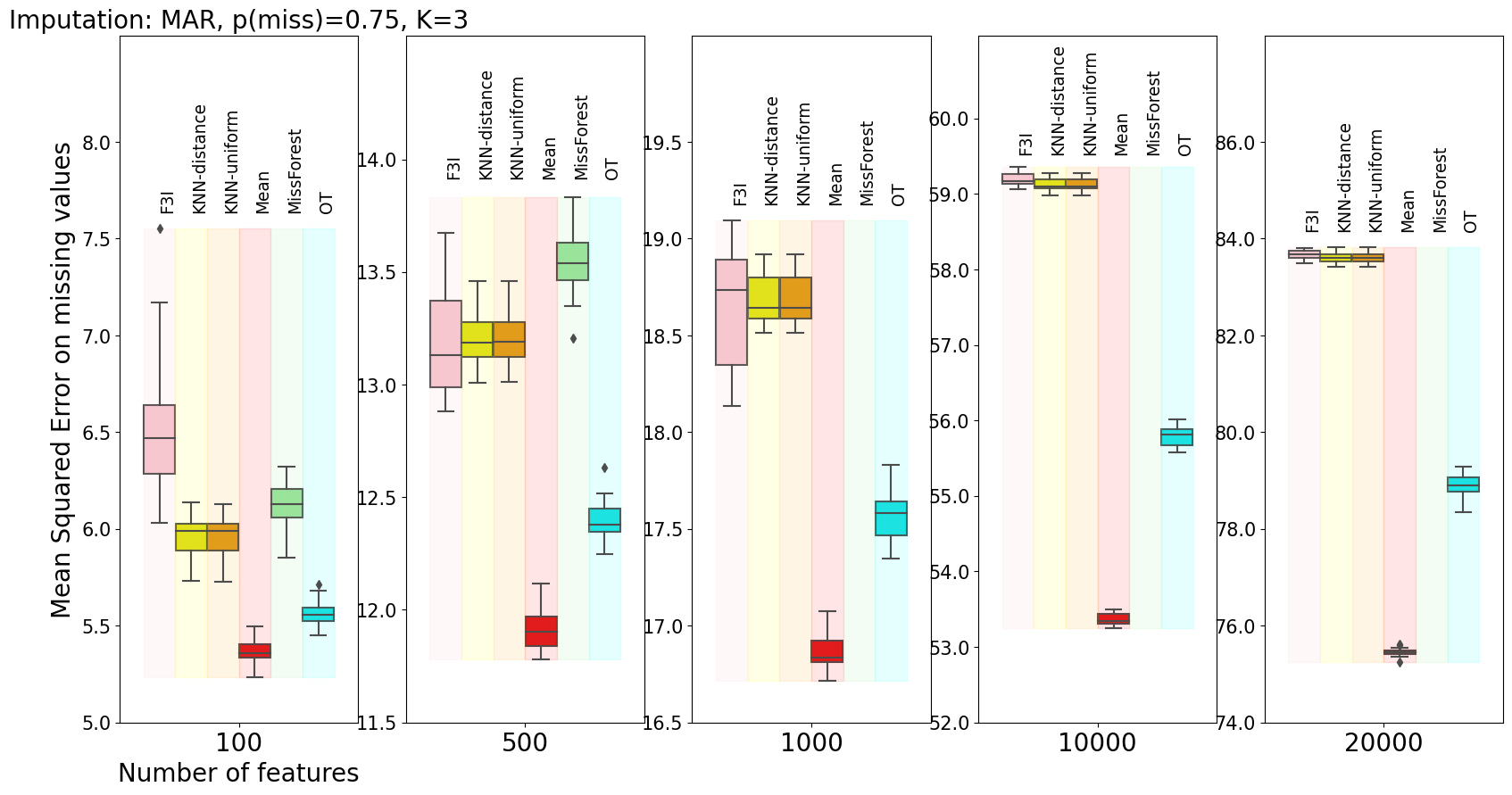}
    \includegraphics[width=\textwidth]{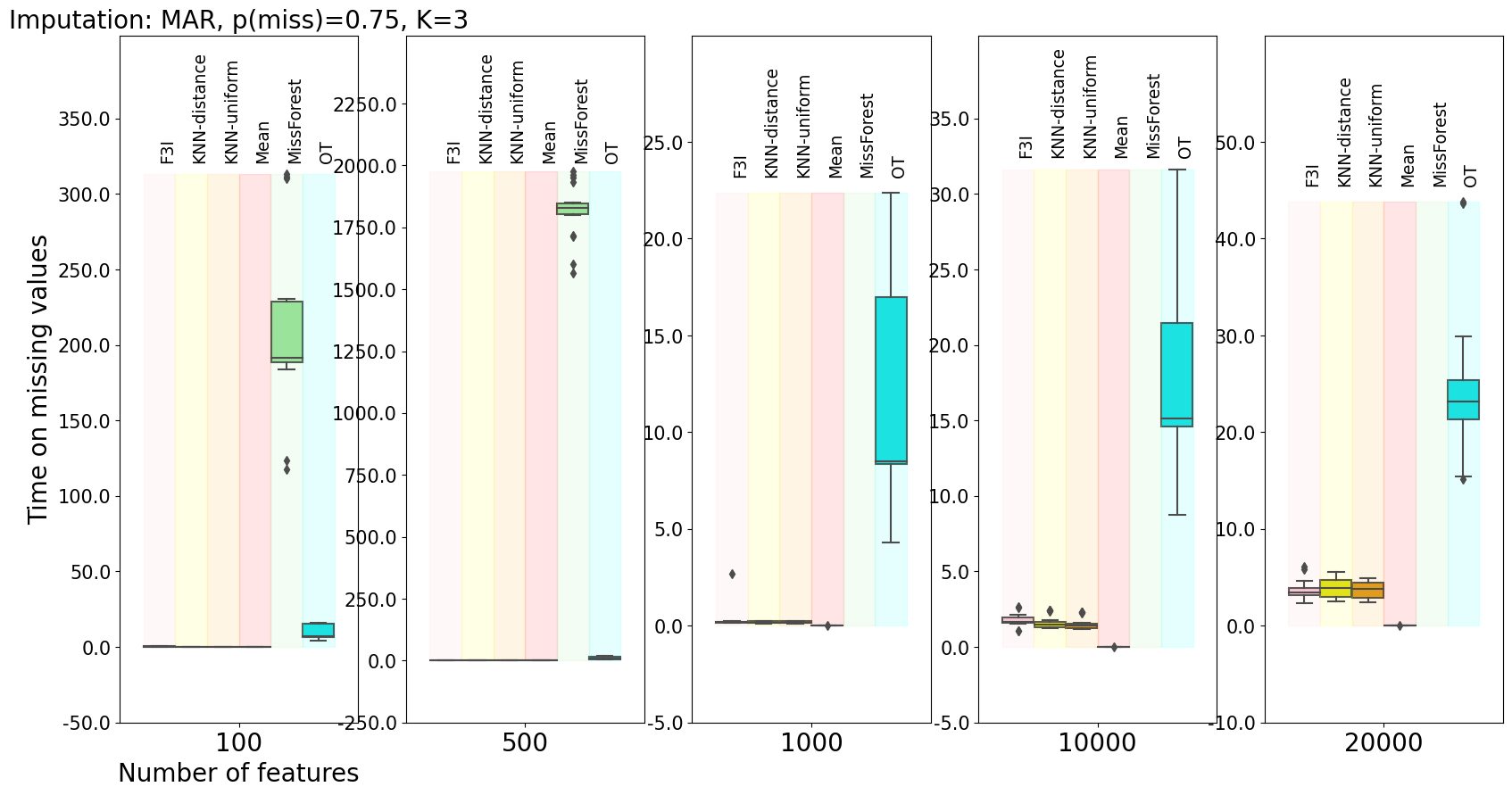}
    \caption{Imputation on $2$ synthetic data sets $\times$ $10$ different random seeds for generating missing values for F3I, K-nearest neighbor imputers~\cite{troyanskaya2001missing} (uniform or distance-based weights), mean imputation, MissForest~\cite{stekhoven2012missforest} and Optimal-Transport imputer~\cite{muzellec2020missing}.}
    \label{fig:synthetic_mar_075}
\end{figure}

\begin{figure}
    \centering
    \includegraphics[width=\textwidth]{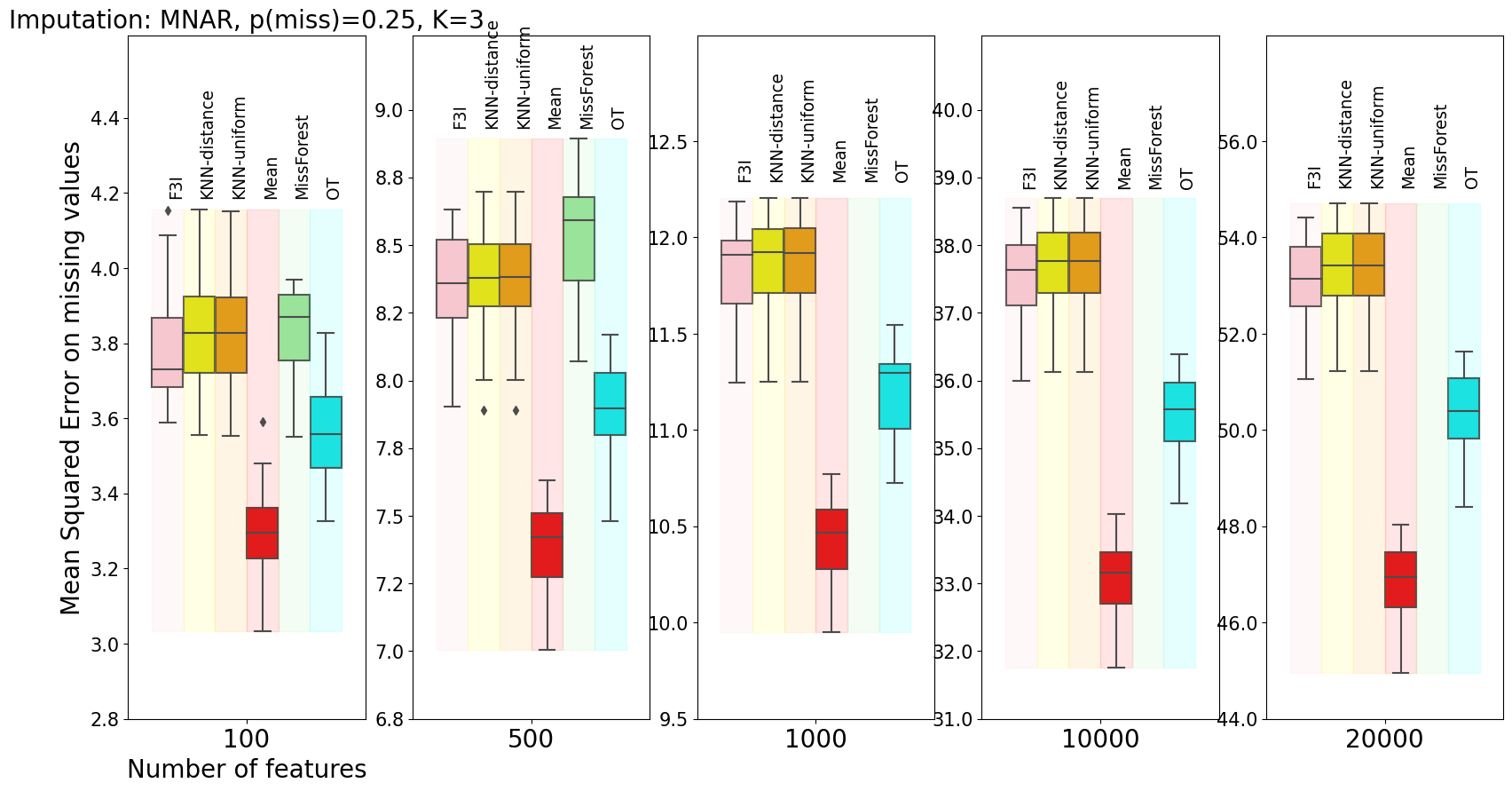}
    \includegraphics[width=\textwidth]{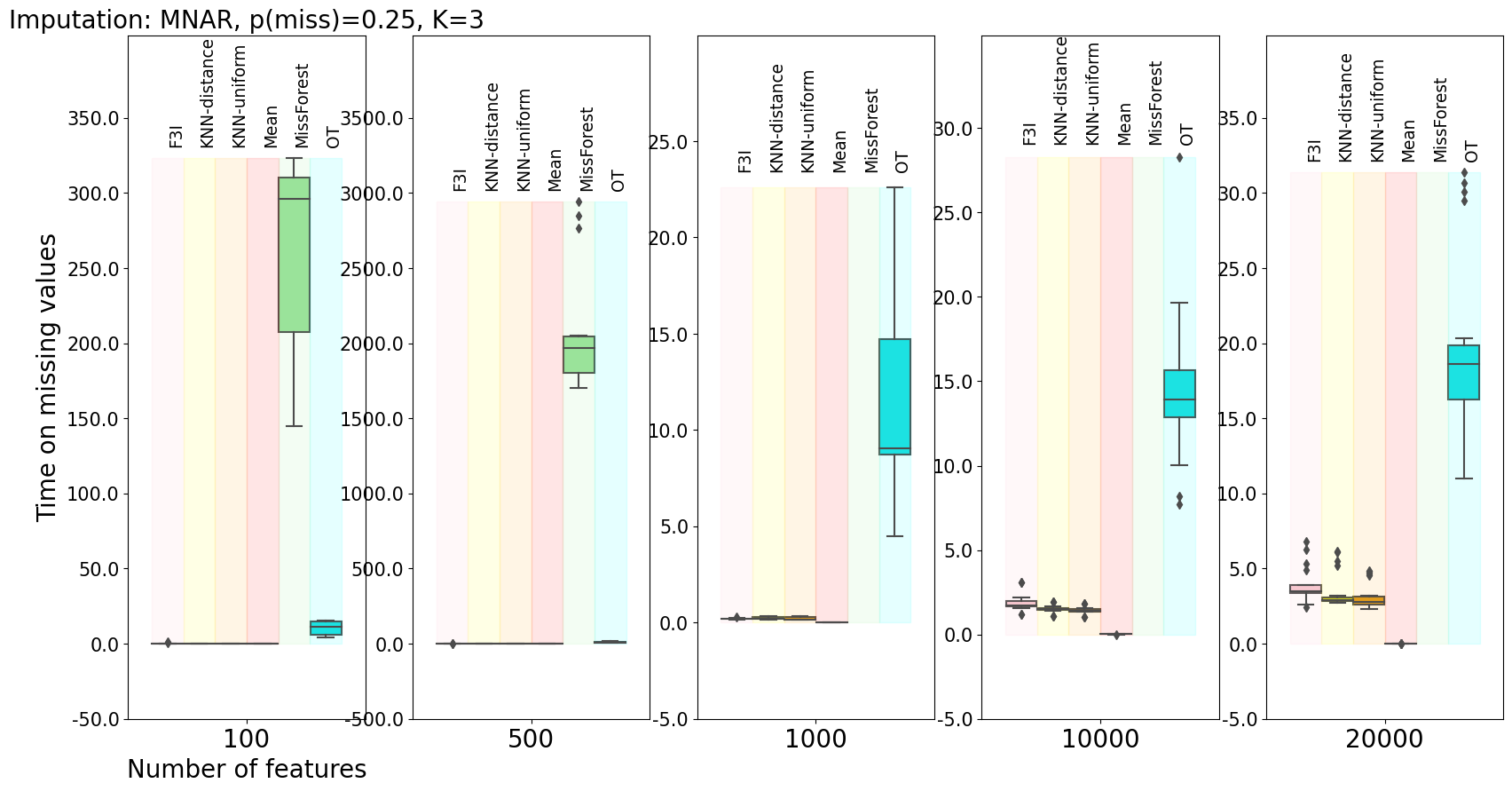}
    \caption{Imputation on $2$ synthetic data sets $\times$ $10$ different random seeds for generating missing values for F3I, K-nearest neighbor imputers~\cite{troyanskaya2001missing} (uniform or distance-based weights), mean imputation, MissForest~\cite{stekhoven2012missforest} and Optimal-Transport imputer~\cite{muzellec2020missing}.}
    \label{fig:synthetic_mnar_025}
\end{figure}

\begin{figure}
    \centering
    \includegraphics[width=\textwidth]{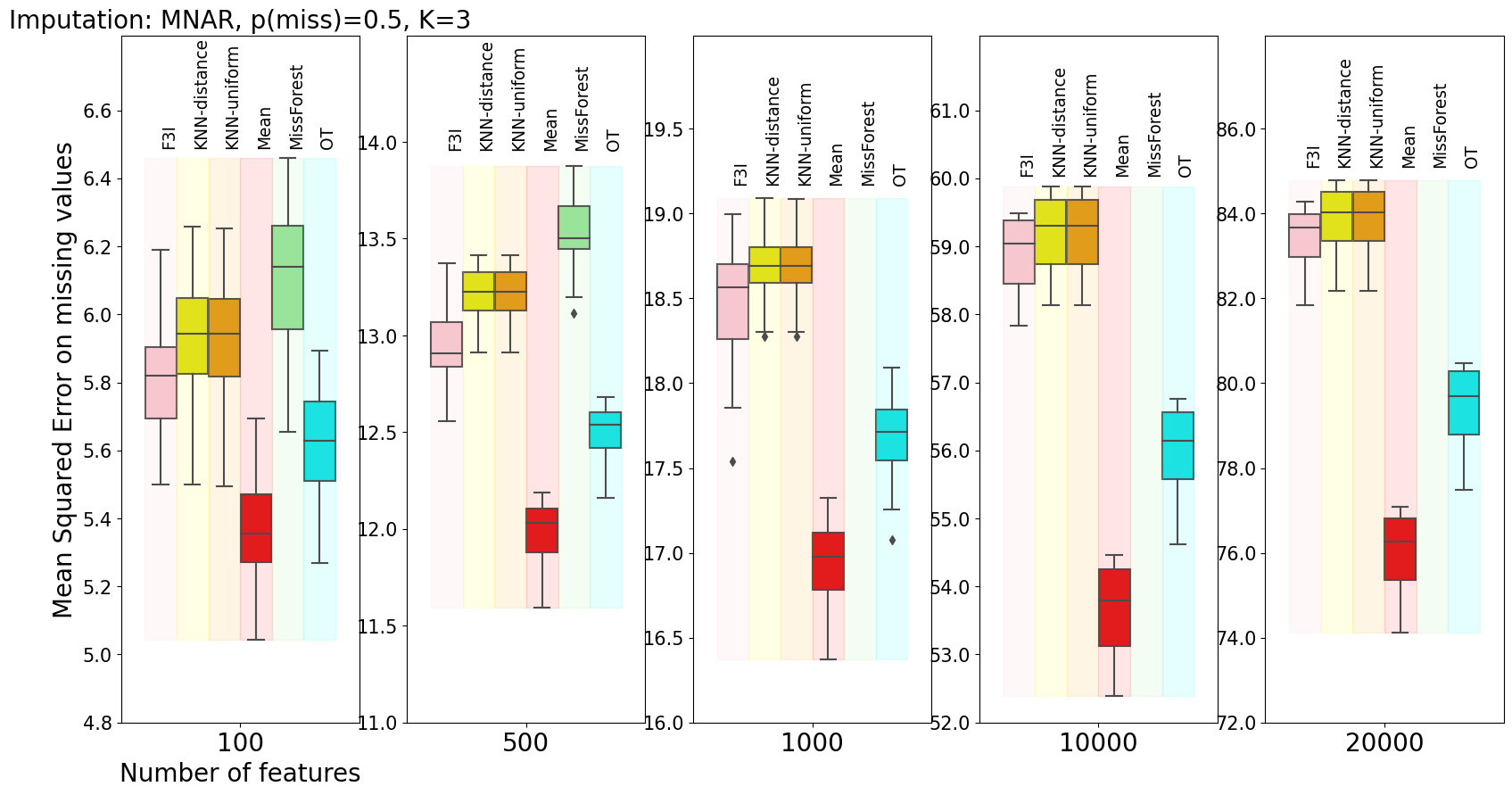}
    \includegraphics[width=\textwidth]{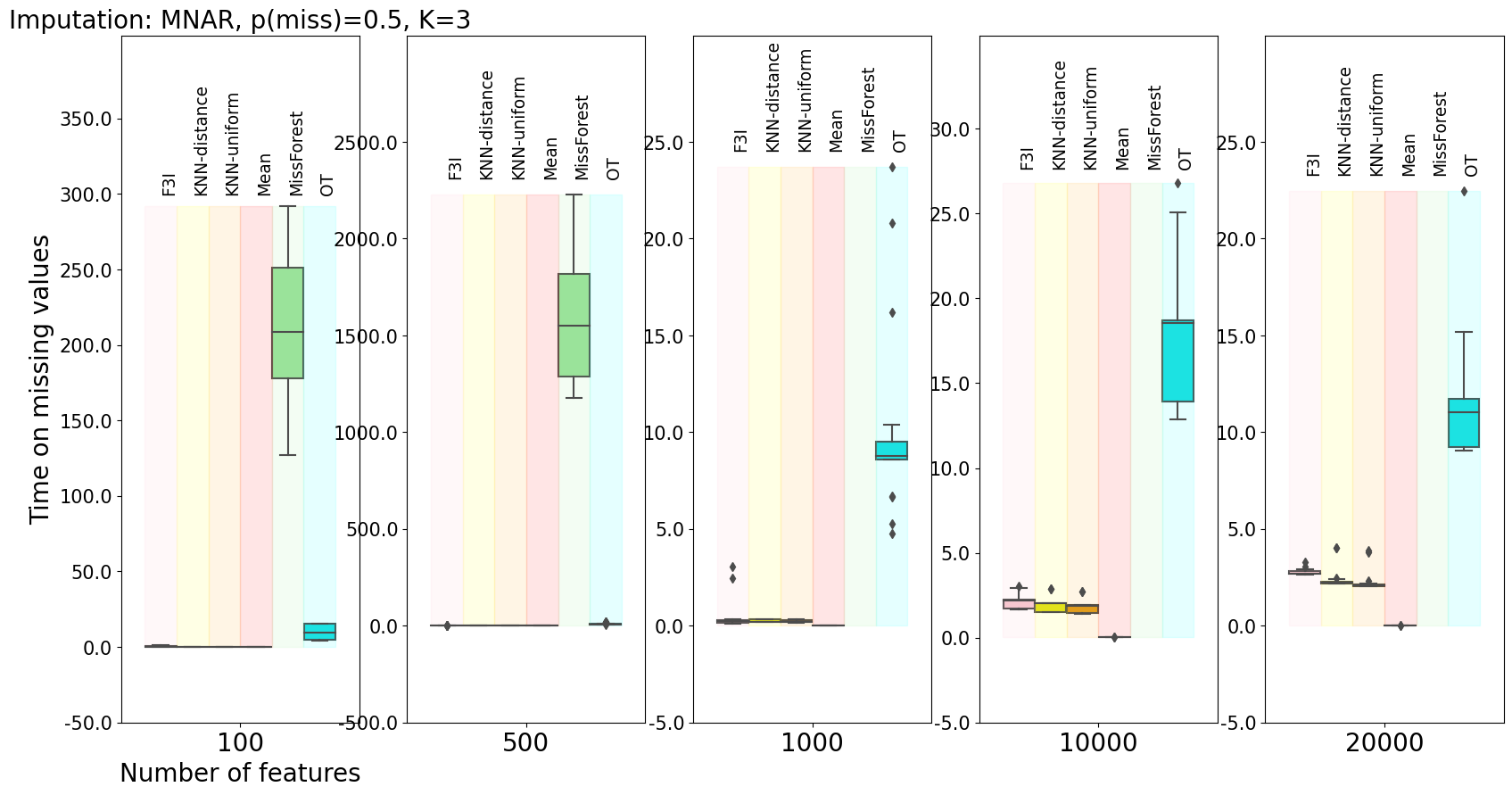}
    \caption{Imputation on $2$ synthetic data sets $\times$ $10$ different random seeds for generating missing values for F3I, K-nearest neighbor imputers~\cite{troyanskaya2001missing} (uniform or distance-based weights), mean imputation, MissForest~\cite{stekhoven2012missforest} and Optimal-Transport imputer~\cite{muzellec2020missing}.}
    \label{fig:synthetic_mnar_05}
\end{figure}

\begin{figure}
    \centering
    \includegraphics[width=\textwidth]{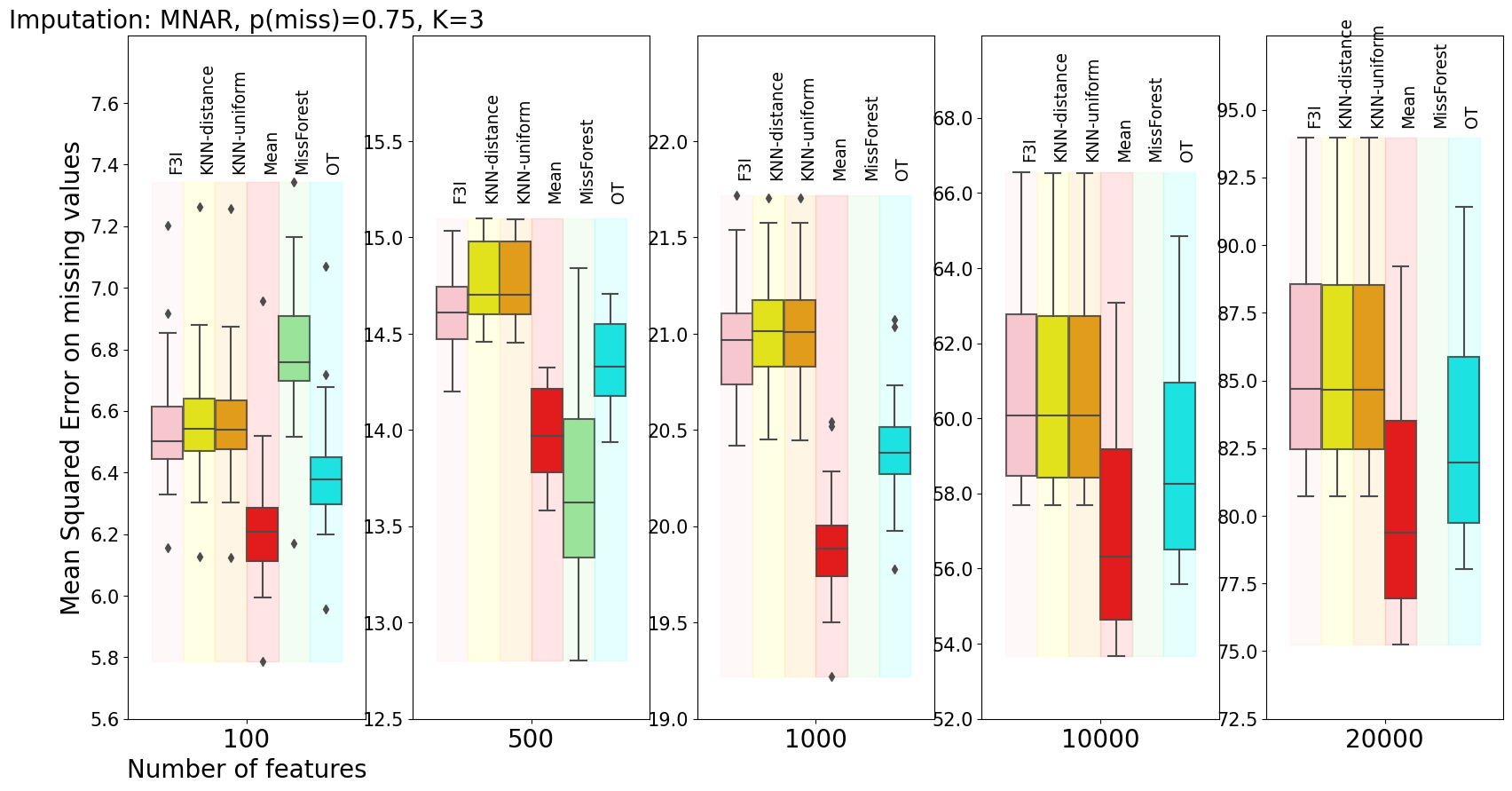}
    \includegraphics[width=\textwidth]{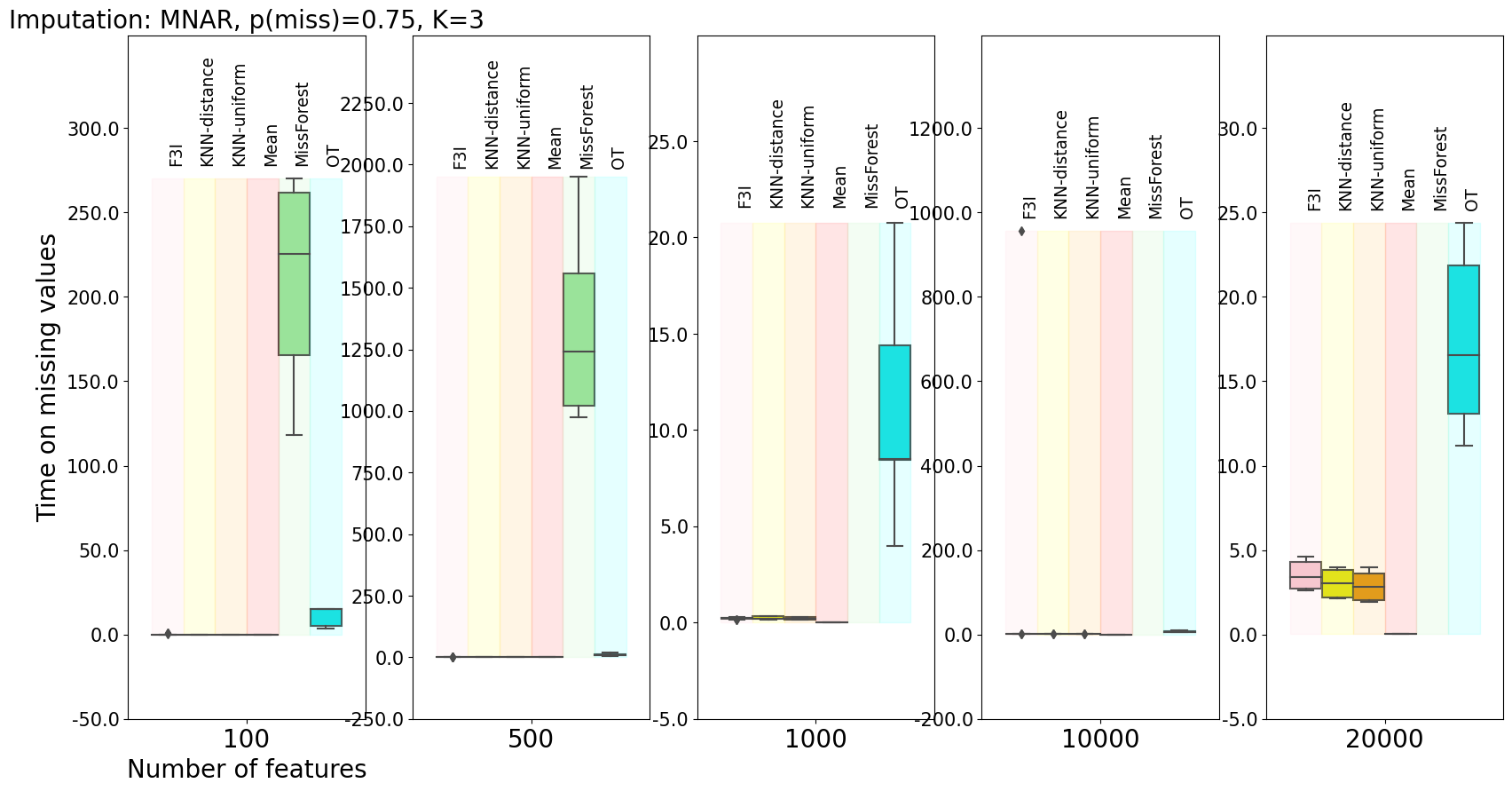}
    \caption{Imputation on $2$ synthetic data sets $\times$ $10$ different random seeds for generating missing values for F3I, K-nearest neighbor imputers~\cite{troyanskaya2001missing} (uniform or distance-based weights), mean imputation, MissForest~\cite{stekhoven2012missforest} and Optimal-Transport imputer~\cite{muzellec2020missing}.}
    \label{fig:synthetic_mnar_075}
\end{figure}

We first note that not-MIWAE has a significantly worse imputation error than all other algorithms; see Figure~\ref{fig:notmiwae_worst}. We then do not report the results for not-MIWAE in all other cases. Moreover, from $F=1,000$, the runtime of MissForest is too long to be run. The remainder of the boxplots for the mean squared error and imputation runtime across iterations and data sets can be found in Figures~\ref{fig:synthetic_mcar_025}-\ref{fig:synthetic_mnar_075}. 

Second, we note the superior performance of F3I, nearest neighbor, and mean imputers regarding the computational cost of imputation across missingness frequencies $p^\text{miss}$, missingness mechanisms (MCAR, MAR, MNAR), and numbers $F$ of features. This makes F3I a competitive approach when the number of features is huge (for instance, $F \in \{10,000; 20,000\}$). 

Third, as a general rule across missingness mechanisms and frequencies, the performance of F3I is close to the ones of other nearest-neighbor imputers and sometimes better when the number of features is large. It might be because F3I considers the same neighbors across features past the initial imputation step. This allows us to impute perhaps more reliably missing values, contrary to the other NN imputers where neighbor assignation is performed feature-wise. Moreover, we notice that the performance of F3I and all baselines are on par (that is, boxplots overlap) for data where the missing values are generated from a MNAR mechanism (Figures~\ref{fig:synthetic_mnar_025}-\ref{fig:synthetic_mnar_075}) regardless of the missing frequency. 

Fourth, F3I performs worst for data generated by a MCAR or a MAR mechanism, which is the setting where mean imputation works best. This might make sense, as F3I tries to capture a specific missingness pattern that depends on neighbors in the data set. The (perfectly) random pattern might be the most difficult to infer for F3I and other nearest-neighbor imputers.

\subsubsection{Validation of theoretical results (joint imputation-classification task)}

We implement the joint imputation-classification training with the log-loss function and sigmoid classifier $\ell(\bm{x}) \triangleq -y\log C_{\bm{\omega}}(\bm{x})$ mentioned in Example~\ref{lem:pcgrad_example} (Appendix~\ref{subapp:joint_training}), where $y \in \{0,1\}$ is the binary class associated with sample $\bm{x} \in \mathbb{R}^F$. To implement PCGrad-F3I, we chain the imputation phase by F3I with an MLP, which returns logits. At time $t$, the imputation part applies at a fixed set of parameters $\bm{\omega}^t$  with the learner losses defined in Equation~\eqref{eq:pcgrad_update}. 
We construct the synthetic data sets for classification as follows. 
We draw two random matrices in the synthetic data set as in Algorithm~\ref{alg:data_generation} corresponding to the item and user feature matrices. We assign binary class labels to each item-user pair using a K-means++ algorithm~\cite{arthur2006k} with $K=2$ clusters on the item-user concatenated feature vectors. 

We first validate the upper bound on the cumulative regret in Theorem~\ref{thm:regret_f3i_joint} for all three missingness mechanisms we studied, similar to what was done for Theorem~\ref{thm:regret_f3i}. As in our proofs (see Appendix~\ref{subapp:joint_training}), we consider the log-loss $\ell$ with the sigmoid classifier $C_{\bm{\omega}}$ and $\beta=0.5$. We consider a missingness frequency of $50\%$. To obtain the value of $\arg\,\max_{\bm{\alpha} \in \triangle_K} \sum_{s=1}^t \overline{G}(\bm{\alpha}, X^{s-1})$, where $\overline{G}$ is defined as
\[\overline{G} : \bm{\alpha} \in \triangle_K, X \in \mathbb{R}^{N \times F} \mapsto (1-\beta)G_\star(\bm{\alpha}, X)-\frac{\beta}{N}\sum_{i \leq N}\ell(\bm{x}_i(\bm{\alpha}))\;,\]
we solve the following optimization problem by solving the related convex problem with function \texttt{minimize} in Python package \texttt{scipy.optimize}~\cite{2020SciPy-NMeth}, considering the $(X^{s-1})_{s \leq t}$ and parameter of the sigmoid classifier $C_\omega$ in the last epoch in PCGrad-F3I
\[ \min_{\bm{\alpha} \in \mathbb{R}^K} - \sum_{s=1}^t \overline{G}(\bm{\alpha}, X^{s-1}) \quad \text{ such that } \quad \forall k \leq K, \ \alpha_k \geq 0 \quad \text{ and } \quad \sum_{k \leq K} \alpha_k = 1\;.\]

\begin{figure}
    \centering
   \includegraphics[width=0.32\linewidth]{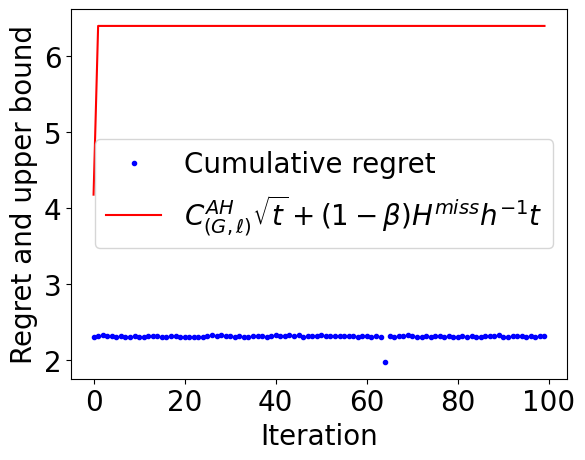}
     \includegraphics[width=0.32\linewidth]{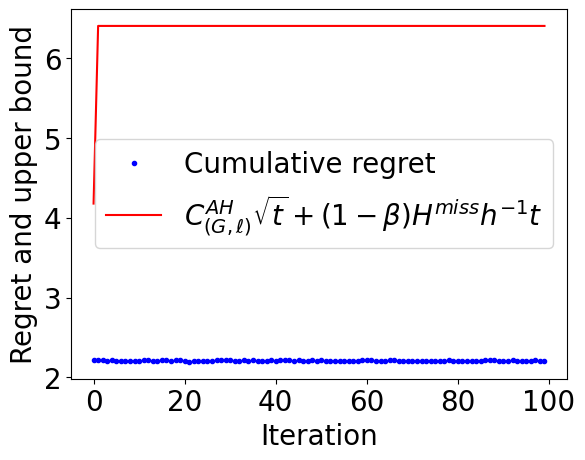}
     \includegraphics[width=0.32\linewidth]{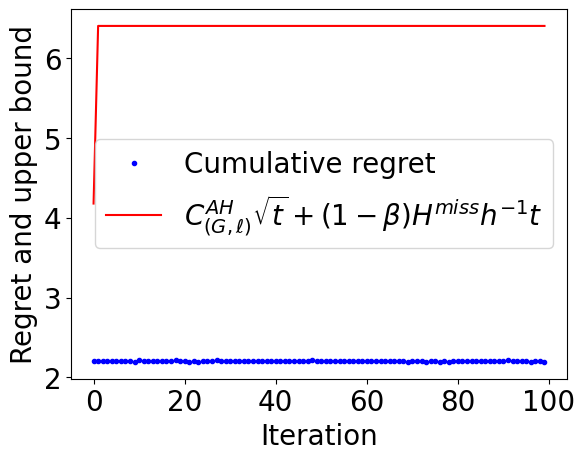}
    \caption{Cumulative regret for F3I and upper bound from Theorem~\ref{thm:regret_f3i_joint} across $100$ iterations for $\beta=0.5$. The blue points are always below the red lines. Left: MCAR setting. Center: MAR setting. Right: MNAR setting.}
    \label{fig:theorem3}
\end{figure}

The gradient and the Hessian matrix of the objective function $\overline{G}$ with respect to $\bm{\alpha}$ are obtained by combining the results from Lemmas~\ref{lem:gradient_G}, Lemma~\ref{lem:hessian_G} and Section~\ref{subapp:joint_training}. Figure~\ref{fig:theorem3} and Table~\ref{tab:theorem3} indeed show that the upper bound reliably holds on the cumulative regret for $\overline{G}$.

\begin{table}[tb]
    \centering
    \caption{Empirical validation of Theorem~\ref{thm:regret_f3i_joint} by comparing the value of the upper bound and the average and standard deviation value of the cumulative regret $\max_{\bm{\alpha} \in \triangle_K} \sum_{s=1}^t \overline{G}(\bm{\alpha}, X^{s-1})-\overline{G}(\bm{\alpha}^s, X^{s-1})$ across iterations where $t$ is the final round for F3I in the last epoch. All values are rounded to the closest second decimal place. The time round is fixed to $T=3$ in PCGrad-F3I and $\beta=0.5$. Theorem~\ref{thm:regret_f3i} states that $\max_{\bm{\alpha} \in \triangle_K} \sum_{s=1}^t \overline{G}(\bm{\alpha}, X^{s-1})-\overline{G}(\bm{\alpha}^s, X^{s-1}) \leq C^\text{AH}_{(G,\ell)}\sqrt{t} + (1-\beta)H^\text{miss}h^{-1}t$ with probability $1-1/2,500 \approx 0.9996$.}
    \label{tab:theorem3}
    \begin{tabular}{llrr}
    \toprule
   Missingness type &  $t$ & Cumulative regret on $\overline{G}$ & $C_{(G,\ell)}^\text{AH}\sqrt{t} + (1-\beta)H^\text{miss}h^{-1}t$\\
    \midrule
       MCAR  $\quad$(Assumption~\ref{as:mcar})  &  3 & 2.31 $\pm$0.03 & 6.38 $\pm$0.22\\
        MAR $\quad$~ (Assumption~\ref{as:mar})   & 3 & 2.21 $\pm$0.01 & 6.37 $\pm$0.22\\
         MNAR $\quad$(Assumption~\ref{as:mnar})  & 3 & 2.21 $\pm$0.00 & 6.38 $\pm$0.22\\
         \bottomrule
    \end{tabular}
\end{table}

\subsubsection{Empirical performance (joint imputation-classification task)}

Then, we compare the performance of PCGrad-F3I with adding a NeuMiss block~\cite{le2020neumiss} and with performing an imputation by the mean (``Mean'') before the classifier. Similarly to PCGrad-F3I, we chain an imputation part with an MLP classifier, which returns logits. In the baselines, at time $t$, the imputation part applies at a fixed set of parameters $\bm{\omega}^t$ an imputation by the mean, or a shared-weights NeuMiss block, as introduced in~\cite{le2021sa}. The criterion for training the model is the log loss, and we split the samples into training ($70\%$), validation ($20\%$), and testing ($10\%$) sets, where the former two sets are used for training the MLP, and the performance metrics are computed on the latter set. We consider the classical Area Under the Curve (AUC) on the test set (hidden during training) as the performance metric for the binary classification task. In this section, we set the number of imputation rounds in F3I to $T=2$, and we train the MLPs for each imputation approach with the hyperparameter values reported in Table~\ref{tab:hyperparam_joint}. 

\begin{figure}
    \centering
    \includegraphics[width=0.75\textwidth]{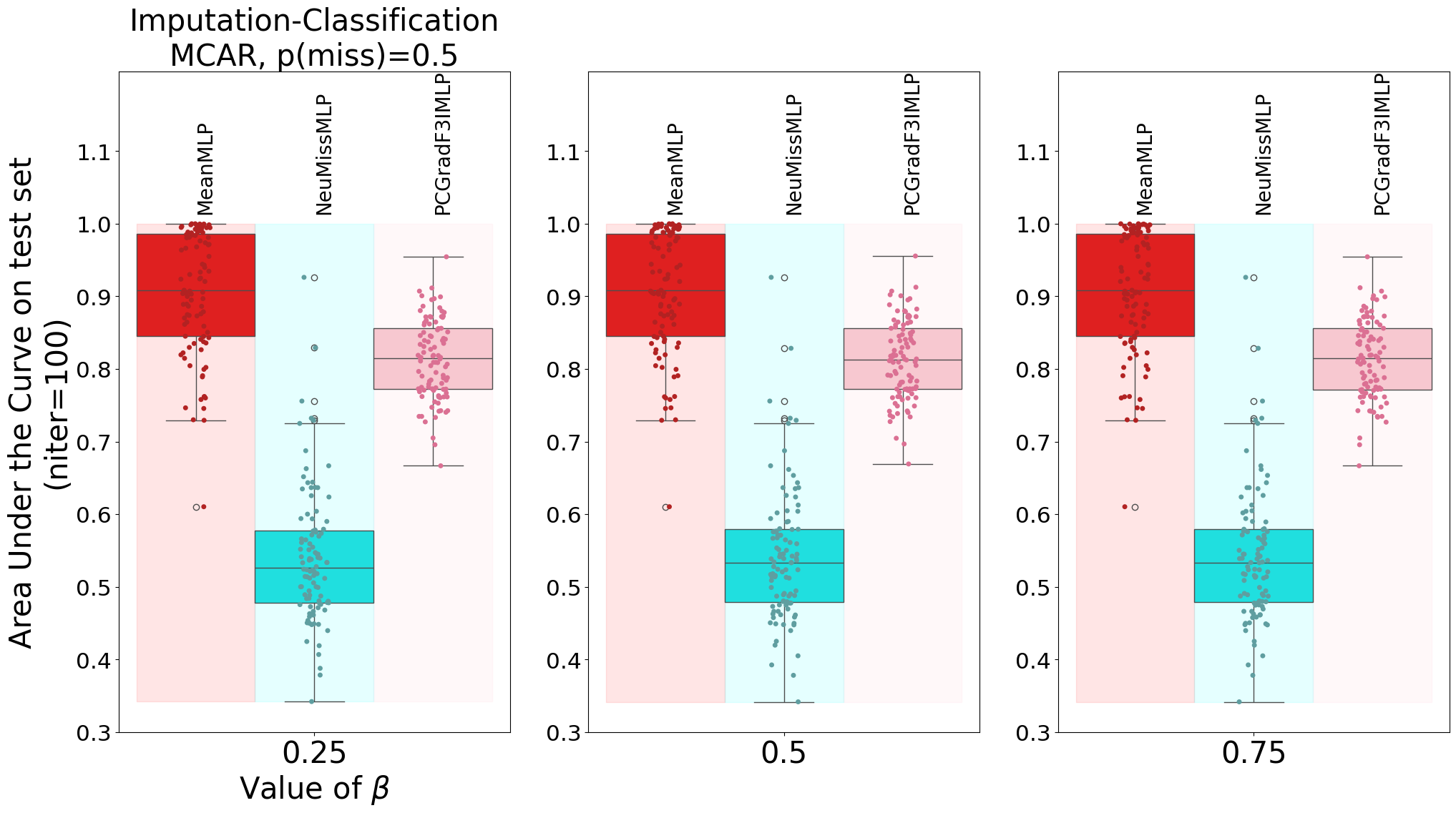}
    \includegraphics[width=0.75\textwidth]{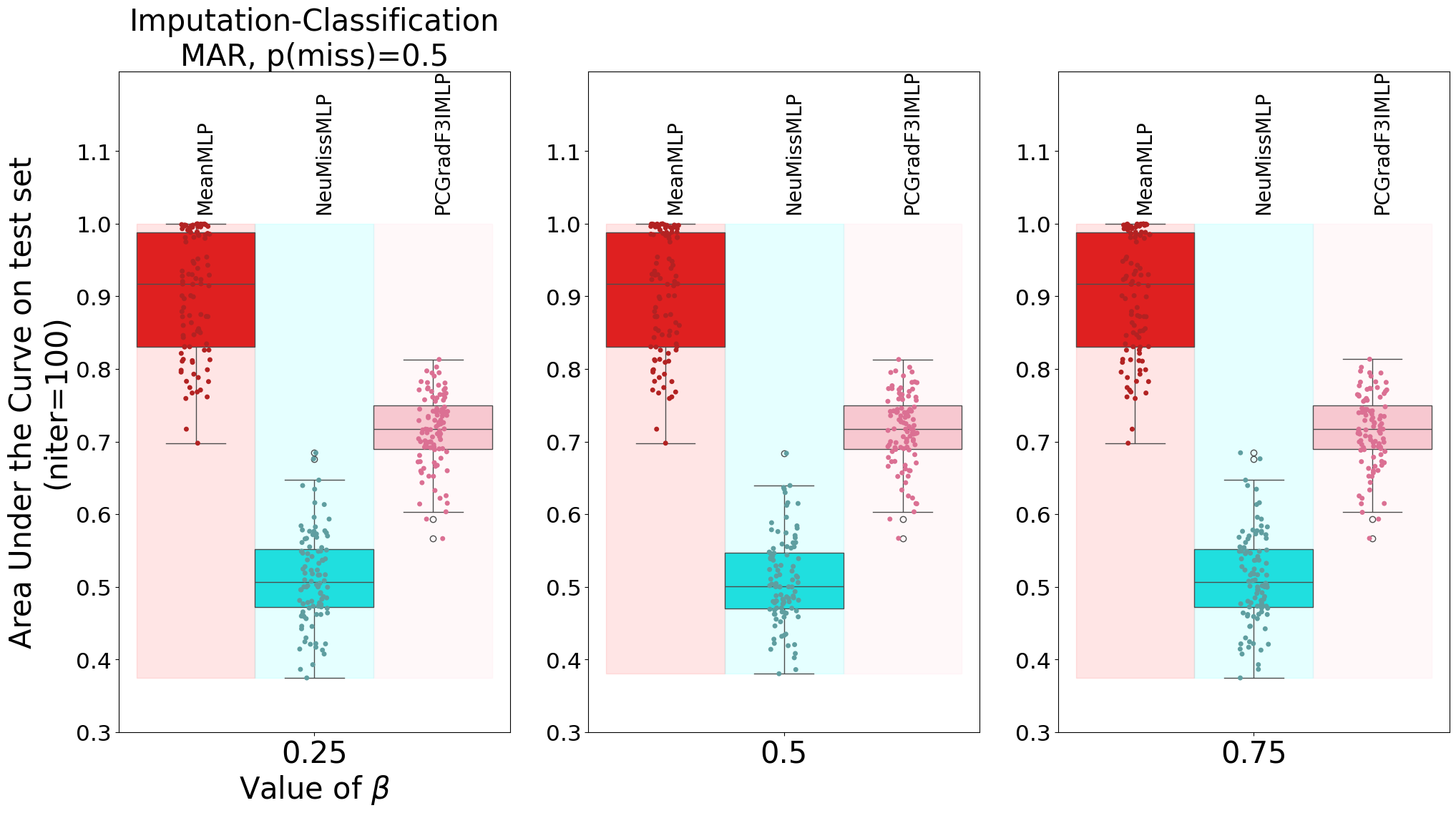}
    \includegraphics[width=0.75\textwidth]{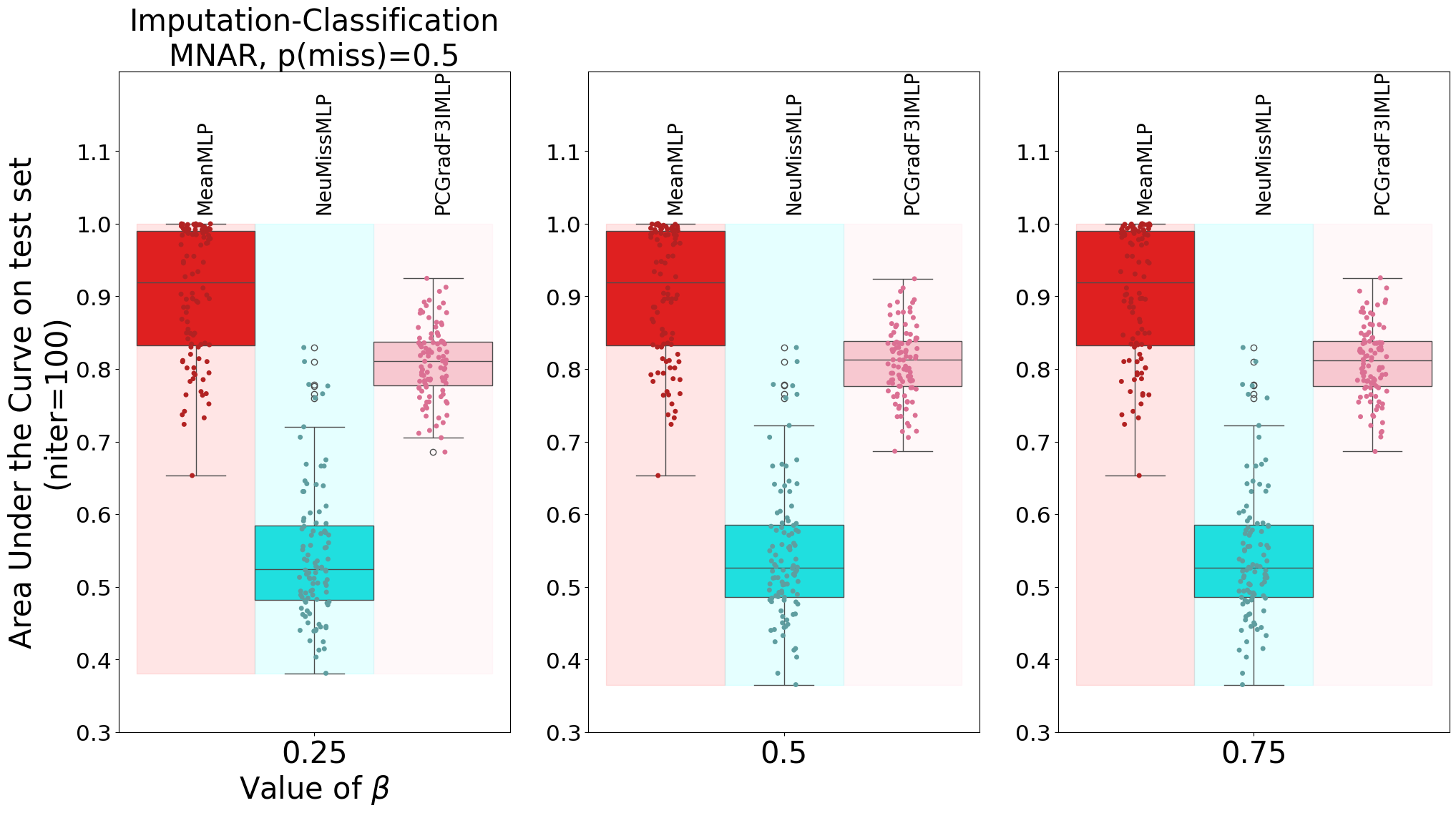}
    \caption{Joint-Imputation on a synthetic data set with MCAR (left), MAR (center) and MNAR (right) missing values and approximate missingness frequency $p^\text{miss}=0.5$, for $\beta \in \{0.25, 0.5, 0.75\}$.}
    \label{fig:synthetic_joint_05}
\end{figure}

\begin{figure}
    \centering
    \includegraphics[width=0.28\textwidth]{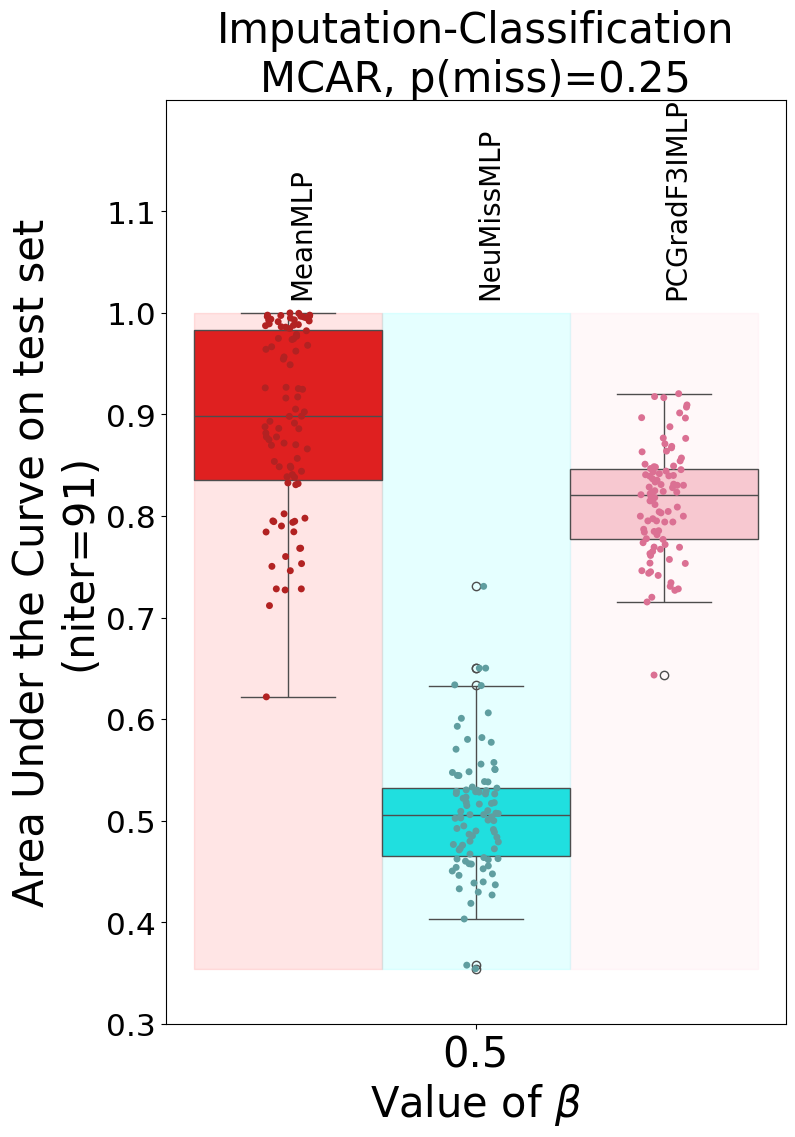}
    \includegraphics[width=0.28\textwidth]{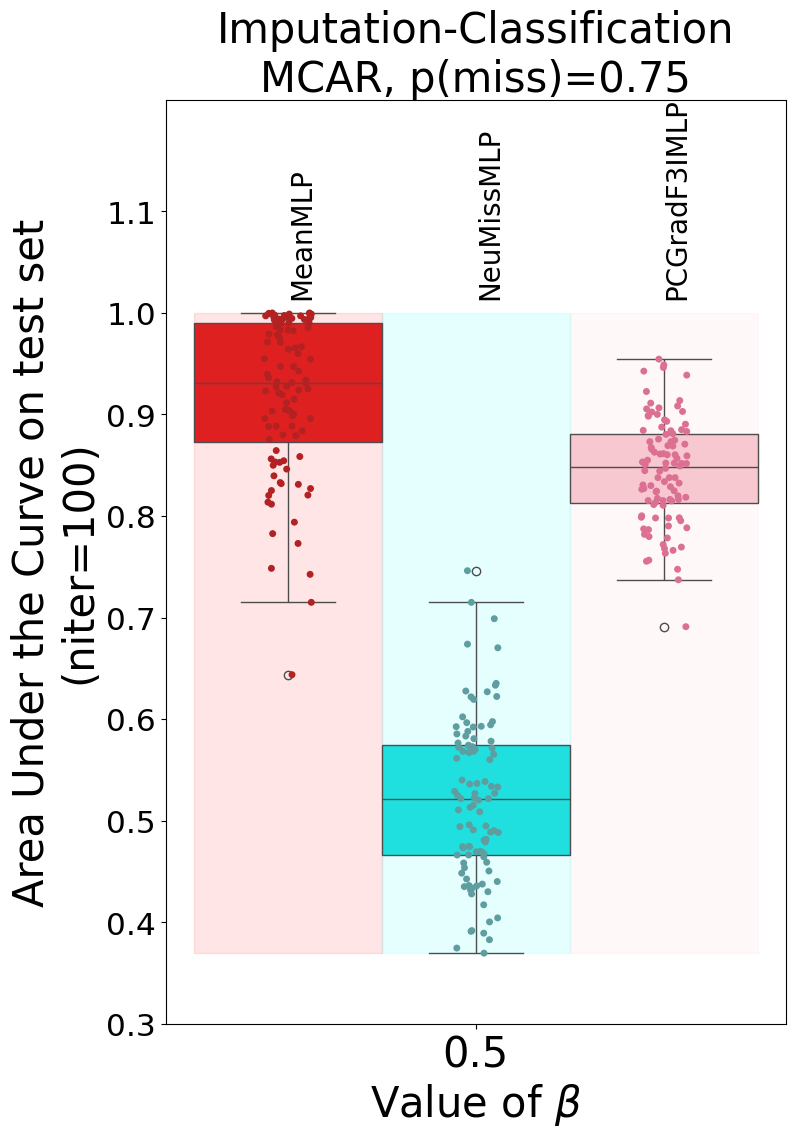}\\
    \includegraphics[width=0.28\textwidth]{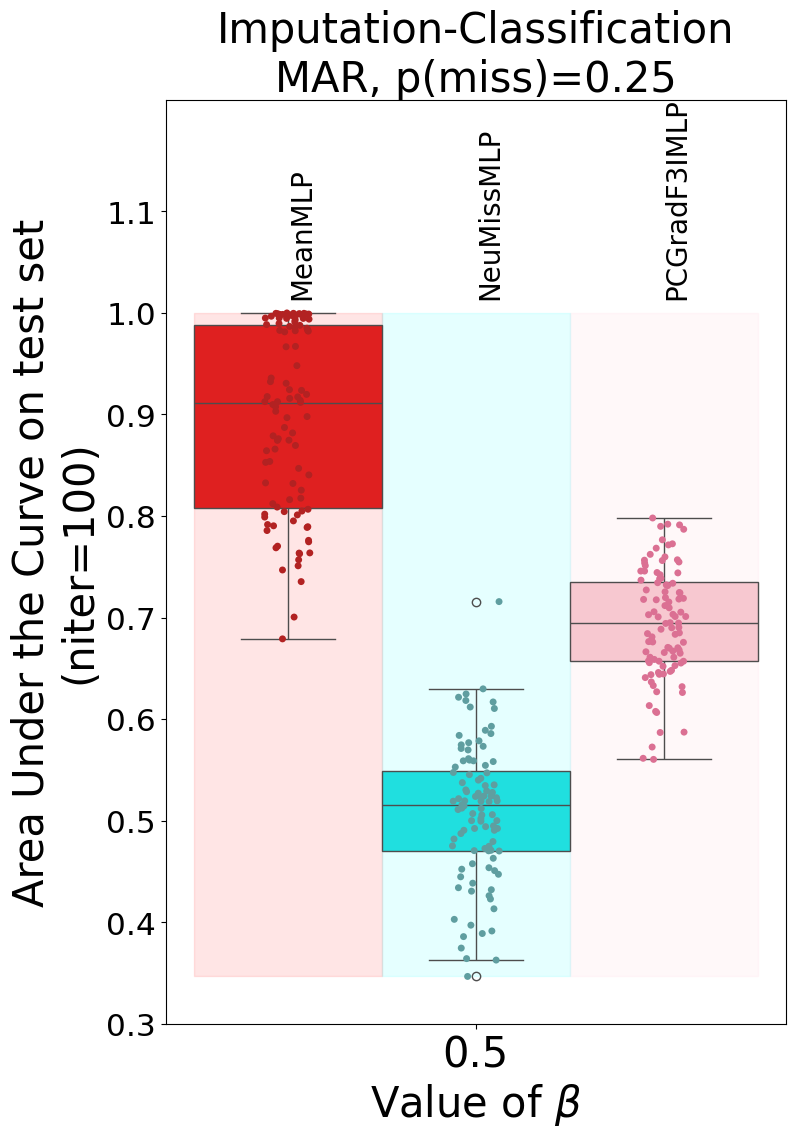}
    \includegraphics[width=0.28\textwidth]{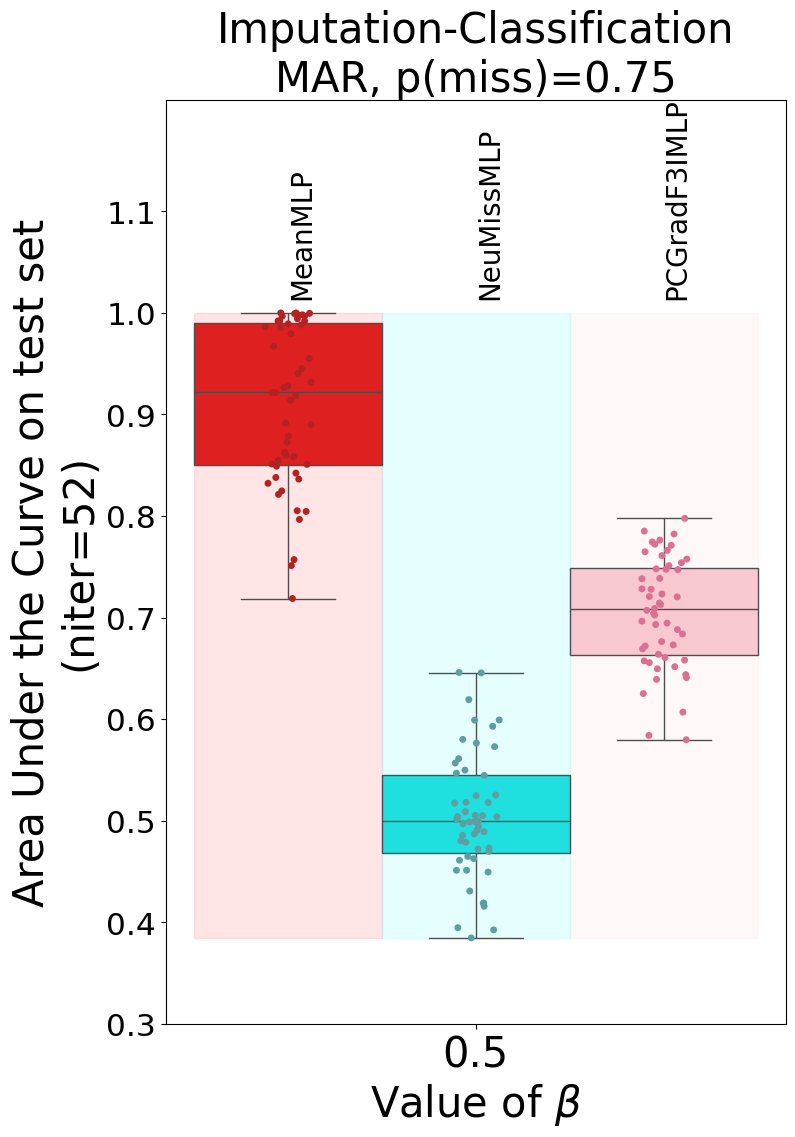}\\
    \includegraphics[width=0.28\textwidth]{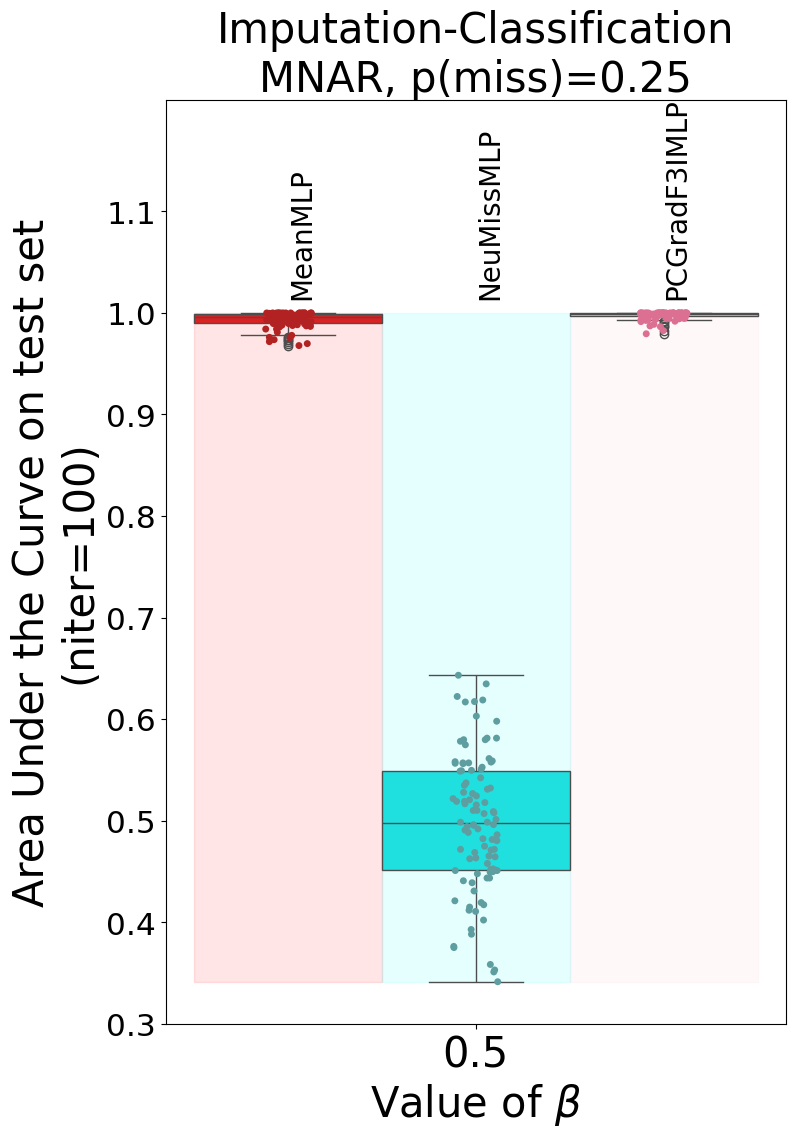}
    \includegraphics[width=0.28\textwidth]{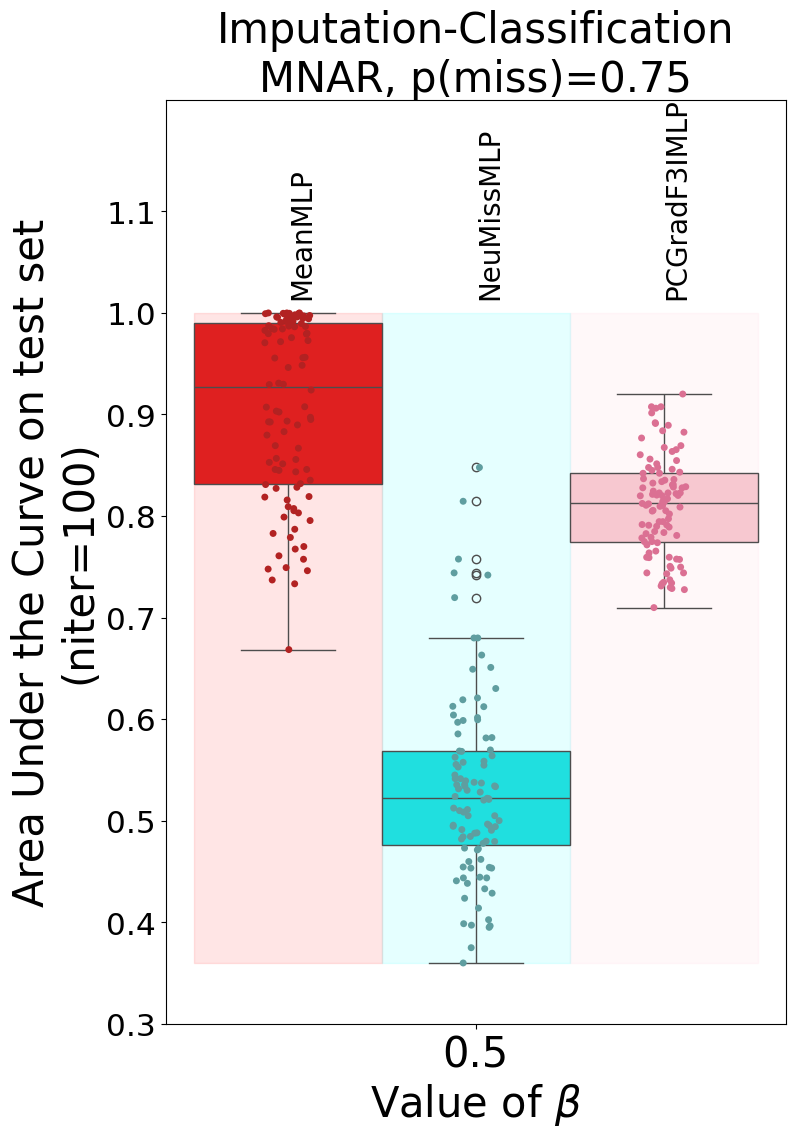}
    \caption{Joint-Imputation on a synthetic data set with MCAR (left), MAR (center) and MNAR (right) missing values and approximate missingness frequency $p^\text{miss} \in \{0.25, 0.75\}$, for $\beta \in \{0.25, 0.5, 0.75\}$.}
    \label{fig:synthetic_joint_025_075}
\end{figure}

Figure~\ref{fig:synthetic_joint_05} shows the results for MCAR (Assumption~\ref{as:mcar}), MAR (Assumption~\ref{as:mar}), and MCAR (Assumption~\ref{as:mnar}) synthetic data with an approximate missingness frequency of $50\%$ and varying values of $\beta \in \{0.25, 0.5, 0.75\}$. Figure~\ref{fig:synthetic_joint_025_075} shows the corresponding results when the approximated missingness frequency is in $\{25\%, 75\%\}$. There is a significant improvement in PCGrad-F3I over the baseline NeuMiss. However, the imputation by the mean remains the top contender on the synthetic Gaussian data sets, as already noticed for imputation in the previous paragraph, even if PCGrad-F3I is sometimes on par regarding classification performance.

\begin{table}[tb]
    \centering
    \caption{Hyperparameter values for the training of the MLP block for each imputation method (NeuMiss, mean imputation, F3I).}
    \label{tab:hyperparam_joint}
    \begin{tabular}{lrrrr}
    \toprule
  {\bfseries Hyperparameter} &  \# epochs & Weight decay in the optimizer  & Learning rate & MLP depth \\
  \midrule
   {\bfseries Value} & 10 & 0 & 0.01 & 1 layer\\
   \bottomrule
    \end{tabular}
\end{table}

\subsection{Real-life data sets (drug repurposing \& handritten-digit recognition)}\label{subapp:drug_repurposing}

In addition to the synthetic Gaussian data sets, we also evaluate the performance of F3I on real-life data for drug repurposing or handwritten-digit recognition on the well-known MNIST data set~\cite{lecun1998mnist}. 

Drug repurposing aims to pair diseases and drugs based on their chemical, biological, and physical features. However, those features might be missing due to the incompleteness of medical databases or to a lack or failure of measurement. Table~\ref{tab:dd_datasets} reports the sizes of the considered drug repurposing data sets, which can be found online as indicated in their corresponding papers. A positive drug-disease pair is a therapeutic association (that is, the drug is known to treat the disease). In contrast, a negative one is associated with a failure in treating the disease or the emergence of toxic side effects.

\begin{table}[ht]
    \centering
    \caption{Overview of the drug repurposing data sets in the experimental study in Section~\ref{sec:experiments}, with the number of drugs, drug features, diseases, disease features, along with the number of positive and negative drug-disease pairs. 
    \label{tab:dd_datasets}
    }
    \begin{tabular}{lrrrr} 
    \toprule
         Name of the data set &  
         $N_\text{drugs}$ ~ ~ ~ $F_\text{drugs}$ & $N_\text{diseases}$ ~ ~ ~  $F_\text{users}$ & Positive pairs &  Negative pairs \\
        \midrule
        Cdataset~\cite{luo2016drug} & 663 ~ ~ ~ 663 & 409 ~ ~ ~ 409 	& 2,532 	& 0 \\
         DNdataset~\cite{gao2022dda} 	& 550 ~~ 1,490 & 360 ~~ 4,516 	& 1,008 	& 0 \\
       Gottlieb~\cite{luo2016drug}    &	593 ~ ~ ~ 593 & 	313 ~ ~ ~ 313 & 	1,933 	& 0\\
       PREDICT-Gottlieb~\cite{gao2022dda} 	  & 593 ~~ 1,779 &  313 ~ ~ ~ 313	& 1,933 	& 0 	\\
       TRANSCRIPT~\cite{reda2023transcript}  &204~ 12,096 & 	116~ 12,096 & 	401 	& 11 	\\
    \bottomrule
    \end{tabular}
\end{table}

\subsubsection{Imputation quality and runtimes (drug repurposing task)} 

Drug repurposing aims to pair diseases and drugs based on their chemical, biological, and physical features. However, those features might be missing due to the incompleteness of medical databases or to a lack or failure of measurement. We consider five public drug repurposing data sets of varying sizes without missing values (see Table~\ref{tab:dd_datasets} in Appendix~\ref{subapp:drug_repurposing}). We add missing values with a MNAR Gaussian self-masking mechanism (Assumption~\ref{as:mnar}). We run each imputation method $100$ times on the drug and the disease feature matrices with different random seeds. Note that the position of the missing values is the same across runs. We considered as baselines the imputation by the feature-wise mean value (Mean), the MissForest algorithm~\cite{stekhoven2012missforest}, K-nearest neighbor (KNN) imputation with distance-proportional weights, where the weight is inversely proportional to the distance to the neighbor~\cite{troyanskaya2001missing}, an Optimal Transport-based imputer~\cite{muzellec2020missing} and finally not-MIWAE~\cite{ipsen2021not}. Hyperparameter values are reported in Table~\ref{tab:hyperparams}.

\begin{figure}[H]
    \centering
    \includegraphics[width=\linewidth]{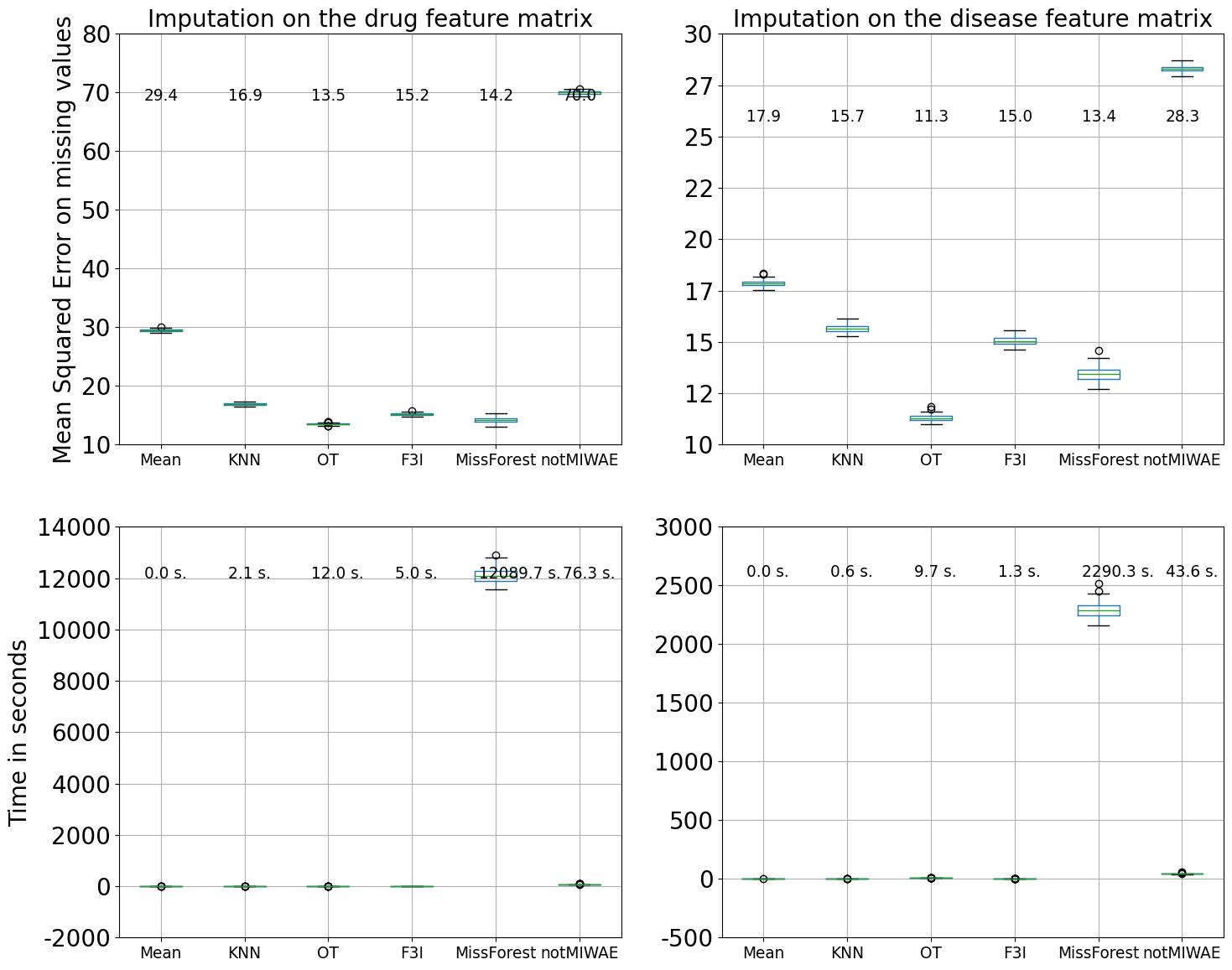}
    \caption{Imputation of missing values in the drug (left) and disease (right) feature matrices for F3I and its baselines in the Cdataset drug repurposing data set~\cite{luo2016drug}. The first row shows boxplots of mean-squared errors (MSE) across each algorithm's $100$ iterations (with different random seeds). In contrast, the second row displays the runtimes (in seconds) across iterations for the imputation step. The average value of MSE and runtime is displayed above each corresponding boxplot. Abbreviations: OT: Optimal Transport-based imputer~\cite{muzellec2020missing}, KNN: KNN imputer with distance-associated weights~\cite{troyanskaya2001missing}, Mean: imputation by the feature-wise mean value.}
    \label{fig:cdataset}
\end{figure}

\begin{figure}[H]
    \centering
    \includegraphics[width=\linewidth]{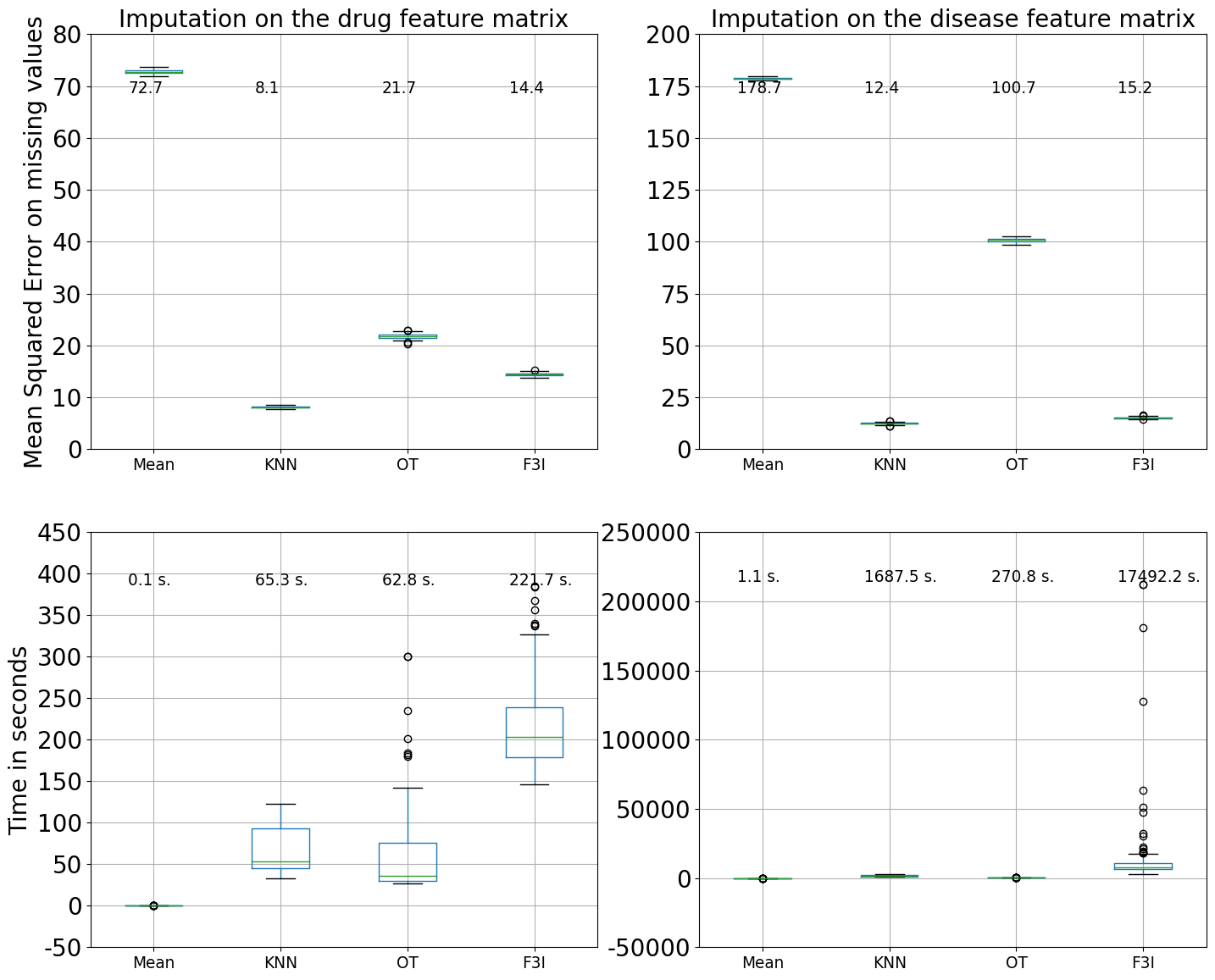}
    \caption{Imputation of missing values in the drug (left) and disease (right) feature matrices for F3I and its baselines in the DNdataset drug repurposing data set~\cite{gao2022dda}. The first row shows boxplots of mean-squared errors (MSE) across each algorithm's $100$ iterations (with different random seeds). In contrast, the second row displays the runtimes (in seconds) across iterations for the imputation step. The average value of MSE and runtime is displayed above each corresponding boxplot. Abbreviations: OT: Optimal Transport-based imputer~\cite{muzellec2020missing}, KNN: KNN imputer with distance-associated weights~\cite{troyanskaya2001missing}, Mean: imputation by the feature-wise mean value. } 
    \label{fig:dndataset}
\end{figure}

\begin{figure}[H]
    \centering
    \includegraphics[width=\linewidth]{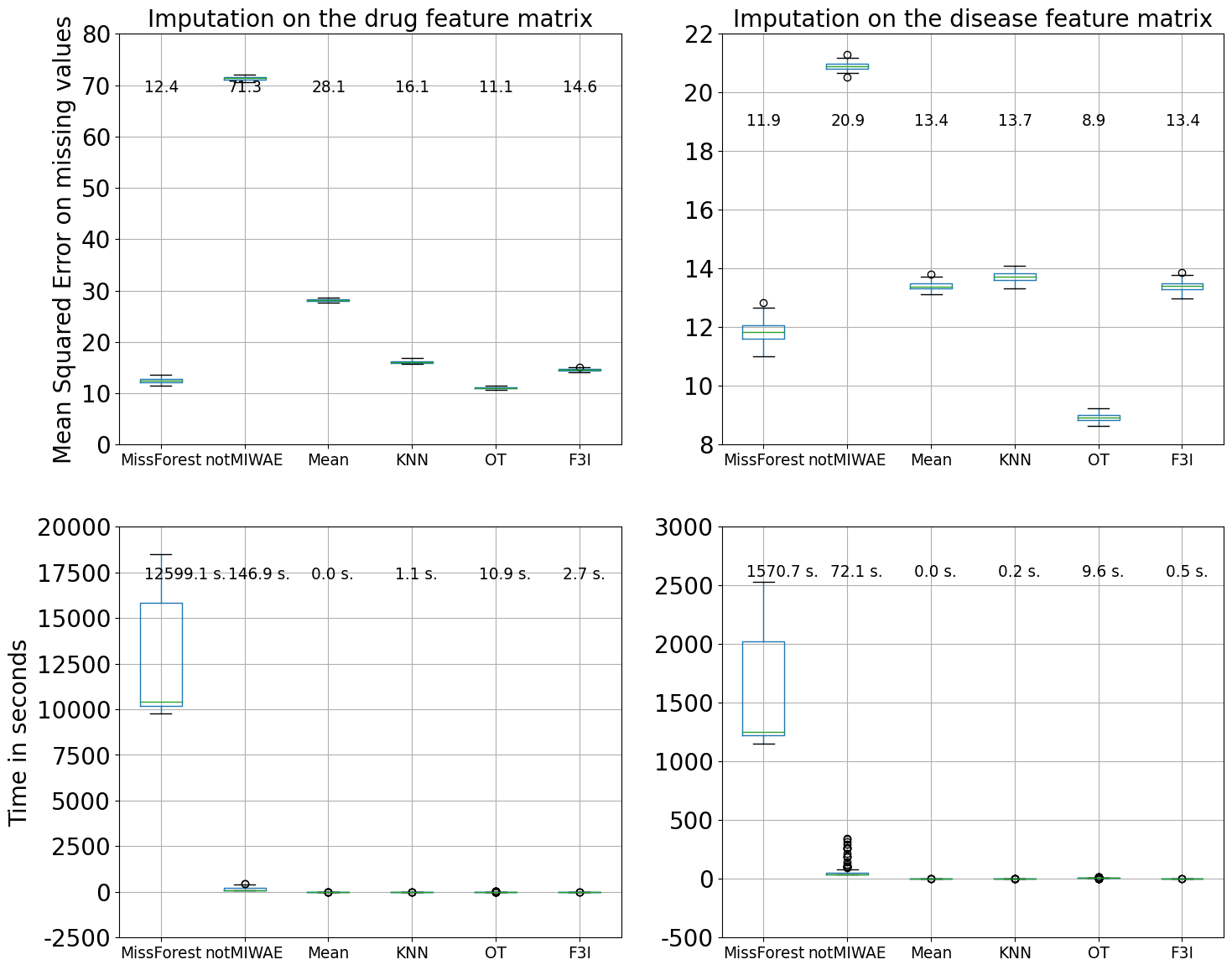}
    \caption{Imputation of missing values in the drug (left) and disease (right) feature matrices for F3I and its baselines in the Gottlieb drug repurposing data set~\cite{luo2016drug}. The first row shows boxplots of mean-squared errors (MSE) across each algorithm's $100$ iterations (with different random seeds). In contrast, the second row displays the runtimes (in seconds) across iterations for the imputation step. The average value of MSE and runtime is displayed above each corresponding boxplot. Abbreviations: OT: Optimal Transport-based imputer~\cite{muzellec2020missing}, KNN: KNN imputer with distance-associated weights~\cite{troyanskaya2001missing}, Mean: imputation by the feature-wise mean value. }
    \label{fig:gottlieb}
\end{figure}

\begin{figure}[H]
    \centering
    \includegraphics[width=\linewidth]{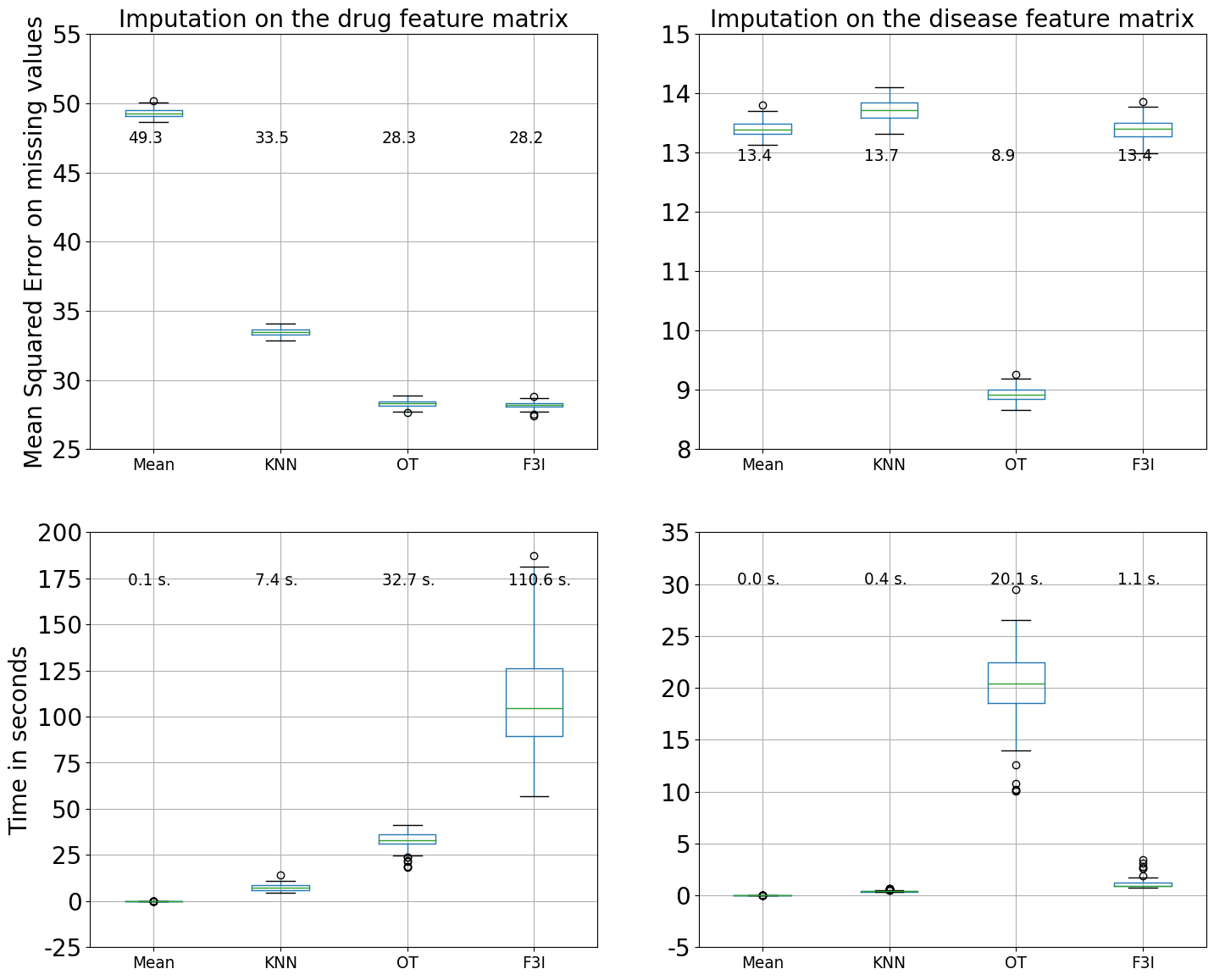}
    \caption{Imputation of missing values in the drug (left) and disease (right) feature matrices for F3I and its baselines in the PREDICT-Gottlieb drug repurposing data set~\cite{gao2022dda}. The first row shows boxplots of mean-squared errors (MSE) across each algorithm's $100$ iterations (with different random seeds). In contrast, the second row displays the runtimes (in seconds) across iterations for the imputation step. The average value of MSE and runtime is displayed above each corresponding boxplot. Abbreviations: OT: Optimal Transport-based imputer~\cite{muzellec2020missing}, KNN: KNN imputer with distance-associated weights~\cite{troyanskaya2001missing}, Mean: imputation by the feature-wise mean value.}
    \label{fig:predict-gottlieb}
\end{figure}

\begin{figure}[H]
    \centering
    \includegraphics[width=\linewidth]{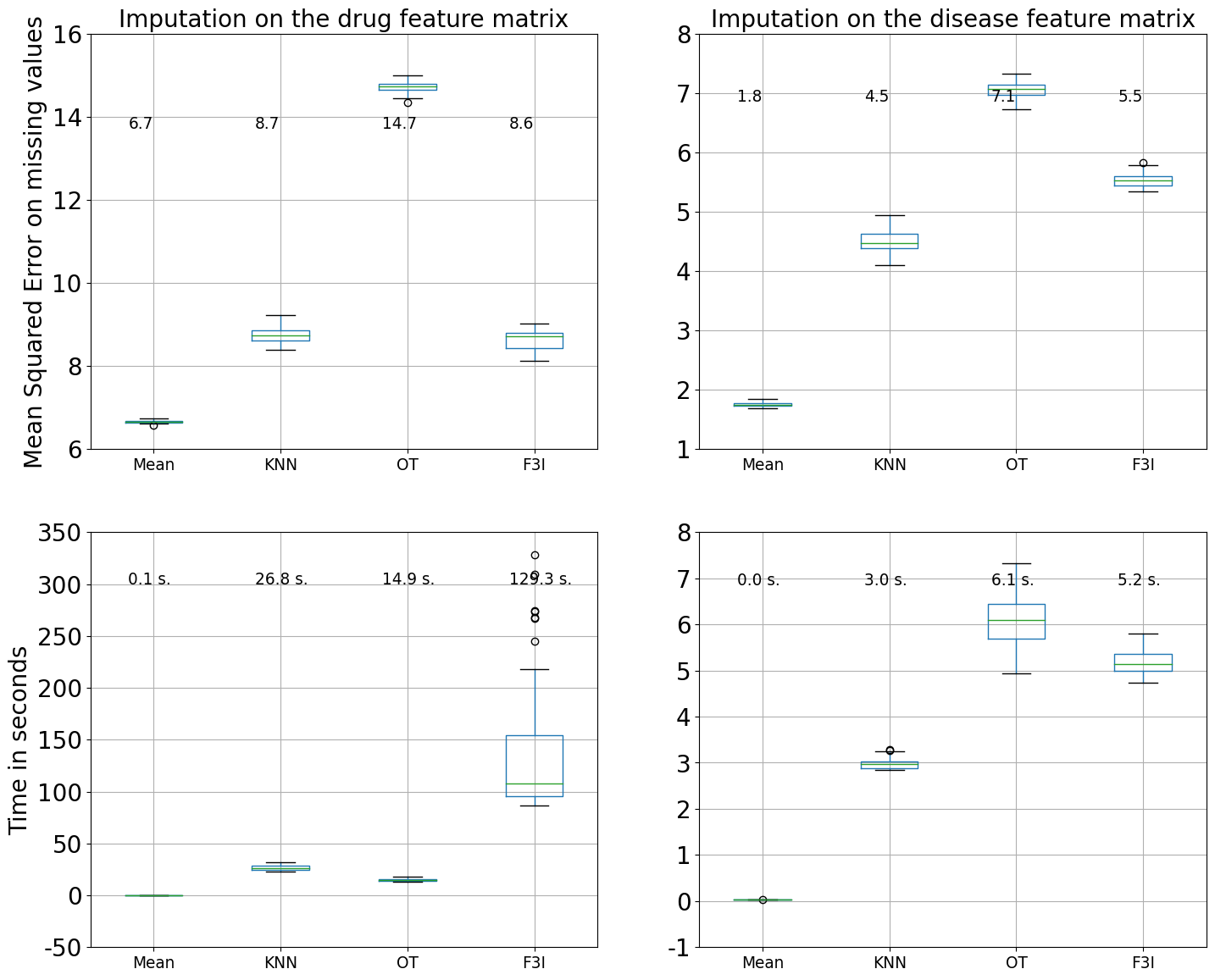}
    \caption{Imputation of missing values in the drug (left) and disease (right) feature matrices for F3I and its baselines in the TRANSCRIPT drug repurposing data set~\cite{reda2023transcript}, restricted to the $9,000$ features with highest variance across samples. The first row shows boxplots of mean-squared errors (MSE) across each algorithm's $100$ iterations (with different random seeds). In contrast, the second row displays the runtimes (in seconds) across iterations for the imputation step. The average value of MSE and runtime is displayed above each corresponding boxplot.} 
    \label{fig:transcript-9000}
\end{figure}

Since the number of features $F\approx 12,000$ in the TRANSCRIPT dataset~\cite{reda2023transcript} is prohibitive for most of the baselines, we reduce the number of features to $9,000$, selecting the features with the highest variance across drugs and diseases. Moreover, MissForest~\cite{stekhoven2012missforest} and not-MIWAE~\cite{ipsen2021not} are too resource-consuming to be run on the largest data sets, DNdataset~\cite{gao2022dda}, PREDICT-Gottlieb~\cite{gao2022dda} and TRANSCRIPT~\cite{reda2023transcript}. Figures~\ref{fig:cdataset}-\ref{fig:transcript-9000} show the boxplots of average mean squared errors and runtimes of each algorithm across the $100$ iterations for both drug and disease feature matrices. Table~\ref{tab:dd_imputation} shows the corresponding numerical results (average values $\pm$ standard deviations across the $100$ iterations).

Overall, F3I has a runtime comparable to the fastest baselines, that is, the Optimal Transport-based imputer~\cite{muzellec2020missing} (OT in plots), the imputation by the mean value (Mean) and the k-nearest neighbor approach~\cite{troyanskaya2001missing} (KNN), while having a performance in imputation which is on par with the best state-of-the-art algorithm MissForest~\cite{stekhoven2012missforest}, as reported by several prior works~\cite{emmanuel2021survey,joel2024performance}. However, MissForest is several orders of magnitude slower than F3I and sometimes cannot be run at all (for instance, for the highly-dimensional TRANSCRIPT data set). The Optimal Transport imputer also performs well across the data sets and often competes with our contribution F3I in imputation and computational efficiency.

\subsubsection{Classification quality (handwritten-digit recognition task)}\label{app:mnist}

Again, we compare the performance of PCGrad-F3I with a simple mean imputation or NeuMiss~\cite{le2020neumiss,le2021sa}, chaining the corresponding imputation part with an MLP as previously done on synthetic data sets in Subsection~\ref{subapp:synthetic}. The training procedure of the full architecture is provided in the code. We consider the MNIST dataset~\cite{lecun1998mnist}, which comprises grayscale images of $25 \times 25$ pixels. We restrict our study to images annotated with class 0 or class 1 to get a binary classification problem. We remove pixels at random with probability 50\% using a MCAR mechanism (Assumption~\ref{as:mcar}). 

\begin{table}[tb]
    \centering
    \caption{Fine-tuned hyperparameter values using Optuna to train the MLP block for each imputation method (NeuMiss, Mean imputation, PCGradF3I) on the MNIST data set. K, T, $\beta$ and $\eta$ are F3I-specific parameters, whereas all remaining parameters are common to all three methods and belong to the MLP block.}
    \label{tab:hyperparam_joint_MNIST}
    \begin{tabular}{lrrrrrrr}
    \toprule
  {\bfseries Hyperparameter} &  Number of epochs  & MLP depth & K  & T  & $\beta$ & $\eta$ \\
  \midrule
   {\bfseries Value} & 5 & 5 layers & 17  & 13 & 0.71 & 0.053 \\
   \bottomrule
    \end{tabular}
\end{table}

\begin{table}[tb]
    \centering
    \caption{Area Under the Curve (AUC) values (average $\pm$ standard deviation) in the testing subset in MNIST (hidden during the training phase) and corresponding tuned hyperparameter values (rounded up to the $2^\text{nd}$ decimal place for values in $\mathbb{R}$) for $N=100$ iterations. NeuMiss has been trained on the same number of epochs and the same MLP architecture as PCGradF3I and the mean imputation followed by the MLP (Mean imputation). Those are the same results displayed in the third column of Table~\ref{tab:joint_MNIST_PREDICT_main}.}
    \label{tab:joint_MNIST}
    \begin{tabular}{lclr}
        \toprule
        Type & $p^\text{miss}$ & Algorithm   & AUC\\
         \midrule
        MCAR (Assumption~\ref{as:mcar}) & $50\%$ & GRAPE~\citep{you_handling_2020} & \textbf{1.00 $\pm$0.00}\\
         & & K-NN~\citep{troyanskaya2001missing} & 0.93 $\pm$0.17 \\
         & &  Mean  & 0.64 $\pm$0.18 \\
        & & NeuMiss~\cite{le2020neumiss} & \underline{0.99 $\pm$0.07} \\
          & & PCGradF3I (\textbf{ours}) & \underline{0.99 $\pm$0.09}\\
          & & RF-GAP~\citep{rhodes2023geometry} & \textbf{1.00 $\pm$0.00}\\
         \bottomrule
    \end{tabular}
\end{table}

\textbf{Hyperparameter tuning and importance.} We employed the Optuna framework to optimize our model's hyperparameters~\cite{akiba2019optuna}. The optimization process focused on tuning several key parameters: $\beta, \eta, T, K$, and the depth of the classifier MLP. For the hyperparameter search, we utilized Optuna's default Tree-structured Parzen Estimator-based sampler~\cite{hyperparamTPE}, conducting 50 trials to explore the parameter space. The dataset was evenly divided into three portions, with 34\% allocated for training, 33\% for validation, and 33\% for testing. During the optimization process, we aimed to maximize the logarithm of the Area Under the Curve (AUC) scores from the Receiver Operating Characteristic (ROC) curve on the validation set. After identifying the optimal hyperparameter configuration, we constructed the final model. Final hyperparameter values are reported in Table~\ref{tab:hyperparam_joint_MNIST}. 

We also estimated the importance of each hyperparameter on the objective function using the functional analysis of variance, or fANOVA~\cite{pmlr-v32-hutter14}. fANOVA estimates the percentage of variance in the classification performance on the validation set explained by each hyperparameter, given a regression tree and a hyperparameter space. Then, the larger the percentage, the greater impact on the classification performance on the validation set. The corresponding values are listed in Table~\ref{tab:importance_hyperparams}. 

\begin{table}[htb]
    \centering
    \caption{Percentages of variance of the Area Under the Curve (AUC) on the validation set explained by each hyperparameter finetuned on the MNIST data set for PCGradF3I. Hyperparameters are listed in the order of decreasing explained variance.}
    \label{tab:importance_hyperparams}
    \begin{tabular}{lrrrrrr}
    \toprule
  {\bfseries Hyperparameter} &  Number of epochs & $\eta$ & K & T  & $\beta$  & MLP depth    \\
  \midrule
   {\bfseries Explained variance (\%) } & 38.1 &  32.6 & 12.5  & 9.6  & 4.1 & 2.9   \\
   \bottomrule
    \end{tabular}
\end{table}

We observe that the number of training epochs for the MLP and the regularization factor $\eta$ for the weight vector $\bm{\alpha}$ account for approximately $38\%$ and $33\%$ respectively of the variability of the performance across the entire hyperparameter space. 
The number $K$ of nearest neighbours chosen for the convex combination is also relatively important. This is expected, as the number of neighbors controls the quality of the estimation of the data distribution in the imputation algorithm. 
Surprisingly, the number of iterations $T$ explains less than $10\%$ of the variance in the performance, demonstrating that 
F3I probably reaches the early stopping criterion (Line 16 in Algorithm~\ref{alg:online_F3I}) very quickly, without exhausting the budget $T$. Another surprising observation is the relatively low importance for the classification performance of $\beta$ and the depth of the MLP. This shows that even a relatively simple classifier can achieve higher classification accuracy with good imputation quality on the MNIST data set restricted to the classes $0$ and $1$.

\textbf{Training.} This optimized model underwent training using the designated training set, followed by performance evaluation. We assessed its performance by measuring the AUC score of the ROC curve on the test set, repeating this evaluation process across 100 iterations to ensure robust results. 

\begin{figure}
    \centering
    \includegraphics[width=0.19\linewidth]{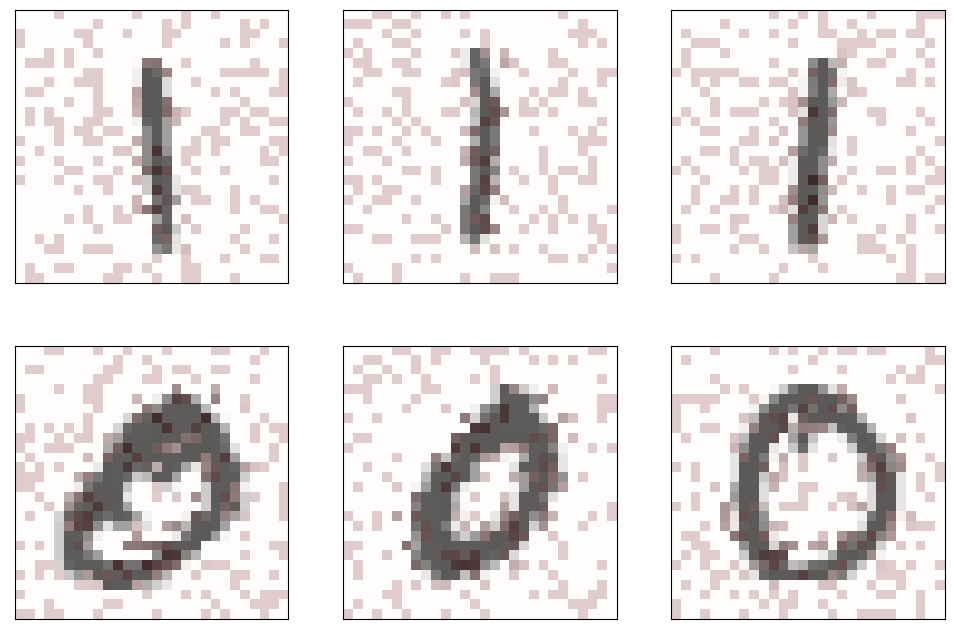}
    \hspace{0.5cm}
    \includegraphics[width=0.19\linewidth]{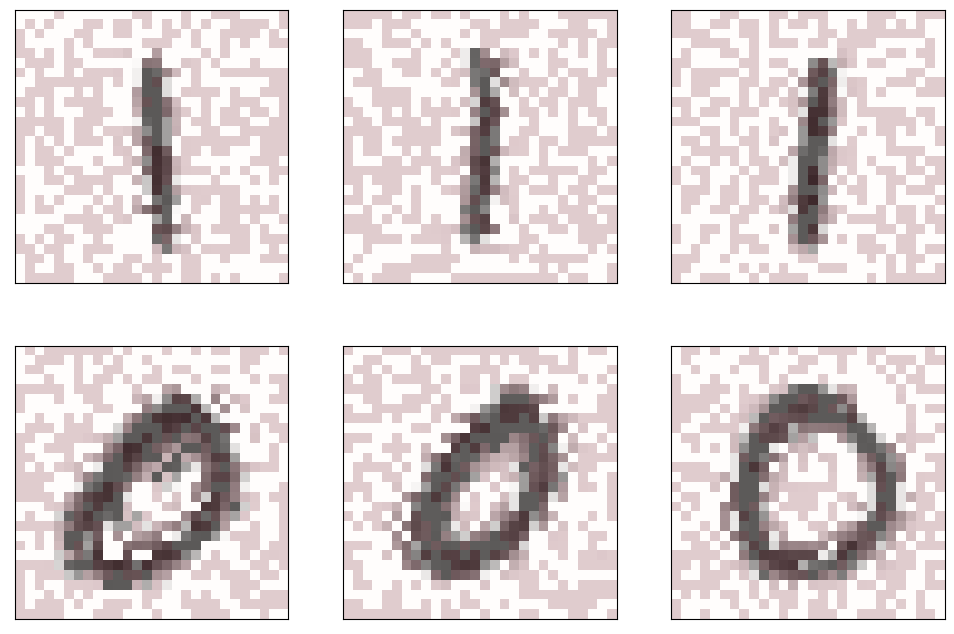}
    \hspace{0.5cm}
    \includegraphics[width=0.19\linewidth]{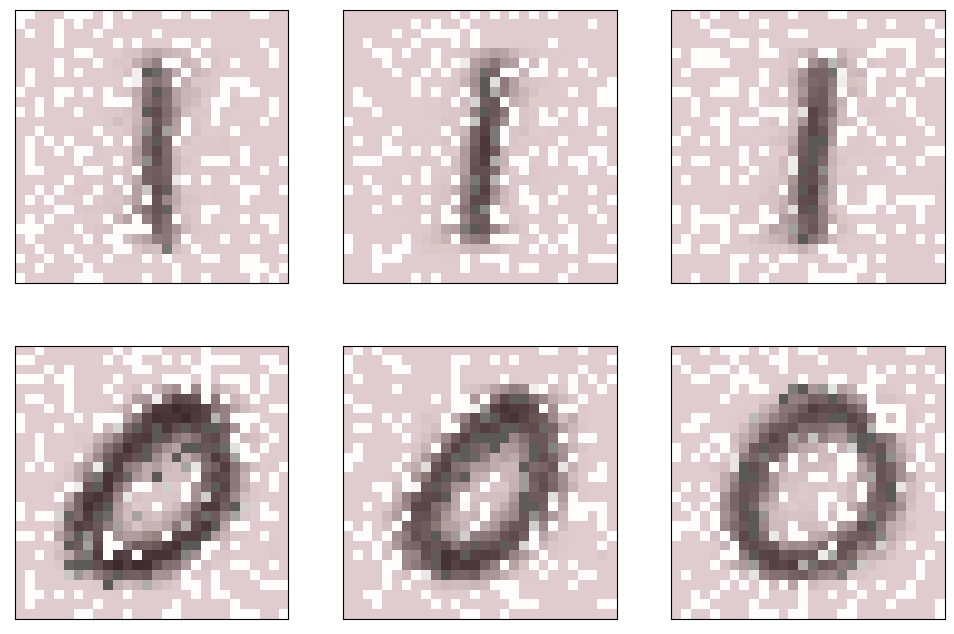}

    ~ \\

    \includegraphics[width=0.19\linewidth]{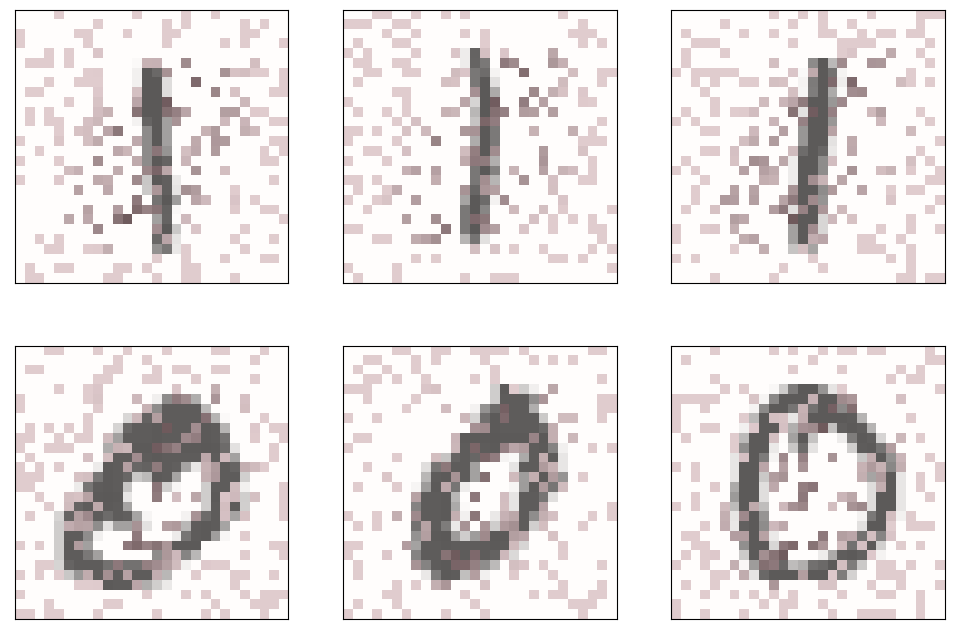}
    \hspace{0.5cm}
    \includegraphics[width=0.19\linewidth]{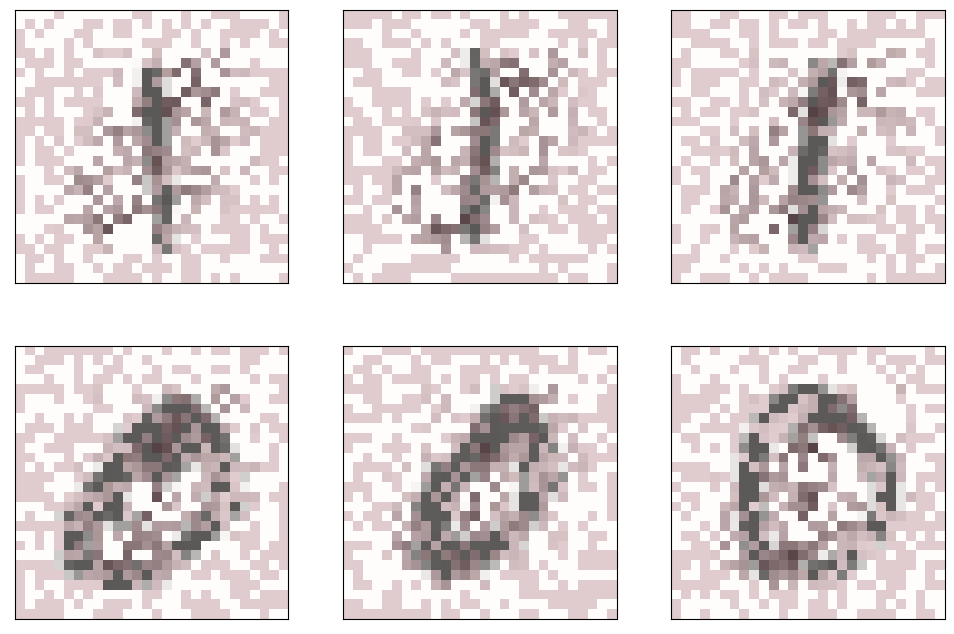}
    \hspace{0.5cm}
    \includegraphics[width=0.19\linewidth]{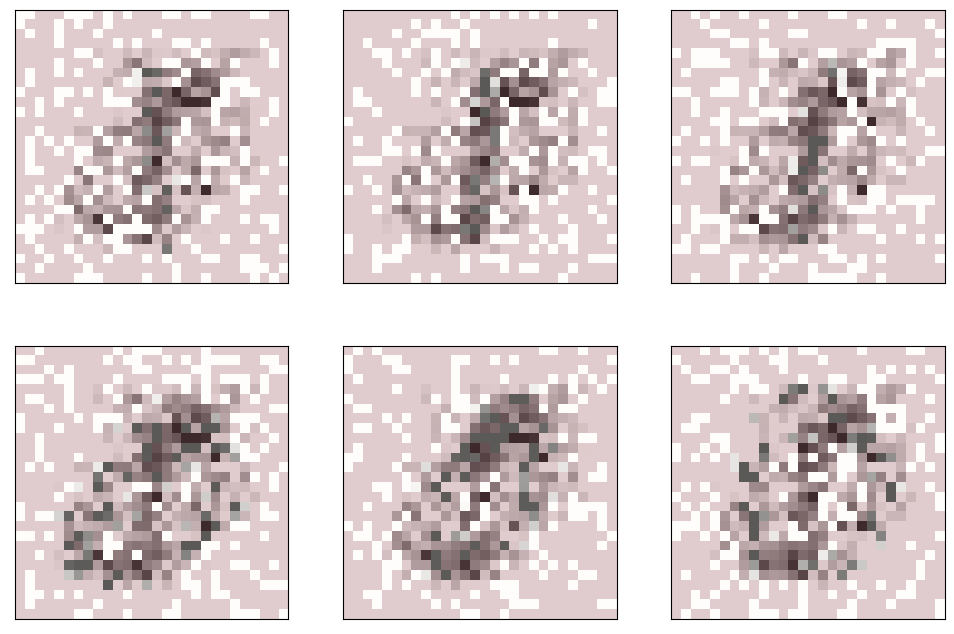}
        \caption{Imputed grayscale images by F3I (first two rows) or mean imputation (last two rows) for the first $6$ samples (trained on the first 600 samples of MNIST with the hyperparameters in Table~\ref{tab:hyperparam_joint_MNIST}) with MCAR-missing pixels, with missingness frequencies in $\{25\%, 50\%, 75\%\}$. Columns 1 to 3 correspond to $p^\text{miss} = 25\%$, columns 4 to 6 to $p^\text{miss}=50\%$, and columns 7 to 9 to $p^\text{miss}=75\%$. Positions of red pixels represent missing pixels during the training phase which are imputed by either F3I or mean imputation.}
    \label{fig:joint_MNIST}
\end{figure}

\begin{figure}
    \centering
    \includegraphics[width=0.19\linewidth]{images/mnist_type=MCAR_pmiss=50_method=f3i_beta=0.142_REDUCE=True.png}
    \hspace{0.5cm}
    \includegraphics[width=0.19\linewidth]{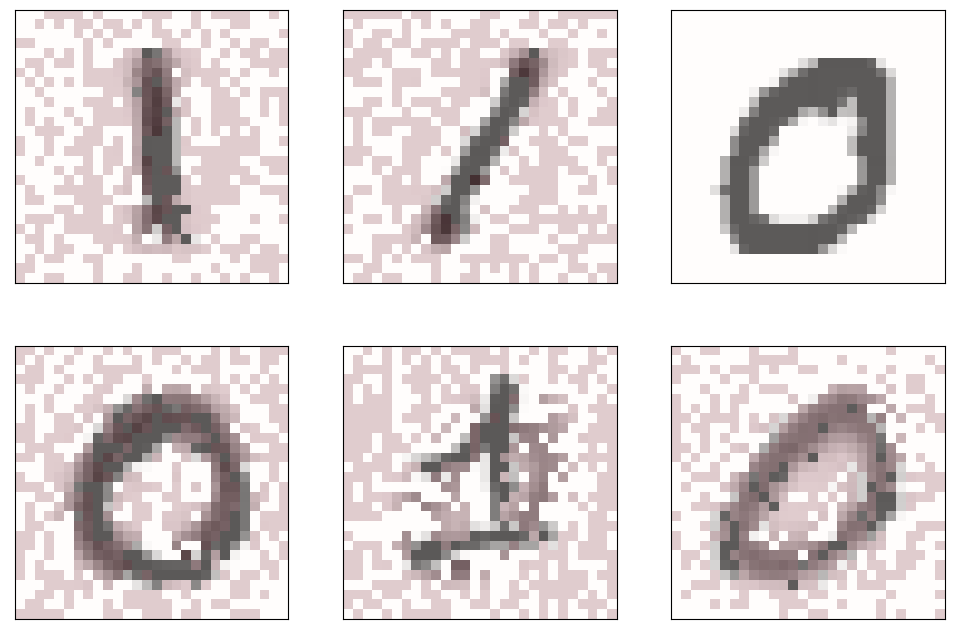}
    \hspace{0.5cm}
    \includegraphics[width=0.19\linewidth]{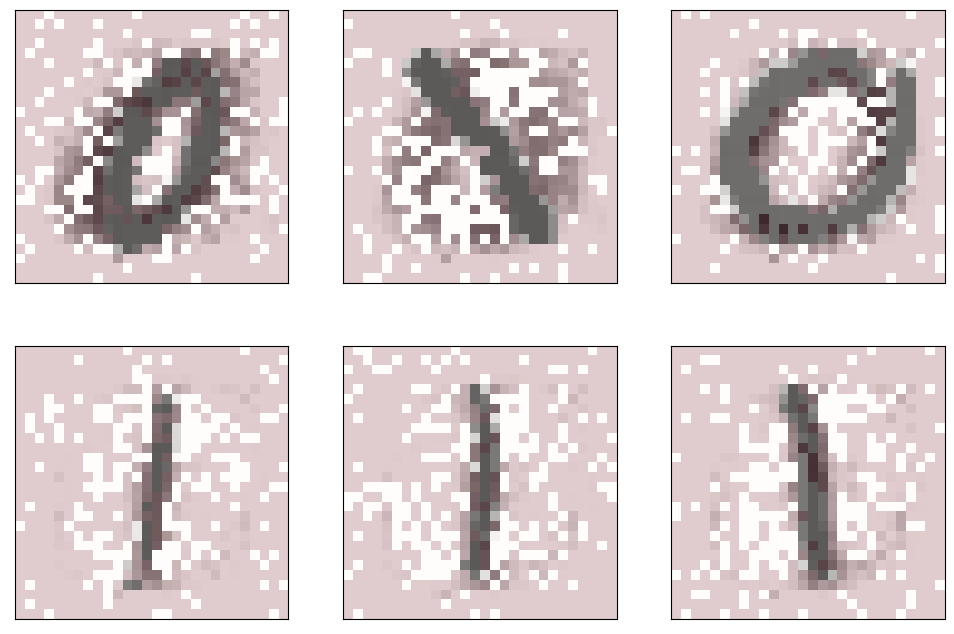}
    
    ~ \\
    
    \includegraphics[width=0.19\linewidth]{images/mnist_type=MCAR_pmiss=50_method=mean_beta=0.142_REDUCE=True.png}
    \hspace{0.5cm}
    \includegraphics[width=0.19\linewidth]{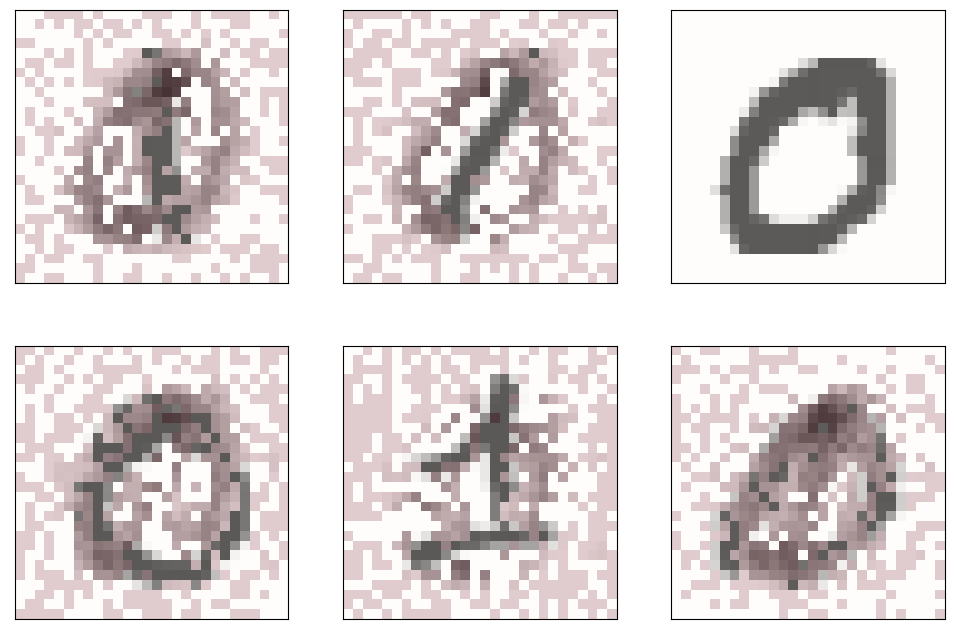}
    \hspace{0.5cm}
    \includegraphics[width=0.19\linewidth]{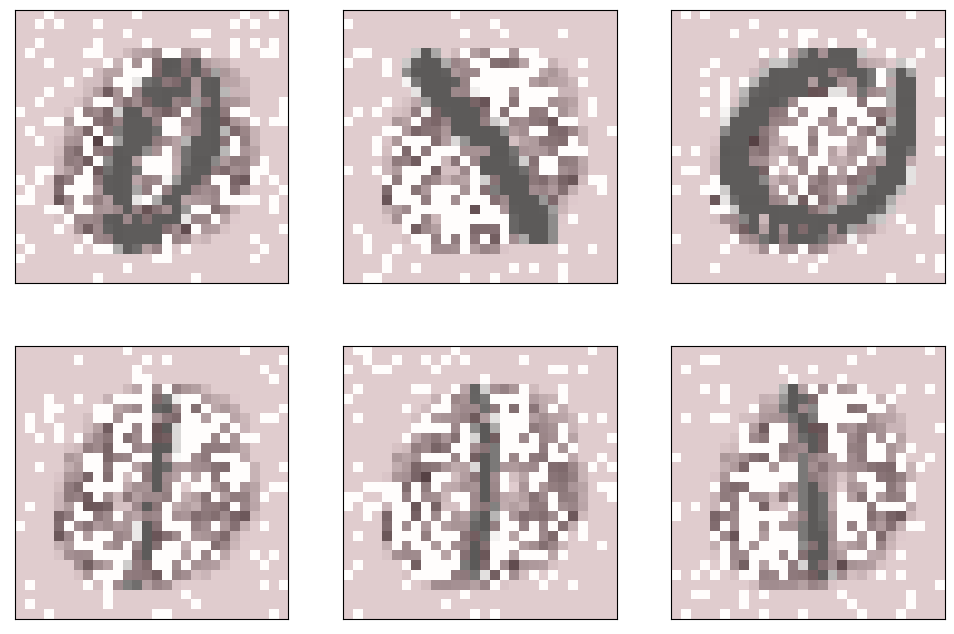}
    \caption{Imputed grayscale images by F3I (first two rows) or mean imputation (last two rows) for the first $6$ samples (trained on the first 600 samples of MNIST with the hyperparameters in Table~\ref{tab:hyperparam_joint_MNIST}). Positions of red pixels represent missing pixels during the training phase which are imputed by either F3I or mean imputation. Columns 1 to 3 correspond to MCAR-missing pixels (Assumption~\ref{as:mcar}), columns 4 to 6 to MAR-missing pixels (Assumption~\ref{as:mar}), and columns 7 to 9 to MNAR-missing pixels (Assumption~\ref{as:mnar}). Note that in one of the samples with MAR-missing pixels, no pixel is missing, which is due to randomness in the generation of missing pixels.}
    \label{fig:joint_MNIST3}
\end{figure}

\textbf{Results.} We report the numerical results in Table~\ref{tab:joint_MNIST}. We also display the imputed images for the first $6$ samples in MNIST by F3I or by mean imputation, trained on the first 600 samples.~\footnote{The NeuMiss network~\cite{le2020neumiss} does not perform imputation, only classification or regression.} We vary $p^\text{miss} \in \{25\%, 50\%, 75\%\}$ in Figure~\ref{fig:joint_MNIST}. 
Finally, we modify the missingness mechanism in Figure~\ref{fig:joint_MNIST3}, switching the MCAR missingness mechanism to MAR (Assumption~\ref{as:mar}) or MNAR (Assumption~\ref{as:mnar}). 

As mentioned in the main text, PCGradF3I beats the mean imputation and NeuMiss regarding classification accuracy according to Table~\ref{tab:joint_MNIST}. It also preserves a good imputation of the MNIST images compared to the mean imputation, even when the number of missing values increases (see Figure~\ref{fig:joint_MNIST}). Even if one still can distinguish between ones and zeroes with the mean imputation, there is a higher confidence in the predicted labels when looking at F3I-imputed images. 
Moreover, F3I turns out to be more robust to the different types of missingness mechanisms compared to the mean imputation, as illustrated by Figure~\ref{fig:joint_MNIST3}. For missing-completely-at-random pixels, both methods fare good regarding imputation. However, the performance of the mean imputation is limited in the case of MAR or MNAR-missing pixels (Columns 4 et 6), as most samples represent both a 0 and a 1.

\begin{table}
    \centering
    \caption{Overview of the drug repurposing data set PREDICT for joint imputation-classification experiments, with the number of drugs, drug features, percentage of missing drug data, diseases, disease features, percentage of missing disease features, along with the number of positive and negative drug-disease pairs. 
    }\label{tab:dd_datasets2}
    \begin{tabular}{lrrrrrr} 
    \toprule
         Data set &  $N_\text{drugs}$ & $F_\text{drugs}$ ($\%$ missing) & $N_\text{diseases}$ &  $F_\text{users}$ ($\%$ missing) & Pos &  Neg \\
        \midrule
        PREDICT~\cite{reda2023predict} & 1,150 & 1,642 ~ (24) & 1,028 & 1,490 ~ (26) & 4,627 	& 132 \\
        PREDICT (reduced) & 175 & 326 ~ (36) & 175  & 215 ~ (60) & 454 	&  0 \\
    \bottomrule
    \end{tabular}
\end{table}

\subsubsection{Classification quality (drug repurposing task)} 

\paragraph{Joint imputation and repurposing~ } This time, we consider the drug (item) and disease (user) feature matrices, along with the drug-disease association class labels from another drug repurposing data set, which natively includes missing values in the drug and disease feature matrices. This data set, named PREDICT~\cite{reda2023predict}, is further described in Table~\ref{tab:dd_datasets2}. All unknown drug-disease associations are labeled $0.5$, whereas positive (respectively, negative) ones are labeled $+1$ (resp., $-1$). To restrict the computational cost, we restricted the data set to its first $500$ ratings (\textit{i}.\textit{e}., drug-disease pairs) and to the $350$ features with highest variance across all drugs and diseases. We also add other missing values to the data set via a MCAR mechanism --as we might have lost some missing values when reducing the data set-- and run a hyperparameter optimization procedure, similarly to what has been done on the MNIST data set (see Subsection~\ref{app:mnist}). See Table~\ref{tab:hyperparam_joint_PREDICT} for the selected hyperparameter values.

The corresponding numerical results compared to the mean imputation and NeuMiss (with the same architecture of MLPs) is displayed in Table~\ref{tab:joint_PREDICT}. This table shows that PCGradF3I performs slightly better than NeuMiss on this very difficult data set, while being significantly better than the naive approach relying on the imputation by the mean value. Those results confirm what we observed on the MNIST data set (see Table~\ref{tab:joint_MNIST}).

\begin{table}[tb]
    \centering
    \caption{Fine-tuned hyperparameter values using Optuna to train the MLP block for each imputation method (NeuMiss, Mean imputation, PCGradF3I) on the PREDICT data set. K, T, $\beta$ and $\eta$ are F3I-specific parameters, whereas all remaining parameters are common to all three methods and belong to the MLP block.}
    \label{tab:hyperparam_joint_PREDICT}
    \begin{tabular}{lrrrrrrr}
    \toprule
  {\bfseries Hyperparameter} &  Number of epochs  & Learning rate & MLP depth & K  & T  & $\beta$ & $\eta$ \\
  \midrule
   {\bfseries Value} & 10  & 0.01 & 1 layer & 12  & 25  & 0.246 &  0.008 \\
   \bottomrule
    \end{tabular}
\end{table}

\begin{table}[tb]
    \centering
    \caption{Area Under the Curve (AUC) values (average $\pm$ standard deviation) in the testing subset in PREDICT (hidden during the training phase) and corresponding tuned hyperparameter values (rounded up to the $2^\text{nd}$ decimal place for values in $\mathbb{R}$) for $N=100$ iterations. NeuMiss has been trained on the same number of epochs and the same MLP architecture as PCGradF3I and the mean imputation followed by the MLP (Mean imputation). Those are the same results displayed in the fourth column of Table~\ref{tab:joint_MNIST_PREDICT_main}.}
    \label{tab:joint_PREDICT}
    \begin{tabular}{lclr}
        \toprule
        Type & $p^\text{miss}$ & Algorithm   & AUC\\
         \midrule
        MCAR (Assumption~\ref{as:mcar}) & $50\%$ & GRAPE~\citep{you_handling_2020} & 0.49 $\pm$0.07\\
         & & K-NN~\citep{troyanskaya2001missing} & 0.47 $\pm$0.07 \\
         & &  Mean  & 0.48 $\pm$0.00 \\
        & & NeuMiss~\cite{le2020neumiss} & 0.50 $\pm$0.01 \\
          & & PCGradF3I (\textbf{ours}) & \underline{0.51 $\pm$0.01}\\
          & & RF-GAP~\citep{rhodes2023geometry} & \textbf{0.53 $\pm$0.13}\\
         \bottomrule
    \end{tabular}
\end{table}


\begin{table}[H]
    \centering
    \caption{Mean Squared Errors and runtimes (average $\pm$ standard deviation) of the imputation of missing values on drug feature matrices, for 100 random seeds, rounded to the closest $1^{st}$ decimal place. Best values are in bold type, second best values underlined.}
    \label{tab:impute_dd_main}
    \begin{tabular}{lrr}
        \toprule
         Algorithm/Data set & MSE $\downarrow$ & Runtime (sec.) $\downarrow$\\
         \midrule
         \textbf{Cdataset} & & \\
         \midrule
         KNN  & 16.9 $\pm$0.2 & \underline{2.0 $\pm$0.2} \\
         Mean & 29.4 $\pm$0.2 & \textbf{0.0 $\pm$0.0}\\
         MissForest & \underline{14.2 $\pm$0.4} & 12,089.7 $\pm$268.8\\
         not-MIWAE & 70.0 $\pm$0.3 & 76.3 $\pm$3.8\\
         Optimal Transport &  \textbf{13.5 $\pm$0.2} &  12.0 $\pm$0.9\\
         F3I &  15.2 $\pm$0.2 &  5.0 $\pm$0.9\\
         \midrule
         \textbf{DNdataset} & & \\
         \midrule
         F3I (ours) &  \underline{14.4 $\pm$0.3} &  221.7 $\pm$59.3\\
         KNN  & \textbf{8.1 $\pm$0.2} & 65.3 $\pm$27.1 \\
         Mean & 72.7 $\pm$0.4 & \textbf{0.1 $\pm$0.1}\\
         MissForest & - & -\\
         not-MIWAE & - & -\\
         Optimal Transport &  21.7 $\pm$0.5 &  \underline{62.8 $\pm$55.1}\\
         \midrule
         \textbf{Gottlieb} & & \\
         \midrule
         KNN  & 16.1 $\pm$0.2 & \underline{1.1 $\pm$0.2} \\
         Mean & 28.1 $\pm$0.2 & \textbf{0.0 $\pm$0.0}\\
         MissForest & \underline{12.4 $\pm$0.4} & 12,599.1 $\pm$3,115.5\\
         not-MIWAE & 71.3 $\pm$0.3 & 146.9 $\pm$111.5\\
         Optimal Transport &  \textbf{11.1 $\pm$0.2} &  10.9 $\pm$2.6\\
         F3I &  14.6 $\pm$0.2 &  2.7 $\pm$0.4\\
         \midrule
         \textbf{PREDICT-Gottlieb}  & & \\
         \midrule
         KNN  & 33.5 $\pm$0.3 & \underline{7.4 $\pm$1.7} \\
         Mean & 49.3 $\pm$0.3 & \textbf{0.1 $\pm$0.0}\\
         MissForest & - & -\\
         not-MIWAE & - & -\\
         Optimal Transport &  \underline{28.3 $\pm$0.3} &  32.7 $\pm$4.9\\
         F3I &  \textbf{28.2 $\pm$0.2} &  110.6 $\pm$28.2\\
         \midrule
         \textbf{TRANSCRIPT}  & & \\
         \midrule
         KNN  & 8.7 $\pm$0.2 & 26.8 $\pm$2.4 \\
         Mean & \textbf{6.7 $\pm$0.0} & \textbf{0.1 $\pm$0.0} \\
         MissForest & - & -\\
         not-MIWAE & - & -\\
         Optimal Transport &  14.7 $\pm$0.1 &  \underline{14.8 $\pm$1.0}\\
         F3I &  \underline{8.6 $\pm$0.2} &  129.3 $\pm$52.9\\
         \bottomrule
    \end{tabular}
    \label{tab:dd_imputation}
\end{table}

\section{Experiments complementary to the main text}

We report here tables of numerical results related to experiments described in the main text (Section~\ref{sec:experiments_main}).

\subsection{Imputation-only task}\label{subapp:onlyImpute}

We study the imputation quality--without any downstream task. We resorted to the framework HyperImpute~\cite{jarrett2022hyperimpute} to implement and run the benchmark for an imputation task across different performance metrics (including RMSE) on the four standard data sets BreastCancer~\citep{breast_cancer_wisconsin_(diagnostic)_17}, Diabetes (from scikit-learn~\cite{pedregosa2011scikit}), HeartDisease~\citep{heart_disease_45}, Ionosphere~\citep{ionosphere}, including only the top baselines based on Table~\ref{tab:imputation-benchmark} to obtain more robust estimates of the performance (over $100$ runs instead of $10$ runs in the main text). Those top baselines are: GAIN~\citep{yoon2018gain}, HyperImpute~\citep{jarrett2022hyperimpute}, MIRACLE~\citep{kyono2021miracle} and NewImp~\citep{chen2024rethinking}. We considered again the scenario MNAR in the framework HyperImpute to add missing values. We report in Table~\ref{tab:imputation-benchmark-2} the corresponding numerical results across $100$ runs with different random seeds. 

As written in the main text, those results on a larger set of runs confirm our observations, and show that F3I is competitive imputation-wise while being dramatically faster than baselines.

\begin{table}
    \centering
    \caption{Average and standard deviation values of imputation quality metrics (rounded to the closest second decimal place) and runtime across 100 different random seeds. HeartDisease has native missing values, which is why the Wasserstein distance cannot be computed. RMSE: root mean square error. MAE: mean average error. WD: Wasserstein distance. Runtime is in seconds. TDM failed on the Gottlieb data set. Bold type is the top performer, underline denotes the second best (and corresponding percentage of deterioration of performance across metrics compared to the top performer).}
    \label{tab:imputation-benchmark-2}
    \begin{tabular}{lrrrr}
     \toprule
        Data set & RMSE $\downarrow$ &  MAE $\downarrow$ & WD $\downarrow$ & Runtime $\downarrow$\\
        \midrule
       BreastCancer   & & & &  \\
       \midrule
       F3I (ours) & \textbf{0.08 $\pm$0.03} & \textbf{0.03 $\pm$0.01} & \textbf{0.07 $\pm$0.02} & \textbf{0.18 $\pm$0.06}\\
       GAIN  & 0.27 $\pm$0.03 & 0.10 $\pm$0.02 & 0.24 $\pm$0.05 & 45 $\pm$27\\
       HyperImpute & (+$213\%$) \underline{0.26 $\pm$0.03} & \underline{0.09 $\pm$0.02} & \underline{0.22 $\pm$0.05} & \underline{32 $\pm$13} \\
       MIRACLE & 4.44 $\pm$0.48 & 4.32 $\pm$0.44 & 10.4 $\pm$1.51 & 189 $\pm$39\\
       NewImp & 415 $\pm$172 & 300 $\pm$152 & 726 $\pm$386 & 1,323 $\pm$245\\
        \midrule
      Diabetes\\
      \midrule
       F3I (ours) &  (+$10\%$) \underline{0.34 $\pm$0.05} & \underline{0.28 $\pm$0.04} & \underline{0.64 $\pm$0.15} & \textbf{0.12 $\pm$0.03}\\
       GAIN  & 0.54 $\pm$0.07 & 0.45 $\pm$0.07 & 0.87 $\pm$0.18 & \underline{26 $\pm$13}\\
       HyperImpute & \textbf{0.32 $\pm$0.05} & \textbf{0.25 $\pm$0.04} & \textbf{0.57 $\pm$0.15} & 43 $\pm$25\\
       MIRACLE & 5.61 $\pm$0.56 & 5.49 $\pm$0.55 & 13.22 $\pm$2.15 & 149 $\pm$42\\
       NewImp & 2.56 $\pm$0.71 & 1.83 $\pm$0.54 & 4.30 $\pm$1.19 & 1,016 $\pm$204 \\
       \midrule
        HeartDisease &  &   &  & \\
      \midrule
       F3I (ours) &  \textbf{0.14 $\pm$0.05} & \textbf{0.07 $\pm$0.03} & - & \textbf{0.13 $\pm$0.04}\\
       GAIN  & 0.30 $\pm$0.07 & 0.18 $\pm$0.05 &-  & 15.89 $\pm$4.47\\
       HyperImpute & (+$79\%$) \underline{0.24 $\pm$0.07} & \underline{0.13 $\pm$0.04} & - & \underline{35 $\pm$27}\\
       MIRACLE & 5.04 $\pm$0.74 & 4.84 $\pm$0.64 &  - & 101 $\pm$28\\
       NewImp & 361 $\pm$174 & 239 $\pm$139 &-  & 914 $\pm$223\\
       \midrule
       Ionosphere & & & &  \\
       \midrule
       F3I (ours) & (+$11\%$) \underline{0.23 $\pm$0.05} & \underline{0.16 $\pm$0.04} & \underline{0.31 $\pm$0.09} & \textbf{0.20 $\pm$0.04}\\
       GAIN  & 0.47 $\pm$0.05 & 0.35 $\pm$0.05 & 0.53 $\pm$0.12 & \underline{22 $\pm$13}\\
        HyperImpute & \textbf{0.22 $\pm$0.07} & \textbf{0.14 $\pm$0.04} & \textbf{0.27 $\pm$0.07} & 110 $\pm$92\\
       MIRACLE &  5.30 $\pm$0.48 & 5.21 $\pm$0.46 & 12.6 $\pm$1.65 & 100 $\pm$6\\
       NewImp & 0.61 $\pm$0.37 & 0.47 $\pm$0.22 & 1.02 $\pm$0.54 & 1,126 $\pm$63\\
       \bottomrule
    \end{tabular}
\end{table}

\subsection{Joint imputation-binary classification task}

We also looked at the trend in performance in PCGradF3I on the BreastCancer and Ionosphere data sets for increasing values of $\beta$, related to the importance of the classification task, in Table~\ref{tab:joint_MNIST_PREDICT_main2}. As expected, increasing values of $\beta$ improve the performance of the classifier, reaching a plateau in the average AUC value. 

\begin{table}
    \centering
    \caption{Average and standard deviation Area Under the Curve (AUC) values on a held-out testing set across runs on the joint imputation-classification task (MCAR scenario, $p^\text{miss}=0.5$). $\beta$ is the weight of the classification task in PCGradF3I. Bold type is the top performer, underline denotes the second best. Results for the BreastCancer data set are computed over 100 runs, over 60 runs for the Ionosphere data set.}
    \label{tab:joint_MNIST_PREDICT_main2}
    \begin{tabular}{lrrrr}
        \toprule
        Data set / $\beta$  & 0.14 &  0.25 & 0.50 & 0.75 \\
       \midrule
       PCGradF3I (Ionosphere) & 0.778$\pm$ 0.174  & \textbf{0.785$\pm$ 0.174} & \textbf{0.785$\pm$ 0.176} & \underline{0.783$\pm$ 0.174} \\
       PCGradF3I (BreastCancer) & \underline{0.699$\pm$ 0.142} & \textbf{0.700$\pm$ 0.143} & \textbf{0.700$\pm$ 0.142} & \textbf{0.700$\pm$ 0.142}\\
       \bottomrule
    \end{tabular}
\end{table}

\end{document}